%% file: main.tex
\documentclass[lettersize,journal]{IEEEtran}
\input{packages}

\input{commands}
\usepackage{graphicx}
\usepackage{placeins}
\usepackage{lipsum}
\usepackage{cleveref}
\usepackage[caption=false, font=footnotesize]{subfig}

\begin{document}

\title{Multistep Belief Space Dynamics Learning For Risk-Aware Control}

\author{Jason Gibson$^1$, Bogdan Vlahov$^1$, Patrick Spieler$^2$, Evangelos A. Theodorou$^1$
\thanks{$^{1}$ Autonomous Control and Decision Systems Lab, Georgia Institute of Technology, Atlanta GA 30313 USA (\textit{Corresponding author: jgibson37@gatech.edu)}}%
\thanks{$^{2}$Patrick Spieler are with the NASA Jet Propulsion Laboratory, California Institute of Technology,
        Pasadena, CA 91125, USA
        {\tt\small \{patrick.spieler\}@jpl.nasa.gov}}%
    \thanks{The research was carried out at the Jet Propulsion Laboratory, California Institute of Technology, under a contract with the National Aeronautics and Space Administration (80NM0018D0004). This work was  supported by Defense Advanced Research Projects Agency (DARPA). Approved for Public Release, Distribution Unlimited. \copyright 2026. All rights reserved.} %
}

\markboth{Journal of \LaTeX\ Class Files,~Vol.~14, No.~8, August~2021}%
{Shell \MakeLowercase{\textit{et al.}}: A Sample Article Using IEEEtran.cls for IEEE Journals}


\maketitle

\begin{abstract}
As autonomous vehicles move from a simplified research setting to practical use, there exists a large gap between the dynamic behavior of a human driving and an autonomous system.
Risk-aware behavior needs to naturally develop in order to scale to the demands of the real world.
A major issue for risk-aware planning and control has been predicting how dynamical uncertainty evolves through time and optimizing plans that account for this without being overly conservative.
Here, we present a learning framework to predict distributional dynamics that can be optimized in real time for \ac{MPC}.
We explore the importance of structure when learning distributional dynamics for use in \ac{MPC}.
A rigorous ablation study is conducted on a large dataset of real world off-road driving that shows the impact of deviations from our proposed structure.
Furthermore, we deploy our learned model and planning stack on a full sized vehicle in challenging off-road conditions.
Our planning architecture is able to naturally regulate the speed of the vehicle based on the environment and consistently demonstrates intelligent behavior over miles of diverse terrain.
\href{https://youtu.be/CaPhMgXMPG4}{Video}
\end{abstract}


\section{Introduction}

Autonomous navigation has been studied extensively in recent years, but most of the work has been devoted to structured environments like on-road autonomous driving \cite{Nahavandi2025}.
These structured environments are easy to setup experimentally and can have complicated interaction, but most of the testing in research settings lack the diversity of the real world environment.
These issues only increase in complexity when comparing closed-loop performance for these tasks \cite{Dauner2023PartingWithMisconceptions}.
Similar issues impact research for off-road driving.
Typically, planning and controls systems are tested repeatedly on the same small courses.
This can cause algorithms that perform well in these semi-structured environments but may not necessarily translate to the greater variance presented in the real world.
Our autonomous agents must be robust to disturbances and perturbations of the environment and should not fail under slight variations \cite{Cobbe2019QuantifyingGeneralization}.

\begin{figure}[t]
\centering
\includegraphics[width=\linewidth]{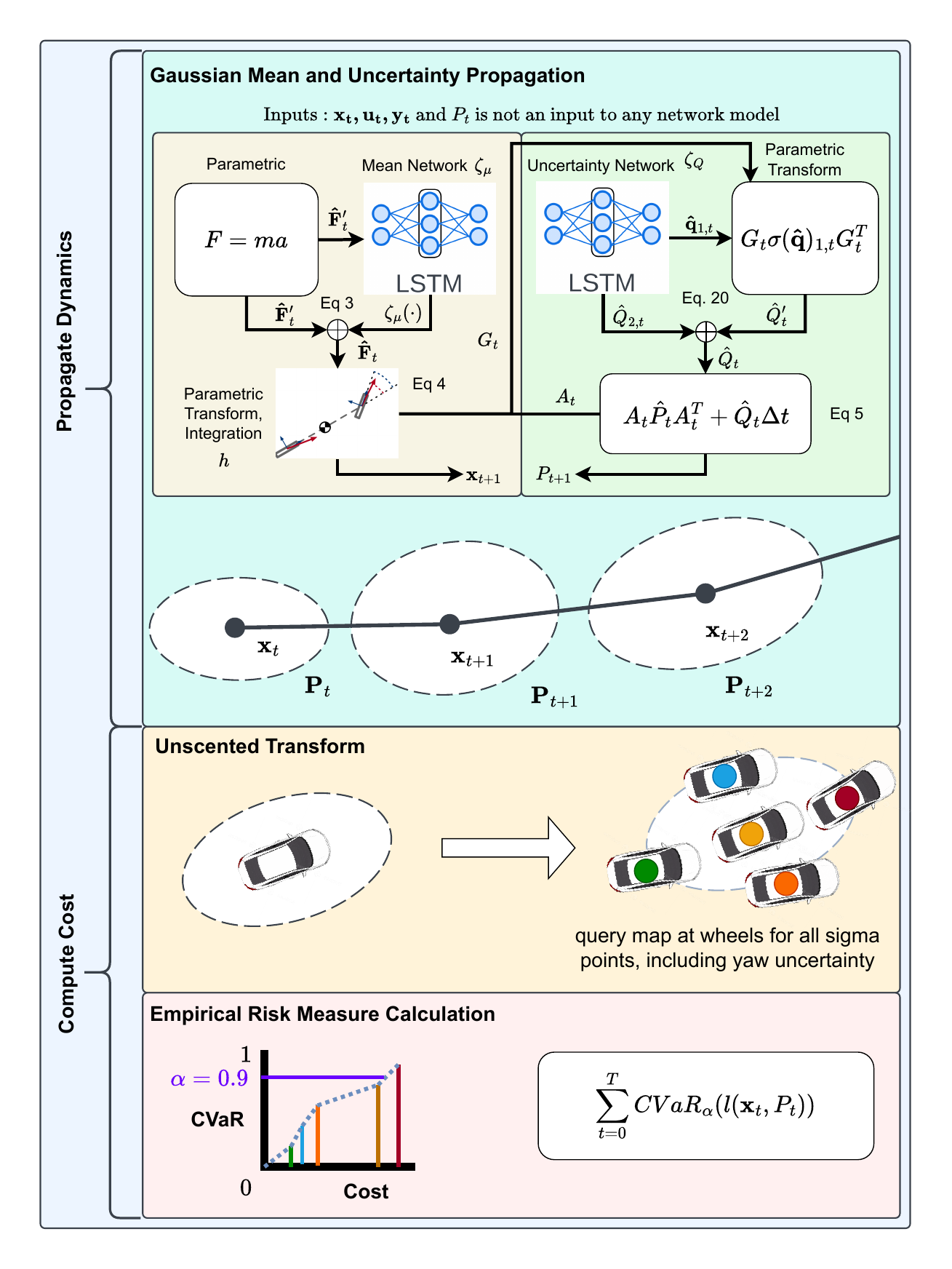}
\caption{Architecture of uncertainty propagation and risk measure computation for Model Predictive Path Integral (MPPI). 
A Gaussian distribution is propagated through time using both parametric and neural network components.
The cost of a trajectory is calculated using a risk measure computed on the empirical distribution of costs generated from sampling using the unscented transform.
}
\end{figure}

Our focus is on autonomous off-road driving at high speeds in varied terrain for miles at a time.
The environment in which our vehicle operates is diverse, unstructured, and unable to be reproduced in a small-scale test.
We want to develop autonomous agents that are able to adapt their behavior to a wide variety of new environments and conditions.
Parts of the navigation will be trivial while others will be extremely difficult.
We want a system that can adapt between these levels of difficulty and automatically make an informed decision about its trajectory and speed.
We focus on an application of navigation to a given GPS waypoint where no prior map exists, which can often be the case during rescue operations or in unmapped areas.
Incorporating additional pre-planning is orthogonal to this work and would integrate with our framework as a higher-level planner that would guide our system.
This is a domain that will have an outsized impact on disaster response, agricultural, mining, and extraterrestrial exploration, among others \cite{Jorge2019}.
Fundamentally, off-road autonomous driving is a great application for the development of planning architectures and algorithms because the vehicle cannot avoid the complexity of the real world.
Off-road autonomous driving encompasses difficult perceptual problems and cascading errors of that perceptual system must be handled gracefully in the planning stack.
Only algorithms that can operate well under a wide variety of conditions can navigate miles of off-road terrain successfully without failure.

Our work builds on substantial prior work on autonomous planning and control in the off-road driving space \cite{2021Subosits,AutoRally,Gibson2022,Vlahov2023,han2024dynamicsmodelsaggressiveoffroad}.
The focus is on the trajectory optimization problem for a fixed horizon and with a learned, uncertainty-aware representation for the dynamics.
The standard approach is to formulate this as a trajectory optimization task, and the literature is dominated by sampling-based approaches like \ac{MPPI} \cite{williams2017}.
This is because of the nonlinearity of the off-road dynamics, but mostly the discontinuous or multimodal nature of the cost functions.
Sampling-based methods often perform well in challenging tasks, provided that there are sufficient samples to optimize with.
The advent of GPUs has made this possible for many vehicles \cite{mppi-generic}.
Unfortunately, these methods can struggle with being overly aggressive, but we use uncertainty to mitigate this issue.

We believe that uncertainty-aware planning is the key to generalization of planning and control.
Uncertainty allows the system to make an informed decision and trade off risk and speed in an intelligent manner.
This behavior is similar to a human driving in a challenging off-road environment would drive quickly or even aggressively in open areas, while driving slowly and precisely in constrained ones \cite{Osman2010}.
Exactly how the driver modulates their speed occurs naturally as a function of the environment and vehicle.
Planning and control architectures must be able to replicate this behavior in novel configurations without arduous tuning and heuristics.

A core part of being able to optimize with uncertainty in mind is the ability to propagate well calibrated uncertainty through time.
We combine parametric and learned approaches to learn the dynamics of the vehicle in a hybrid architecture that is distributional.
Parametric approaches focus on using known physics-based equations and tend to generalize well in low data regimes.
Unfortunately, they also tend to overly simplify and give poor results for complex interactions.
Pure learning approaches tend to outperform parametric approaches for in-distribution data, but can fail when exposed to novel inputs.
Hybrid approaches combine the two approaches and will be the focus of this work.
They tend to also improve performance computationally since a smaller learned model is needed.
Furthermore, dynamics modeling for use in control provides additional complications when combining learned components.
The dynamics must be stable and cannot have exploitable edge cases under a wide variety of possible inputs, especially when using sampling-based optimizers.
Even small drifts can result in poor behavior when the controller can decrease the cost by doing something impossible.
Hybrid approaches help mitigate this issue through the stability provided by the parametric equations.

In a similar vein, the planning stack is based around creating principled, general models to predict human-understandable values that relate to safety constraints.
We prioritized general models over explicit modes often found in literature \cite{Dauner2023PartingWithMisconceptions} because the general principle scales with the complexity and diversity of situations we encountered.
These models are critical for determining safe speeds and orientations for complex maneuvers with infinite possible variations.
An example of this is a model for the suspension forces of the vehicle.
It is easier to design a cost function for something intuitive like the maximum allowed force on the tire than to classify geometric terrain as traversable or not for all possible speeds and orientations.
These models tend to generalize well since the system can use its knowledge of the environment and actions to arrive at good, emergent behaviors.
A great example of this is allowing the vehicle to roll backwards, while in forward gear, down a hill when the robot cannot proceed forward.
This saves the time that would be required from a gear shift but only works when the hill is steep enough to induce motion.

By combining the uncertainty with the principled approximations used in the cost function, we are able to achieve intelligent behavior in a wide variety of situations.
The structure provided by physics-based principles in these approximations aids in generalization and gives robustness to noise that is difficult to handle otherwise.
An example of this is the vehicle making a sharp turn.
The vehicle could slide into a nearby obstacle or potentially roll over depending on the exact speed and elevation.
The exact same turn could be acceptable in some locations given a specific risk profile or never acceptable in others.
In essence, we want the robot to behave conservatively near constraints and aggressively when unconstrained.
Uncertainty should only be penalized as it relates to the environment; a problematic amount of uncertainty is impossible to define without context.

Furthermore, we want the risk the robot is willing to take to be easily tunable with a small number of parameters.
The general algorithm should be usable in a wide range of situations that range from risky to safety critical.
Prior, versions of the software for this project relied on numerous speed and density-based rules to encourage good behavior. 
These were implemented as upper limits of speed as a function of distance to or density of potential hazards and served as a proxy for controller dynamical uncertainty.
They were difficult to tune, generalized poorly to new environments, and were overly conservative since directionality was difficult to capture.
Our implementation of uncertainty is able to drive faster with improved safety compared to the prior approach.

This paper presents a risk-aware planning architecture, based on \ac{MPC}, for autonomous off-road driving that demonstrates shrewd behavior on a real full-size vehicle.
The focus is on learning dynamical uncertainty that is directional, specific, and related to the constraints at issue for the control task.
These are combined with an \ac{MPC} that does uncertainty-aware trajectory optimization in real time that allows the vehicle to adapt its behavior to the environment at hand.
The structure of our learning and trajectory optimization problem provides a single setup that could handle driving miles at a time without interventions in diverse scenarios \cref{fig:long_traverse} with essentially no retuning of the planning architecture.
Our main contributions of this work are,
\begin{enumerate}
    \item A novel framework for learning computationally efficient distributional dynamics on a general nonlinear system
    \item An enlightening ablation study of adding and removing structure of the learning problem applied to real-world data of an off-road vehicle in challenging environments
    \item Validation of the improved performance and context-aware emergent behavior of planning on real-world hardware when using uncertainty
\end{enumerate}

The paper is outlined as follows. 
First in \cref{sec:belief_dynamics_learning}, we will explain the distributional dynamics and how we provide structure to the learning problem in the distributional and mean sense.
Second, we will cover the control architecture in \cref{sec:control_architecture} and explain how the cost function is constructed and evaluated in \cref{subsec:cost_function_design}. 
Finally, we will provide an ablation study that highlights the important parts of the structured learning problem in \cref{subsec:ablation_studies} and show the practical ability of the entire architecture in a diverse set of environments in \cref{subsec:hardware_results}.

\section{Related Works}

Our work borrows heavily from prior work on using physics-based parametric equations for vehicle modeling in drifting scenarios in 3D \cite{Goh2020TowardAutomatedVehicleControl, 2021Subosits}.
Physics-based tire models like Pacejka \cite{pacejkaModel, Pacejka2005-ed} or Fiala \cite{2021Subosits} are commonly used and give structure to the complex interactions.
Our work uses a simplification of these parametric equations but matches the underlying structure.
Related work on parametric models of 3D dynamics models has been done on the same vehicle \cite{han2024dynamicsmodelsaggressiveoffroad,Gibson2022,gibson2024dynamicsmodelingusingvisual}.
These works focus on mean prediction in multiple terrains and use parametric models augmented with learning.
A key learning from our work is how important the structure of the learning problem is; we see the same trend in publications like \cite{Chrosniak2024DeepDynamics} where constants for tire models are predicted using a neural network or papers that used physics-informed learning \cite{Djeumou2023AutonomousDrifting}.

The difference in our approach from the related literature is the specific structure of the dynamics learning problem and its experimental validation.
Our approach to learning the process noise was inspired from \cite{Russell2021MultivariateUncertainty} where a Kalman filter's measurement noise covariance is predicted using a neural network.
A similar work \cite{Revach2022KalmanNet} learns a \ac{RNN} to predict the Kalman gain to get a better mean estimate and only track the covariance implicitly whereas we rely on the structure of the covariance update and focus on the full distribution's accuracy.
An alternative approach predicts process noise, measurement noise, and Kalman gain as a policy \cite{Luo2022LearningBasedNoiseTracking}.
They predict a noise vector and transform it into the matrix form of the process noise instead of predicting the matrix directly.
We avoided other approaches to uncertainty propagation like \ac{PC} \cite{OHagan2013PC} or \ac{GPR} for computational reasons though they are heavily used in related publications \cite{Trivedi2023ProbabilisticDynamicModeling,Trivedi2024DataDrivenSampling,Ko2007GP-UKF}.
Quantile methods \cite{Chen2020QuantileSurvey} were ignored since it was unclear how to propagate them through time with structure \cite{zhang2020quantilepropagation} in a computationally efficient manner, but their additional flexibility can be important for heavily skewed distributions.
Very similar work has been done propagating a Gaussian distribution with \acp{GP} \cite{Trivedi2023ProbabilisticDynamicModeling,Ko2007GP-UKF,Trivedi2024DataDrivenSampling}, but those works use the \ac{GP} to estimate the process noise of each output and time-step independently rather than jointly.
The other common approach for distribution propagation is to use an ensemble of networks predicting a diagonal Gaussian distribution at each time step, first presented in \cite{Chua2018DeepReinforcementLearning}.
This parameterization is different from ours since it lacks structure from the linearization of the dynamics and is more focused on approximating aleatoric vs epistemic uncertainty for active exploration in \ac{MBRL} \cite{Lee2023TerrainAwareKinedynamic,Kim2023BridgingActiveExploration}.
Versions of this use the unscented transform to reduce the learning problem to learning nonlinear mean dynamics that is reused on each sigma point \cite{Parwana2024RiskAwareMPPIHybrid} in the style of the \ac{UKF}.
While, this version might result in more accurate propagation of the dynamics due to the unscented transform, it is computationally intensive.

Multiple approaches exist to add distributional understanding to nonlinear \ac{MPC} that are similar to ours, but all deviate in how the distribution is represented or propagated when compared to our work.
We focus on related works using \ac{MPPI} but other work exists that tackle the optimization with other methods \cite{Hakobyan2023DistributionallyRobust}, which do not allow for general cost functions of our form.
\ac{MPPI} and \ac{CS} are combined in \cite{Yin2022TrajectoryDistributionControl}, where the \ac{CS} is used to ensure a specific terminal covariance in the sampling, but the cost of a trajectory is still computed only at the mean.
Approximations that do not use \ac{MCS} rely on approximations or simplifications, like ellipsoidal approximations or quadratic cost, to represent risk-aware costs \cite{Mohamed2023TowardsEfficientMPPI,Trivedi2024DataDrivenSampling}.
Chance constraint-based reformulations are explored in \cite{Trivedi2024DataDrivenSampling,Trivedi2023ProbabilisticDynamicModeling} but require simplification of constraints and would be implemented as soft penalties anyway.
A common approach for \ac{MCS} is to use the unscented transform \cite{Parwana2024RiskAwareMPPIHybrid,Lee2023TerrainAwareKinedynamic,Kim2023BridgingActiveExploration}. 
Our work also deviates from applying a penalty on the state covariance directly like in \cite{Trivedi2023ProbabilisticDynamicModeling,Lee2023TerrainAwareKinedynamic,Kim2023BridgingActiveExploration}.
The penalty on risk makes logical sense in active exploration for \ac{MBRL} \cite{Lee2023TerrainAwareKinedynamic}, especially when a separation of aleatoric vs epistemic uncertainty is available \cite{Kim2023BridgingActiveExploration}.
Our work only really provides an estimation of aleatoric uncertainty, but could be used as a component in an ensemble similar to these works.
We found that a penalty on uncertainty directly was difficult to tune in practice for diverse scenarios; large uncertainty is unavoidable at high speeds and it should be possible to drive in open areas with low risk.
Possible risk measures include \ac{CVaR} \cite{Wang2021AdaptiveRiskSensitive, Yin2023RiskAwareMPPICVAR} or maximum cost \cite{Parwana2024RiskAwareMPPIHybrid}.
\cite{Yin2023RiskAwareMPPICVAR} uses the CVaR with a hand-tuned threshold as a penalty function which can be difficult to tune.

\section{Methods: Belief Dynamics Learning}
\label{sec:belief_dynamics_learning}

\textbf{Notation and Conventions:}
All bolded values are vectors, $\dot{x}$ is the derivative of $x$, and $\hat{x}$  indicates the predicted value of $x$.
Velocities are all in body frame, $v^x, v^y$ are body velocities in $x, y$ respectively at the center of mass.
Angular rotations are represented with Euler angles using $\phi, \theta, \psi$ for roll, pitch, and yaw, respectively. 
$c$ represents constants or vectors of constants when needed, the subscript differentiates them.

We model the system as a discrete-time nonlinear dynamics of the form
\begin{align}\label{eq1:dynamics_discrete}
    \vx_{t+1} &= f(\vx_t, \vu_t, \vy_t, w_t),
\end{align}
where $\vx_t \in \mathbb{R}^{n_s}$, $\vu_t \in \mathbb{R}^{n_y}$ are the state and controls respectively, $\vy_t \in \mathbb{R}^{n_y}$ are state-dependent sensor readings that could impact the dynamics (such as an elevation reading), and $w_t$ is unknown noise.
A key note is that during training the values of $\vy_t$ are independent from the prediction of $\hat{\vx}_t$.
The values of $\vy_t$ during training are always the values seen at the true state $\vx_t$ during dataset collection.
At runtime the values of $\vy_t$ are determined using $\hat{\vx}_t$.
This can cause a slight mismatch in the learned dynamics and their performance in practice, but using prediction-based values invalidates the trajectory loss due to compounding error anyway.

Our goal is to approximate open-loop stochastic dynamics with a compact belief-space representation over some prediction horizon $T$. 
We  model the uncertainty as a Gaussian distribution to simplify the learning process and leverage known equations for propagation of Gaussians with linearized dynamics.
A parametric approximation $\hat{f}$ of~\eqref{eq1:dynamics_discrete} is an important building block for linearization and defining the relationships between variables.
Our work will show that using some reasonable approximation of parametric dynamics defines a relationship between variables that can provide needed structure to the learning problem.
That is combined with some learned compensation $\zeta$, represented by a neural network.
The stochastic dynamics is  approximated by
\begin{align}
    \hat{\vF}'_t &= \hat{f}(\hat{\vx}_t, \vu_t, \vy_t) \label{eq:general_force_equation}\\
    \hat{\vF}_t &= \hat{\vF}_t' + \zeta_\mu(\hat{\vx}_t, \vu_t, \vy_t, \hat{\vF}_t') \label{eq:general_hybrid_mean_update}\\
    \dot{\hat{\vx}}_{t} &= h(\hat{\vF}_t) \label{eq:general_mean_derivative_with_transform}\\
    \hat{P}_{t+1} &= A_t \hat{P}_t A_t^T + \hat{Q}(\hat{\vx}_t, \vu_t, \vy_t, \hat{\vF}_t) \Delta t,\label{eq:general_unc_update}
\end{align}
where $\vF$ represents force, $h$ is a function that maps the predicted forces and previous states into the correct frame for the derivative using kinematics and known transforms, $A_t$ is the jacobian of the parametric dynamics including $h$, and $\hat{Q}(\cdot)$ is a learned function we will define later in \cref{subsec:process_noise_learning}.
We found that prediction in force space can greatly improve performance when known geometric transforms exist (see \cref{subsubsec:structure_ablation}).
The derivatives of the states are predicted as outputs to encourage smoothness when integrated with forward Euler integration.

Logically and experimentally, we see that the use of a recurrent architecture, like an \ac{LSTM}, would help with prediction; see \cref{subsubsec:structure_ablation} for a comparison with other network architectures.
The network architecture is an \acp{I-LSTM} first proposed in \cite{Mohajerin2019} where the initial hidden and cell state of the \ac{LSTM} is the output of another \ac{LSTM} running over some historical buffer of values.
We shall refer to the \ac{LSTM} predicting the initial hidden and cell state as the initialization network whereas the \ac{LSTM} used for dynamics compensation will be the predictor network.
There are also additional fully connected layers that transform the input and predictor hidden state into an output vector of correct size, referred to as the output networks.
They are helpful in providing some ability to make the outputs nonlinear functions of the local context.
The initial state and thus initial hidden\slash cell states are the same for all samples, so the initialization network only runs once for each full computation of \ac{MPPI}.

\subsection{Process Noise Learning}
\label{subsec:process_noise_learning}

\cref{eq:general_unc_update} gives a structured method to propagate the uncertainty, but not how we should predict the process noise matrix $\hat{Q}(\cdot)$.
First, we will reduce the size of the covariance matrix to a subset of the state space $\tilde{\vx} \in \mathbb{R}^{n_{\tilde{x}}}$ where $n_{\tilde{x}} \leq n_x$.
We denote the subset of equations that predict the values in $\tilde{\vx}$ with $\hat{f}_{\tilde{x}}$ and we linearize the parametric dynamics including the transform around $\hat{\tilde{\vx}}_t$ to get
\begin{align}
    A_t = \partial h(\hat{f}_{\tilde{x}}(\cdot)) / \partial \tilde{\vx}_t
\end{align} and use that to propagate the covariance as in \cref{eq:general_unc_update}.
This reduction is helpful for computational reasons and may not significantly impact the accuracy on those states if the covariance matrix is sparse.
If all variables are tightly coupled and the matrix is dense, then using the entire state vector may be advisable.
Second, we will split the prediction of the process noise into a hybrid structure like we have for the mean dynamics.
The process noise is predicted using a neural network $\zeta_Q(\cdot)$ whose output is split into a structured and unstructured component.
The structured output $\vq_t \in \mathbb{R}^{n_q}$ is made of independent noise values for two types of values: states not included in $\tilde{\vx}$ denoted by $\vpi^{\tilde{x}}$ and outputs in the force space denoted by $\pi^F$ concatenated into $\vpi = [\vpi^{\tilde{x}}, \vpi^F]$.
These values are transformed using,
\begin{align}
    G = \partial h(\hat{f}_{\tilde x}(\cdot)) / \partial \vpi
\end{align}
which relies on the parametric dynamics $\hat{f}$ (\cref{eq:general_force_equation}) and transform function $h$ (\cref{eq:general_mean_derivative_with_transform}) to relate these values to the uncertainty of values in $\tilde{\vx}$.
The unstructured output $\hat{Q}' \in \mathbb{R}^{(n_{\tilde{x}}-1)n_{\tilde{x}}/2}$ is predicted as an upper-diagonal matrix and is turned into a symmetric matrix by copying values reflected across the diagonal.
We predict the process noise $\hat{Q}$ from \cref{eq:general_unc_update} using,
\begin{align}
    \begin{bmatrix} \hat{\vq}_{t}, \hat{Q}_{t}'\end{bmatrix} &= \zeta_{Q}(\hat{\vx}_t, \vu_t, \vy_t, \hat{\vF}_t)\\
    \hat{Q}_t &= G_t \left(\sigma(\hat{\vq}_{t}) \odot c_\sigma\right) G_t^T + \kappa(\hat{Q}_{t}') \odot \vc_\kappa \label{eq:specific_unc_update}
\end{align}
Where $\odot$ is the Hadamard product (element-wise multiplication of values) and $\sigma(\cdot)$ is the sigmoid function which is scaled by parameter vector $c_\sigma$.
The diagonal values in $\hat{Q}'$ are also modified by the scaled sigmoid function while the off-diagonal elements are modified using $\tanh$, denoted by $\kappa: \mathbb{R}^{(n_{\tilde{x}}-1)n_{\tilde{x}}/2} \rightarrow \mathbb{R}^{n_{\tilde{x}} \times n_{\tilde{x}}}$.
This was done to clip the values in a continuous way.
The values for $c_\sigma$ will fit to be small for variables that should not be represented and can be removed to simplify computation.
We can simplify the architecture by removing the prediction of $\hat{\vq}$ and only predicting $\hat{Q}'_t$.
$A$ and $G$ provide structure for relationships between the different variables that helps to simplify the learning process.
The ablation study in \cref{subsubsec:structure_ablation} will show that the more structured version outperform networks without structure.

We used the state estimator to initialize $\vx_0$ but compared multiple methods to predict an initial $P_0$. 
A logical choice would be the predicted covariance from the state estimator for $P_0$, but its accuracy was dubious over all the collected data.
Therefore, we compared predicting the initial covariance using the initialization network or meta-learning them from data.
We also included a baseline where the diagonals were set to constant small fixed values ($1e-5$).
In order to predict a positive definite initial matrix, we match the structure used with $\hat{Q}_2$ and $\hat{\vq}$ to ensure symmetric positive definite matrices.

The loss function we will use is the \ac{NLL} loss function summed across the entire time horizon $T$ with additional regularization,
\begin{align}
    \mathcal{L} = \sum_{t=0}^T &\frac{1}{2}(\tilde{\vx}_t - \hat{\tilde{\vx}}_t)^T \hat{P}_t^{-1} (\tilde{\vx}_t - \hat{\tilde{\vx}}_t)\nonumber \\
        &+ \frac{1}{2} \ln{|\hat{P}_t|} + \|\vc_\pi(\pi_t - \hat{\pi}_t)\|_2
    \label{eq:loss_function}
\end{align}
Note that the \ac{NLL} is only on the subset of states in the covariance matrix while those outside of it are still penalized with a L2 loss scaled by $\vc_\pi$.
A trajectory-based loss function provides multiple important properties as outlined in \cite{Gibson2022} but summarized in three main points.
First, it encourages stable propagation of the mean state and belief dynamics, since the accumulated error of recursive model prediction is explicitly accounted for.
Second, it allows us to parameterize the dynamics however we choose and only require state labels for learning.
Third, our loss function penalizes correct integration, so our forward Euler step is learned to be stable at a specific $\Delta t$.
However, we still see some oscillations in practice that can cause instability in learning with stiff parametric equations.

\subsection{Closed Loop Uncertainty Approximation}

\begin{figure*}
    \centering
    \begin{subfloat}[HW Open Vs Closed Loop\label{fig:closed_loop_xy_plot}]{
    \centering
    \includegraphics[page=2, trim={0cm, 0cm, 0cm, 2.5cm}, clip, width=0.23\textwidth]{Figures/trajectory_results/closed_loop_HW_nice_figures_vertical_processed.pdf}
    }
    \end{subfloat}
    \begin{subfloat}[HW Mean Trajectory \label{fig:hw_mean_trajectory}]{
    \centering
    \includegraphics[page=1, trim={0cm, 0cm, 0cm, 2.5cm}, clip, width=0.23\textwidth]{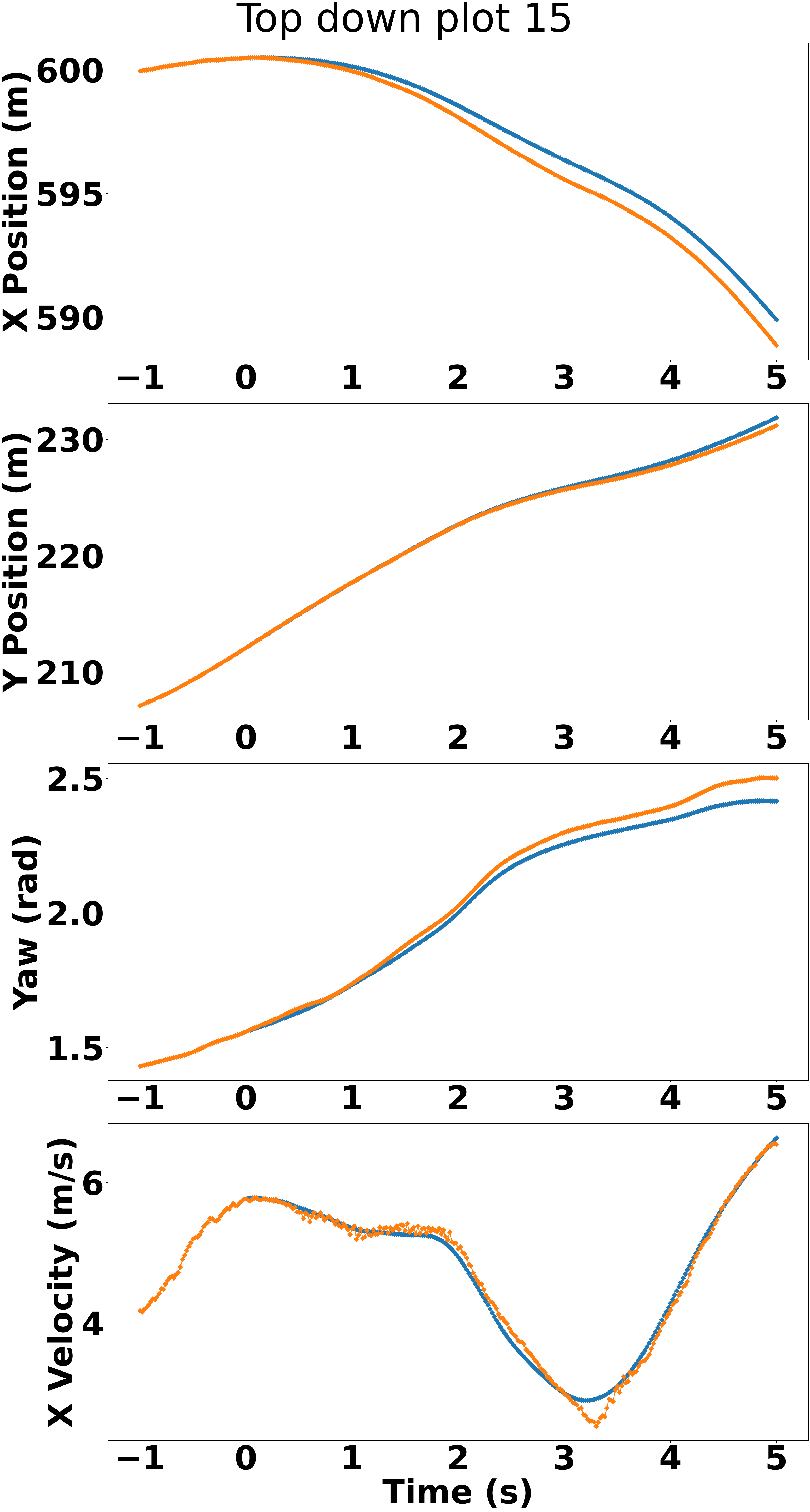}
    }
    \end{subfloat}
    \begin{subfloat}[HW Open-Loop Sigma Points \label{fig:hw_open_loop_sigma}]{
    \centering
    \includegraphics[page=3, trim={0cm, 0cm, 0cm, 2.5cm}, clip, width=0.23\textwidth]{Figures/trajectory_results/closed_loop_HW_nice_figures_vertical_processed.pdf}
    }
    \end{subfloat}
    \begin{subfloat}[HW Closed-Loop Sigma Point \label{fig:hw_closed_loop_sigma}]{
    \centering
    \includegraphics[page=4, trim={0cm, 0cm, 0cm, 2.5cm}, clip, width=0.23\textwidth]{Figures/trajectory_results/closed_loop_HW_nice_figures_vertical_processed.pdf}
    }
    \end{subfloat}
    \caption{
        State trajectory and sigma point plots of the HW model with and without the closed-loop gains. 
        In \cref{fig:closed_loop_xy_plot} True state values are in orange, the predicted trajectory by the model is in blue, the green circles are the open-loop $xy$ uncertainty ellipses, and the purple indicates the closed-loop uncertainty.
        \cref{fig:hw_open_loop_sigma,fig:hw_closed_loop_sigma} show plots of the states and the sigma points are shown in their respective colors along with a red line indicating the mean error of the model.
    }
    \label{fig:closed_loop_trajectories}
\end{figure*}

What the learning pipeline captures is the open-loop uncertainty as a Gaussian distribution. 
This often leads to overly large uncertainties at the end of the trajectory, which is problematic for planning and incorrect when applied in an \ac{MPC} fashion.
Uncertainty tends to correlate with speed, so this would prevent high-speed operation and navigation in tight spaces.
We should expect that the uncertainty size is limited through \ac{MPC} and that the vehicle can track tighter than an open-loop prediction would indicate.
To reduce the covariance at runtime only, we apply feedback gains to the state Jacobian in \cref{eq:general_unc_update},
\begin{align}
    A_t = \frac{\partial \hat{f}_{\tilde{x}}}{\partial \tilde{\vx}_t} - K \frac{\partial \hat{f}_{a_u}}{\partial \tilde{\vu}_t}
\end{align}
Where $\hat{f}_{a_u}$ is a simpler form of the dynamics since these values for $K$ are hand-tuned.
For our specific system, we treat the controls as directly controlling acceleration in body x frame $\dot{v^x_t}$, and the steering angle $\delta$.
Since we have no state trajectory to track during the sampling process, we can treat the state error as zero, so the mean trajectory is not modified by this change.
An example propagation of the open-loop vs closed-loop is shown in \cref{fig:closed_loop_xy_plot}.
This should be seen more as a heuristic to move in the direction of a closed-loop uncertainty rather than a true estimate of what it should be.
Learning this properly would require a dataset tracking a fixed trajectory in a variety of conditions; this dataset was never collected.

\subsection{Specific Hybrid Dynamics}

The vehicle modeled is a Polaris S4 1000 RZR.
The equations in this section represent the parametric components of the dynamics $\hat{f}$ and transforms $h$ that are used in \cref{eq:general_force_equation,eq:general_mean_derivative_with_transform} and are the building blocks of $A, G$ in \cref{eq:general_unc_update,eq:specific_unc_update}.
The specifics of which variables are used in $\tilde{\vx}_t$ and $\vpi_t$ are outlined in the results section when describing the baseline model \cref{sec:results}.
We will be outlining equations for five semi-independent systems.
The first three are the delay systems that are used to predict components of the physical state of the vehicle.
Those are the steering angle $\delta$, the current brake pressure $b$, and the engine RPM $e$.
These create the basis for inputs to the main parametric model that predicts forces in the body $x,y$ and a rotational force for yaw.
The final model is a simple spring-mass-damper system to model suspension that is primarily used in the costing and does not impact the main parametric model for forces.
The state will be 16 values in three groups $(p^x, p^y, \psi, v^x, v^y, \dot{\psi}), (b, e, \delta, \dot{\delta}), (p^z, \phi, \theta, \dot{p}^z, \dot{\phi}, \dot{\theta})$ where the groups are predicted using the main force model, the delay models, and the suspension system respectively.
Predictions from the force model will depend on the delay models but the force model is independent of the suspension model other than using its estimated pitch and roll for the transform from body location to wheel location for querying the elevation map.

\subsubsection{Delay Models}
For the delay systems, the controls are given by $u^t$ for throttle, $u^b$ for brake, and $u^\delta$ for the steering angle.
The steering angle command is tracked by an internal controller in the steering assembly that could not be modified and adds delay.
Three separate models are learned to predict states that correspond to the physical state of the vehicle that are used in the force model.
Sensor readings giving correct values for these are collected from the CAN bus during operation of the vehicle to provide labels.
We will be using the same delay models for all comparisons; they are trained using \ac{MSE} loss on values they predict and remain fixed during training of the other networks unless otherwise noted.
The delay models are outlined below,
\begin{align}
    \dot{e}_t &= \zeta_e(\hat{v}^x_t, u^t_t)\\
    \dot{b}_t &= \left\{ \begin{array}{ll}
            \gamma\!\left(\hat{x}^b_t - u^b_t; c_{b+}\right) & \hat{x}^b_t \leq u^b_t \\
            \gamma\!\left(\hat{x}^b_t - u^b_t; c_{b-}\right)&\text{otherwise}
        \end{array} \right.\\
    \ddot{\delta}_t &= \zeta_\delta(\hat{v}^x_t, \delta_t, \dot{\delta}_t, u^\delta_t) + \gamma\!\left(\delta_t - u^\delta_t; c_\delta\right)
\end{align}
Where $\gamma(\alpha; \vc) = tanh(\vc[0]\alpha + \vc[1] \alpha^2)\vc[2]$ and was chosen for the brake and steering delays as a smooth function to implement a maximum rate.
The delay models are self-contained except for the engine and steering delay systems taking in the predicted $\hat{v}^x_t$.
These models are trained independently from the the main parametric model and we replace the x velocity estimate $\hat{v}^x_t$ with the true values from the state estimator during training.
The physical brake actuator has different dynamics in increasing or decreasing the brake pressure, so those cases are separated.

\subsubsection{Parametric Force Model}
The parametric force model is based on a bicycle model, where tire forces and yaw rate are predicted using a hybrid architecture.
It is critical that the errors of the delay models are part of the training process of the force model, especially when delay models tend to have consistent errors.
The effects on performance are explored in the \cref{subsec:ablation_studies}.
The force model differs from prior work on this vehicle \cite{Gibson2022,Han2023ModelPredictiveControl,han2024dynamicsmodelsaggressiveoffroad} by including an RPM prediction, rolling resistance, and using the elevation map normals rather than the approximated pitch from the elevation map.
The model is identical to the one published concurrently in \cite{gibson2024dynamicsmodelingusingvisual}, but that work only includes mean dynamics.
Normals $\eta = [\eta_x, \eta_y, \eta_z]$ are queried at the wheels using a smoothed elevation map, then rotated into the body frame and averaged across the four wheels.
The elevation values are generated using \cite{Fan2021, GVOM, atha2024fewshotsemanticlearningrobust} and have inherent error; see those publications for related discussion of failure cases.

The force equations are inspired from \cite{2021Subosits} but use a slight modification of the Pacejka tire model from \cite{2020Metzler},
\begin{align}
    F_x &= \left(P\left(e\right) P\left(u^{t}\right) - \tilde{P}\left(b; v^x\right) - \beta\left(v^x\right)\right) \eta_z \label{eq:x_front_force}\\
    F_{yf} &= \left(c_D\text{sin}\left(c_C\text{tanh}\left(c_B\alpha_R\right)\right)\right) \eta_z \label{eq:y_front_force}\\
    F_{yb} &= \left(c_D\text{sin}\left(c_C\text{tanh}\left(c_B\alpha_F\right)\right)\right) \eta_z \label{eq:y_back_force}\\
    F_r &= \left(\frac{v^x}{c_L} \tilde{\delta}\right) c_{r} - c_{r, d} \dot{\psi},\label{eq:yaw_rate_force}
\end{align}
where $\vf = [F_x, F_{yf}, F_{yb}, F_r]$ defines a vector of forces, $P$ defines a quadratic polynomial function, $v^x, v^y, \dot{\psi}$ are the body velocities and yaw rate respectively, and $c_D, c_C, c_B$ etc. are all fit constants.
We convert the steering angle $\delta$ from degrees the steering wheel is rotated as reported on the CAN bus to a wheel angle by scaling by a constant value, $\tilde{\delta} = \delta / c_\delta$.
The tire slip angles for front and rear wheels are given by,
\begin{align}
    \alpha_F &= \text{arctan}\left(\frac{v^y + c_L \dot{\psi}}{\text{max}(c_{max},v^x)}\right) - \tilde{\delta},& \label{tire3}\\
    \alpha_R &= \text{arctan}\left(\frac{v^y - c_R \dot{\psi}}{\text{max}(c_{max},v^x)}\right),&\label{tire4}
\end{align} where $c_{max}, c_R, c_F$ are learned parameters. 
These forces are the most important part of the parametric model, because they capture the complex interactions between variables for tire force and naturally encourage the vehicle to predict a stable slide.
We can only get an unrealistic sideways slide if the networks give consistently poor predictions.
We find modeling the engine torque with the delayed RPM state important because there is a large delay between actuating the throttle $u^t$ and the force $F_x$ changing.
The polynomial product $P(e_t) P(u^t_t)$ provides structure since it approximates the surface that tends to show up when learning torque curves of gasoline engines \cite{2010Shakouri}.
The yaw rate model does not use the wheel torques because the two-wheel approximation gave poor results when attempted.
We kept this model since the empirical yaw rate error is small and the model is very stable.

An important detail is proper stiffness of the force equations for $F_x, F_{yf}, F_{yb}$ when learning dynamics for control.
The vehicle must be able to stop when navigating close to obstacles, so we have modified the brake in \cref{eq:x_front_force} for maximum stiffness without oscillations at lower speeds.
$\tilde{P}$ in \cref{eq:x_front_force} multiplies the force by $v^x$ when the value would result in a sign change of $v^x$ to damp out oscillations.
We also modify the tire slip angles in \cref{tire3,tire4} with $c_{max}$, creating a trade-off between the slip angle accuracy and the stiffness.
This modification reduced issues with oscillatory behavior at low speeds that interfered with planning by allowing small amounts of movement in the body y direction while braking.

The forces are transformed into accelerations in body frame using \cite{2021Subosits},
\begin{equation}
\begin{aligned}
    \dot{v}^x &= \frac{(1 + \cos\tilde{\delta}) F_x - F_{yf} \sin\tilde{\delta}}{m} - c_{x, d} (v^x)^2  - c_{x, g}\eta_x + v^y \dot{\psi}, \\
    \dot{v}^y &= \frac{F_{yb} + \cos\tilde{\delta} F_{yf} + F_x \sin\tilde{\delta}}{m} - c_{y, d} (v^y)^2  - c_{y, g}\eta_y - v^x \dot{\psi}, \\
    \dot{\psi} &= F_r,\quad \dot{\vp} = R(\psi)\vv
\end{aligned}
\label{eq:hybrid_transform}
\end{equation}
where $c_{\cdot, d}, c_{\cdot,g}$ are the drag and gravity coefficients, $m$ is the vehicle mass, and $R(\psi)$ is the rotation matrix given by the yaw.

\subsubsection{Suspension System}
We use a purely parametric model to predict the propagation of the roll $\phi$, pitch $\theta$, z position $p^z$, and their corresponding rates.
This model serves two primary purposes: first, it gives an approximation of forces from the suspension system for the cost function and second, it provides the $p^z, \phi, \theta$ that are used to determine the $p^x, p^y$ values to query for the wheels.
Knowing when and how much to slow down for an arbitrary geometric obstacle was one of the most challenging tasks in this project.
The suspension system is the general model that provides us with a generalizable constraint through the estimated forces acting of the vehicle.
Unlike previous work \cite{2021Subosits}, we did not couple suspension and forward dynamics since elevation map errors made the dynamics very noisy.
We kept these dynamics purely parametric since they performed well enough in practice and were simpler to tune.

We model the suspension forces as a spring-mass-damper system connected to the elevation map.
The system ignores gravity and behaves identically in the up and down directions for force.
The spring-mass-damper is modeled in world frame and given by,
\begin{align}
    \dot{h}_t &= -v_{x, t} \cos(\tilde{\delta}) \eta_x - v_{x, t} \sin(\tilde{\delta}) \eta_y\\
    F_{w, u} &= -c_k (z_{w} - \tilde{z}_{w}) - c_{\dot{k}} (\dot{z}_{w} - \dot{h}_t) \label{eq:suspension_spring}
\end{align}
where $z_{w}$ is the position of the bottom of the wheel in world frame, $\dot{z}_w$ is the velocity in world frame at the wheel center location, and $\tilde{z}_{w}$ is the height of the elevation map in the world frame at the location of the wheel.
$F_{w, f}, F_{w_s}$ are approximations of the wheel frame $x$ and $y$ tire forces and are only used in the cost function.
%
The suspension system is propagated by approximating a decoupled system for roll and pitch,
\begin{align}
    v^z &= \frac{1}{|\mathcal{W}|}\sum_{w \in \mathcal{W}}F_{w, u} / c_{w, m}\\
    \dot{\phi} &= \frac{1}{|\mathcal{W}|}\sum_{w \in \mathcal{W}} F_{w, u} c_{cg, y} / c_{I_{xx}}\\
    \dot{\theta} &= \frac{1}{|\mathcal{W}|}\sum_{w \in \mathcal{W}}-F_{w, u} c_{cg, x} / c_{I_{yy}}
\end{align}
where $\mathcal{W}$ is the set of wheels and $|\mathcal{W}|$ is the number of wheels, so we are averaging the force over all wheels.
$c_{cg, x}, c_{cg, y}$ are the distance from the wheel to the center of gravity in body $x$,$y$ respectively.
The rest of the constants are hand tuned from human driving data to give reasonable performance.
Note that $c_{w, m} \neq c_{m}$ from earlier, $v^z$ is in world frame and $p^z, \phi, \theta$ are computed using Euler integration of the computed rates.

\section{Control Architecture}
\label{sec:control_architecture}

Our control architecture is based around \ac{MPPI}, a sampling-based stochastic optimizer that can handle nonlinear, discontinuous dynamics and cost functions.
The algorithm can be computed quickly using parallelization on the GPU \cite{mppi-generic}.
We are using a modification on \ac{MPPI}, Colored \ac{MPPI} \cite{Vlahov2023}.
This changes the standard \ac{MPPI} algorithm to sample in frequency space, resulting in correlated noise when transformed back into the time domain.
The main impact of this sampling procedure is increased probability to sample complex maneuvers such as sharp turns that require sign consistency to achieve.

\ac{MPPI} \cite{williams2017, Vlahov2023} computes the optimal control sequence based on a weighted average over multiple samples,
\begin{align}
    \vu_t^* &= \mathbb{E}_{V \sim P(\theta)}\left(w(V) \vv_t\right)\\
    w(V) &= \frac{1}{\eta} \exp \left(\frac{1}{\lambda} S(V, \hat{\vx}_0)\right)\\
    S(U, \hat{\vx}_0) &= \Phi(\hat{\vx}_{0:T}) + \sum_{t=0}^{T}l(\hat{\vx}_t, \vu_t)
\end{align}
Where $\vu_t^*$ is the optimal control for each time step, $V$ is a control trajectory sampled from the distribution $P(\theta)$ with $\vv_t$ being a single timestep, $\Phi(\cdot)$ is the terminal cost that depends on the entire trajectory and $l(\cdot)$ is the running cost.
We will modify the $l$ function in the following section to be a distributional cost, but the update equations remain unchanged.

\subsection{Unscented Transform}

When designing our distributional system, we wanted to allow for general cost functions similar to what was used in the past \cite{Vlahov2023}.
These are not easily lifted into distributional values without a sampling approach since they are generally discontinuous.
The Unscented Transform \cite{2000UKF} provides a structured approximation of the value of applying a nonlinear function to a distribution.
It creates a set of deterministic samples by perturbing the mean based on the covariance matrix and runs each sample through the function.
In this case, the function will be the nonlinear cost function defined \cref{subsec:cost_function_design}.


\begin{figure*}
    \centering
    \begin{subfloat}[Trajectories]{
    \centering
    \includegraphics[page=1, trim={0cm, 0cm, 0cm, 2.5cm}, clip, width=0.7\textwidth]{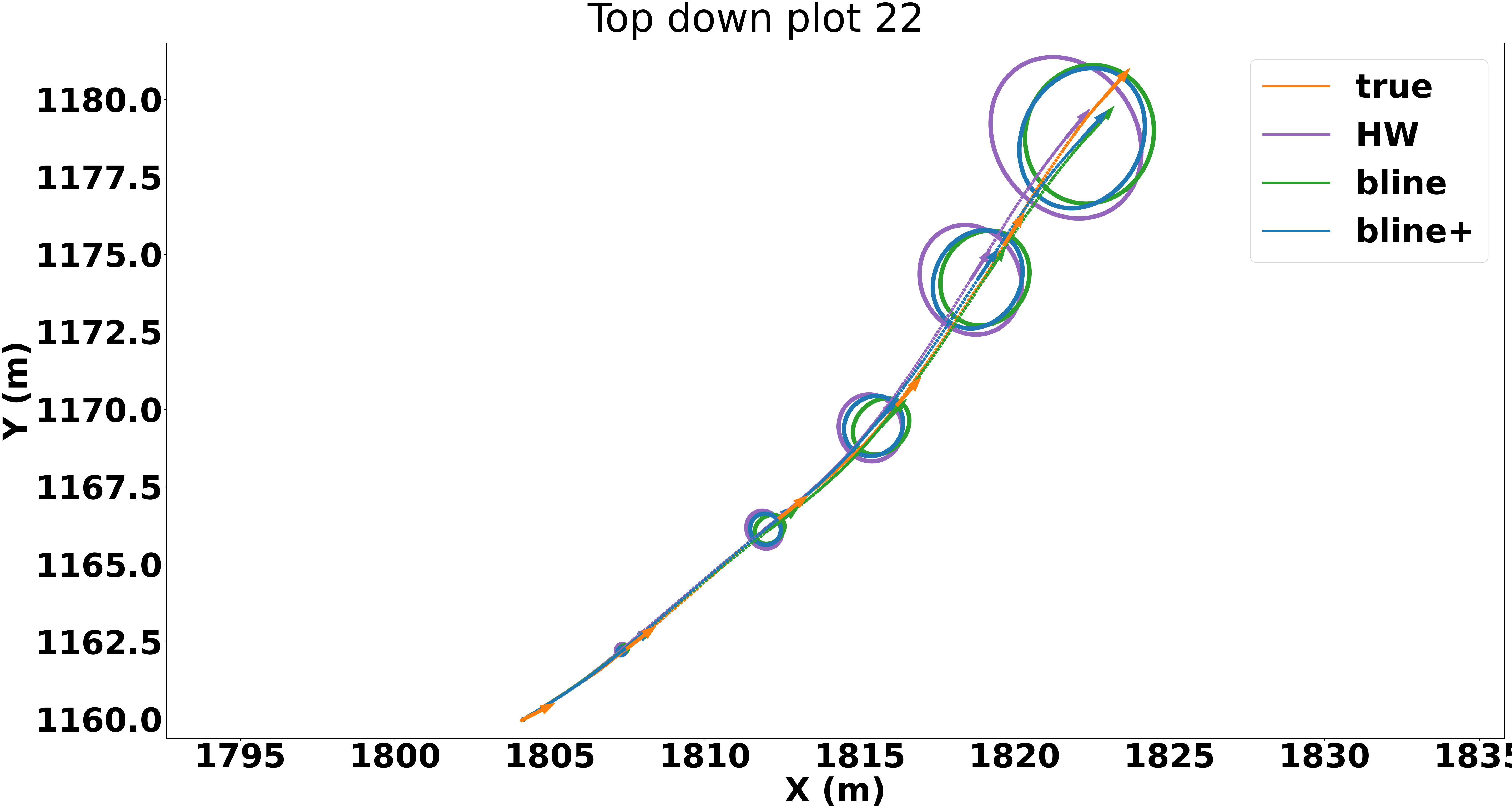}
    }
    \end{subfloat}
    \begin{subfloat}[HW Sigma Points]{
    \centering
    \includegraphics[page=2, trim={0cm, 0cm, 0cm, 2.5cm}, clip, width=0.3\textwidth]{Figures/trajectory_results/final_comparison_subset_combined_results_nice_figures_processed.pdf}
    }
    \end{subfloat}
    \begin{subfloat}[Baseline Sigma Points]{
    \centering
    \includegraphics[page=3, trim={0cm, 0cm, 0cm, 2.5cm}, clip, width=0.3\textwidth]{Figures/trajectory_results/final_comparison_subset_combined_results_nice_figures_processed.pdf}
    }
    \end{subfloat}
    \begin{subfloat}[Baseline+ Sigma Points]{
    \centering
    \includegraphics[page=4, trim={0cm, 0cm, 0cm, 2.5cm}, clip, width=0.3\textwidth]{Figures/trajectory_results/final_comparison_subset_combined_results_nice_figures_processed.pdf}
    }
    \end{subfloat}
    \caption{A comparison of open loop trajectories and sigma points from the three different models used in \cref{fig:final_comparisons} and explained in \cref{subsubsec:network_recommendations}.
    The top image shows the different models mean and uncertainty ellipse in their respective colors along with the ground truth in orange.
    The bottom row shows the graphical representation of the sigma points.
    }
    \label{fig:trajectories}
\end{figure*}

To compute the unscented transform, we perform a Cholesky decomposition using the Cholesky–Banachiewicz algorithm \cite{choleskyDecomp}. 
The added benefit of using the unscented transform is that the cost function calculations are parallelized on the GPU across both samples and time \cite{mppi-generic}.
It is important to note that the decomposition is only used in costing and does not impact the propagation of the distribution.
The decomposition $L$ and samples $\tilde{\vx}_{j, t}$ ($j = 1, ... n_{\tilde{x}}$) are given by,
\begin{align}
    \hat{P}_t &= L L^T\\
    \tilde{\vx}_{j, t} &= \tilde{\vx}_t + c_\sigma L_j\\
    \tilde{\vx}_{n_{\tilde{x}} + j, t} &= \tilde{\vx}_t - c_\sigma L_j
\end{align}
Where $L_j$ is a given row the lower-triangular matrix $L$ and $c_\sigma$ is a user-determined constant.
The mean $\tilde{\vx}$ is also included in the sampling to give $n_{s} = 2 n_{\tilde{x}} + 1$ sigma points.
We can inflate the distribution by changing the tuning parameter $c_\sigma$ to give higher safety margins.
In this work, we use a $c_\sigma$ of 2.0 nominally for increased safety unless otherwise noted.
If $\hat{P}$ is not positive definite, we replace all $L_j$ with a zero vector.
This rarely occurs and is typically in the first couple of steps, so the impact on the algorithm is minimal.
The experiments shown in \cref{subsec:hardware_results} demonstrate the impact of this value.
The order of the Cholesky decomposition can give different sparsity patterns of sigma points, and thus change the costing.
Our cost function cares most about velocity uncertainty and the impact of yaw on wheel queries in the map.
So we order our decomposition $p^x, p^y, \psi, v^x$ with the first two being arbitrary.
You can see the range of values sigma points take on in \cref{fig:trajectories} in the bottom graphs.

\subsection{Empirical Risk Measure Computation}

Given the unscented transform points, we can compute different risk measures on the empirical distribution coming from the unscented transform.
Each sampled point has its cost evaluated independently and in parallel.
The appropriate function to calculate the final cost is a choice left to the user and depends on the risk tolerance and desired behavior.
Our method allows for using VaR, CVaR, mean, max, min, etc.
In this work, we used CVaR \cite{Wang2021AdaptiveRiskSensitive} with $\alpha = 1$ or taking the maximum sigma point as the cost for each time point.
We approximate the CVaR using the following,
\begin{align}
    \text{CVaR}^\alpha_t(\cdot) &= \text{VaR}^\alpha_t(\cdot) + \frac{1}{n_s (1-\alpha)} \sum_{i=0}^{n_s} (l_{i, t} - \text{VaR}^\alpha_t(\cdot))^+
\end{align}
Where $l_{i, t}$ is the running cost of the $i$th sigma point for a given trajectory at a specific timestep $t$, $n_s = 2 n_{\tilde{n}_x} + 1$ is the number of sigma points, and $()^+$ denotes only using positive values.
\ac{VaR} can be thought of as the $\alpha$ quantile of the empirical distribution of costs, using linear interpolation as needed, and \ac{CVaR} as the mean of the costs above that point.
Note that we compute the \ac{CVaR} for each timestep independently.

\subsection{Cost Function Design}
\label{subsec:cost_function_design}

The cost function used to drive the vehicle in challenging off-road scenarios is shown below \cite{Vlahov2023}.
\begin{align}
    l(\hat{\vx}, \hat{P}) &= \textit{Force}(\hat{\vx}) + \textit{Accel}(\hat{\vx})\\
    &+ \hat{d}\Bigl(\textbf{Body}(\hat{\vx}, \hat{P}) + \textbf{Speed}(\hat{\vx}, \hat{P})\nonumber\\
    &+ \textbf{Wheel}(\hat{\vx}, \hat{P}) + \textbf{Rollover}(\hat{\vx}, \hat{P})\nonumber\\
    &+ \textbf{Slip}(\hat{\vx}, \hat{P})\Bigr)\nonumber\\
    \Phi(\hat{\vx}_{t:t+T}) &= \textit{CostToGo}(\hat{\vx}_{t:t+T})\\
    \hat{d} &= \left(\sqrt{\hat{v}_{x,t}^2 + \hat{v}_{y, t}^2}\right) \Delta t \label{eq:distance_scaling}
\end{align}
Where the elements in bold are computed using the perturbed outputs from the unscented transform on $[x, y, \psi, v^x]$ and $\hat{d}$ is the estimated distance traveled in a single time step using mean values for body velocities.
All costs are scaled to have zero cost below a minimum value, a high penalty once the value goes above a maximum threshold, and a linear or quadratic scaling between the min and max values.
Quadratic scaling occurs for slip, roll, pitch, forces and speed while the rest are scaled linearly.
This helps keep the magnitude of the costs similar and the penalty can be seen as enforcing constraints.
In traditional discrete-time costing approaches, it is only the number of time points spent in a state that matters, so going quickly through a high cost region is preferred to going slowly.
We want to encourage slower driving when the cost is high, and using distance traveled ($\hat{d}$) removes this bias.
\textit{CostToGo} is computed as an optimal cost to the waypoint based on a lattice planner using Dijkstras.
It depends on the entire horizon since it is finding the optimal time point in the trajectory to connect back the lattice. 
We will break the cost function into four major components, grid-cell-based traversability costs, physics-based constraints, soft penalties to encourage good behavior, and speed limits.

\subsubsection{Traversability Costs}
The first major component of the distributional cost is the grid-cell-based traversability costs.
They are split into wheel and body costs since some hazards are unique to each.
Wheel maps are queried at the front of the wheel, and the body is queried at the geometric center. 
All queries are computed using the full 6 DOF estimate of the vehicle coming from the suspension system.
The maps are split into soft costs or risk and lethal for states that have a high probability of causing a crash that should be avoided.
All penalties are also scaled by the absolute value of velocity like in \cref{eq:distance_scaling} but without the $\Delta t$, so that hitting an obstacle at lower speeds is preferred.
The impact of yaw uncertainty should not be overlooked; a small yaw error could greatly impact the location of the wheels with a $3\si{m}$ body length. 
This fact makes Closed loop uncertainty critical for navigating tight spaces with wheel hazards on either side.

\subsubsection{Physics-based Constraints}
Second is the approximated physics-based constraints that operate as heuristics for correct behavior.
The rollover cost is an attempt to estimate the minimum normal force on the left or right side of the vehicle to estimate a rollover.
A penalty function is applied when this drops below a specific threshold, but is treated as a soft penalty when above that.
The rollover is calculated as
\begin{align}
    a &= \frac{(v^x)^2 \tan(\tilde{\delta})}{c_l}\\
    s &= \frac{2c_h}{c_w} \left(\frac{a}{c_g} + \cos(\theta) \sin(\phi)\right)\\
    \textbf{Rollover}(\vx) &= \min\left(\cos(\theta) \cos(\phi) + s,\right.\\
    &\qquad\quad\ \left.\cos(\theta) \cos(\phi) - s \right)\nonumber
\end{align}
where $c_h$ is the height to the center of gravity, $c_w, c_l$ are the width and length of the vehicle respectively, and $c_g = 9.81$ is gravity. 
Gravity is used to normalize the force between $[0,1]$ g's.

The other component of physics-based model costs are the \textit{Force} costs.
These are functions of the suspension forces $F_{w,u}$ from \cref{eq:suspension_spring} and are not distributional since the suspension system is only simulated for the mean.
We convert these forces in to the wheel frame using
\begin{align}
    F_{w, f} &= \frac{F_{w, u}}{\eta_z} \left(\eta_x \cos(\tilde{\delta}) + \eta_y \sin(\tilde{\delta}) - \eta_z \theta\right)\\
    F_{w, s} &= \frac{F_{w, u}}{\eta_z} \left(-\eta_x \sin(\tilde{\delta}) + \eta_y \cos(\tilde{\delta}) + \eta_z \phi\right)
\end{align}
We give different penalties for each $F_{w, u}, F_{w, f}, F_{w, s}$ since forces allowed on each are different.
The forces act in wheel frame and represent the forces on the top, front and side, respectively.
They ensure correct tire orientation and limit the forces on the tires when traversing ditches to prevent damage.

\subsubsection{Safety Speed Limits}
While uncertainty has allowed us to remove most speed limits, a few remained for safety purposes.
We implement our speed as quadratic penalty on the deviation from the max speed if it is above it.
Max speeds can be set through a 2D map, where we take the minimum over each wheel, or through the estimated roll and pitch.
Maximum speeds from traversability are used to set a lower speed through a lethal obstacle that would show up in the map as well. 
There are speed limits as a function of roll and pitch of the vehicle for safety since it can be overly aggressive on vegetative slopes that can alter the estimated suspension system due to perceptual errors.
These are set to a linear function of roll from $0.15 \rightarrow 0.3 rad$ it decreases from $6 \rightarrow 2 m/s$.
Thresholds for pitch are separated based on uphill or downhill, uphill has the linear function from $0.3 \rightarrow 0.8 rad$ with speeds $8 \rightarrow 4 m/s$ and downhill has $0.2 \rightarrow 0.4 rad$ with speeds $4 \rightarrow 2 m/s$.

\subsubsection{Soft Costs}
Third, are costs on bad behavior where $\textit{Accel}$ and $\textbf{SlipCost}$ fit in.
The acceleration penalty is a soft cost on $\dot{v}^x$ when it crosses a specific threshold to encourage smooth driving behavior.
There are separate tunings for accelerating and decelerating to encourage braking over throttle.
Both costs help with smoothing out the motion of the vehicle, but the accelerating cost has an additional benefit in limiting aggressive throttle at zero velocity.
A high throttle at rest can make the vehicle dig the wheels into the dirt and get stuck in loose dirt.
The slip cost is important for removing issues with the dynamics fitting process that can sometimes generate trajectories that tend to drift laterally at low speeds.
This can create poor behavior when the trajectory is blocked in front by an obstacle but a small lateral drift can decrease cost by connecting to the cost to go lattice.
These situations are characterized by small $v^x$ and large $v^y$ so the slip penalty from \cite{williams2017} was taken to make the dynamics harder to exploit in edges cases,
\begin{align}
    \textbf{Slip}(\vx) &= \min\left(\left|\arctan\left(\frac{v^y}{|v^x| + \epsilon}\right)\right|, |v^y| \right)
\end{align}
where $\epsilon = 1.0e^{-3}$ was added for numerical stability.
Using the min with $|v^y|$ gives better performance at very low $v^x$, $v^y$ values, which is especially important with the unscented transform.
We also penalize negative speeds five times more than positive ones to encourage forward motion for all speed limits.

\section{Results}
\label{sec:results}


\begin{figure*}[htbp]
\centering
\begin{subfloat}[Structure Changes 1\label{fig:structure_bad}]{
\centering
\includegraphics[page=1, trim={0cm, 0cm, 0cm, 0cm}, clip, width=0.48\textwidth]{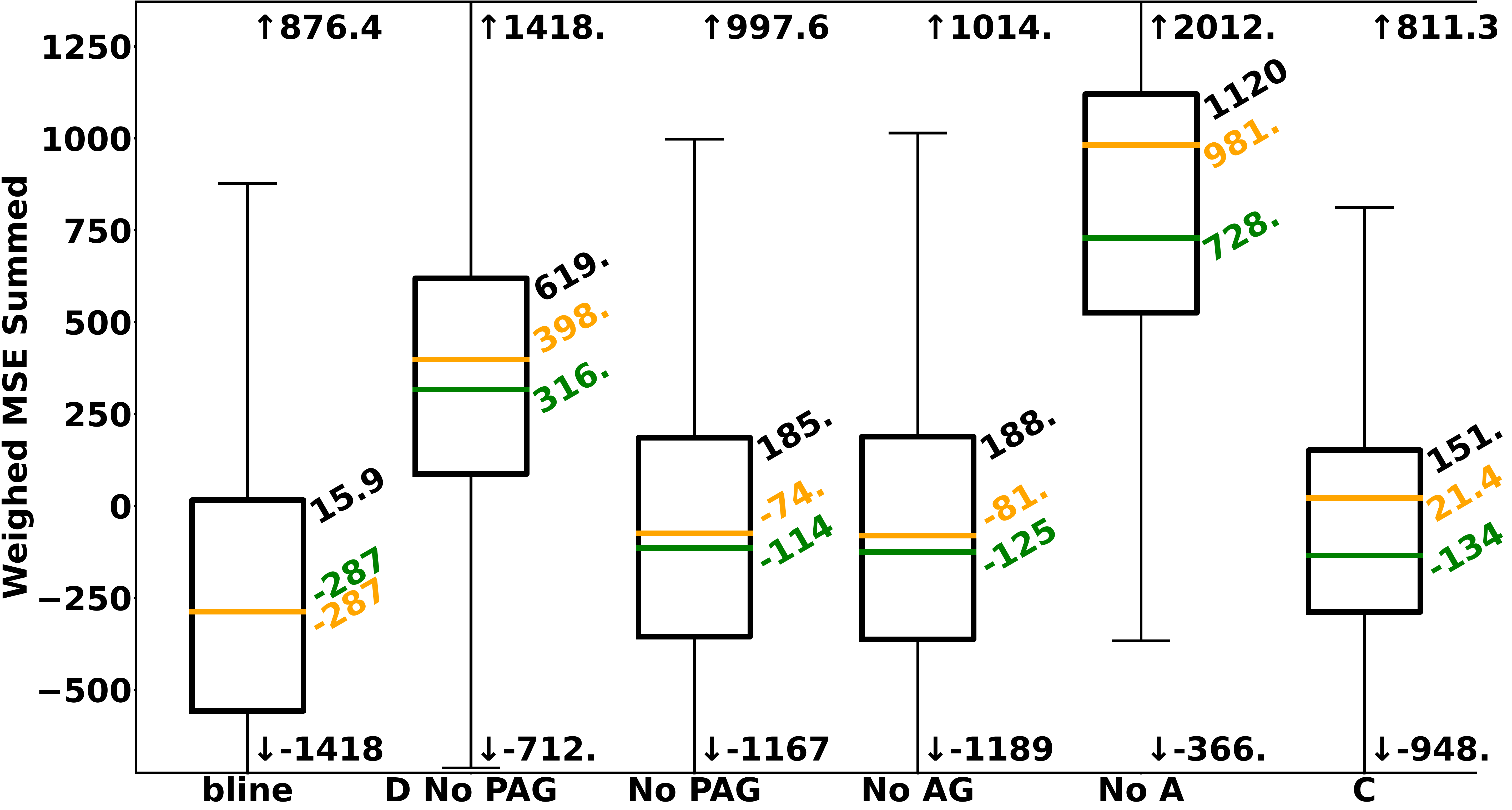}  
}
\end{subfloat}
\begin{subfloat}[Structure Changes 2\label{fig:structure_good}]{
\centering
\includegraphics[page=1, trim={0cm, 0cm, 0cm, 0cm}, clip, width=0.48\textwidth]{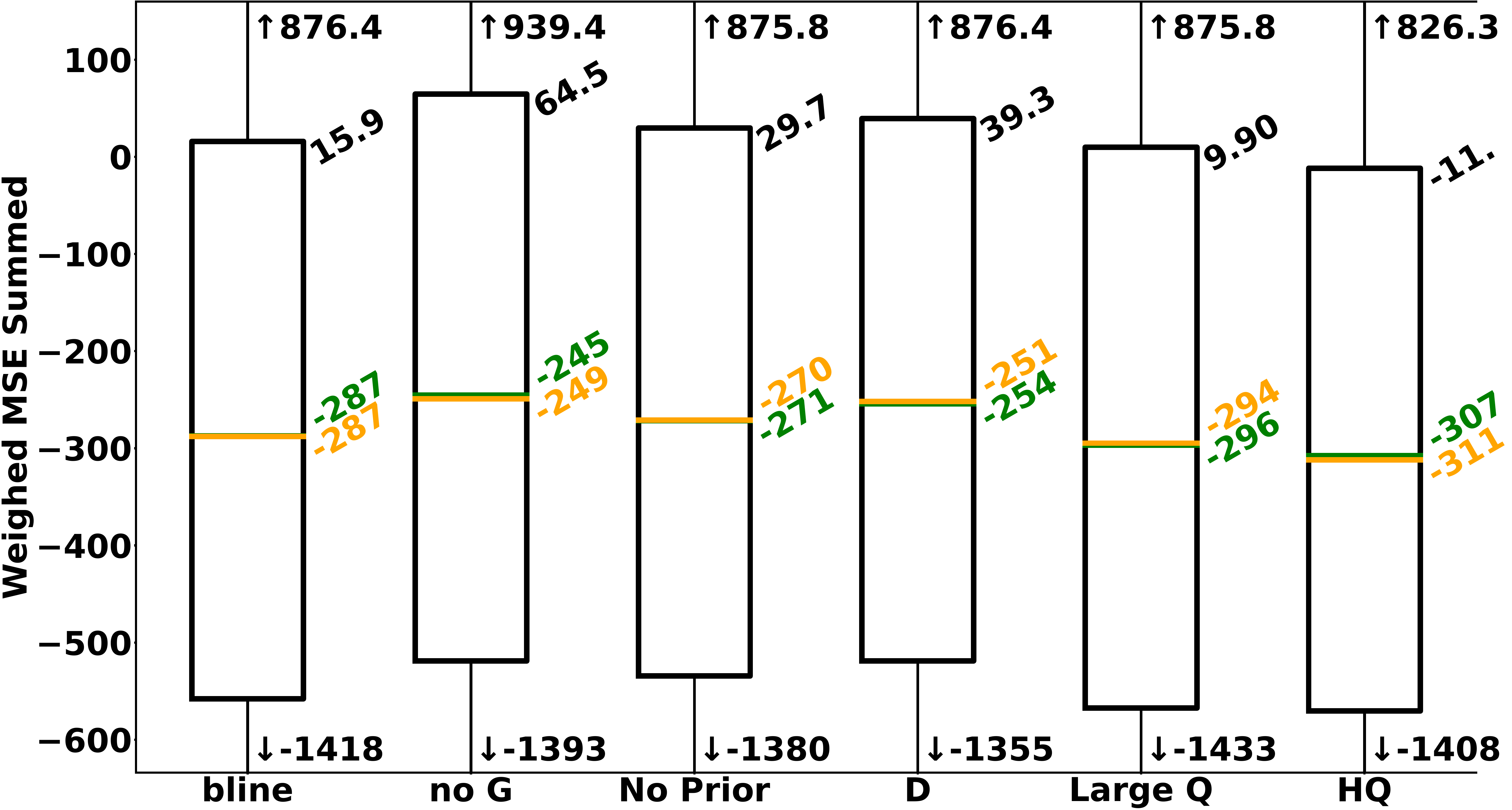}  
}
\end{subfloat}
\begin{subfloat}[Architecture \label{fig:architecture}]{
\centering
\includegraphics[page=1, trim={0cm, 0cm, 0cm, 0cm}, clip, width=0.48\textwidth]{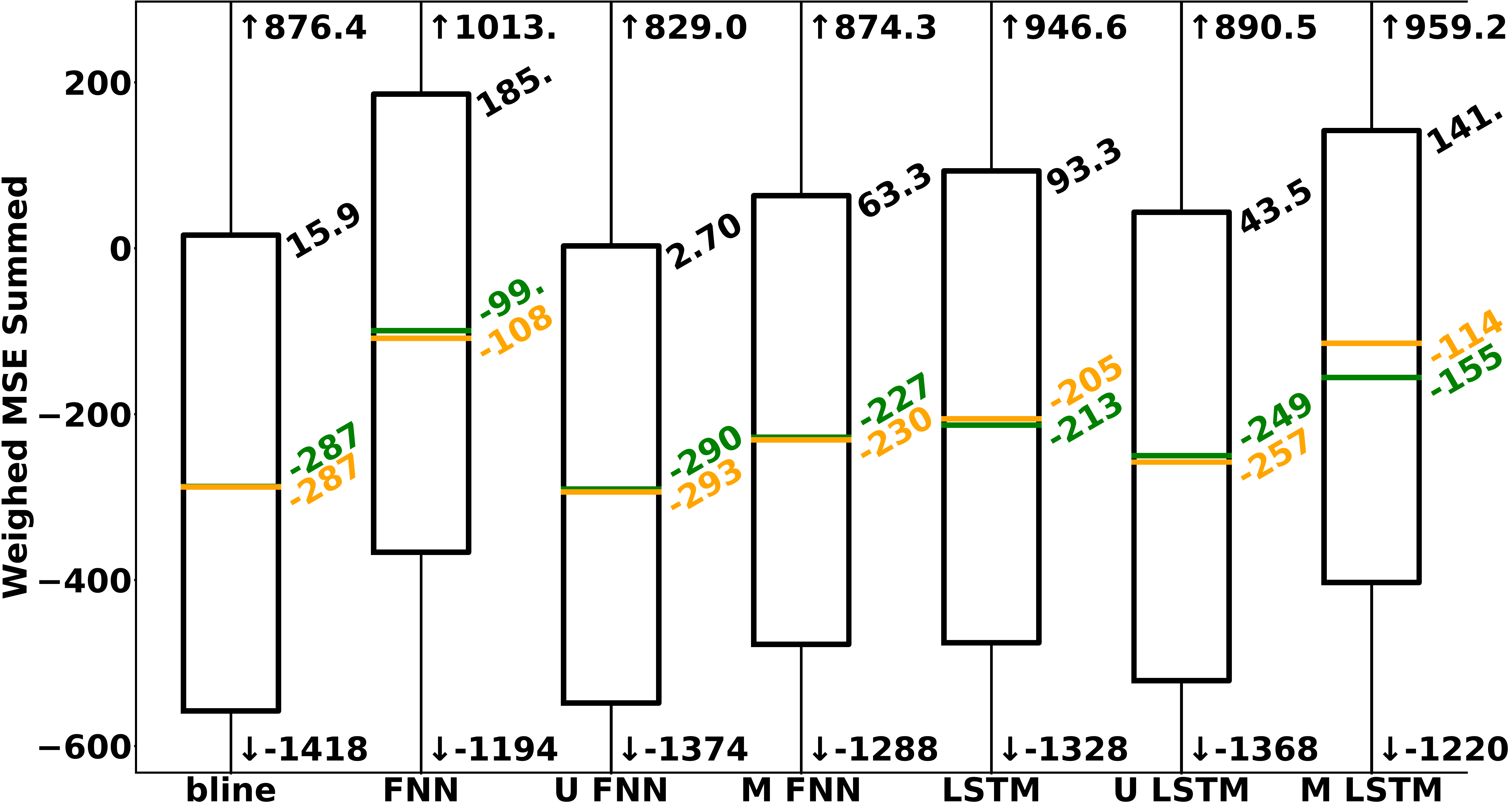}
}
\end{subfloat}
\begin{subfloat}[Initialization \label{fig:init_ablation}]{
\centering
\includegraphics[page=1, trim={0cm, 0cm, 0cm, 0cm}, clip, width=0.48\textwidth]{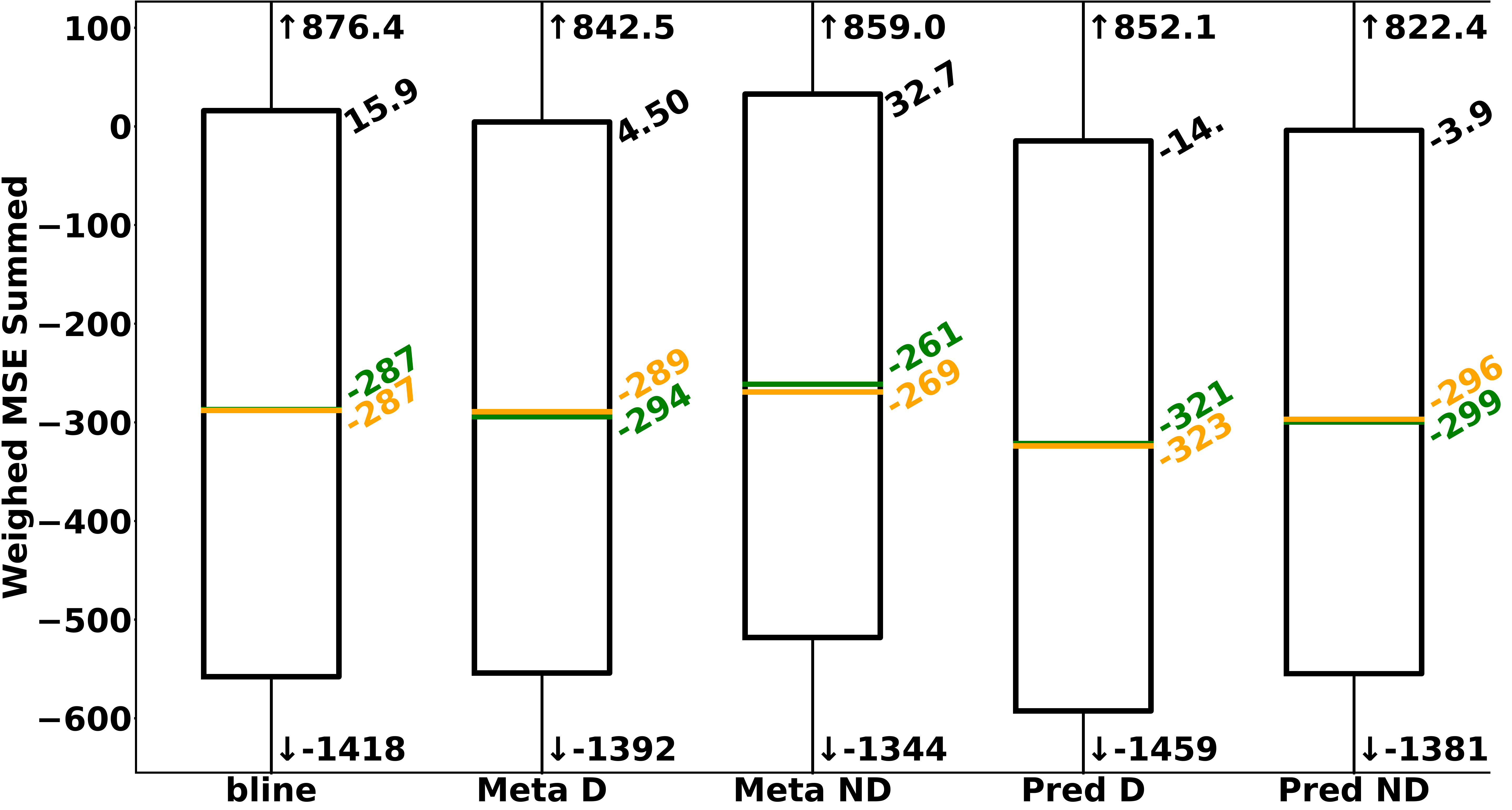}   
}
\end{subfloat}
\begin{subfloat}[Loss Structure \label{fig:structure_loss}]{
\centering
\includegraphics[page=1, trim={0cm, 0cm, 0cm, 0cm}, clip, width=0.48\textwidth]{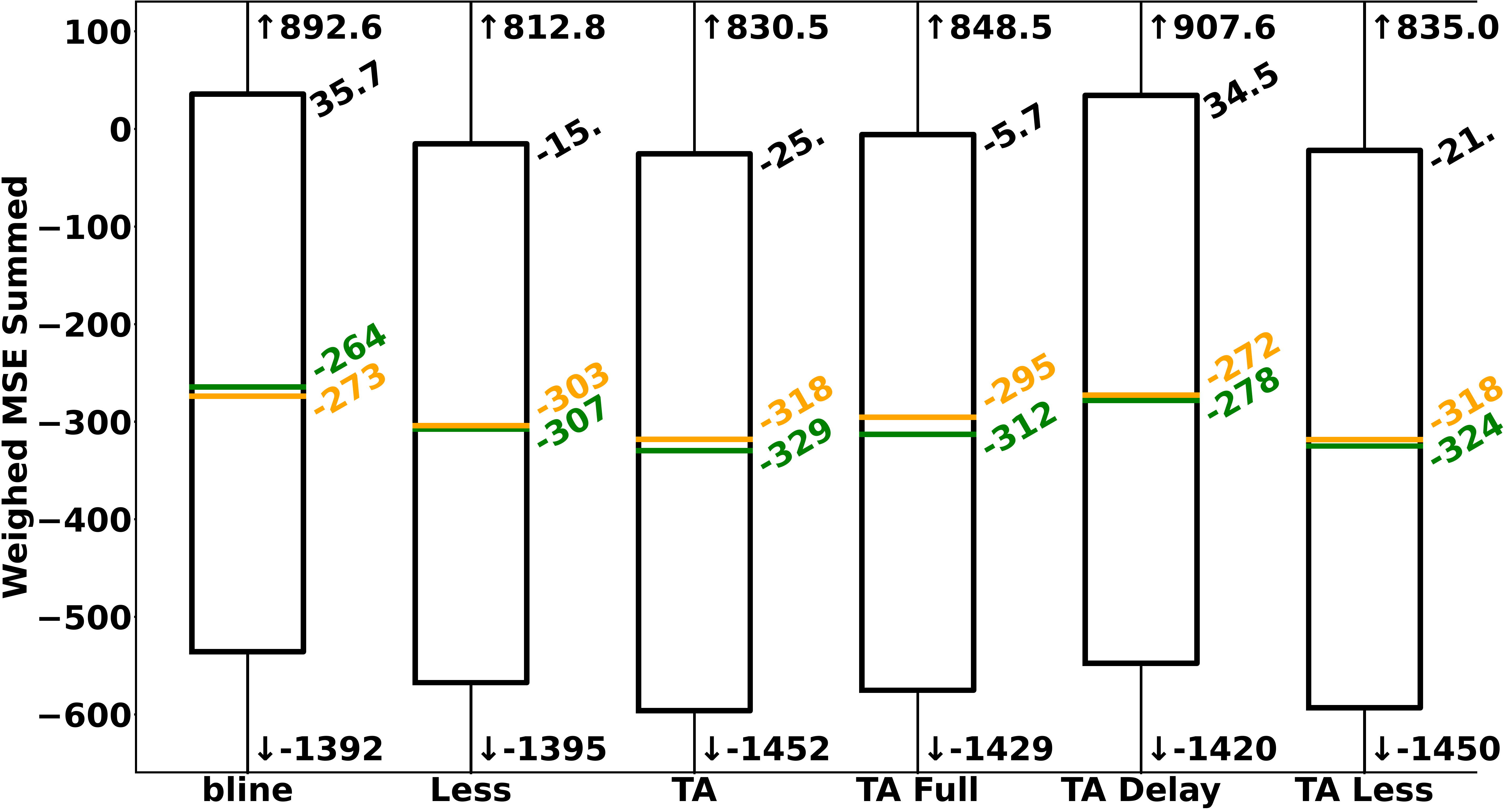}   
}
\end{subfloat}
\begin{subfloat}[Buffer History ($\tau$) \label{fig:tau_ablation}]{
\centering
\includegraphics[page=1, trim={0cm, 0cm, 0cm, 0cm}, clip, width=0.48\textwidth]{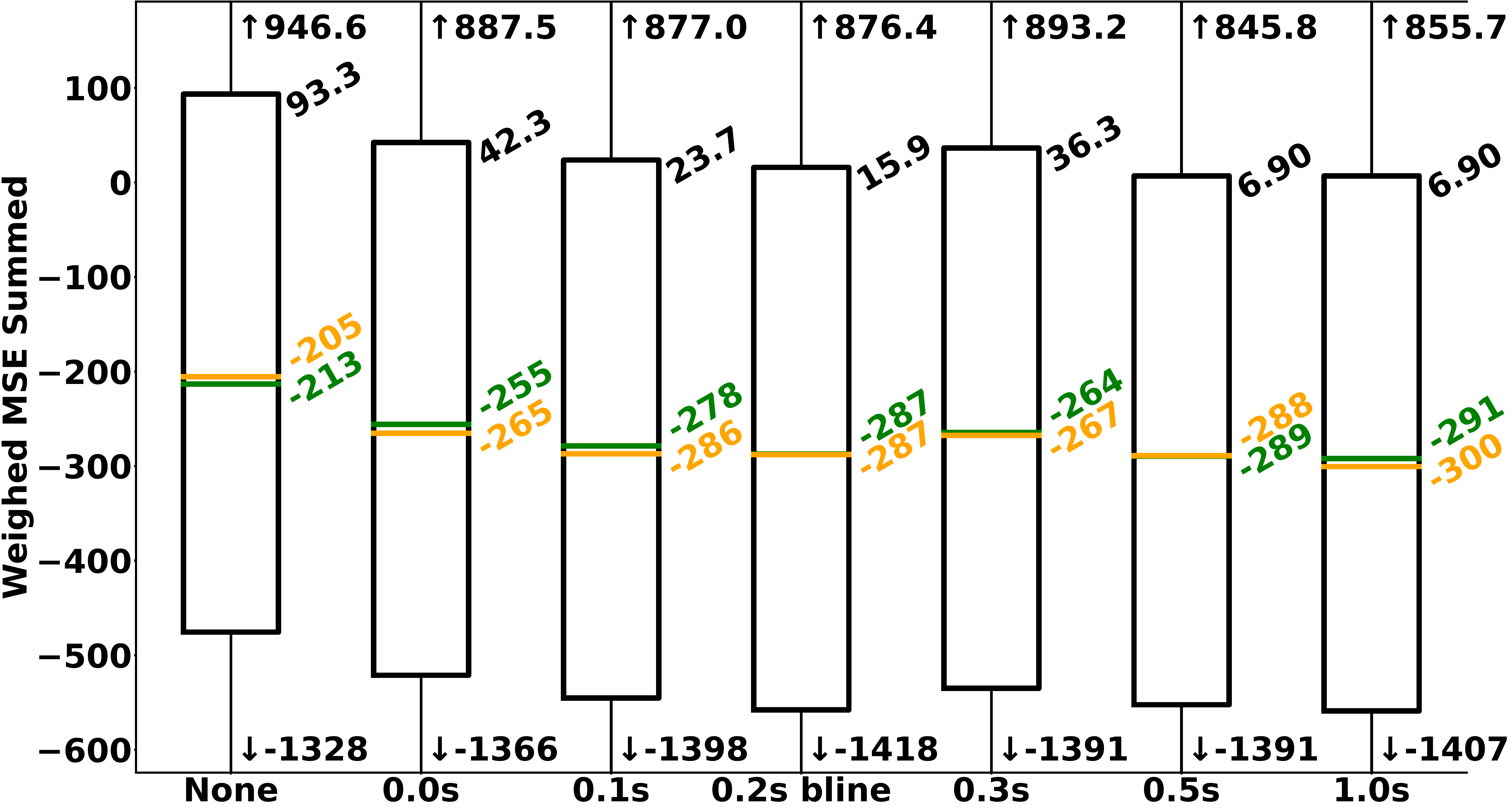} 
}
\end{subfloat}
\caption{
Box plots showing total summed NLL loss for various structural changes to the networks. 
Whiskers are defined as $\pm 1.5 IQR$ and is given with the arrows, the green line defines the median and the orange the mean value. 
The baseline model (bline) is kept consistent.
The explanations of the keys used in shown in the text \cref{subsubsec:structure_ablation}.
}
\label{fig:structure_ablation_results}
\end{figure*}


The results section is broken into two sections, a comparison of methods on a dataset in \cref{subsec:ablation_studies} and hardware results of a similar model running in real time in \cref{subsec:hardware_results}.
Both are critical to learning dynamics for control, as good dataset metrics can hide possible poor performance of the model in closed-loop testing.
Things such as the stiffness of the equations and others mentioned \cref{sec:belief_dynamics_learning} were found and corrected during closed-loop testing but are not easily noticeable in the dataset metrics.
This is also the reason we focus on higher-order statistics than just mean, giving a sense of the loss function distribution using box plots.
For all tests, we compare the summation of the \ac{NLL} loss function over the entire horizon $T$.
We choose this since it gives an estimate of how well on average the Gaussian distribution matches the data while being sensitive to large covariances. 
In practice, we tend to focus on the $75\%$ quantile of error and the deviation between mean and median to determine which is a better model for control.
Ignoring strong outliers is important for cases where velocity is poorly predicted; this can cause a large distance error with a small covariance.
Unless otherwise mentioned, we will use a time horizon $T = 5s$ a history window for intitialziation of $\tau = 0.2s$ and a time increment of $\Delta t = 0.02s$.

Our vehicle is the same as \cite{Gibson2022,gibson2024dynamicsmodelingusingvisual,Vlahov2023,Han2023ModelPredictiveControl,han2024dynamicsmodelsaggressiveoffroad} and has significant on-board compute capabilities (AMD Threadriper 3990X, 8x 32GB DDR4 RAM, and 4x NVIDIA RTX 3080). 
A single GPU is used for the \ac{MPPI} computation.
The \ac{MPPI} algorithm is implemented with $\approx 18K$ samples and $250$ steps per sample, meaning we have to compute the dynamics $\approx 65,000,000$ times a second. 
Odometry is generated using a combination of \cite{Fakoorian2022} for LIO ($10\si{Hz}$) and \cite{nubert2022graph} for low-latency ($20\si{\ms}$ delay, $400\si{Hz}$) odometry.
We perform linear interpolation of odometry messages to provide ground truth labels for training.
Elevation values are generated using a geometric and semantic scheme outlined in \cite{atha2024fewshotsemanticlearningrobust} with GVOM \cite{GVOM} as the voxel mapping building block.
The learning was carried out in Pytorch \cite{PyTorch} using the Adam optimizer \cite{adam}.

The vehicle dataset contains four primary environments.
Most of the data $60\%$ comes from driving in the Mojave Desert near Helendale CA, $30\%$ comes from Halter Ranch near Paso Robles, $10\%$ comes from coastal sage near Oceanside CA and $5\%$ comes from coastal dunes also nearby Oceanside CA.
All environments contain unique challenges and features that can impact performance.
Our learning approach currently does not take into account the environmental context, but future work will extend to this; see \cite{gibson2024dynamicsmodelingusingvisual} to see how that could be accomplished.
It should be noted that our models likely learn an average uncertainty from the entire dataset, meaning our uncertainties are larger than training a model on data only from a specific terrain.

Studies are done in contrast to a consistent baseline model.
Our reduced state space is $\tilde{\vx} = (p^x, p^y, \psi, v^x)$ with $\pi = [b, \delta, v^y, F_x]$.
The models take in state subset $(v^x, v^y, \dot{\psi}, b, e, \delta, \dot{\delta})$, sensor readings $\vy = (u^t, u^b, u^\delta, \eta_x, \eta_y, \eta_z)$, and forces $\hat{\vF}$ as shown in \cref{eq:general_hybrid_mean_update,eq:specific_unc_update} and predicts in the force space \cref{eq:general_hybrid_mean_update}.
The baseline model is identical in structure to the one that was used on hardware but with additional data and improved delay models that were created after the conclusion of the hardware experiments.
The baseline structure uses a predictor network with $4$ hidden states and an output network with an intermediate layer of size $20$ for both mean and uncertainty.
The initialization network has a hidden size of $20$ and an additional output layer of $100$.
The notable difference from the presentation in \cref{eq:specific_unc_update} is that the baseline predicts only $\hat{\vq}$ for process noise.
Additionally, the engine RPM values are scaled by $1.0\mathrm{e}{-3}$ before being put into the network, and delay models are fixed during training unless otherwise noted.
The loss function has a regularization loss constant of $\vc_\pi = 0.2$ on $v^y, \dot{\psi}$ and is the loss that will be used in all comparisons for consistency.

\subsection{Ablation Studies}
\label{subsec:ablation_studies}

We break the ablation studies into 3 major components.
The first, network structure, explores the importance of each component of the structure of the problem.
The second, network size, looks at how the loss changes as a result of changing the size of different layers in the network.
The third summarizes the results of both and walks through the key takeaways in the form of recommendations that are validated by training models.
Overall, most changes result in very similar performance that is limited to the calibration of the uncertainty.
Often models will have different \ac{NLL} but similar errors in state error. 
All models were trained for 50 epochs with the same dataset and the lowest mean loss on the validation set across all epochs was chosen to provide the best comparisons.

\subsubsection{Network Structure}
\label{subsubsec:structure_ablation}

This section covers the set of all experiments represented in \cref{fig:structure_ablation_results} where we compare variations on the structure of the dynamics and how they were trained.
Looking at \cref{fig:structure_bad,fig:structure_good} we see variations on the structure or inputs of the model.
The abbreviations are \textbf{D} stands for direct compensation (mean network does not predict in force space), \textbf{No P} indicates a model without any parametric component, \textbf{No A} indicates a model without the $A_t$ matrix in \cref{eq:general_unc_update}, \textbf{No G} indicates that we only predict $\hat{Q}'$ in \cref{eq:specific_unc_update} and do not use the hybrid structure of $G_t$ and $\vq$ outlined in \cref{eq:specific_unc_update}, \textbf{No Prior} means that the predicted forces $\hat{\vF}$ are not used as an input to the networks (ex. $\zeta_\mu(\vx_t, \vu_t, \vy_t)$), \textbf{C} indicates a combined network architecture -- a single model for mean and uncertainty -- with doubled model size to compensate, \textbf{Large Q} is using all states and forces in $G$, \textbf{HQ} represents using a hybrid $Q$ setup as outlined in \cref{eq:specific_unc_update}.

The first important result is predicting in the force space gives improved results even with a purely learned approach, \textit{D No PAG} vs \textit{No PAG} in \cref{fig:structure_bad} and when comparing the \textit{baseline} to D in \cref{fig:structure_good}.
Predicting in the right space for the mean update is critical for accuracy.
The second point to be made is how important the $A_t$ and $G_t$ matrices are for prediction.
\textit{No AG} vs the \textit{baseline} highlights how much changing the structure of the problem can aid in performance.
We can see the importance of matching structure in the propagation and learned process noise is important when comparing \textit{No A} vs \textit{No AG}. 
Removing the parametric $A_t$ matrix while keeping the forced structure of $G_t$ (\textit{No A}) gives very poor results.
In essence, this means that using only $\hat{\vq}$ enforces a specific structure on the problem that cannot be learned around due to the structure of $G_t$ but can be learned by a network when that structure is removed.
This contrasts with the performance of No G in \cref{fig:structure_good} where we predict only a $\hat{Q}'$.
Together, these facts mean that the structure of $G_t$ provides less useful structure for the problem than that of $A_t$.
The small gap between \textit{No G} and \textit{baseline} motivated the use of the \textit{HQ} structure, which was the best performing change in the structure of the problem.
A hybrid structure seems to always provide better performance than a parametric or purely learned version.
Adding additional values into $\pi$ with \textit{Large Q} in \cref{fig:structure_good} only slightly outperforms the \textit{baseline} but with a more complicated $G_t$ matrix.
Some of the coefficients of these states are fit to 0 with the scaling so we can empirically see that some can be dropped.
The most surpising drop in performance is combining the mean and uncertainty networks into a single \ac{LSTM} shown as C in \cref{fig:structure_bad}.
This occurred even with doubling all parameters of the network size.

Next, we looked at how important using an \ac{LSTM} or initialization of the hidden and cell state was for performance.
The keys in \cref{fig:architecture} denote what part of the model is being replaced with what network structure, \textbf{M FNN} indicates the mean network $\zeta_\mu$ is replaced with a \ac{FNN} and \textbf{U LSTM} indicates the uncertainty network $\zeta_Q$ is replaced with a \ac{LSTM} without initialization, \textbf{All} indicates that both are replaced by the same type of network.
For the \ac{FNN}, we used models with more weights than the \acp{LSTM} to ensure the state augmentation was the important component.
The networks used had widths $128$, $64$ in intermediate layers for \ac{FNN}.
When looking at replacing the \ac{I-LSTM} with either an \ac{LSTM} or a \ac{FNN}, we noticed that the impact was large with respect to the mean compensation network but was minimal to outperforming the \textit{baseline} when looking at the process noise network.
We expect this is because the process noise being computed is less sensitive to the historical values since the covariance matrix captures them implicitly. 
The additional complexity of the uncertainty \ac{LSTM} might degrade performance when looking at \textit{M LSTM} vs \textit{LSTM} (where both are not initialized \acp{LSTM}).
This logically makes sense since the better the prediction of the mean the harder it is to have large outliers in loss.
Typically, these outliers occur when the motion is poorly predicted and the uncertainty remains small; we can see this in the \textit{M LSTM} with the large separation between mean and median values demonstrating skew.

When comparing different methods of initializing the covariance matrix in \cref{fig:init_ablation}, we noted that predicting a diagonal covariance matrix outperforms predicting a full matrix.
\textbf{ND} means we predict a full covariance matrix, \textbf{D} means predicting a diagonal matrix and \textbf{Meta} vs \textbf{Pred} refers to predicting a single meta value for all trajectories or predicting the initial uncertainty using the initialization network.
The added benefits of the off-diagonal element is minimal in this application and could be difficult to fit using our parameterization which explains the poor performance of the full covariance vs the diagonal.
Predicting the values as a function of historical values with \textit{Pred} gave the best results vs using a \textit{Meta} learned values.
In practice these values remain small, but the \textit{Pred} version predicts uncertainty for $v^x$ and $\psi$ that is context aware.
For example, it would predict values with larger or smaller yaw uncertainty based on the sharpness of the turn at initialization.
The meta-learned values tend to be larger consistently, likely to reduce errors when the vehicle is motionless and odometry drifts slightly.

Loss structure in \cref{fig:structure_loss} shows the impact of changing the loss function where, \textbf{Less} means that regularization loss on $\pi$ is removed, \textbf{Full} indicates that either regularization or distributional loss is applied on all states, 
and \textbf{TA} or train all means that delay models are allowed to change during training of the terradynamics models.
Looking at the loss function in \cref{fig:structure_loss}, we see that keeping the delay models fixed, as in the \textit{baseline}, performs worse than letting them continue to fit as in the TA model.
Note that the reported statistics for all comparisons have the regularization loss for consistency.
We emphasize that there is no penalty on the prediction of the values for the delay models when they are allowed to change in all \textit{TA}.
All models achieve about the same accuracy in the prediction of position and velocity so the difference in loss must result from different uncertainty predictions.


Allowing the delay models to change when training the can either increase or decrease their error on the states they are trained to predict based on the weighting.
When we compare the accuracy of the delay states, \textit{TA} has higher error than \textit{TA Full} but also has tighter uncertainty balls, so a lower overall loss.
\textit{TA Full} has the largest improvement in delay states.
For \textit{TA} the delay states are learned to correctly predict the terradynamic outputs without regard to their internal accuracy, so their error can either decrease or increase.
Allowing the delay model weights to change without an explicit penalty of their accuracy can complicate the ability to check these models for accuracy.
Due to the structure of the learning problem, the accuracy of the delay models can increase without an explicit penalty.
For \textit{TA} we see the steering error for $\delta, \dot{\delta}$ at the $75\%$ quantile reduces down about $25\%$ or more; brake errors remain consistent, and engine values double their error.
This indicates that the structure of the problem relies on specific values of $\delta$, but is not sensitive to the exact values of the engine.
Without penalty on the engine state, it becomes a biased additional state that can be used to predict the terradynamics rather than match the underlying engine system correctly, so the increase in error is not relevant under our comparison.
Allowing the delay models to change when training the terradynamics is the most impactful change found during the ablation studies, but in practice we will likely keep a fixed penalty on delay values like with TA Full.
This is because we lose the ability to understand how well the engine subsystem is predicted and complicates verifying the models accuracy.

Finally, we alter the buffer history used for the initialization network in \cref{fig:tau_ablation}.
An important takeaway is the increase in performance from no initialization (\textit{None}), which is the same as the \ac{LSTM} in \cref{fig:architecture}, compared to using $0.0\si{s}$ initialization.
This means initializing the hidden and cell state of the predictor network based on just the initial state and inputs provides most of the benefit.
Increasing the history buffer can decrease the error but has a diminishing return as the horizon increases. 
The impact of this is most focused around mean prediction, and not uncertainty prediction as we saw in the results from the architecture ablation \cref{fig:architecture} above.
Distance errors decrease as $\tau$ increases, \textit{None} vs $\tau=1.0$ have a $20\%$ decrease in error at the end of the trajectory. 
Most of the improvement comes from the body $x$ error as the body $y$ distance errors are very similar.
These results match our intuition that mean prediction does not depend on a long history.

\subsubsection{Network Size}
\label{subsubsec:network_size}

\begin{figure*}[h]
\centering
\begin{subfloat}[Predictor Hidden Size \label{fig:pred_hidden_ablation}]{
\centering
\includegraphics[page=1, trim={0cm, 0cm, 0cm, 0cm}, clip, width=0.48\textwidth]{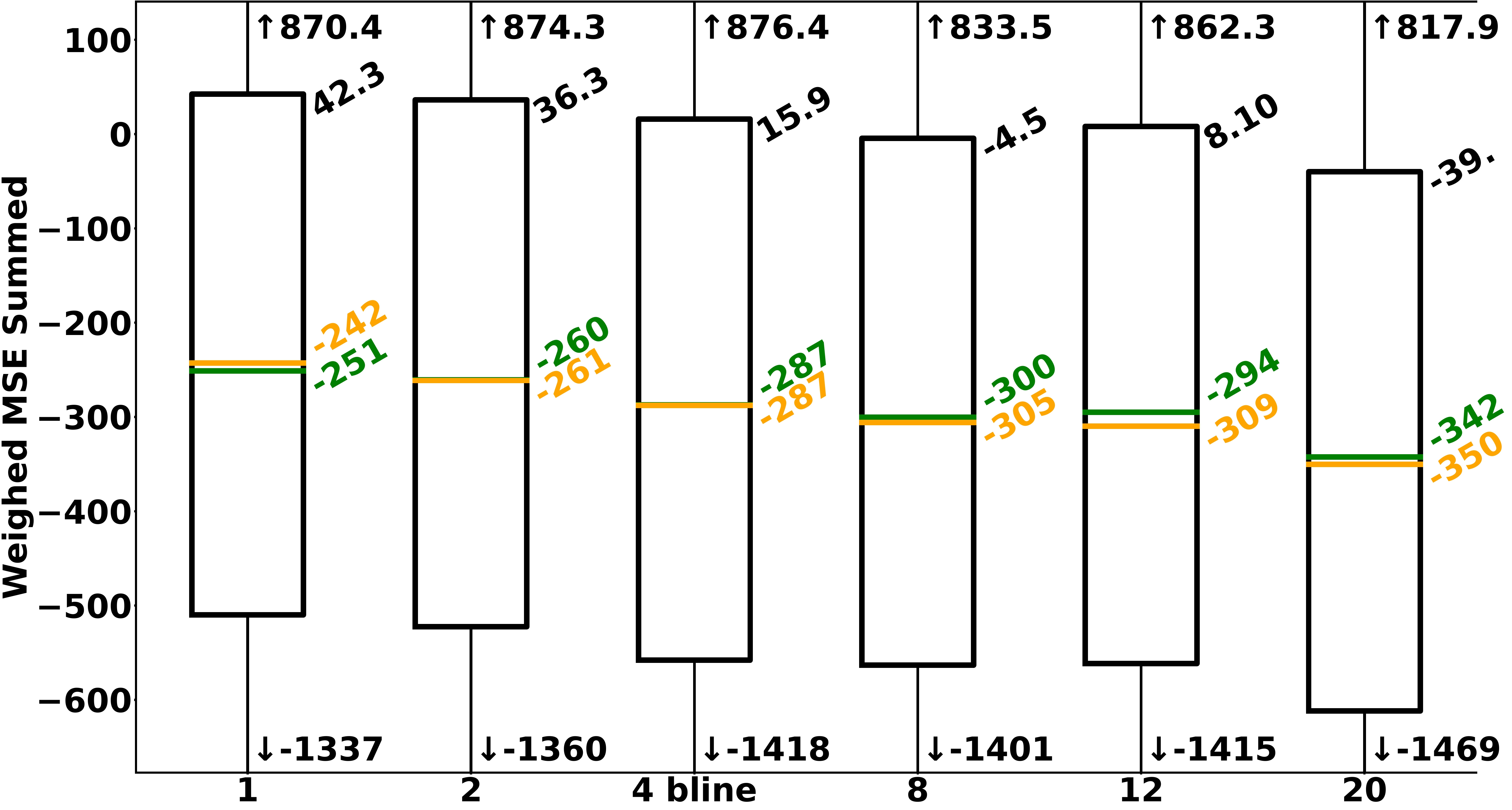}   
}
\end{subfloat}
\begin{subfloat}[Predictor Output Width \label{fig:pred_output_width}]{
\centering
\includegraphics[page=1, trim={0cm, 0cm, 0cm, 0cm}, clip, width=0.48\textwidth]{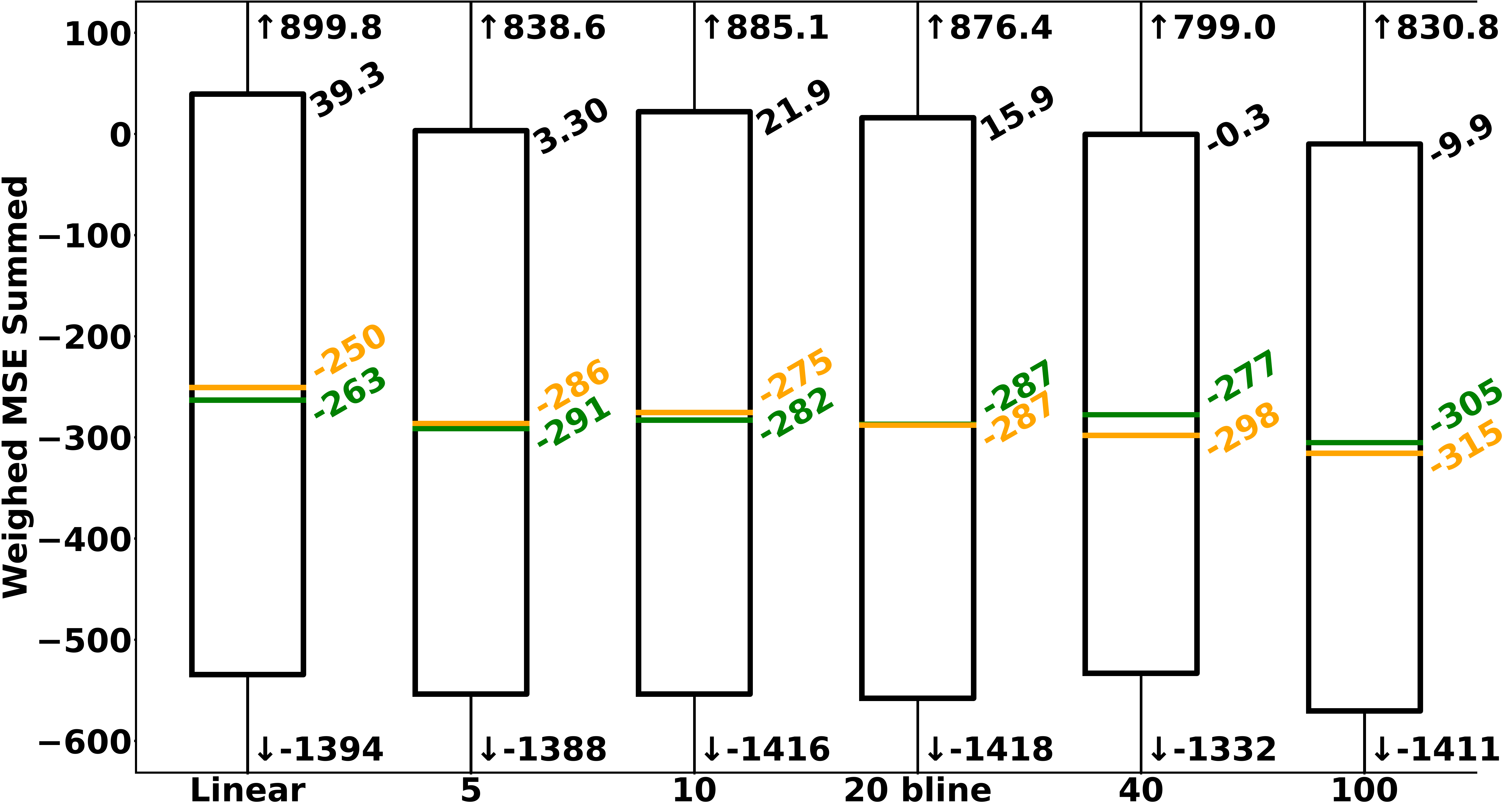}   
}
\end{subfloat}
\begin{subfloat}[Predictor Output Depth \label{fig:pred_output_depth}]{
\centering
\includegraphics[page=1, trim={0cm, 0cm, 0cm, 0cm}, clip, width=0.48\textwidth]{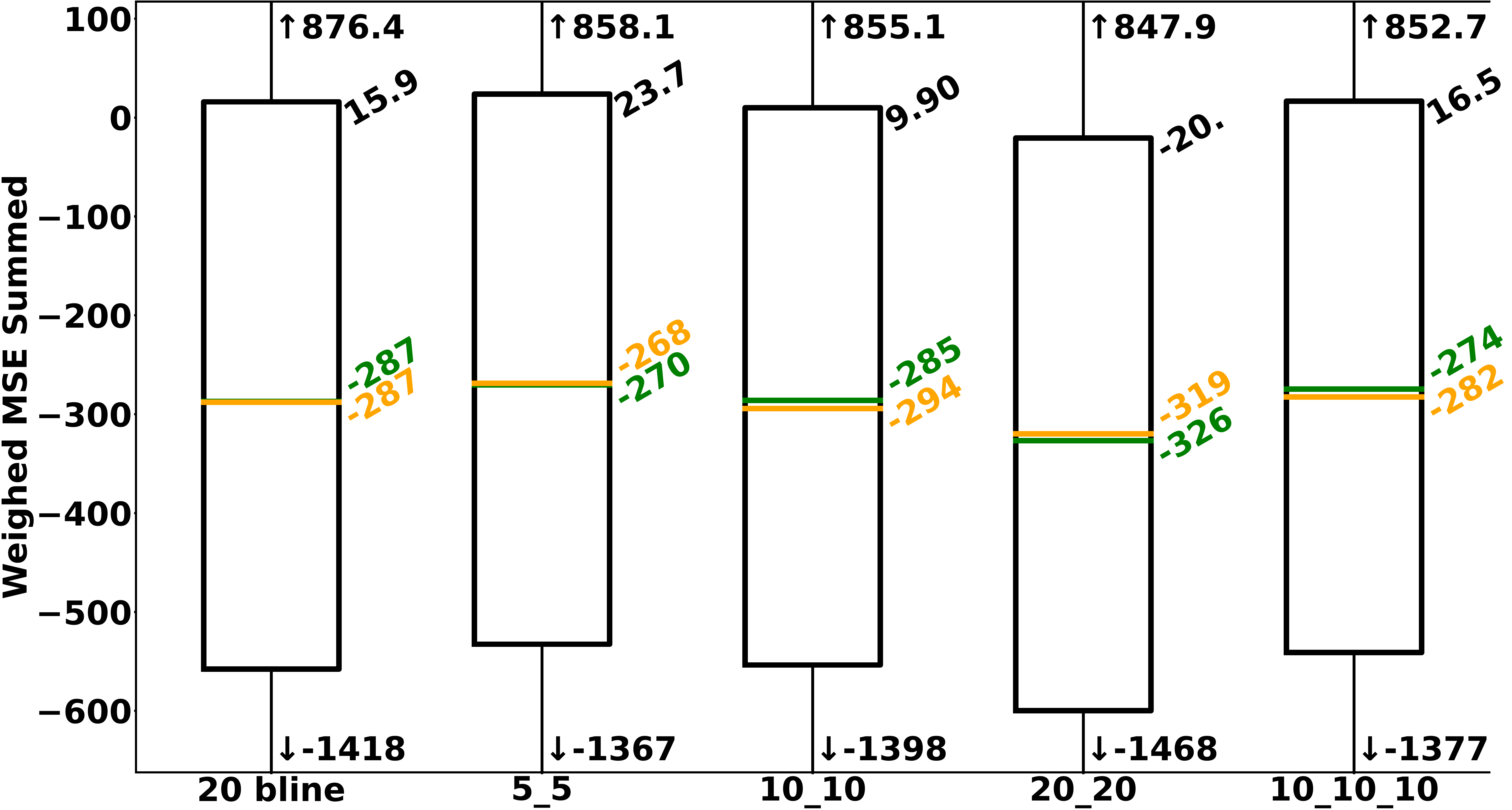}   
}
\end{subfloat}
\begin{subfloat}[Final Comparisons \label{fig:final_comparisons}]{
\centering
\includegraphics[page=1, trim={0cm, 0cm, 0cm, 0cm}, clip, width=0.48\textwidth]{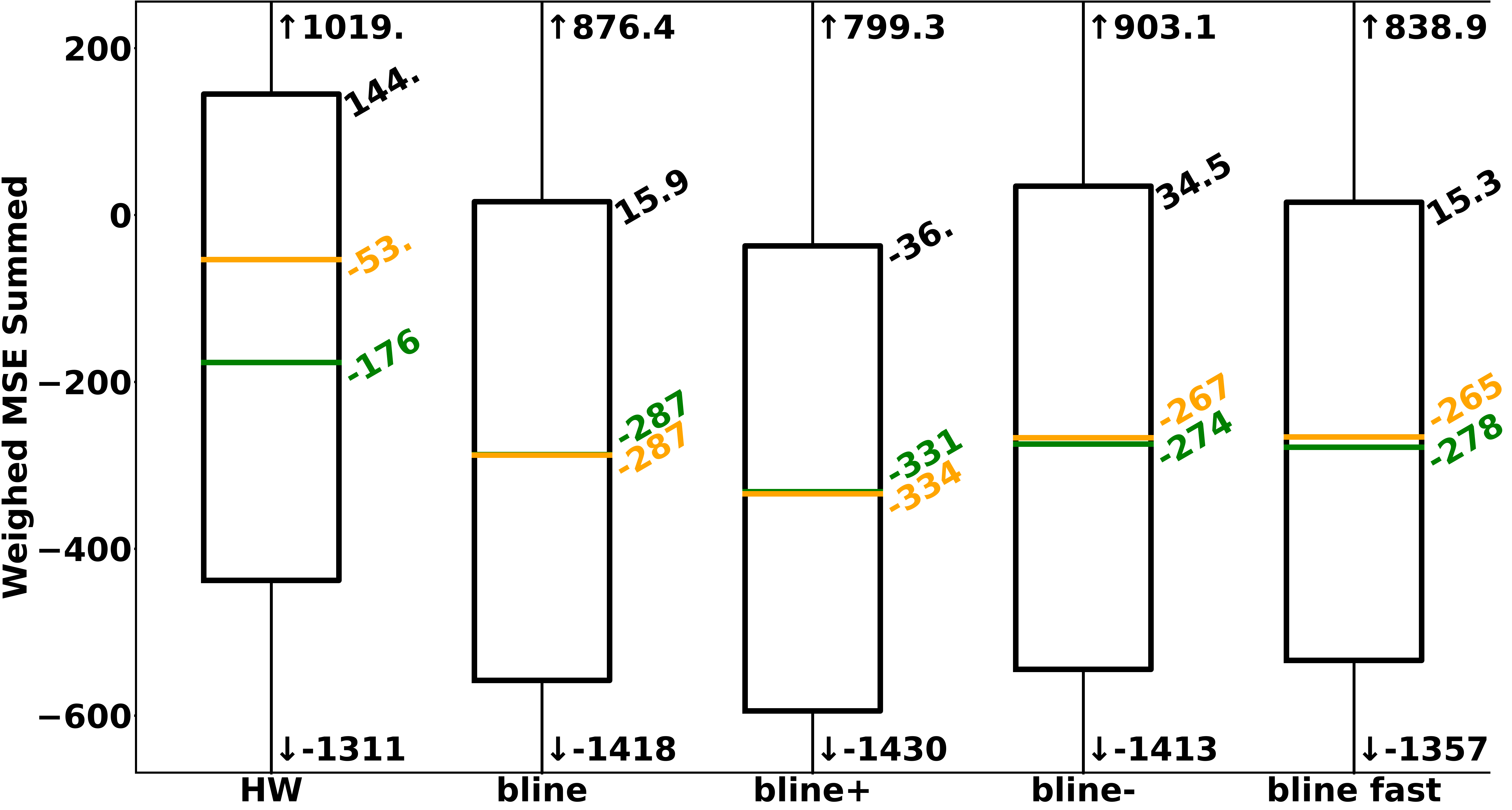}   
}
\end{subfloat}
\caption{
Box plots showing total summed NLL loss for various changes to the size or inputs of the network.
Whiskers are defined as $\pm 1.5 IQR$ and is given with the arrows, the green lines defines the median and the orange the mean value. 
The baseline model (bline) is kept consistent between all graphs.
Predictor Hidden Size \cref{fig:pred_hidden_ablation} shows the impact of varying the hidden size of the predictor network.
Predictor output width \cref{fig:pred_output_width} and depth \cref{fig:pred_output_depth} show the impact of changing the output network of the predictor.
The values are either the size of the output network layer or their multiple layers when separated by an $\_$.
The final comparison \cref{fig:final_comparisons} shows the model ran on hardware (HW), the baseline model, and an improved baseline models from \cref{subsubsec:network_recommendations}.
}
\label{fig:model_complexity_ablation}
\end{figure*}

Overall we saw small changes to the performance of the network as a function of its complexity; varying the size of the models has less impact than modifying the structure of them.
Larger network sizes can give small improvements in prediction accuracy but overall the \textit{baseline} architecture is not sensitive the number of parameters.
In this section and in \cref{fig:model_complexity_ablation}, the number at the bottom indicates either the size of the hidden/cell state of the specified \ac{LSTM} or the intermediate layers of a \ac{FNN} that takes a concatenation of state, control, and sensor readings.
For layer sizes, we always go from the same dimensionality of inputs to the same outputs, but can include additional hidden layers denoted by $\_$.
\cref{fig:pred_hidden_ablation,fig:pred_output_depth,fig:pred_output_width} deal with changing width and depth of the predictor network.

Increasing \ac{LSTM} hidden size generally increases the performance of the predictor network, but can have a strong impact on runtime and memory usage of the model.
Smaller hidden states can still perform almost as well as larger ones in \cref{fig:pred_hidden_ablation} so our architecture is usable even on weaker hardware than was used for this project.
Predictor hidden size is the change that has the largest impact on performance so it should be tuned to the real-time needs of the platform.
When exploring the depth vs width in \cref{fig:pred_output_depth,fig:pred_output_width} of the predictor network, increasing complexity gave inconsistent results in width and depth.

Finally, we look at changing values of the initialization network.
Overall, the impact was barely noticeable and it seemed that changing the complexity of these networks had minimal impact on performance.
Values of the mean and $75\%$ could change by $5-10$ in \ac{NLL} units, but there was no clear trend, just variation model to model.
This matches with the results seen when changing $\tau$, indicating that some initialization is important but specifics on the size/length matter less than it merely existing.

\subsubsection{Recommendations}
\label{subsubsec:network_recommendations}
In summary, we have validated our structured learning problem through a rigorous ablation study.
The optimal model above the \textit{baseline} would be to,
\begin{enumerate}
    \item Update weights of the delay model using \cref{eq:loss_function} without any loss on the delay states
    \item Use the hybrid structure for process noise in \cref{eq:specific_unc_update}
    \item include a prediction of the initial covariance from the initialization network
    \item Decrease the size of the initialization network (8 hidden, linear init output network) since it has little to no impact 
\end{enumerate}
These recommendations were followed to generate the \textit{bline+} model seen in \cref{fig:final_comparisons}.
The last suggestion could reduce accuracy but will speed up running the model slightly.

The simplifications to the \textit{baseline} architecture that can be done without substantial loss to performance would be to 
\begin{enumerate}
    \item Remove the G matrix (only predict $\hat{Q}'$) in \cref{eq:specific_unc_update}
    \item Remove the force as an input to the networks
\end{enumerate}
These recommendations are combined with the ones for \textit{bline+} to create \textit{bline-}.
We note that it performs slightly worse than the \textit{baseline} model but is simpler in structure. 
This is most notable when it comes to the computation of $G_t$ and $G_t (c_\sigma \odot \sigma(\vq)) G_t^T$ that can be cumbersome with complex nonlinear functions that might take many GPU instructions and slow computation time \cite{mppi-generic}.

Our recommendations to increase computation speed would be to keep the changes outlined in bline+, but
\begin{enumerate}
    \item Decrease $\tau$ slightly ($0.1s$)
    \item Reduce all output networks to linear layers
    \item Reduce the size of the predictor hidden / cell state
\end{enumerate}
These recommendations were followed to create the network bline fast.
For this model, we further reduce the initialization network to a hidden size of $4$ and the predictor network to a hidden size of $2$.
This shows that we can get similar performance to the \textit{baseline} model with simpler networks if the structure of the problem is kept.

The \textit{bline}, \textit{bline+}, and \textit{bline-} all have similar distance error statistics, while \textit{HW} and \textit{bline fast} are $10\%$ and $5\%$ worse respectively.
All predictions of values outside of the delay states are very similar for all models.
The differences of the three models are purely in the shape and magnitude of uncertainty they predict as seen in \cref{fig:trajectories}.

\subsection{Hardware}
\label{subsec:hardware_results}

\input{hardware_results}

\section{Conclusion}
In this work, we demonstrate the effectiveness of our approach to learning distributional dynamics and their use for a challenging navigation task.
The structure of the learning and planning problem was key for creating a general system.
Each component of the learning problem was validated through an ablation study on a large driving dataset across multiple environments.
Finally, our planning method was validated over miles of off-road autonomous driving in adverse conditions and showed risk-aware behavior that was environmentally dependent.
These hardware experiments elucidated how choosing appropriate approximations of constraints and modeling their uncertainty can provide the intelligent behavior needed for autonomy deployment in the real world.




\balance
\bibliographystyle{IEEEtran}
\bibliography{references}

\newpage

%
%
%
%

\vfill

\clearpage
\section{Appendix}

\subsection{Additional Ablation Study Results}

The appendix includes additional box plots that are not focused on the MSE loss, to better underscore smaller points made in the results section.
The main point to be made is that uncertainty size and shape is important for the loss function error.
So a smaller loss does not directly imply the mean prediction is better and a similar mean prediction does not imply the uncertainty is improved either.

\begin{figure*}[htbp]
\centering
\begin{subfloat}[Structure Changes 1\label{fig:structure_bad_distance}]{
\centering
\includegraphics[page=5, trim={0cm, 0cm, 0cm, 0cm}, clip, width=0.48\textwidth]{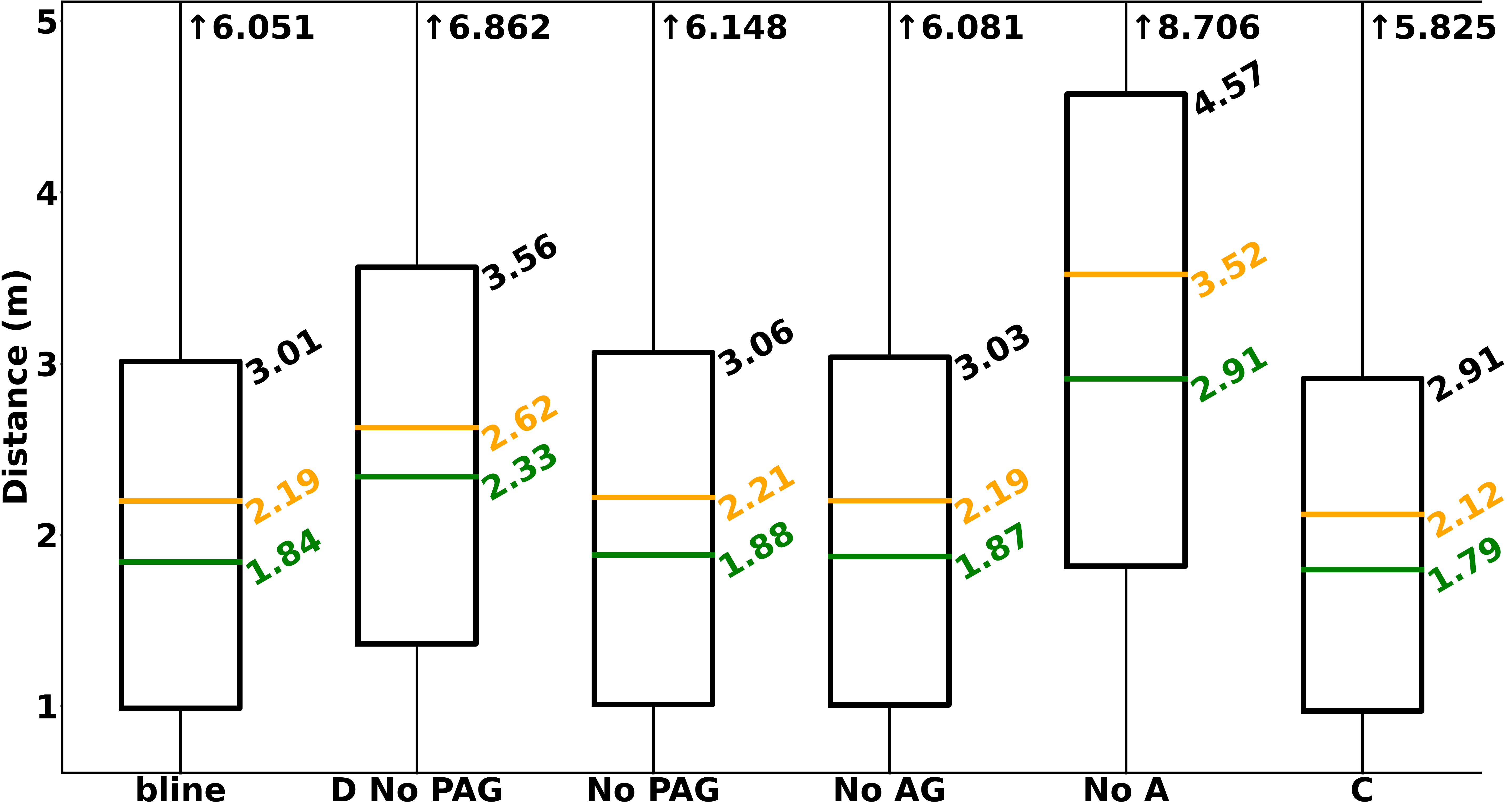}  
}
\end{subfloat}
\begin{subfloat}[Structure Changes 2\label{fig:structure_good_distance}]{
\centering
\includegraphics[page=5, trim={0cm, 0cm, 0cm, 0cm}, clip, width=0.48\textwidth]{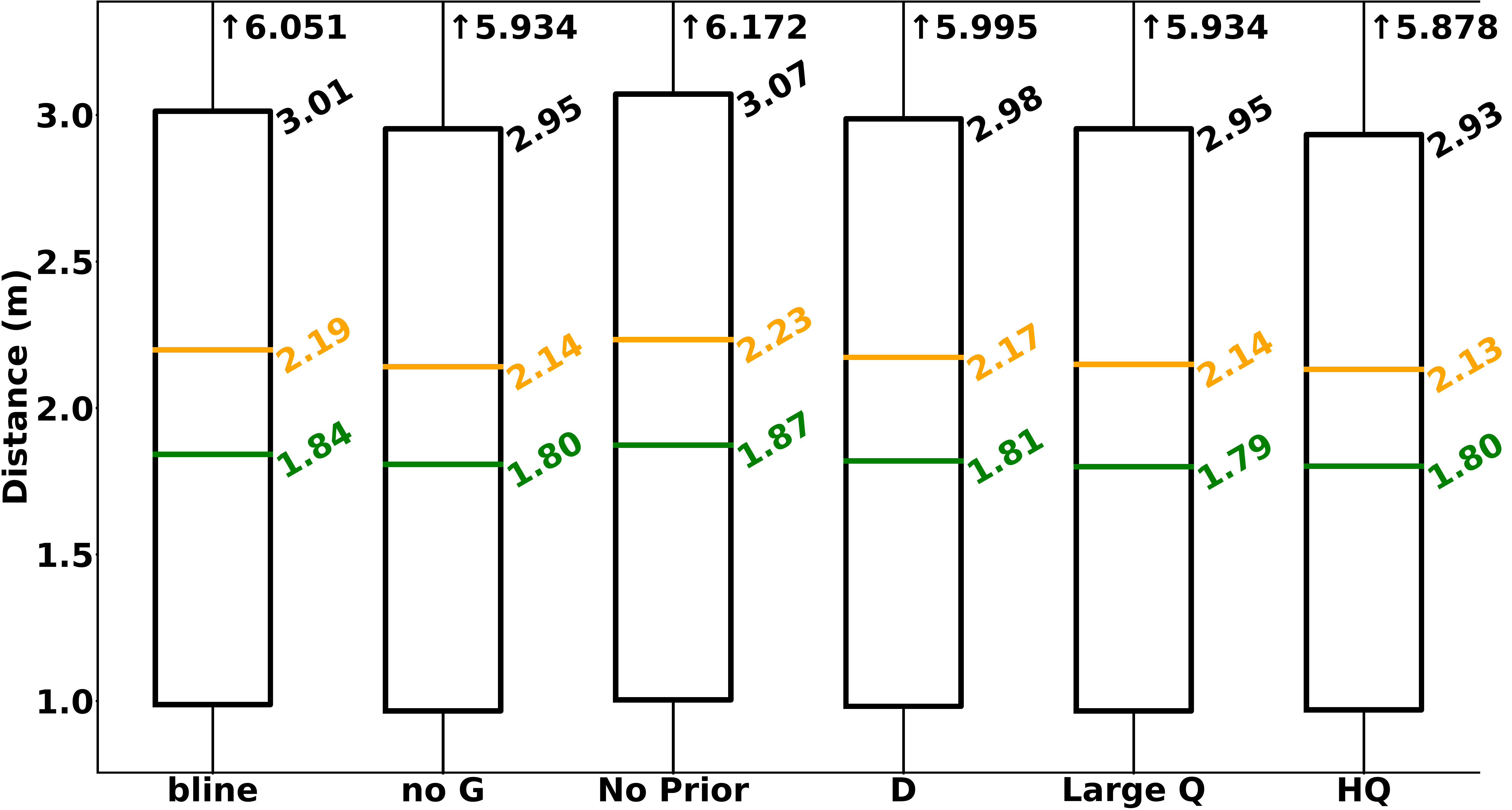}  
}
\end{subfloat}
\begin{subfloat}[Structure Changes 3 \label{fig:structure_weird_distance}]{
\centering
\includegraphics[page=5, trim={0cm, 0cm, 0cm, 0cm}, clip, width=0.48\textwidth]{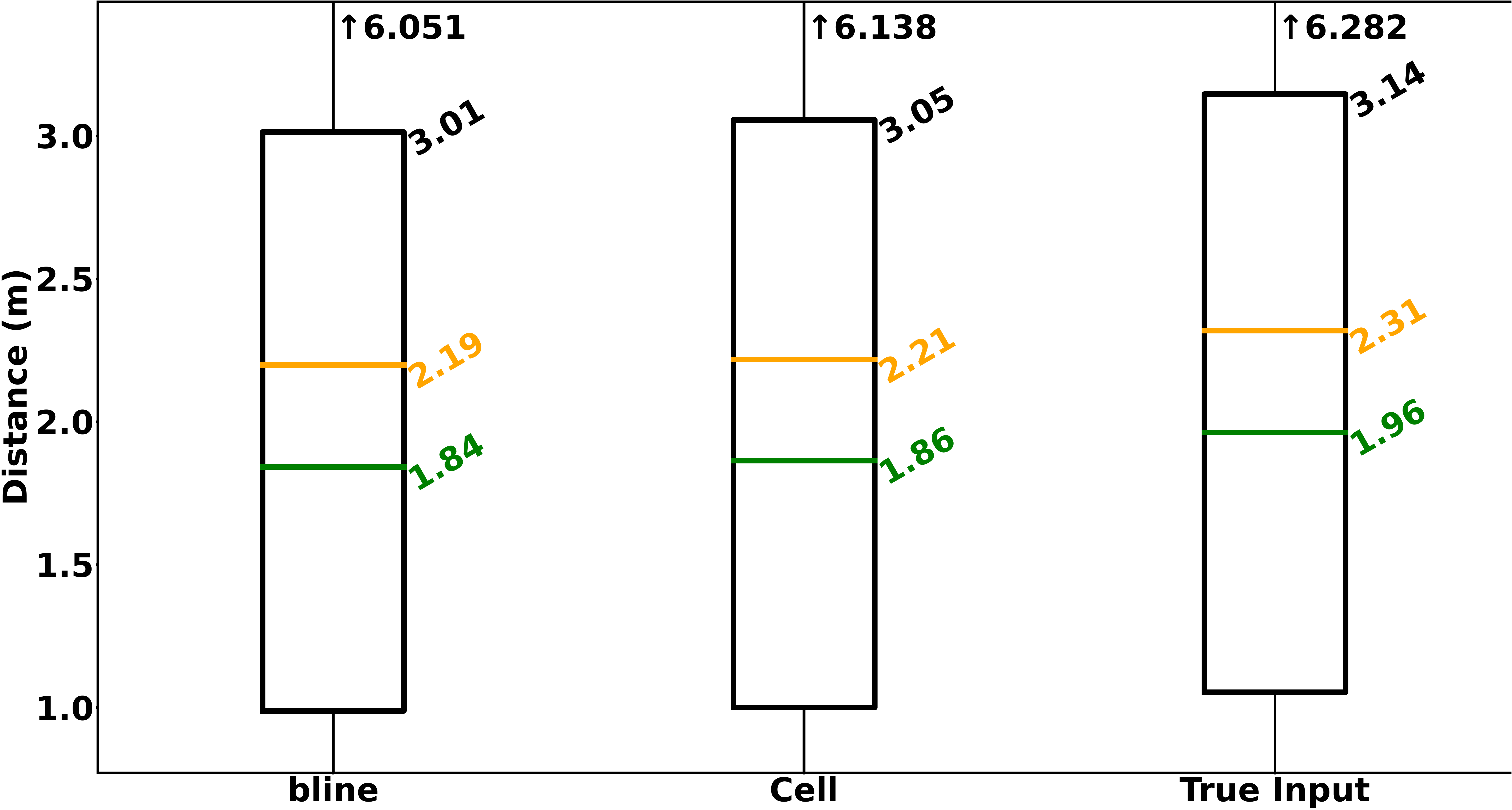}   
}
\end{subfloat}
\begin{subfloat}[Architecture \label{fig:architecture_distance}]{
\centering
\includegraphics[page=5, trim={0cm, 0cm, 0cm, 0cm}, clip, width=0.48\textwidth]{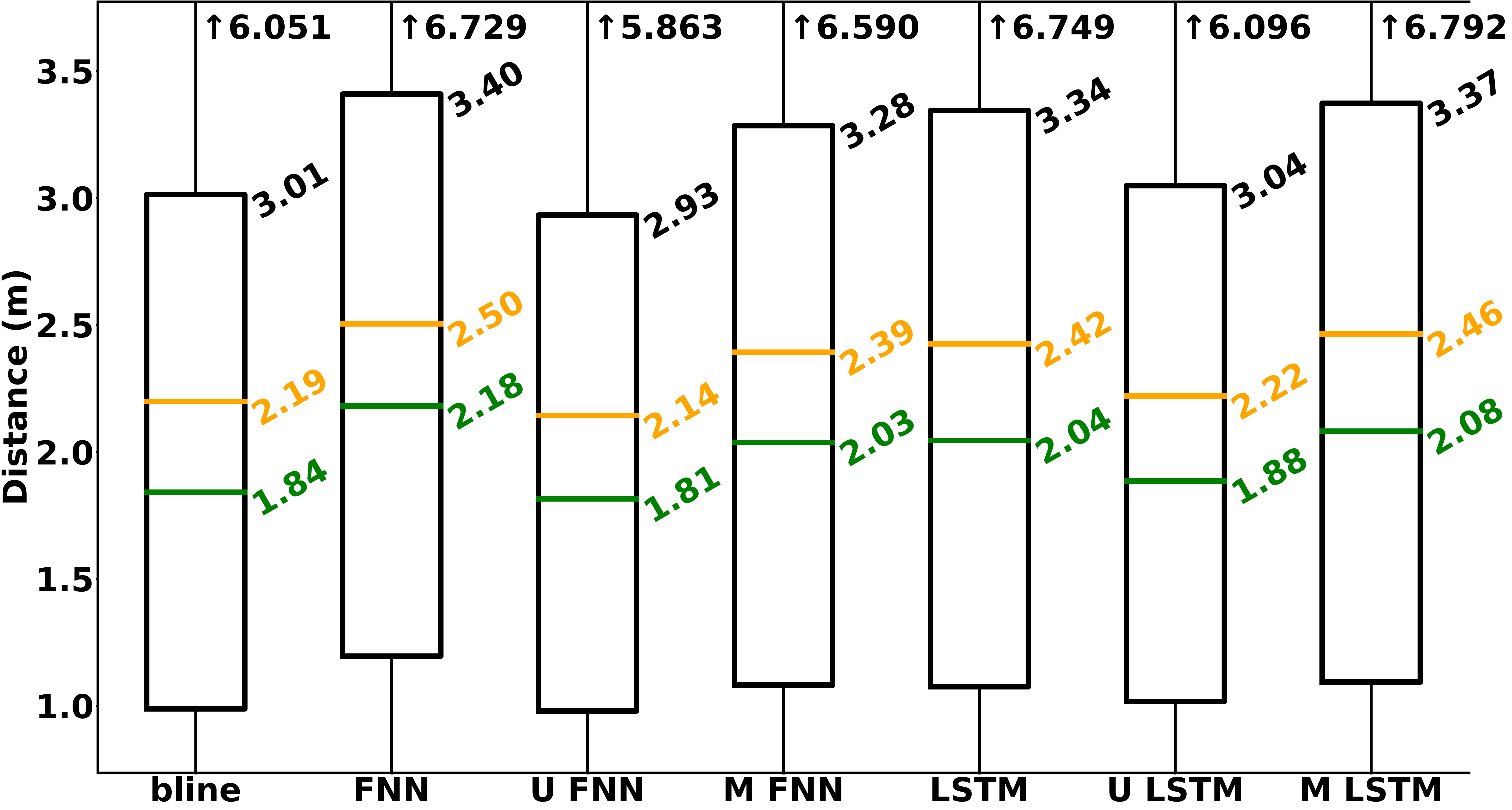}
}
\end{subfloat}
\begin{subfloat}[Initialization \label{fig:init_ablation_distance}]{
\centering
\includegraphics[page=5, trim={0cm, 0cm, 0cm, 0cm}, clip, width=0.48\textwidth]{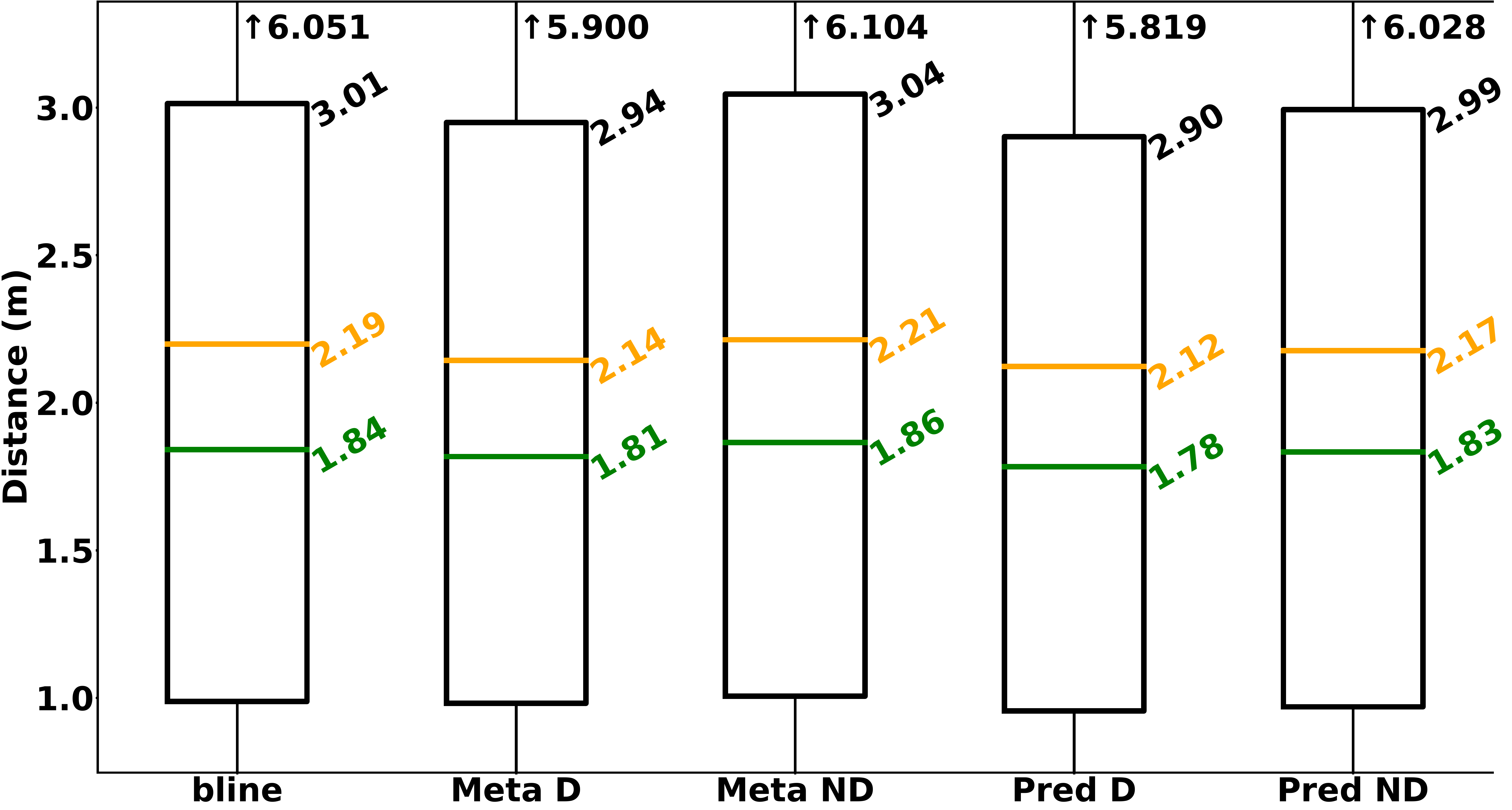}   
}
\end{subfloat}
\begin{subfloat}[Loss Structure \label{fig:structure_loss_distance}]{
\centering
\includegraphics[page=5, trim={0cm, 0cm, 0cm, 0cm}, clip, width=0.48\textwidth]{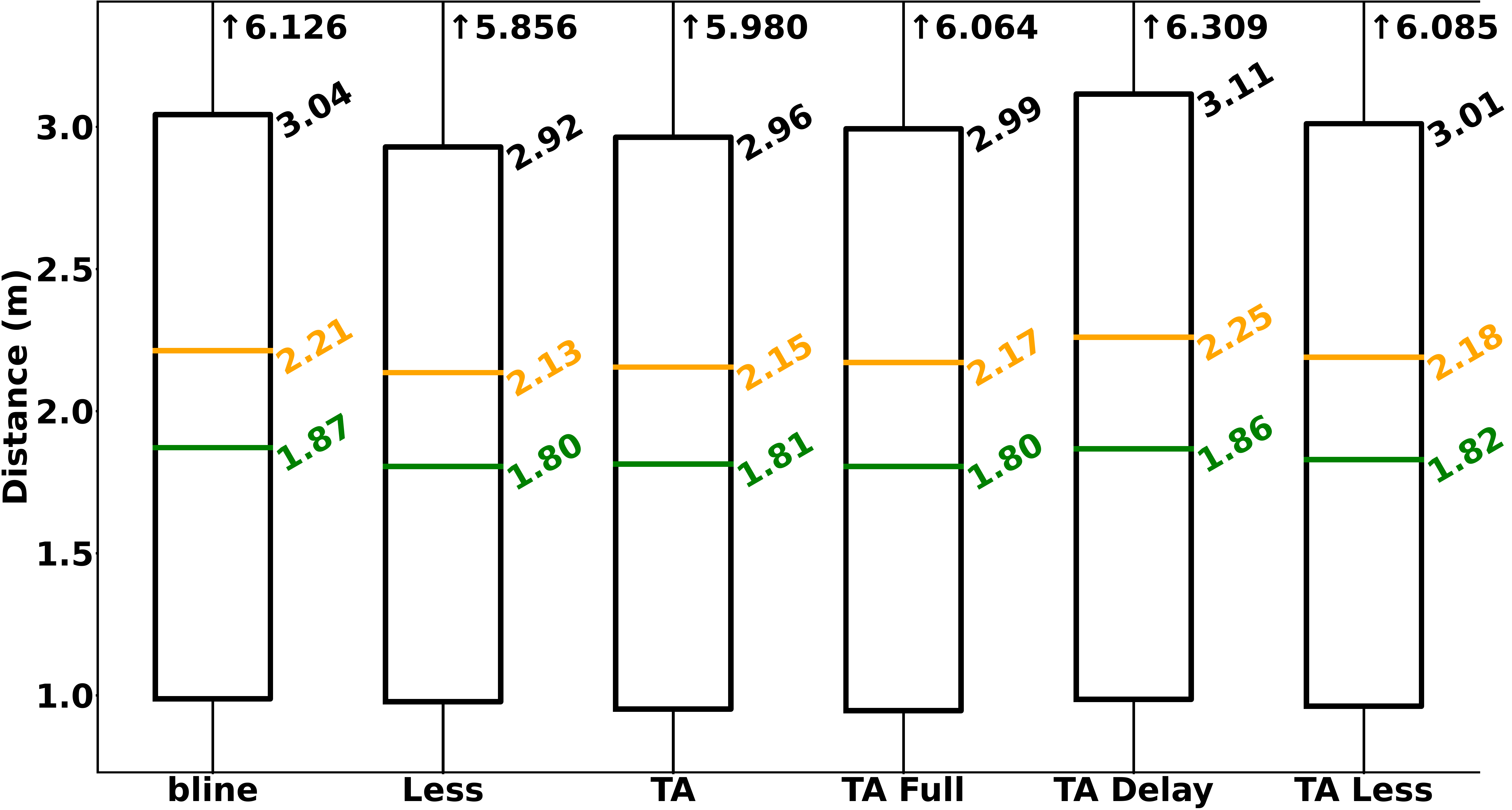}   
}
\end{subfloat}
\begin{subfloat}[Buffer History ($\tau$) \label{fig:tau_ablation_distance}]{
\centering
\includegraphics[page=5, trim={0cm, 0cm, 0cm, 0cm}, clip, width=0.48\textwidth]{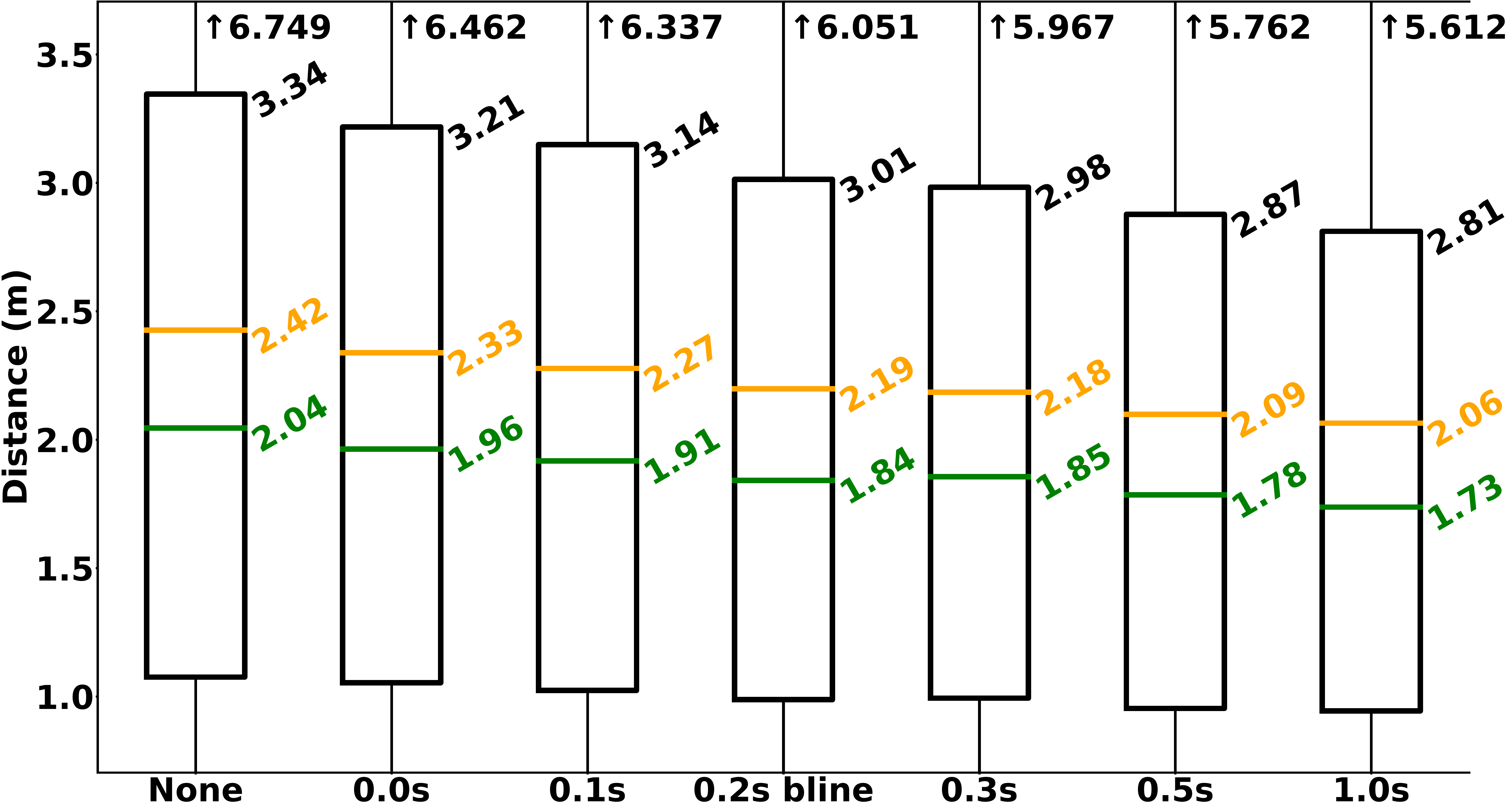} 
}
\end{subfloat}
\begin{subfloat}[Final Comparison \label{fig:final_comparison_distance}]{
\centering
\includegraphics[page=5, trim={0cm, 0cm, 0cm, 0cm}, clip, width=0.48\textwidth]{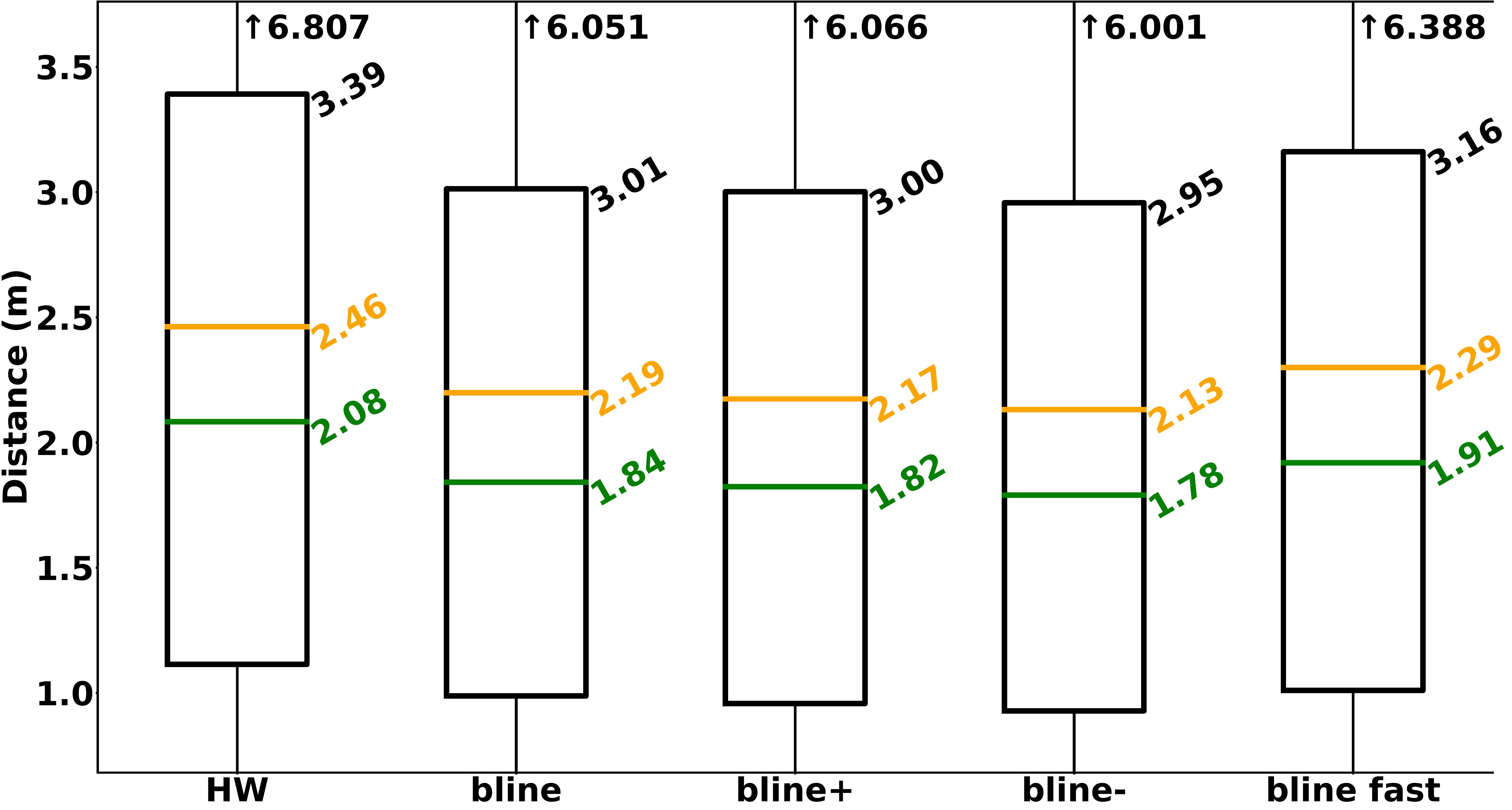}  
}
\end{subfloat}
\caption{
Box plots showing distance error at the end of the $5s$ trajectory for various structural changes to the networks. 
Whiskers are defined as $\pm 1.5 IQR$ and is given with the arrows, the green line defines the median and the orange the mean value. 
The baseline model (bline) is kept consistent.
The explanations of the keys used in shown in the text \cref{subsubsec:structure_ablation}.
}
\label{fig:structure_ablation_results_distance}
\end{figure*}

\begin{figure*}[htbp]
\centering
\begin{subfloat}[Structure Changes 1\label{fig:structure_bad_yaw}]{
\centering
\includegraphics[page=15, trim={0cm, 0cm, 0cm, 0cm}, clip, width=0.48\textwidth]{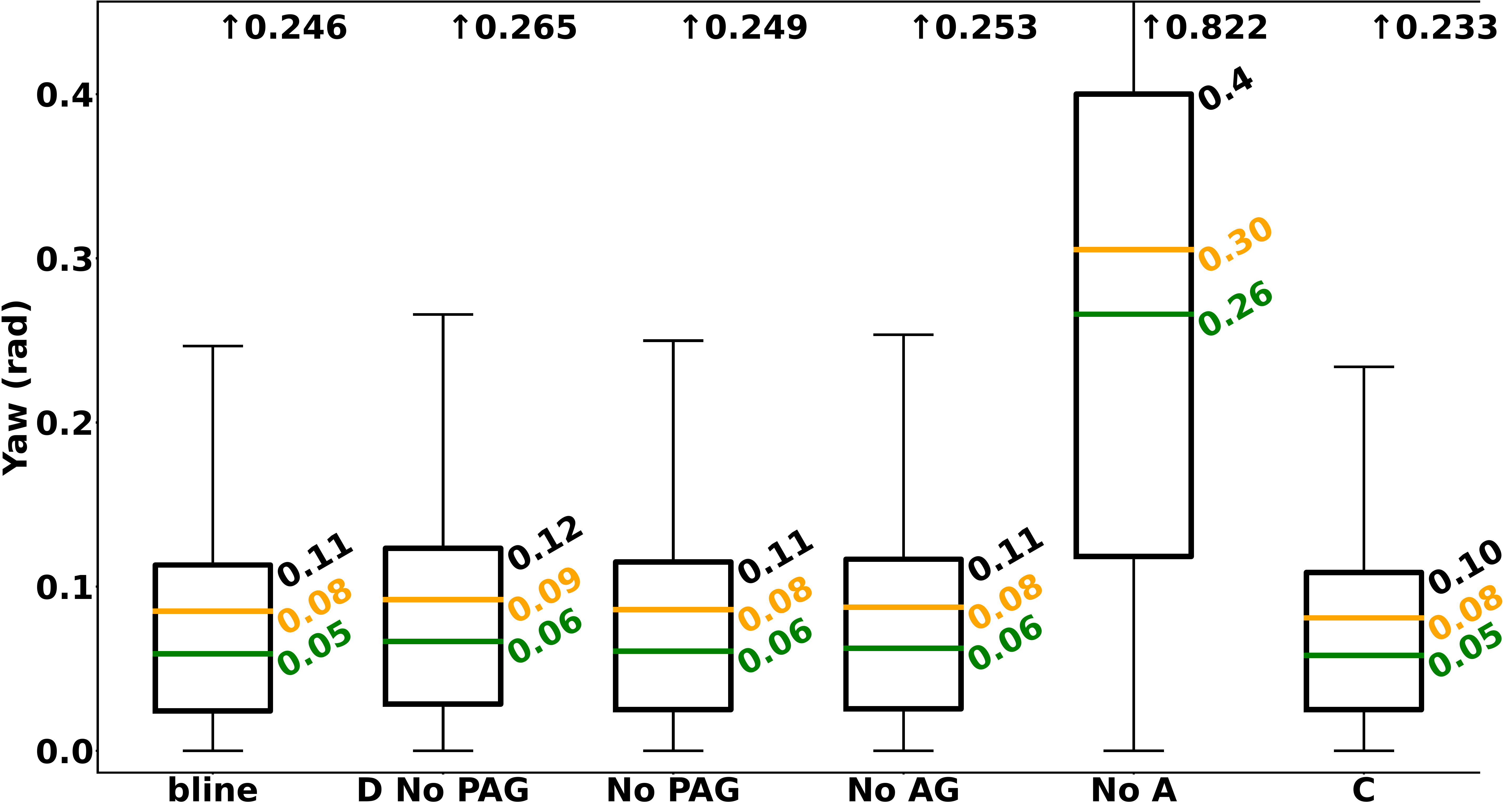}  
}
\end{subfloat}
\begin{subfloat}[Structure Changes 2\label{fig:structure_good_yaw}]{
\centering
\includegraphics[page=15, trim={0cm, 0cm, 0cm, 0cm}, clip, width=0.48\textwidth]{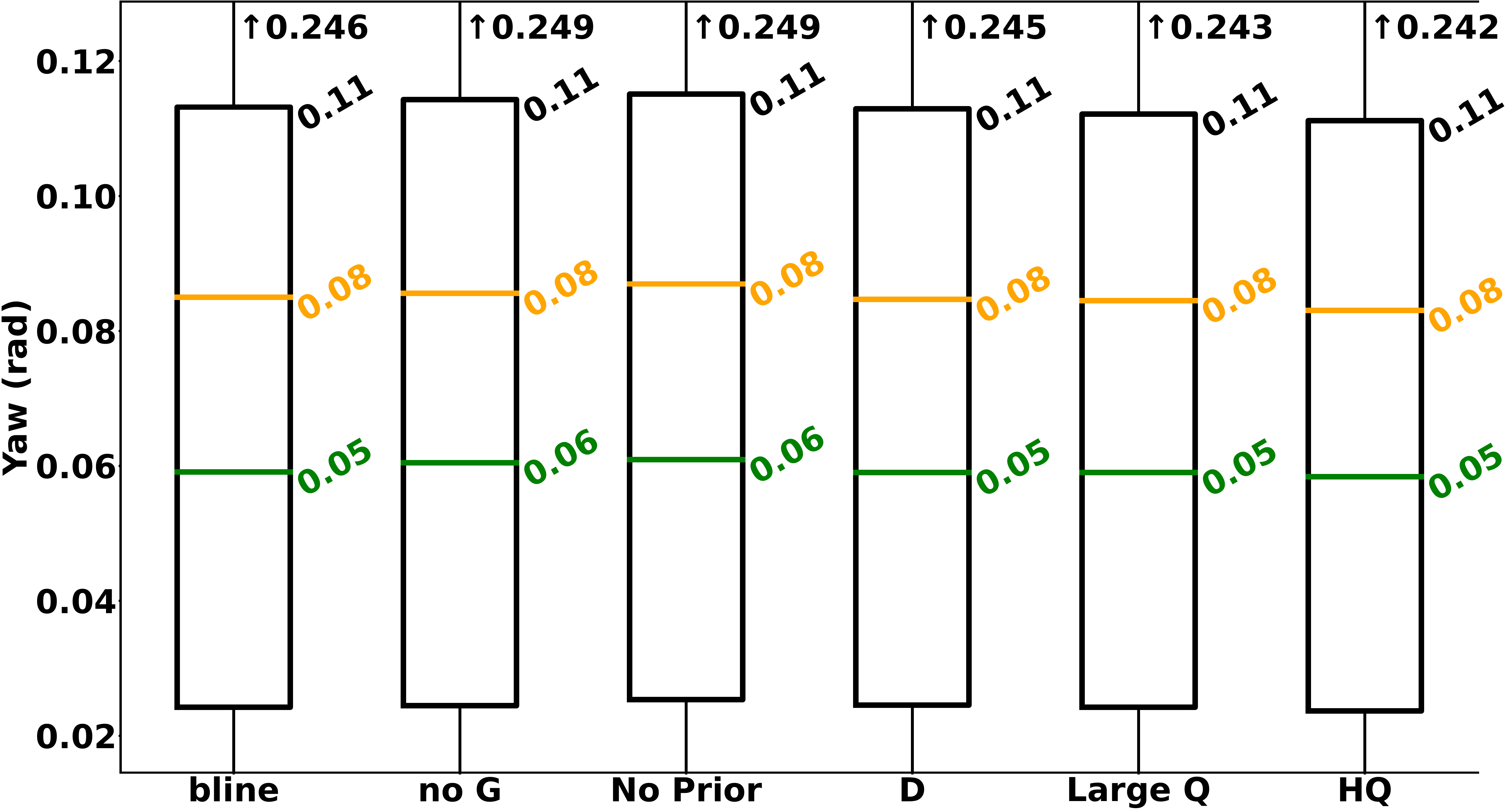}  
}
\end{subfloat}
\begin{subfloat}[Structure Changes 3 \label{fig:structure_weird_yaw}]{
\centering
\includegraphics[page=15, trim={0cm, 0cm, 0cm, 0cm}, clip, width=0.48\textwidth]{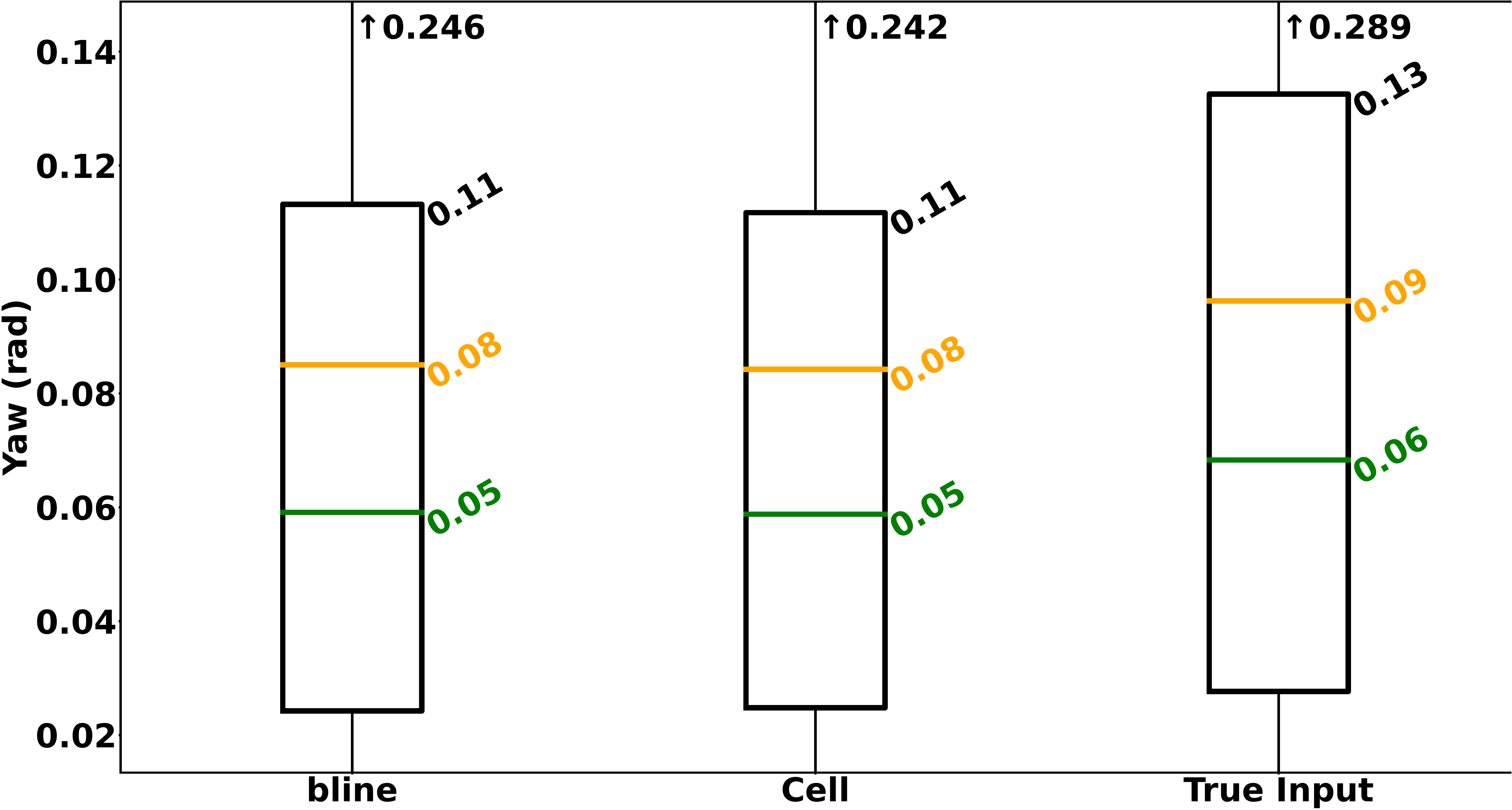}   
}
\end{subfloat}
\begin{subfloat}[Architecture \label{fig:architecture_yaw}]{
\centering
\includegraphics[page=15, trim={0cm, 0cm, 0cm, 0cm}, clip, width=0.48\textwidth]{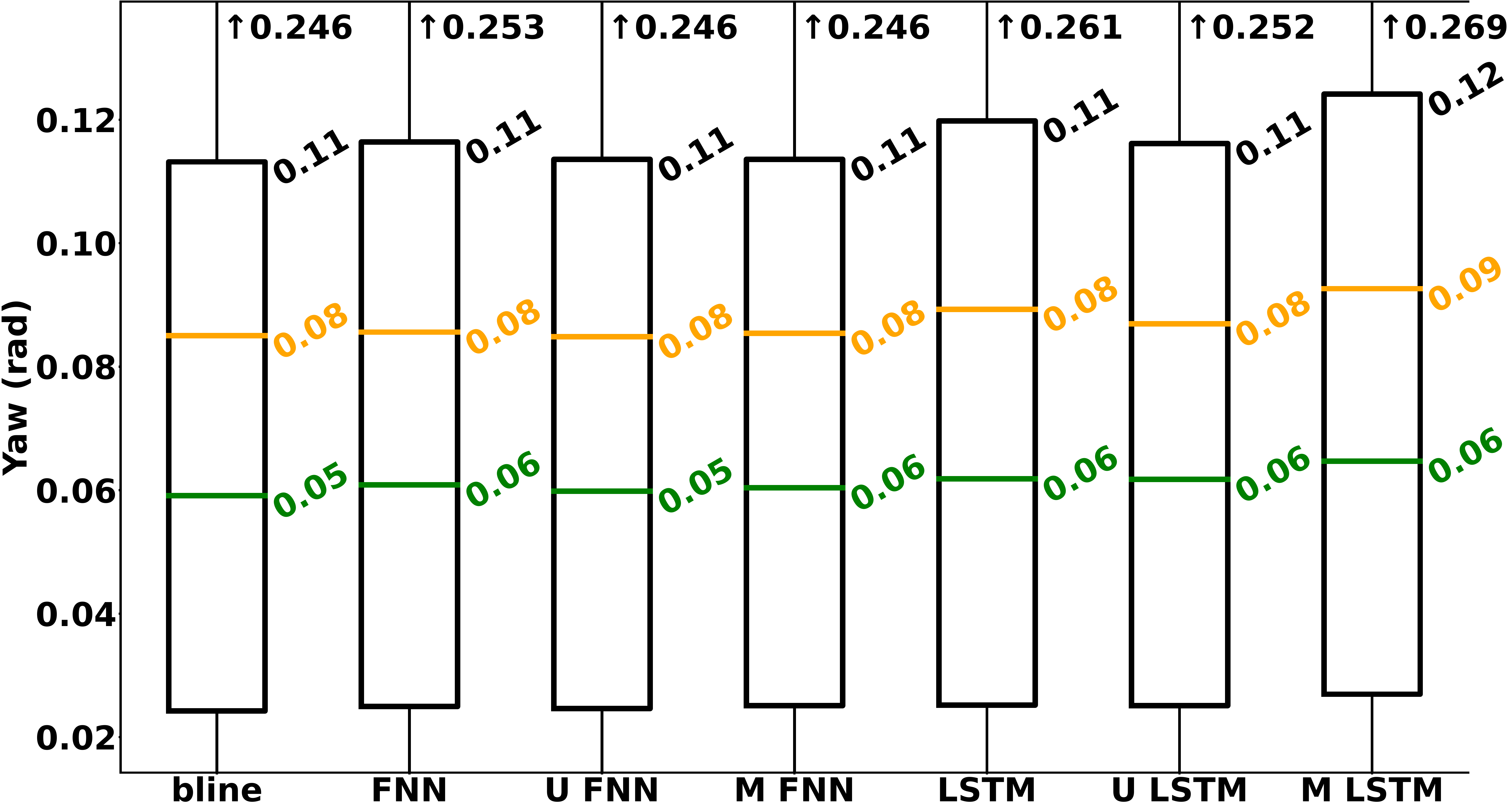}
}
\end{subfloat}
\begin{subfloat}[Initialization \label{fig:init_ablation_yaw}]{
\centering
\includegraphics[page=15, trim={0cm, 0cm, 0cm, 0cm}, clip, width=0.48\textwidth]{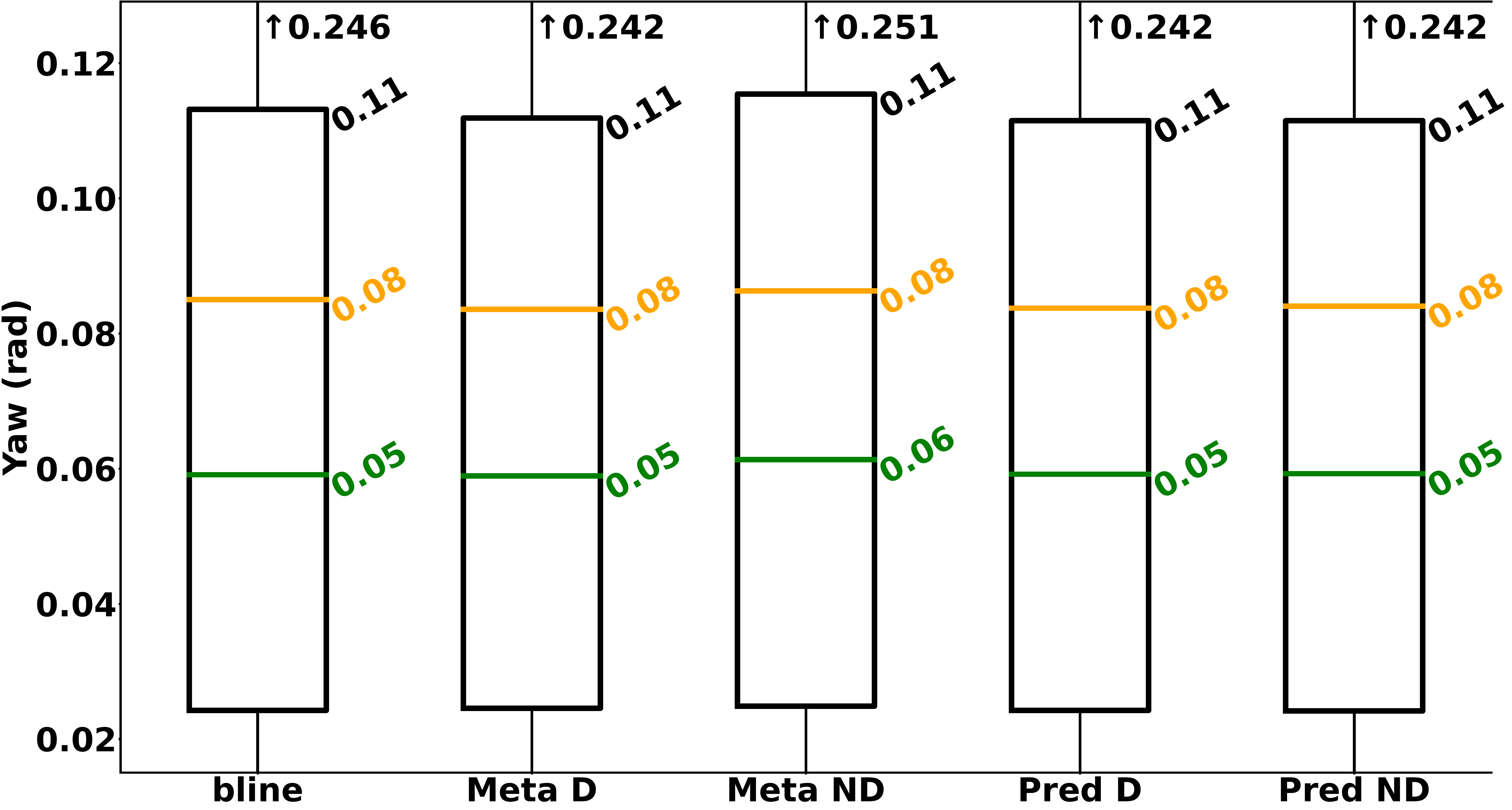}   
}
\end{subfloat}
\begin{subfloat}[Loss Structure \label{fig:structure_loss_yaw}]{
\centering
\includegraphics[page=15, trim={0cm, 0cm, 0cm, 0cm}, clip, width=0.48\textwidth]{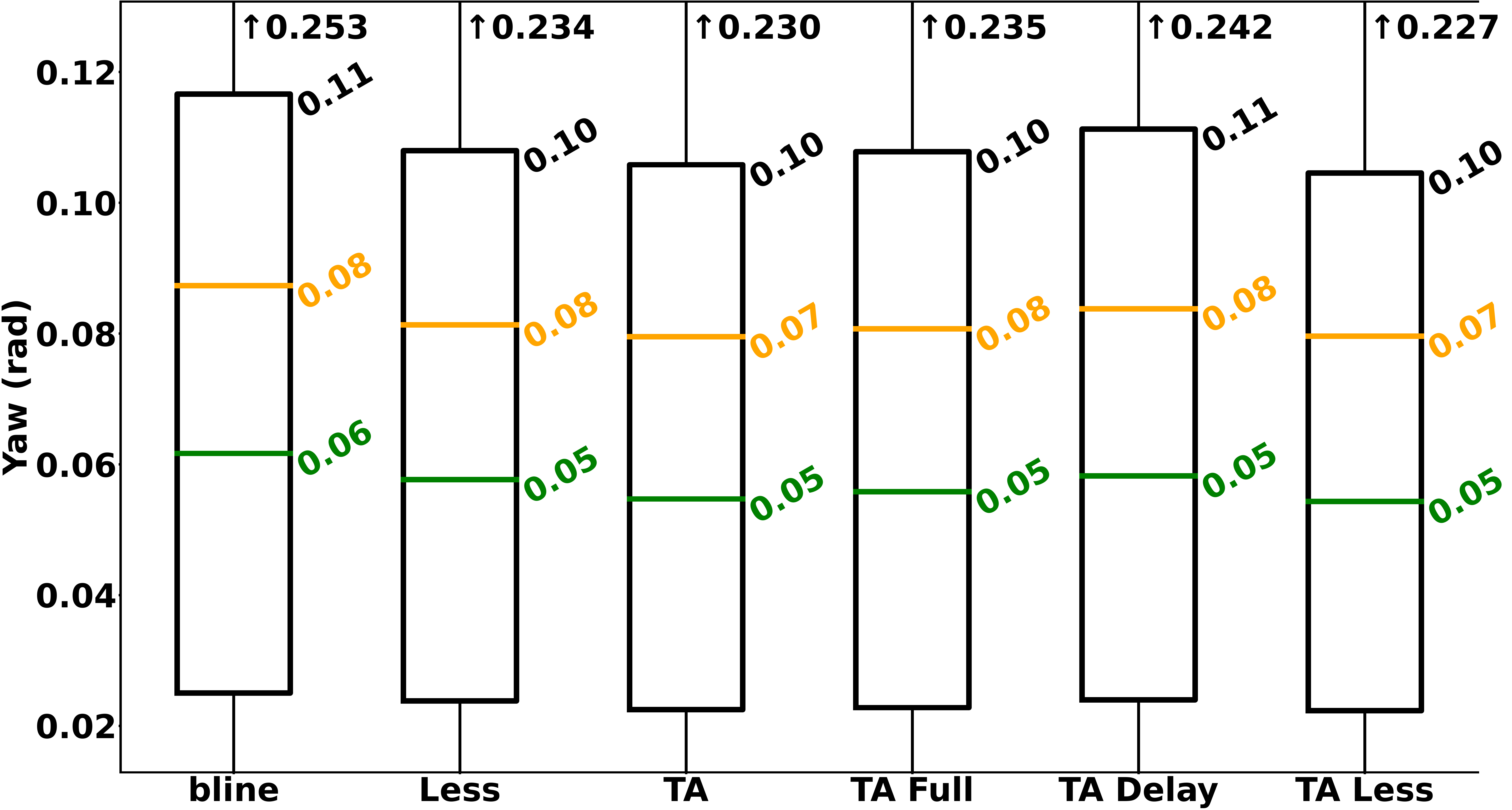}   
}
\end{subfloat}
\begin{subfloat}[Buffer History ($\tau$) \label{fig:tau_ablation_yaw}]{
\centering
\includegraphics[page=15, trim={0cm, 0cm, 0cm, 0cm}, clip, width=0.48\textwidth]{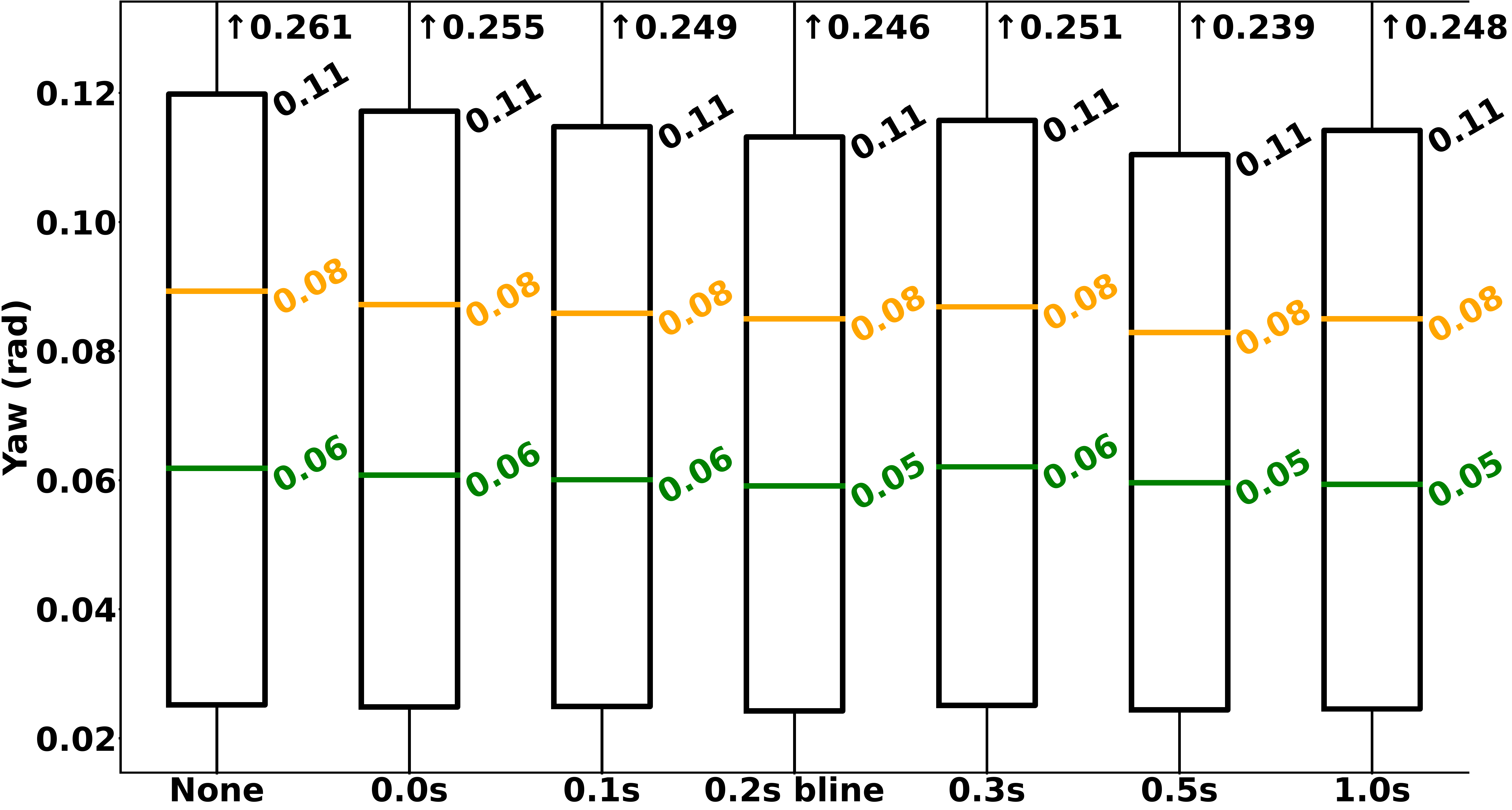} 
}
\end{subfloat}
\begin{subfloat}[Final Comparison \label{fig:final_comparison_yaw}]{
\centering
\includegraphics[page=15, trim={0cm, 0cm, 0cm, 0cm}, clip, width=0.48\textwidth]{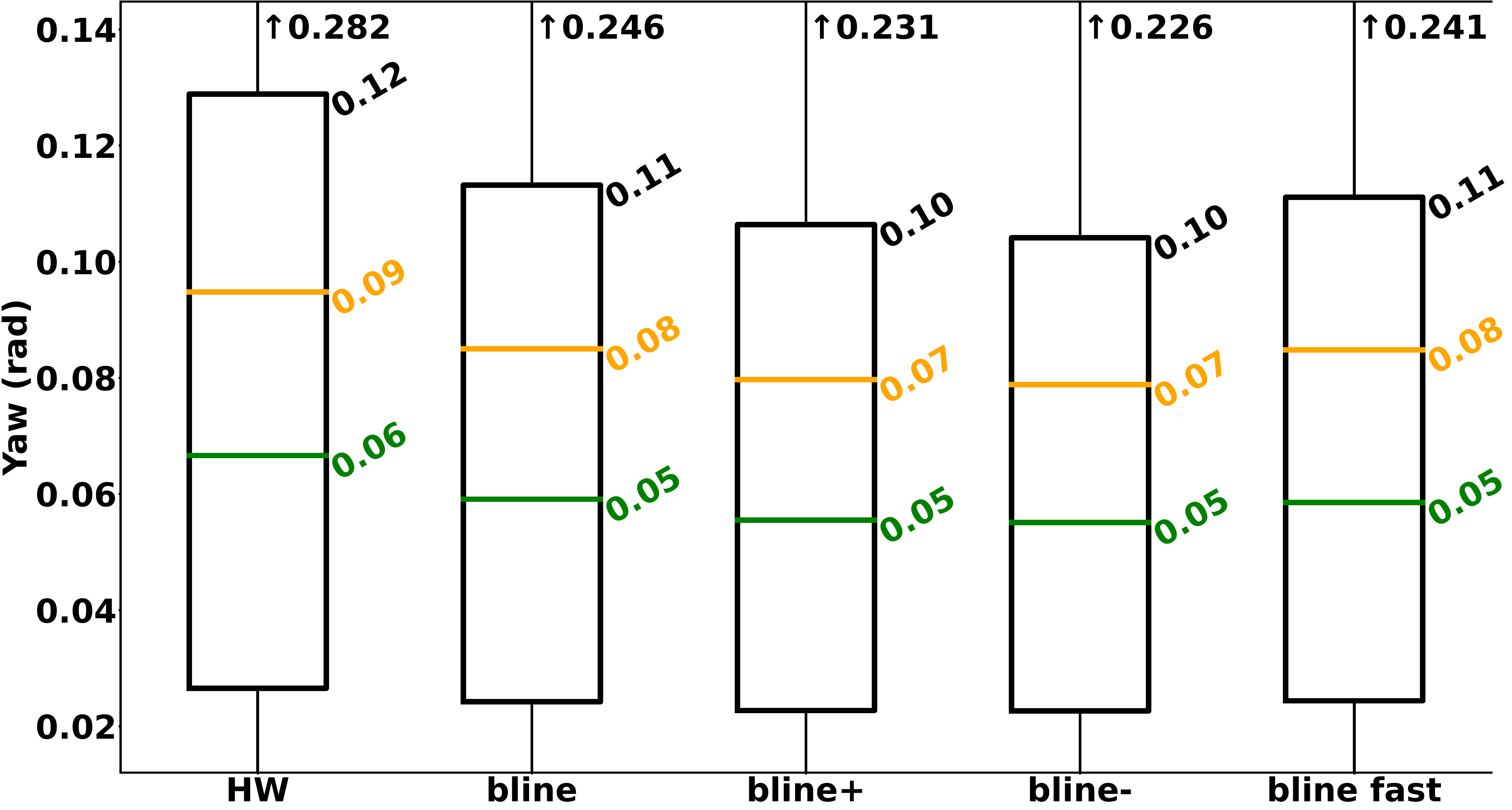}  
}
\end{subfloat}
\caption{
Box plots showing $\psi$ error at the end of the $5s$ trajectory for various structural changes to the networks. 
Whiskers are defined as $\pm 1.5 IQR$ and is given with the arrows, the green line defines the median and the orange the mean value. 
The baseline model (bline) is kept consistent.
The explanations of the keys used in shown in the text \cref{subsubsec:structure_ablation}.
}
\label{fig:structure_ablation_results_yaw}
\end{figure*}

\begin{figure*}[htbp]
\centering
\begin{subfloat}[Structure Changes 1\label{fig:structure_bad_vx}]{
\centering
\includegraphics[page=17, trim={0cm, 0cm, 0cm, 0cm}, clip, width=0.48\textwidth]{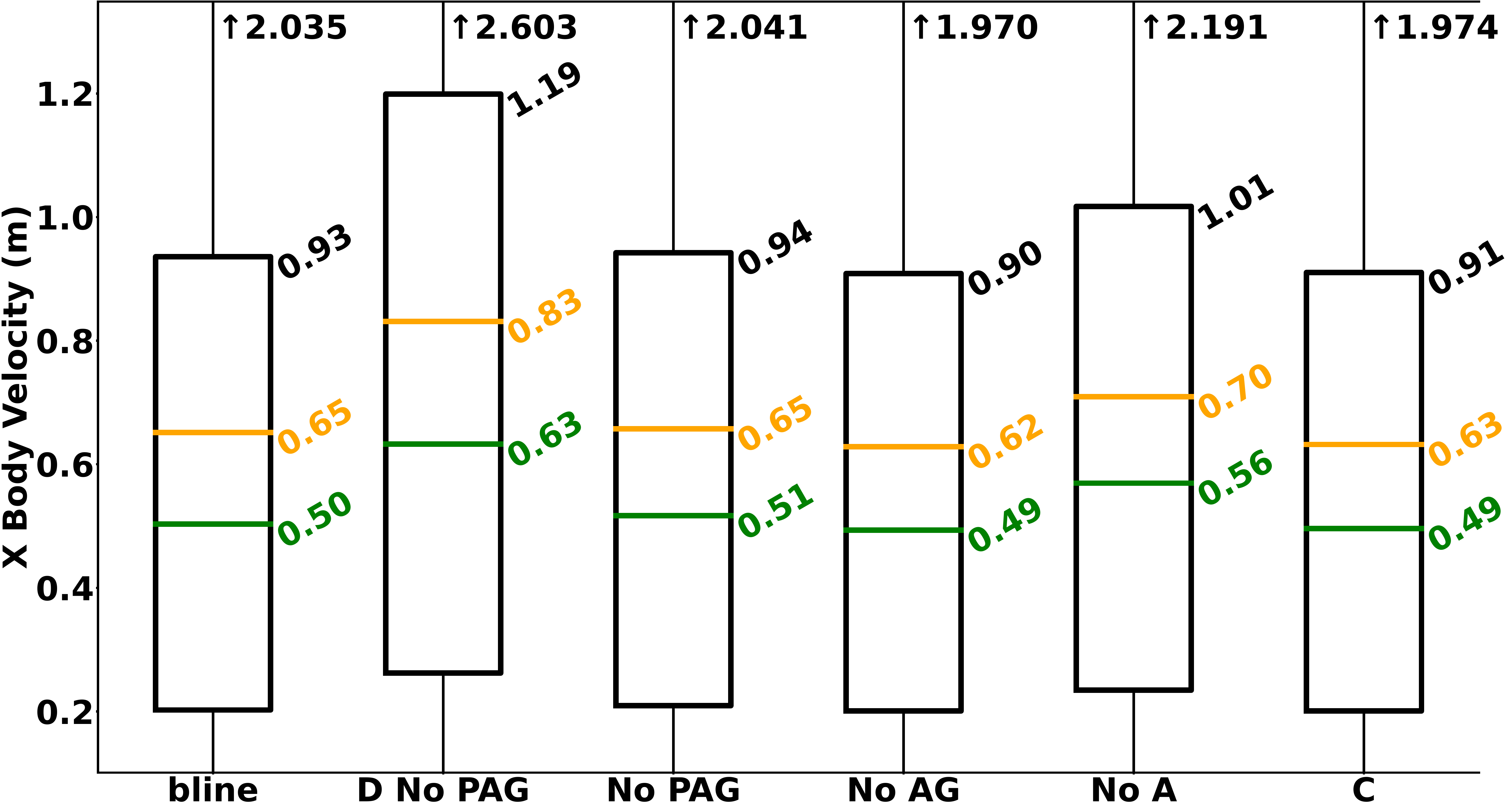}  
}
\end{subfloat}
\begin{subfloat}[Structure Changes 2\label{fig:structure_good_vx}]{
\centering
\includegraphics[page=17, trim={0cm, 0cm, 0cm, 0cm}, clip, width=0.48\textwidth]{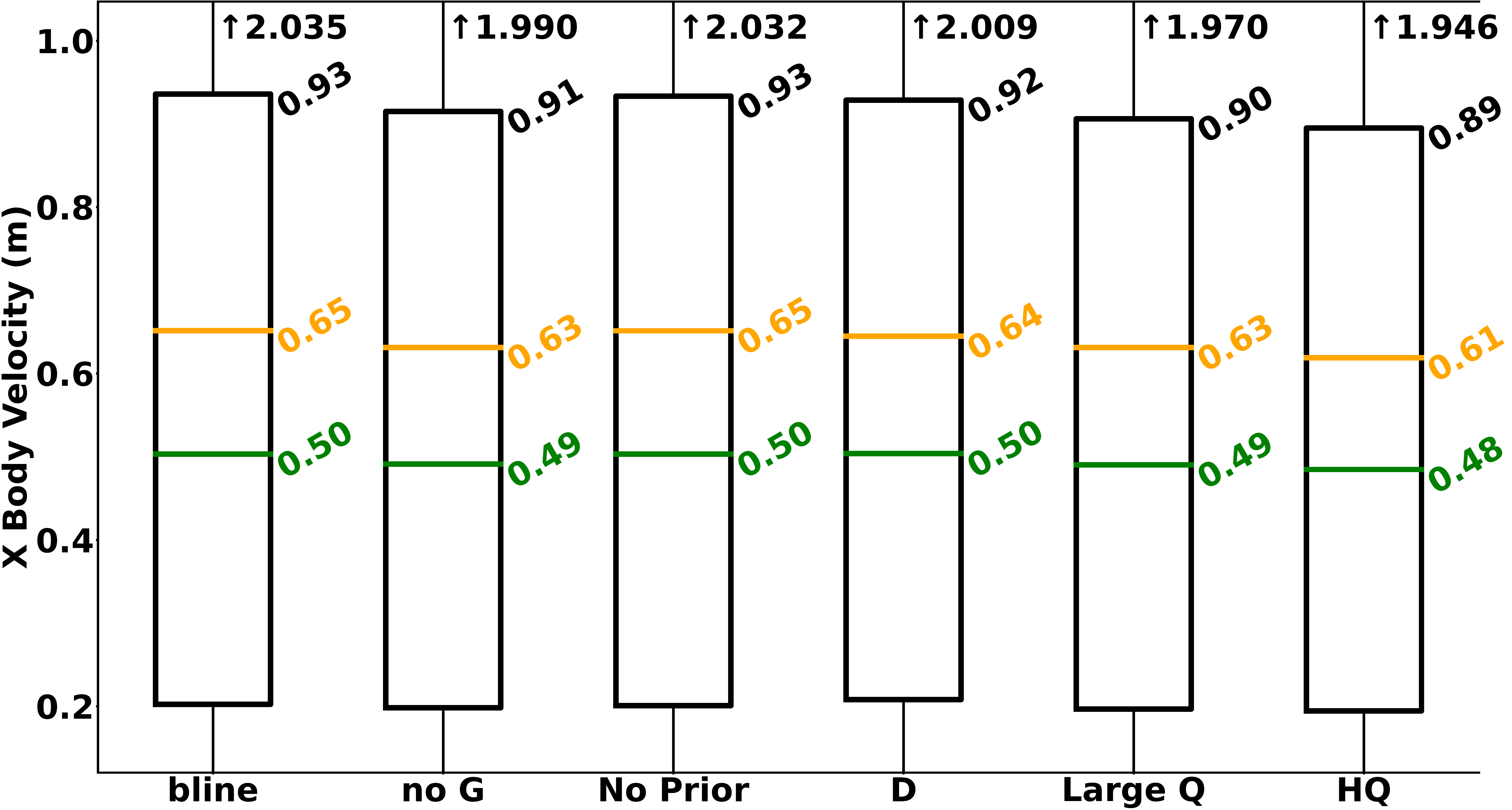}  
}
\end{subfloat}
\begin{subfloat}[Structure Changes 3 \label{fig:structure_weird_vx}]{
\centering
\includegraphics[page=17, trim={0cm, 0cm, 0cm, 0cm}, clip, width=0.48\textwidth]{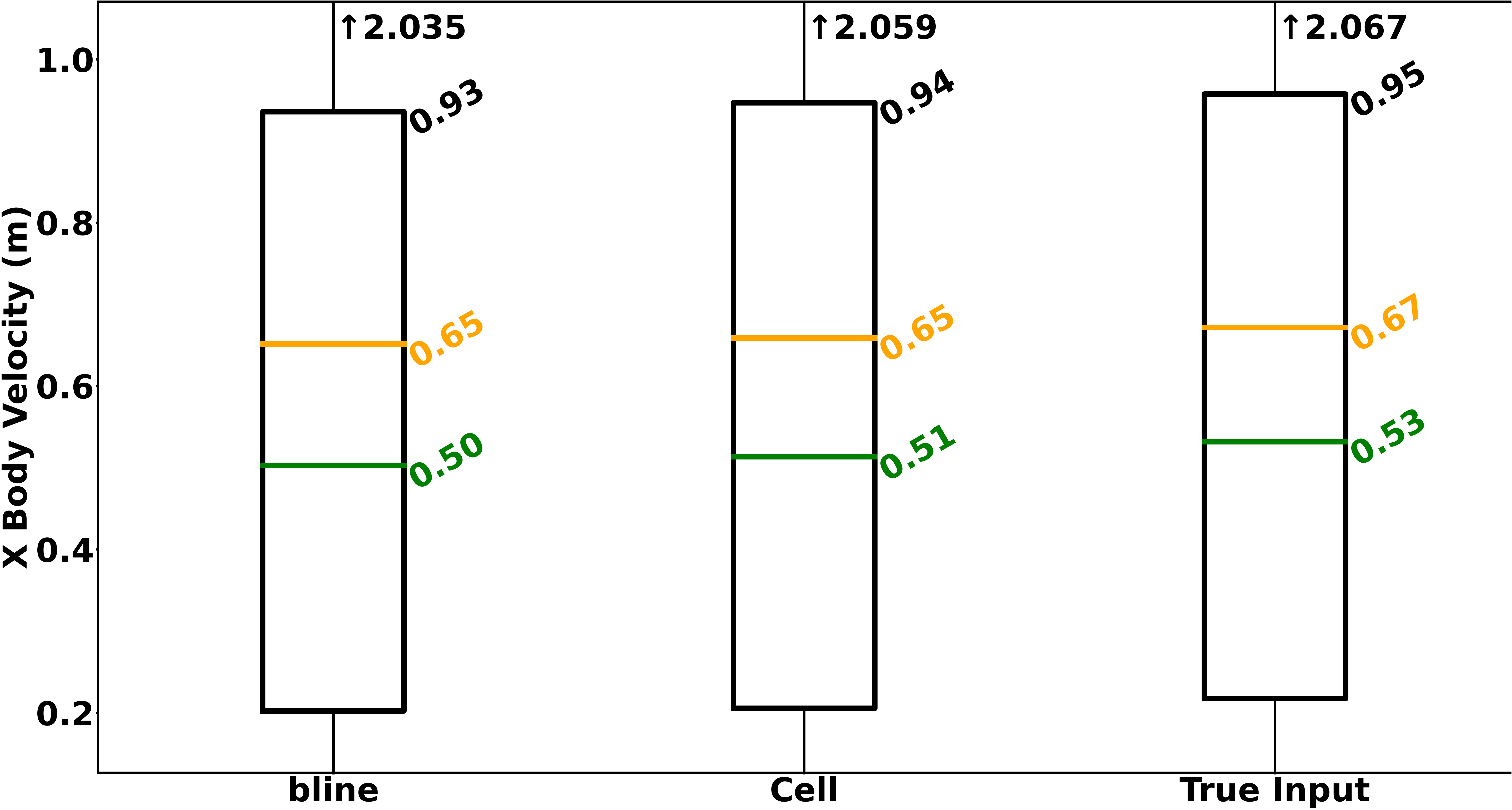}   
}
\end{subfloat}
\begin{subfloat}[Architecture \label{fig:architecture_vx}]{
\centering
\includegraphics[page=17, trim={0cm, 0cm, 0cm, 0cm}, clip, width=0.48\textwidth]{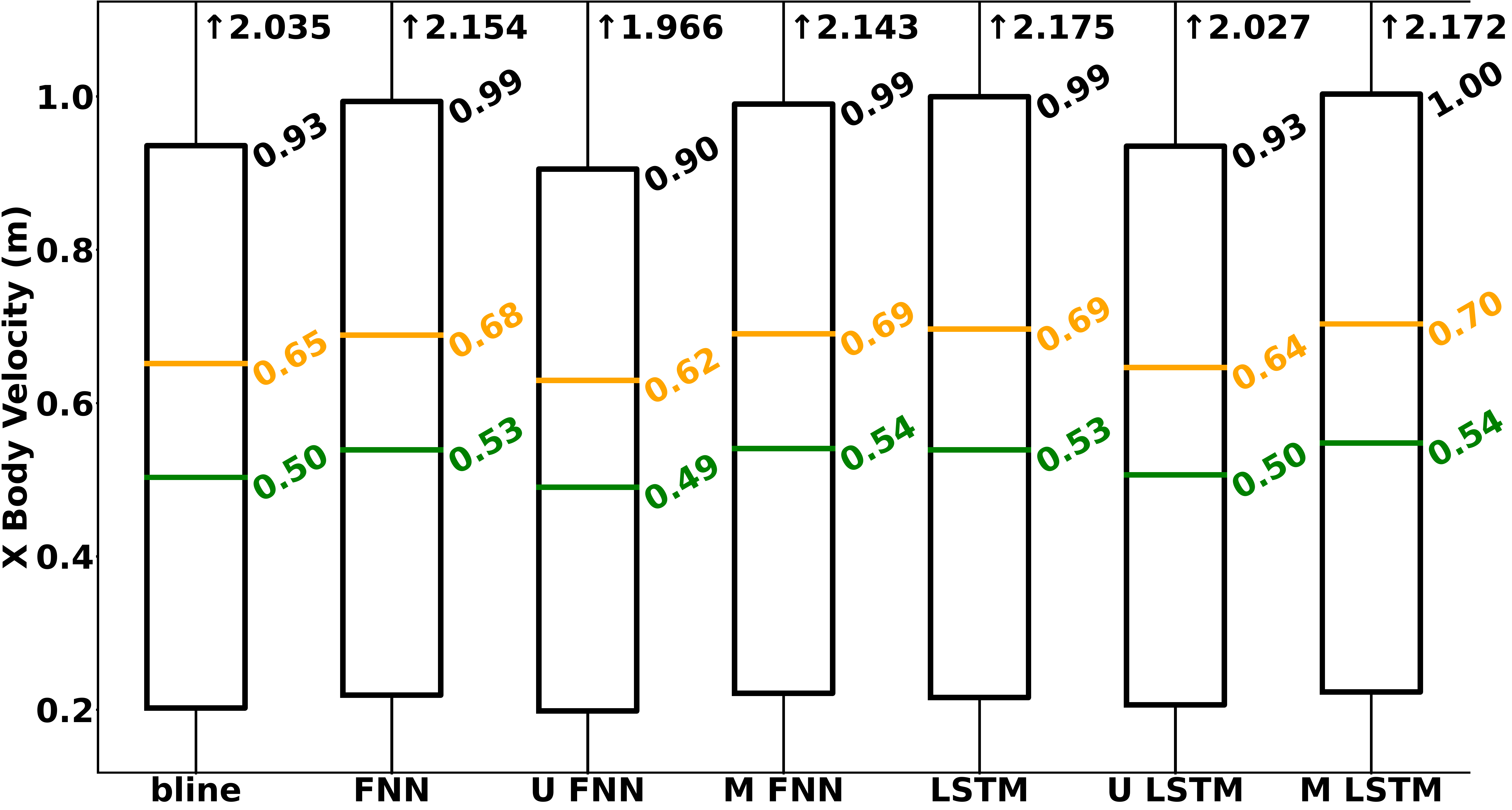}
}
\end{subfloat}
\begin{subfloat}[Initialization \label{fig:init_ablation_vx}]{
\centering
\includegraphics[page=17, trim={0cm, 0cm, 0cm, 0cm}, clip, width=0.48\textwidth]{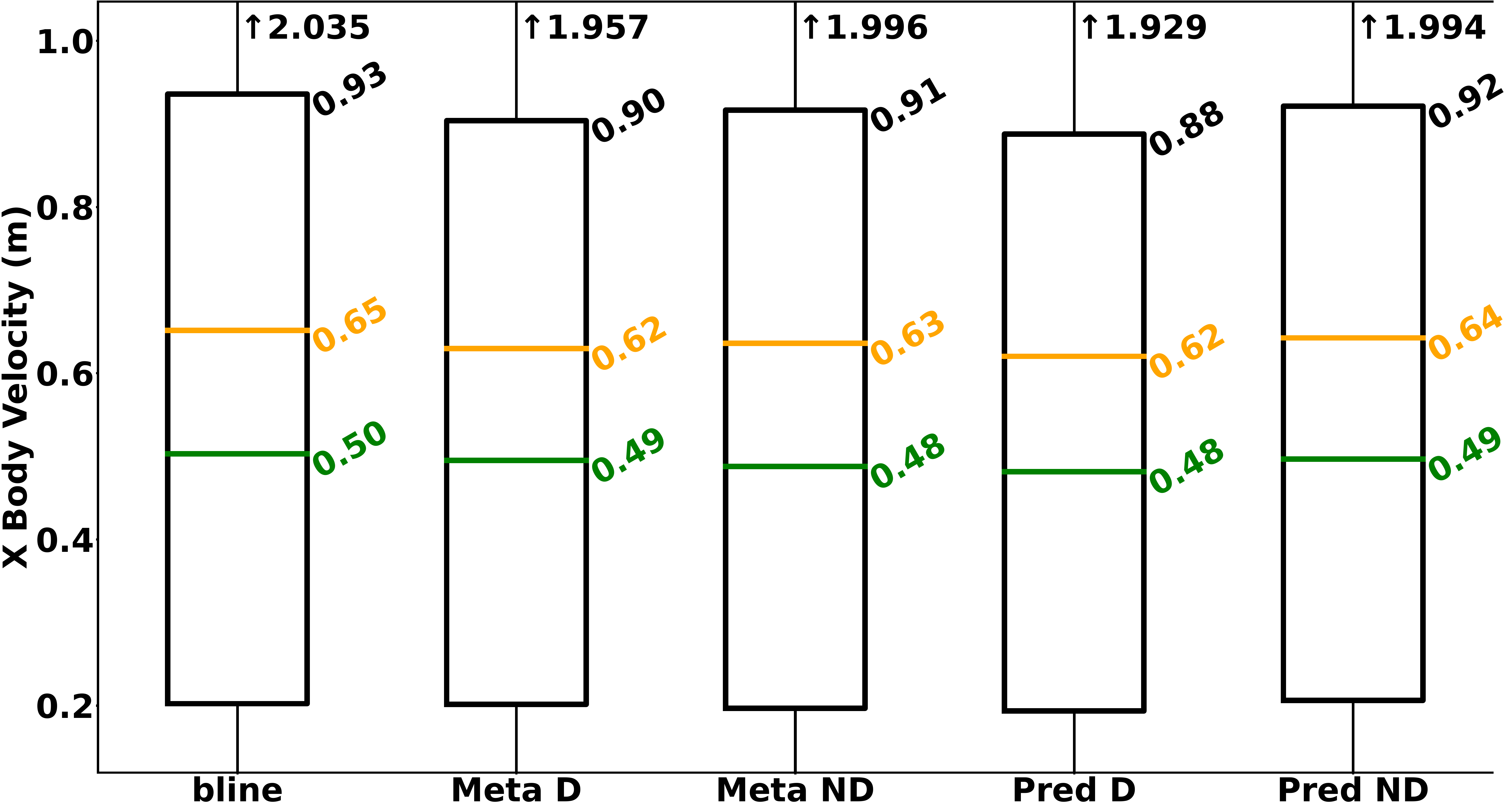}   
}
\end{subfloat}
\begin{subfloat}[Loss Structure \label{fig:structure_loss_vx}]{
\centering
\includegraphics[page=17, trim={0cm, 0cm, 0cm, 0cm}, clip, width=0.48\textwidth]{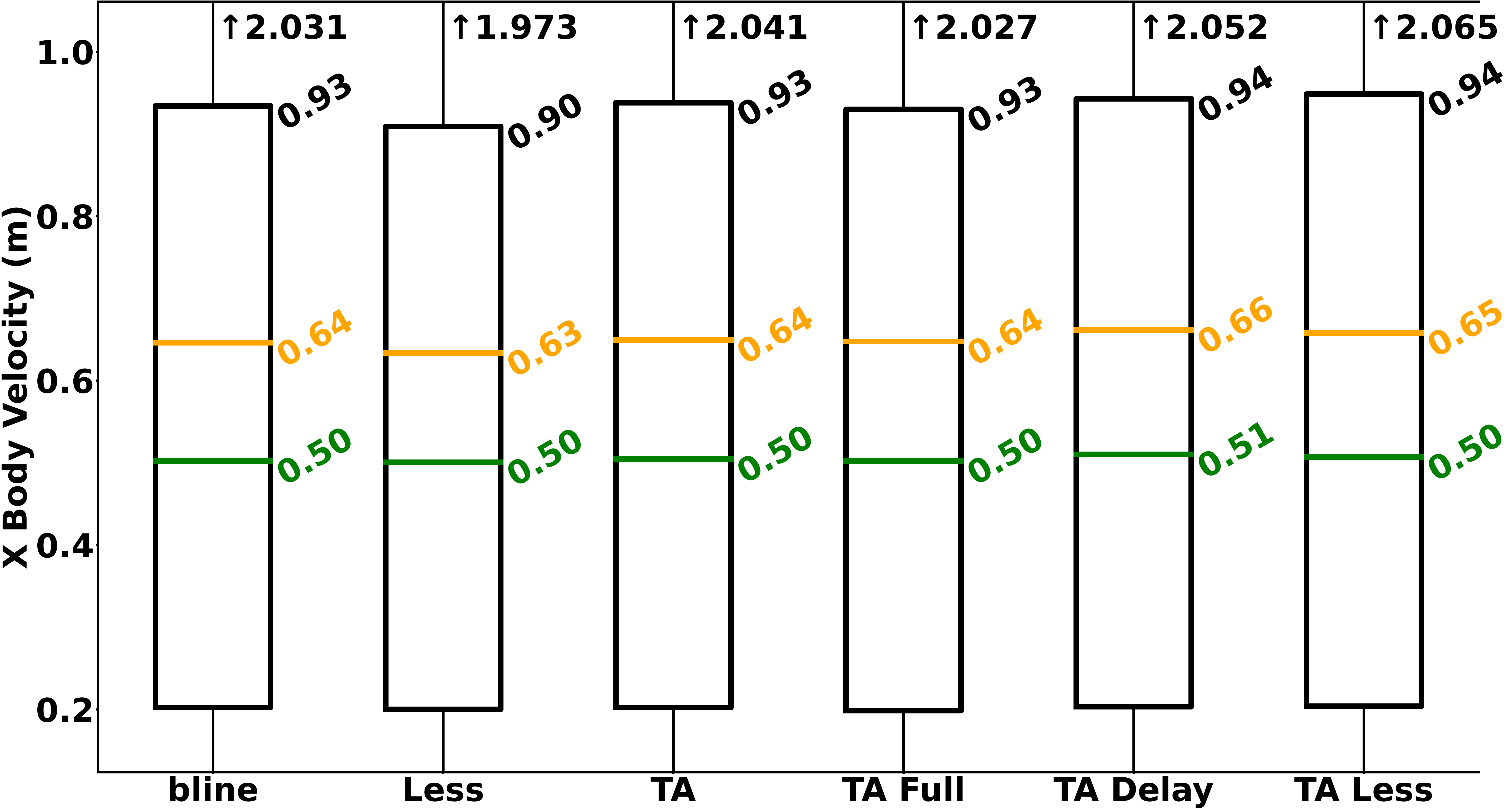}   
}
\end{subfloat}
\begin{subfloat}[Buffer History ($\tau$) \label{fig:tau_ablation_vx}]{
\centering
\includegraphics[page=17, trim={0cm, 0cm, 0cm, 0cm}, clip, width=0.48\textwidth]{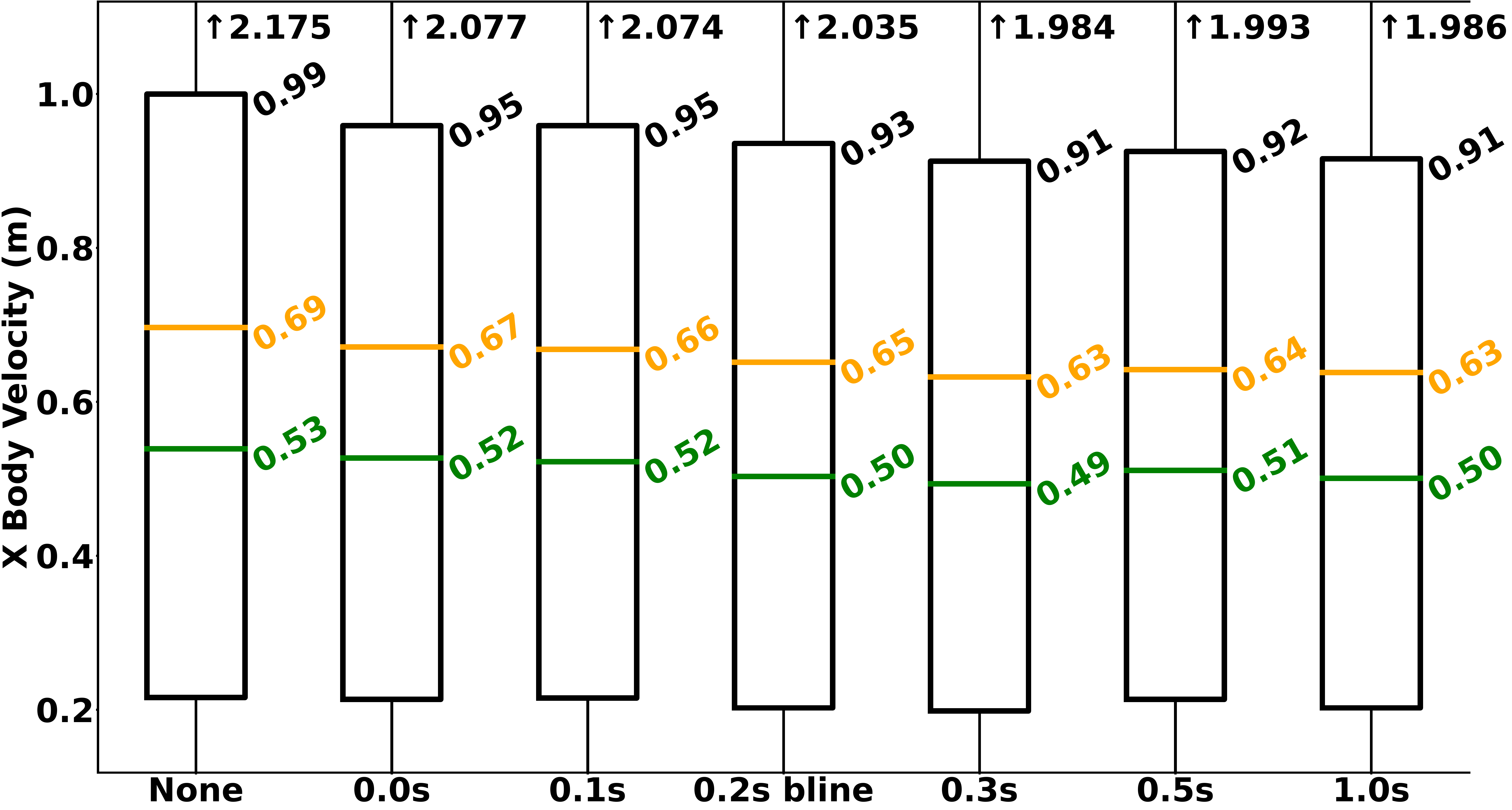} 
}
\end{subfloat}
\begin{subfloat}[Final Comparison \label{fig:final_comparison_vx}]{
\centering
\includegraphics[page=17, trim={0cm, 0cm, 0cm, 0cm}, clip, width=0.48\textwidth]{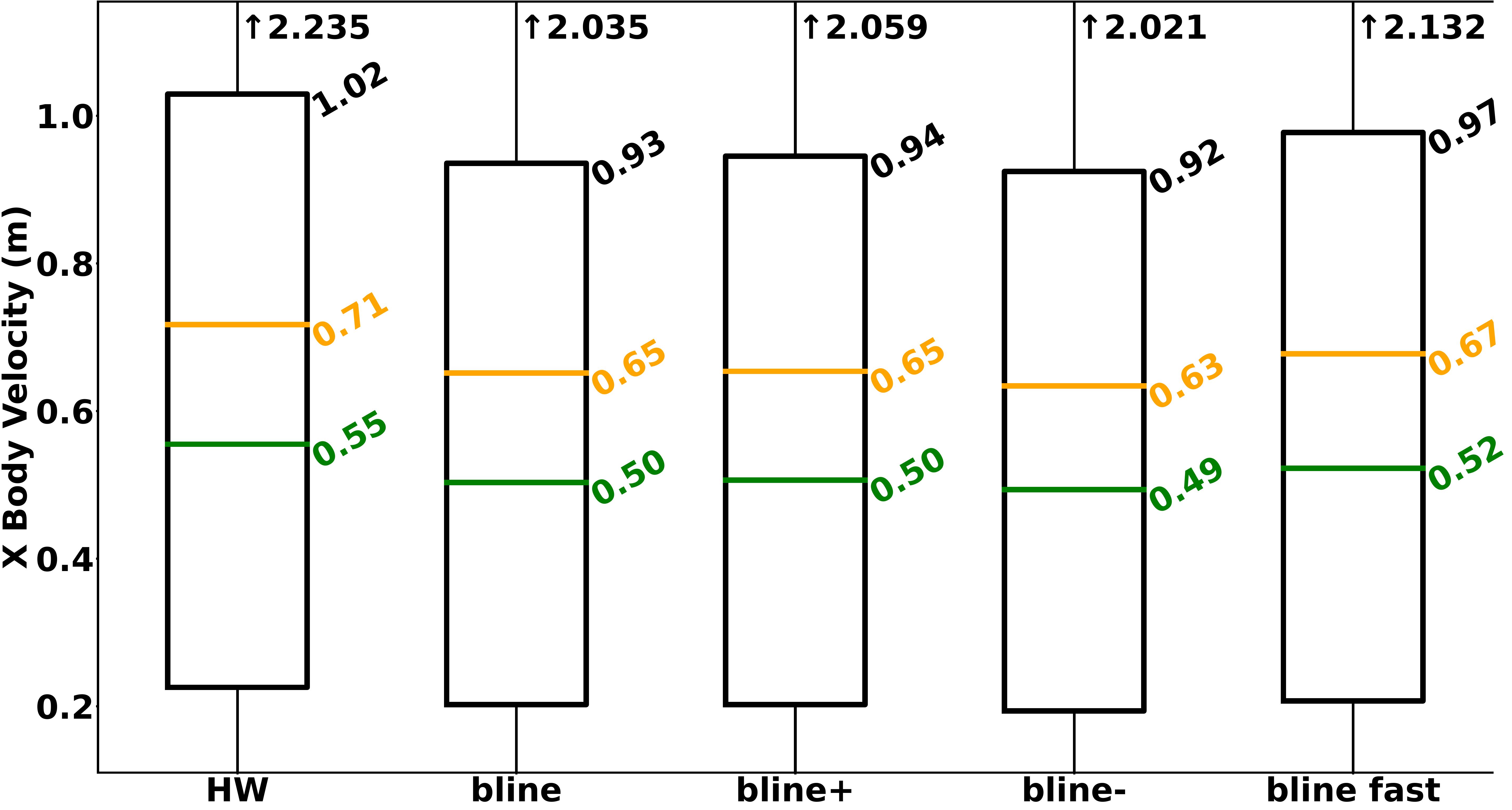}  
}
\end{subfloat}
\caption{
Box plots showing $v^x$ error at the end of the $5s$ trajectory for various structural changes to the networks. 
Whiskers are defined as $\pm 1.5 IQR$ and is given with the arrows, the green line defines the median and the orange the mean value. 
The baseline model (bline) is kept consistent.
The explanations of the keys used in shown in the text \cref{subsubsec:structure_ablation}.
}
\label{fig:structure_ablation_results_vx}
\end{figure*}

\begin{figure*}[htbp]
\centering
\begin{subfloat}[Structure Changes 1\label{fig:structure_bad_vy}]{
\centering
\includegraphics[page=19, trim={0cm, 0cm, 0cm, 0cm}, clip, width=0.48\textwidth]{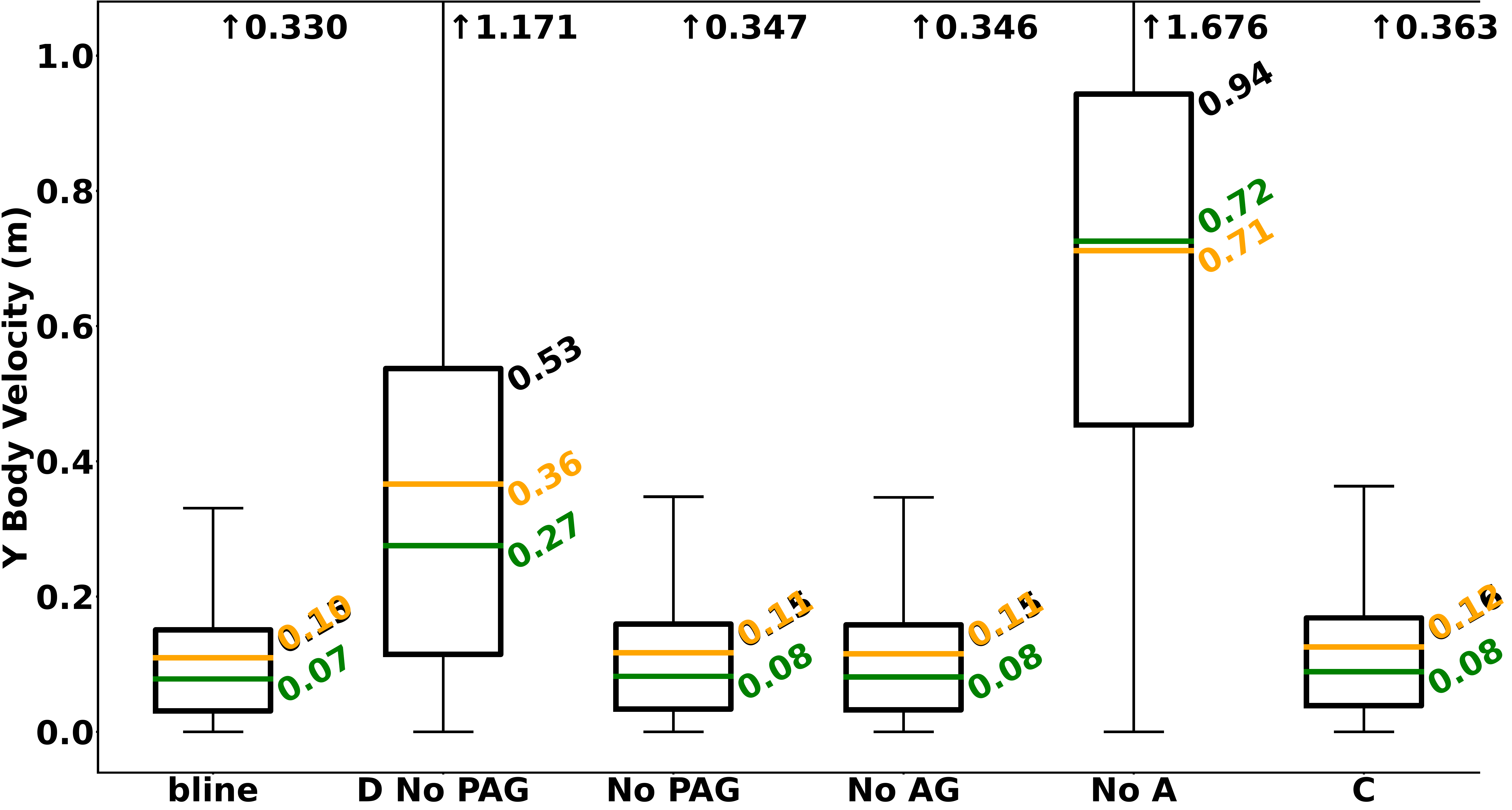}  
}
\end{subfloat}
\begin{subfloat}[Structure Changes 2\label{fig:structure_good_vy}]{
\centering
\includegraphics[page=19, trim={0cm, 0cm, 0cm, 0cm}, clip, width=0.48\textwidth]{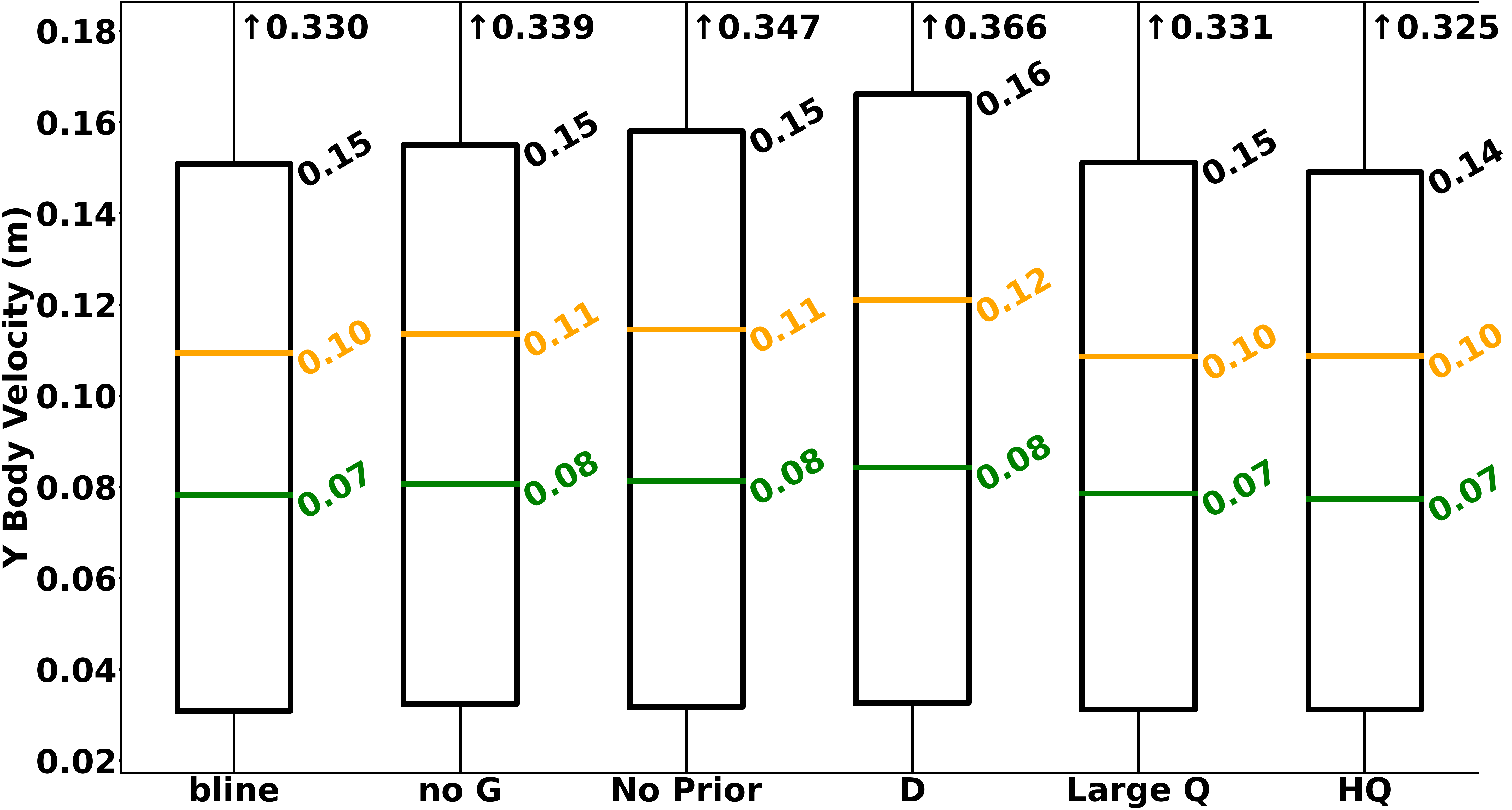}  
}
\end{subfloat}
\begin{subfloat}[Structure Changes 3 \label{fig:structure_weird_vy}]{
\centering
\includegraphics[page=19, trim={0cm, 0cm, 0cm, 0cm}, clip, width=0.48\textwidth]{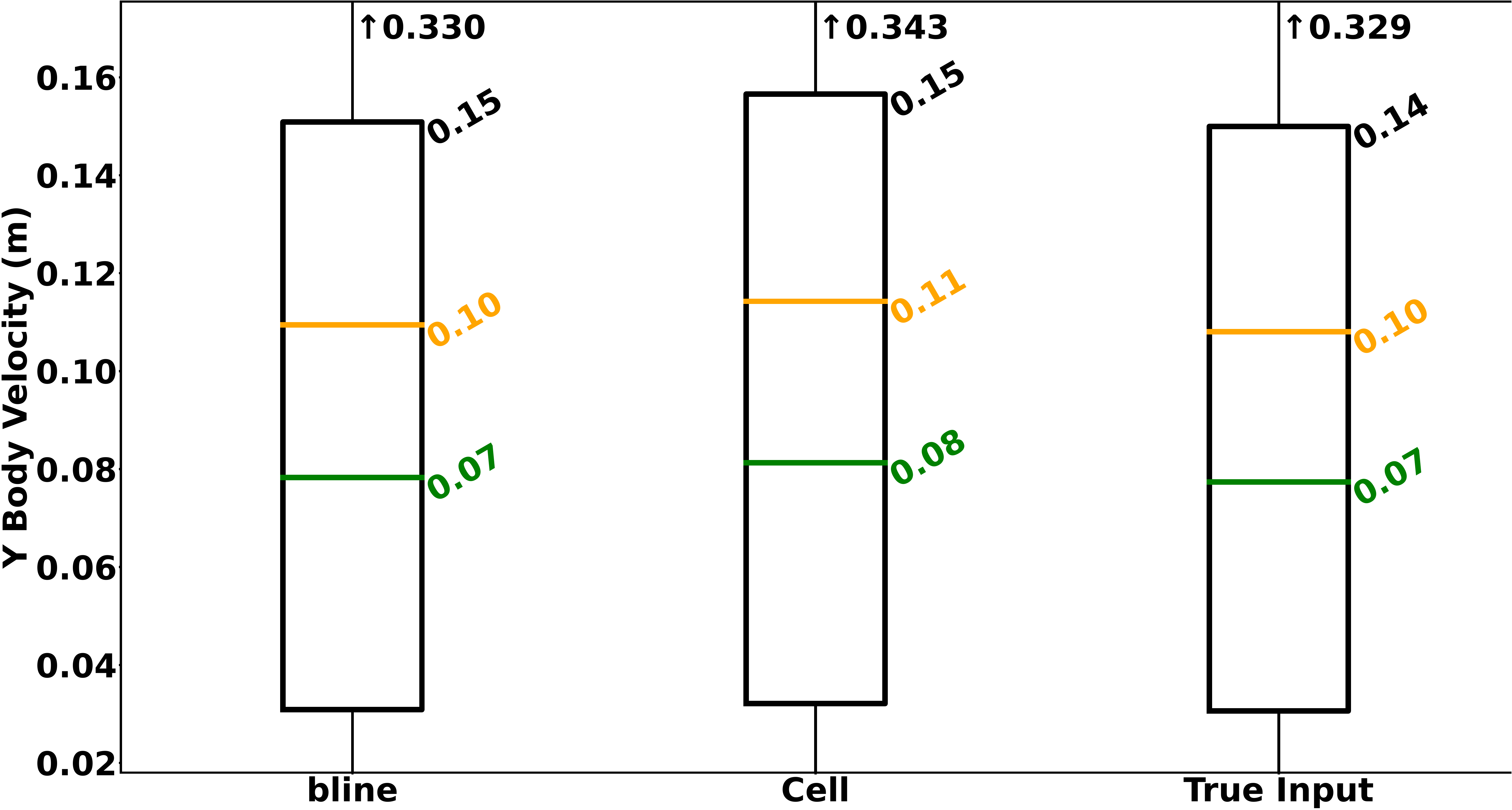}   
}
\end{subfloat}
\begin{subfloat}[Architecture \label{fig:architecture_vy}]{
\centering
\includegraphics[page=19, trim={0cm, 0cm, 0cm, 0cm}, clip, width=0.48\textwidth]{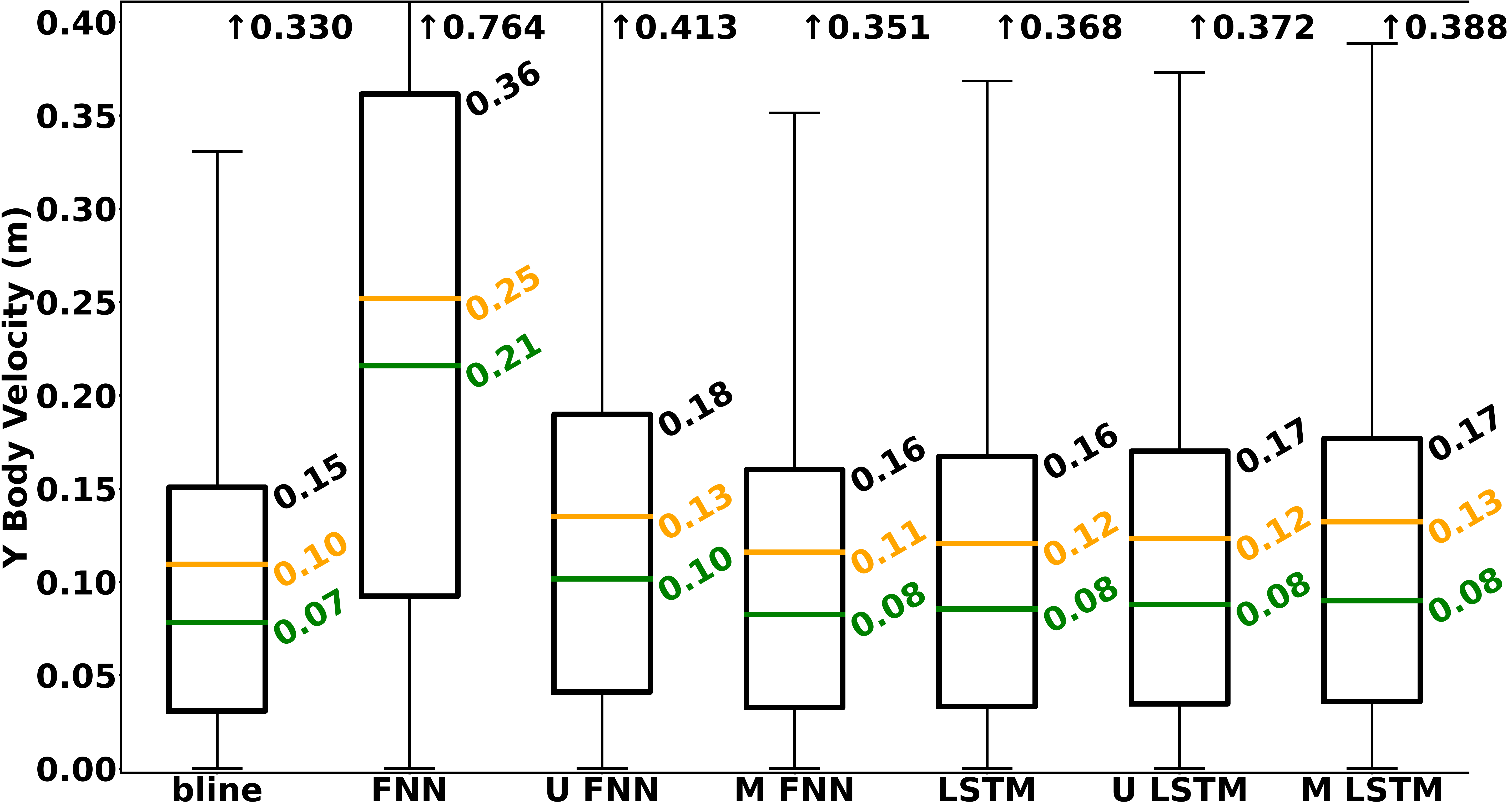}
}
\end{subfloat}
\begin{subfloat}[Initialization \label{fig:init_ablation_vy}]{
\centering
\includegraphics[page=19, trim={0cm, 0cm, 0cm, 0cm}, clip, width=0.48\textwidth]{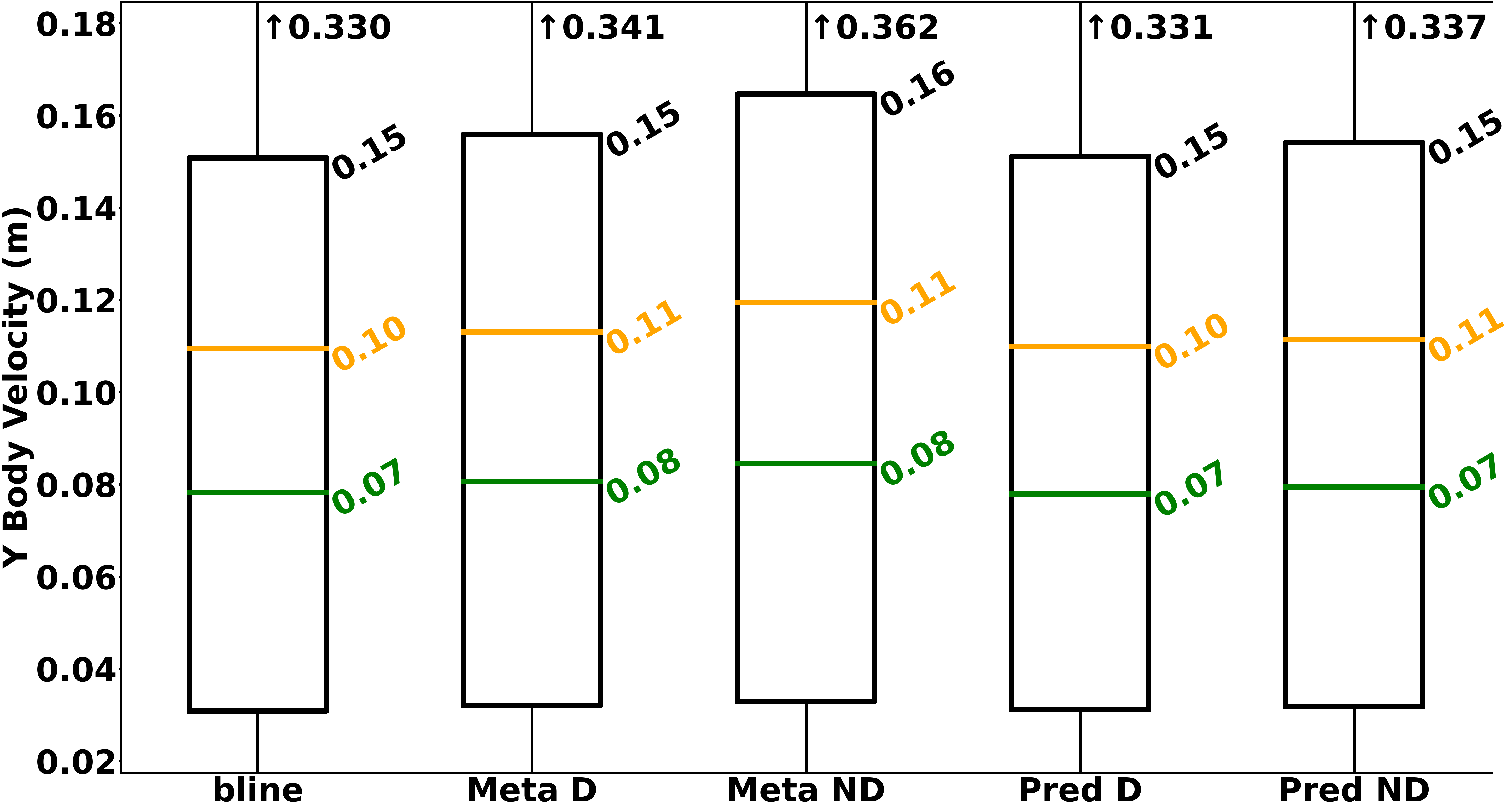}   
}
\end{subfloat}
\begin{subfloat}[Loss Structure \label{fig:structure_loss_vy}]{
\centering
\includegraphics[page=19, trim={0cm, 0cm, 0cm, 0cm}, clip, width=0.48\textwidth]{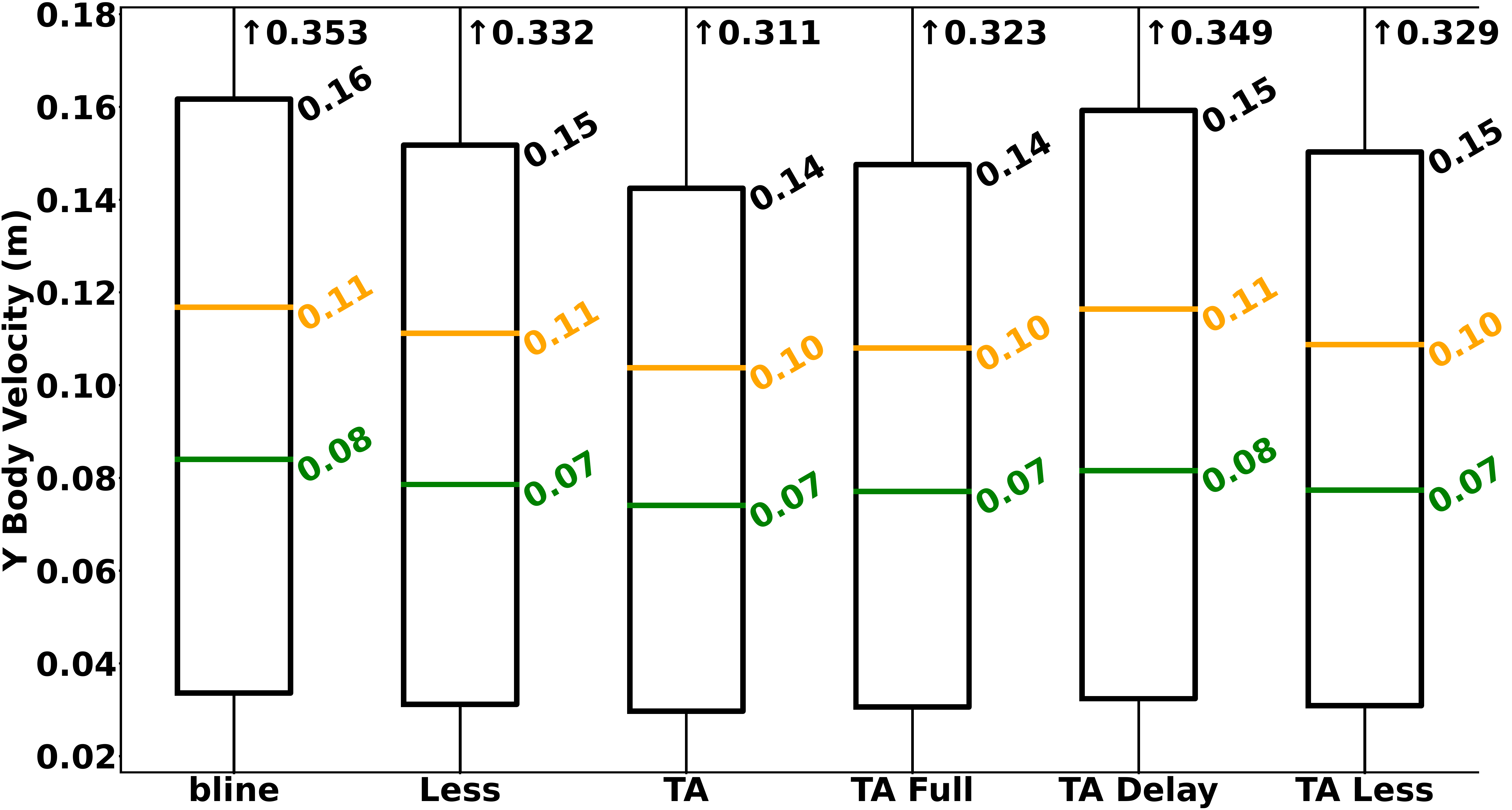}   
}
\end{subfloat}
\begin{subfloat}[Buffer History ($\tau$) \label{fig:tau_ablation_vy}]{
\centering
\includegraphics[page=19, trim={0cm, 0cm, 0cm, 0cm}, clip, width=0.48\textwidth]{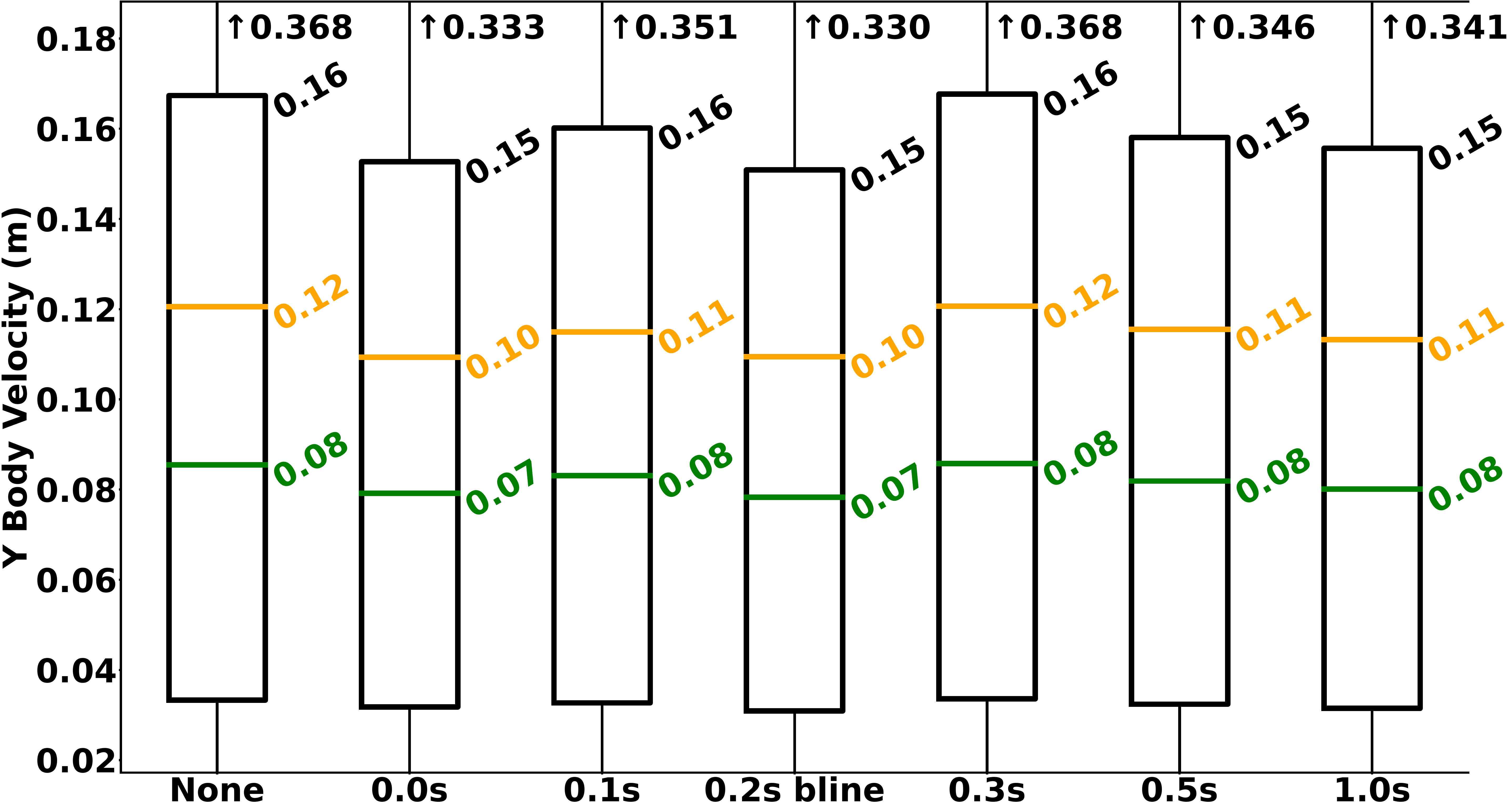} 
}
\end{subfloat}
\begin{subfloat}[Final Comparison \label{fig:final_comparison_vy}]{
\centering
\includegraphics[page=19, trim={0cm, 0cm, 0cm, 0cm}, clip, width=0.48\textwidth]{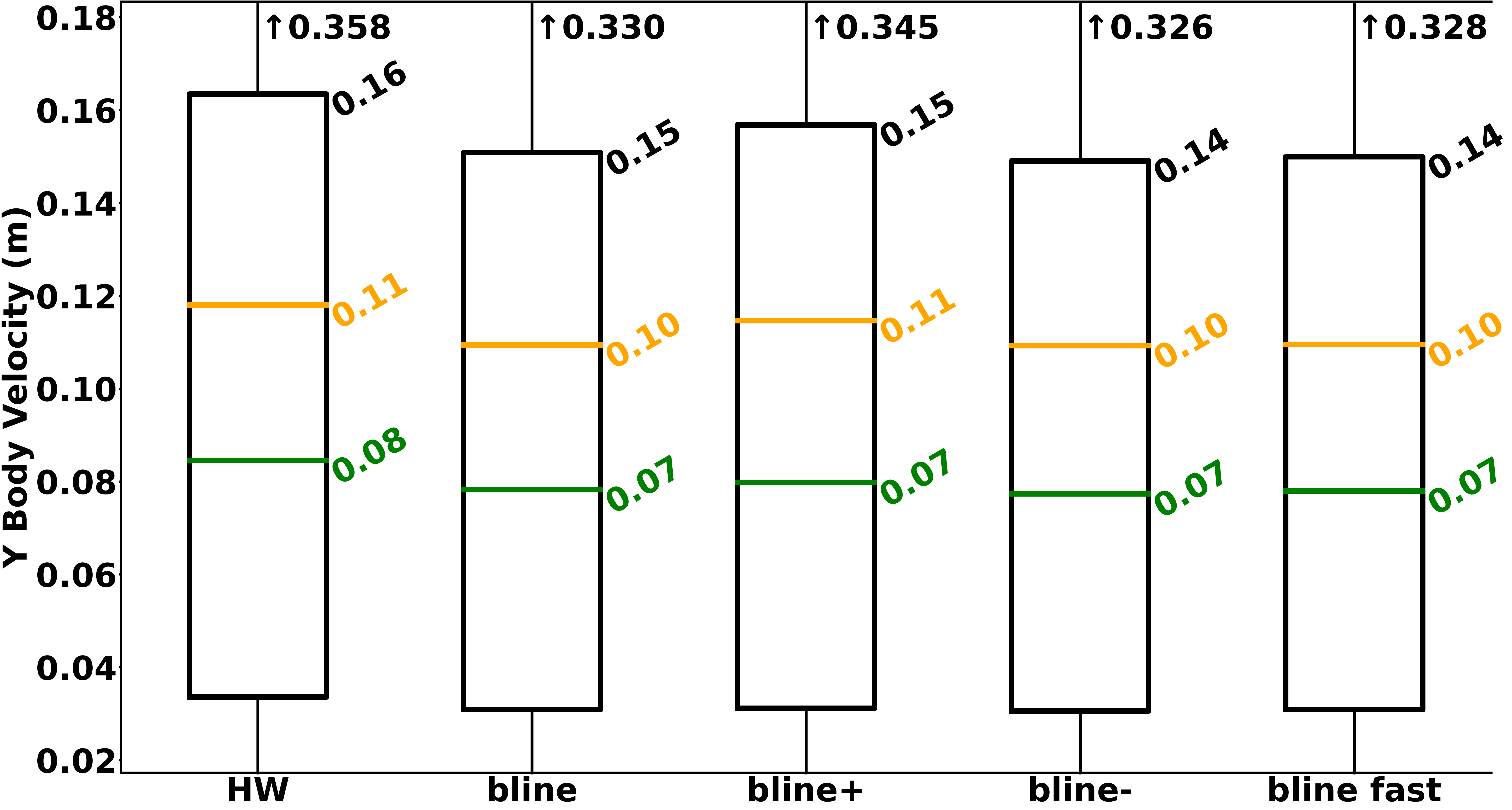}  
}
\end{subfloat}
\caption{
Box plots showing $v^y$ error at the end of the $5s$ trajectory for various structural changes to the networks. 
Whiskers are defined as $\pm 1.5 IQR$ and is given with the arrows, the green line defines the median and the orange the mean value. 
The baseline model (bline) is kept consistent.
The explanations of the keys used in shown in the text \cref{subsubsec:structure_ablation}.
}
\label{fig:structure_ablation_results_vy}
\end{figure*}

\begin{figure*}[htbp]
\centering
\begin{subfloat}[Structure Changes 1\label{fig:structure_bad_yaw_rate}]{
\centering
\includegraphics[page=21, trim={0cm, 0cm, 0cm, 0cm}, clip, width=0.48\textwidth]{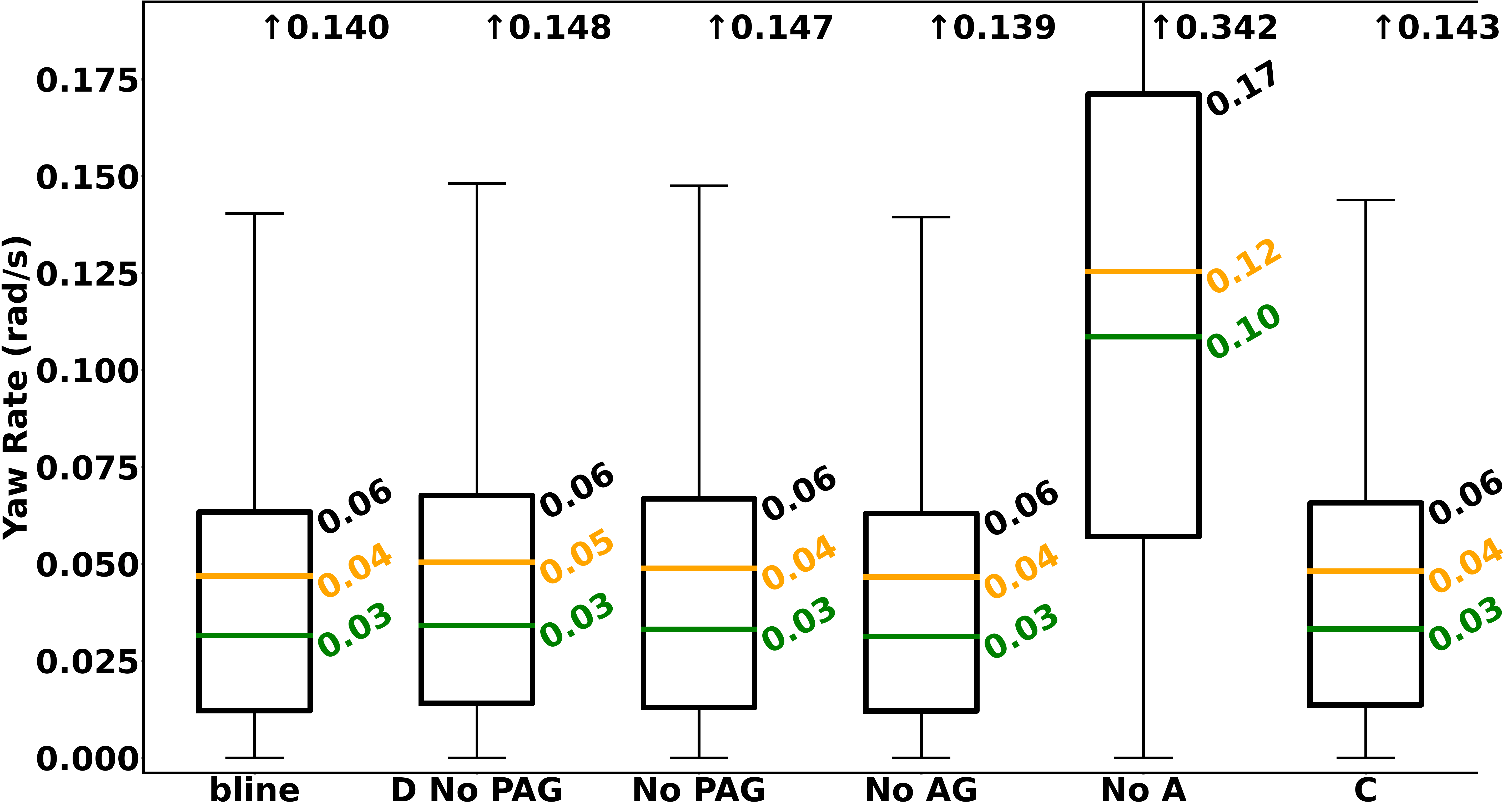}  
}
\end{subfloat}
\begin{subfloat}[Structure Changes 2\label{fig:structure_good_yaw_rate}]{
\centering
\includegraphics[page=21, trim={0cm, 0cm, 0cm, 0cm}, clip, width=0.48\textwidth]{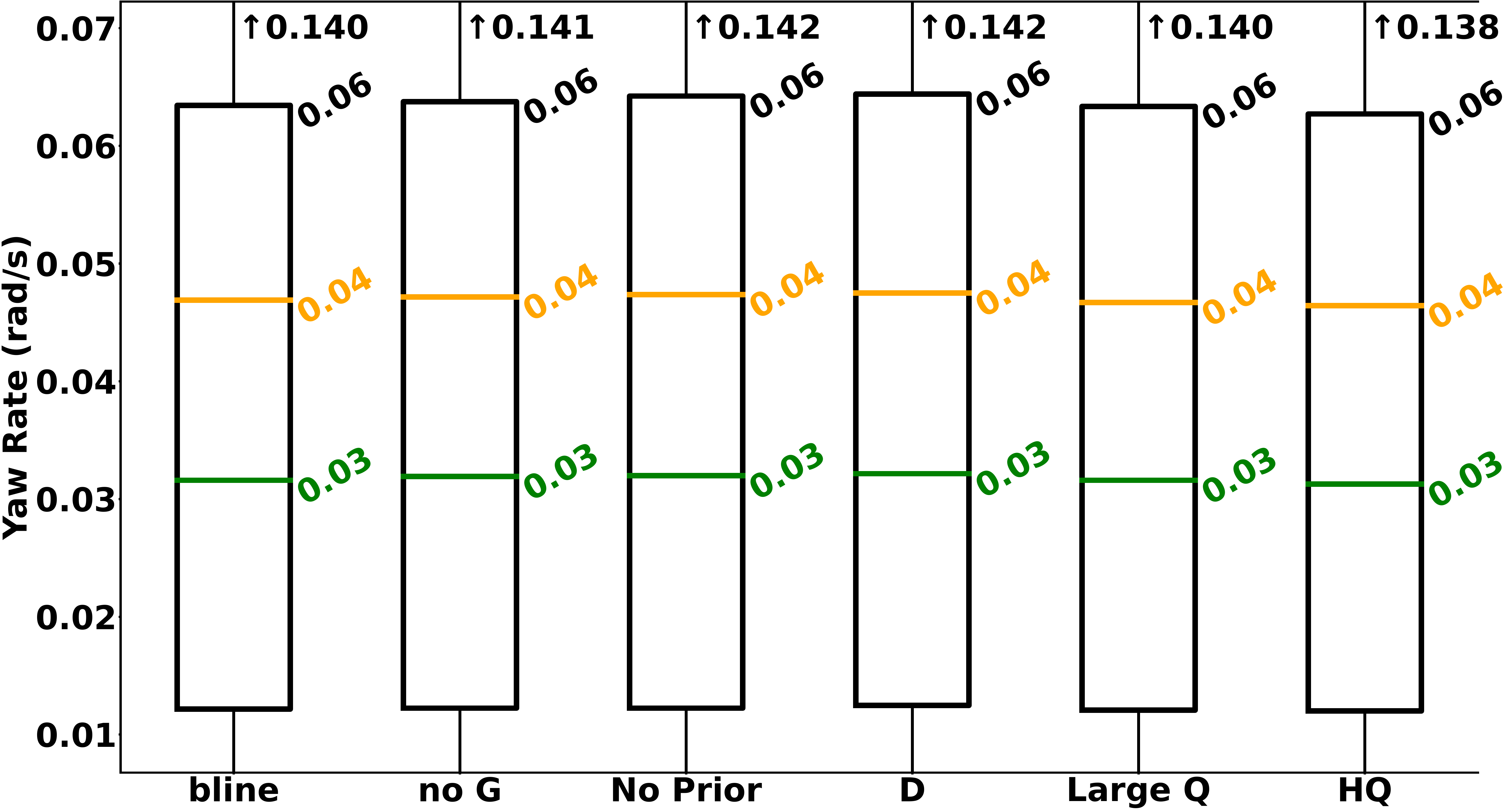}  
}
\end{subfloat}
\begin{subfloat}[Structure Changes 3 \label{fig:structure_weird_yaw_rate}]{
\centering
\includegraphics[page=21, trim={0cm, 0cm, 0cm, 0cm}, clip, width=0.48\textwidth]{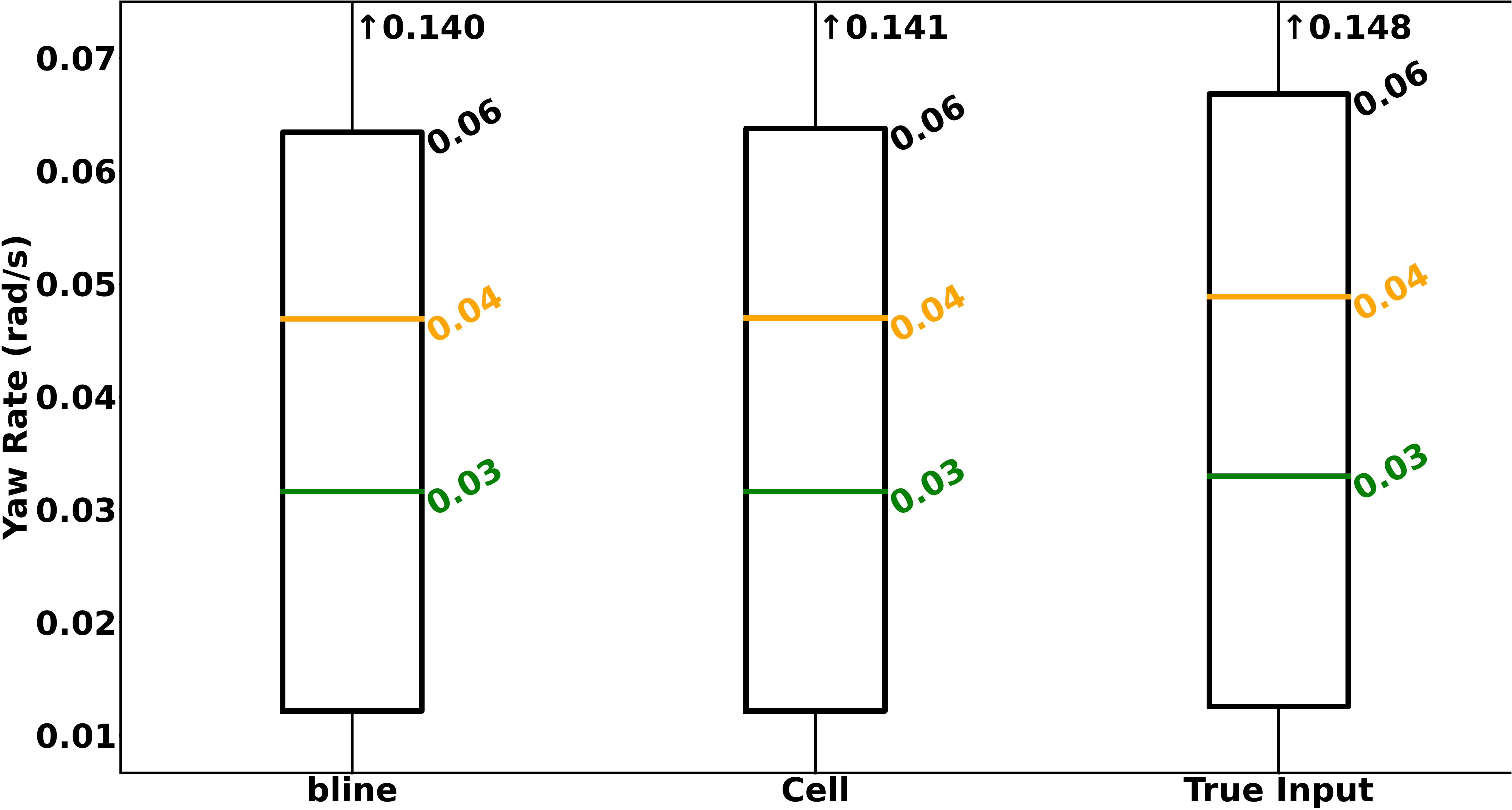}   
}
\end{subfloat}
\begin{subfloat}[Architecture \label{fig:architecture_yaw_rate}]{
\centering
\includegraphics[page=21, trim={0cm, 0cm, 0cm, 0cm}, clip, width=0.48\textwidth]{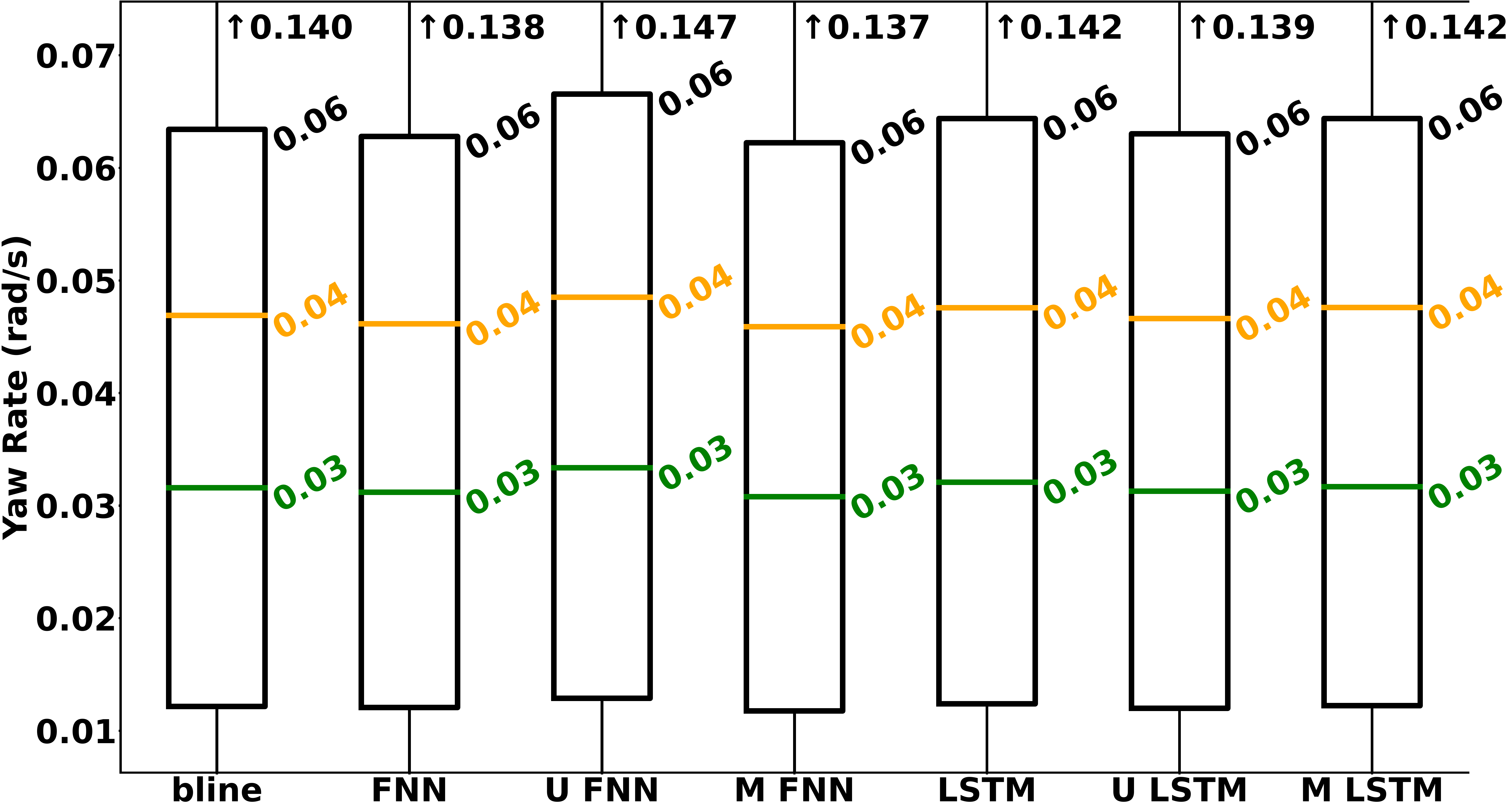}
}
\end{subfloat}
\begin{subfloat}[Initialization \label{fig:init_ablation_yaw_rate}]{
\centering
\includegraphics[page=21, trim={0cm, 0cm, 0cm, 0cm}, clip, width=0.48\textwidth]{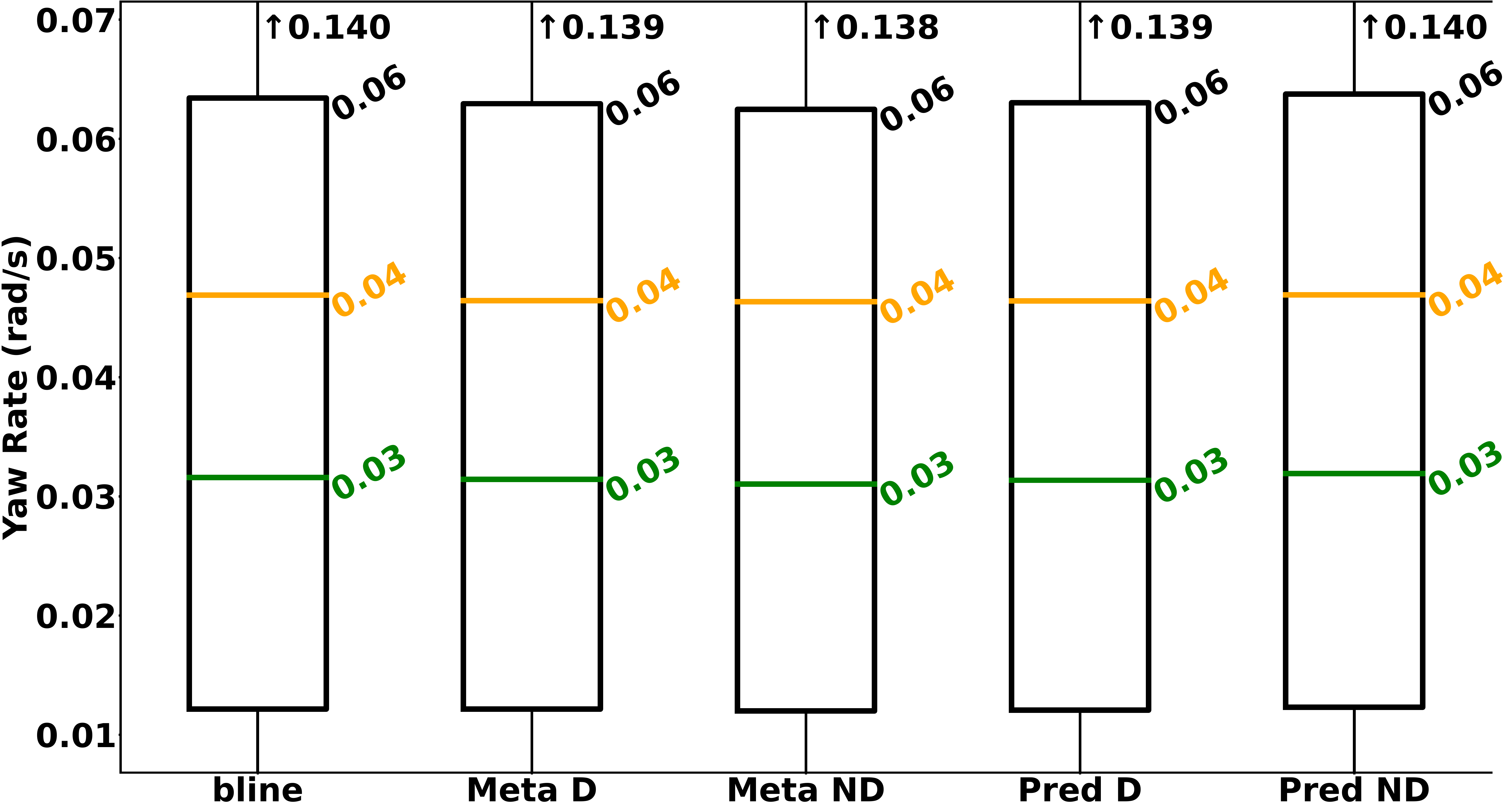}   
}
\end{subfloat}
\begin{subfloat}[Loss Structure \label{fig:structure_loss_yaw_rate}]{
\centering
\includegraphics[page=21, trim={0cm, 0cm, 0cm, 0cm}, clip, width=0.48\textwidth]{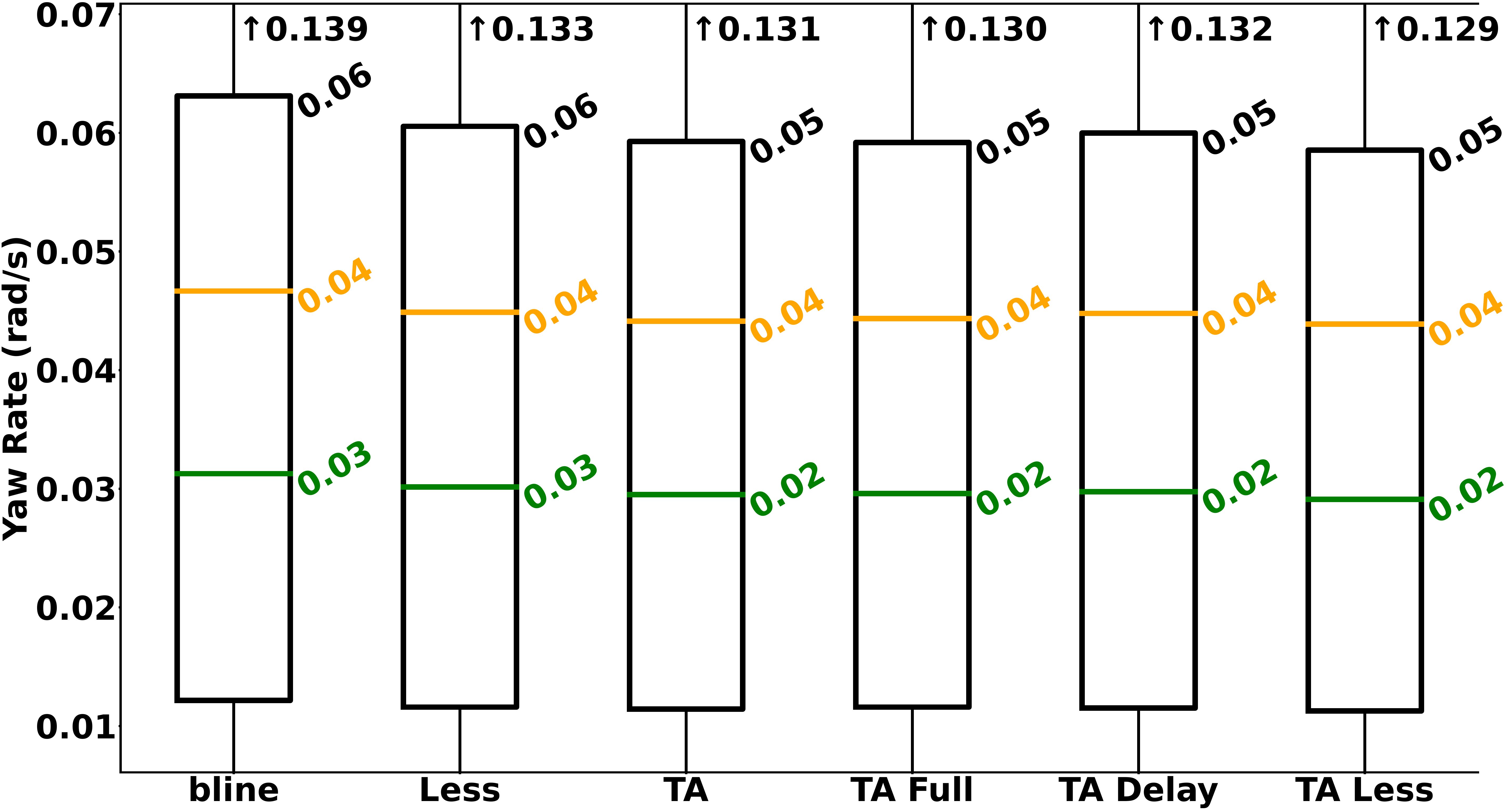}   
}
\end{subfloat}
\begin{subfloat}[Buffer History ($\tau$) \label{fig:tau_ablation_yaw_rate}]{
\centering
\includegraphics[page=21, trim={0cm, 0cm, 0cm, 0cm}, clip, width=0.48\textwidth]{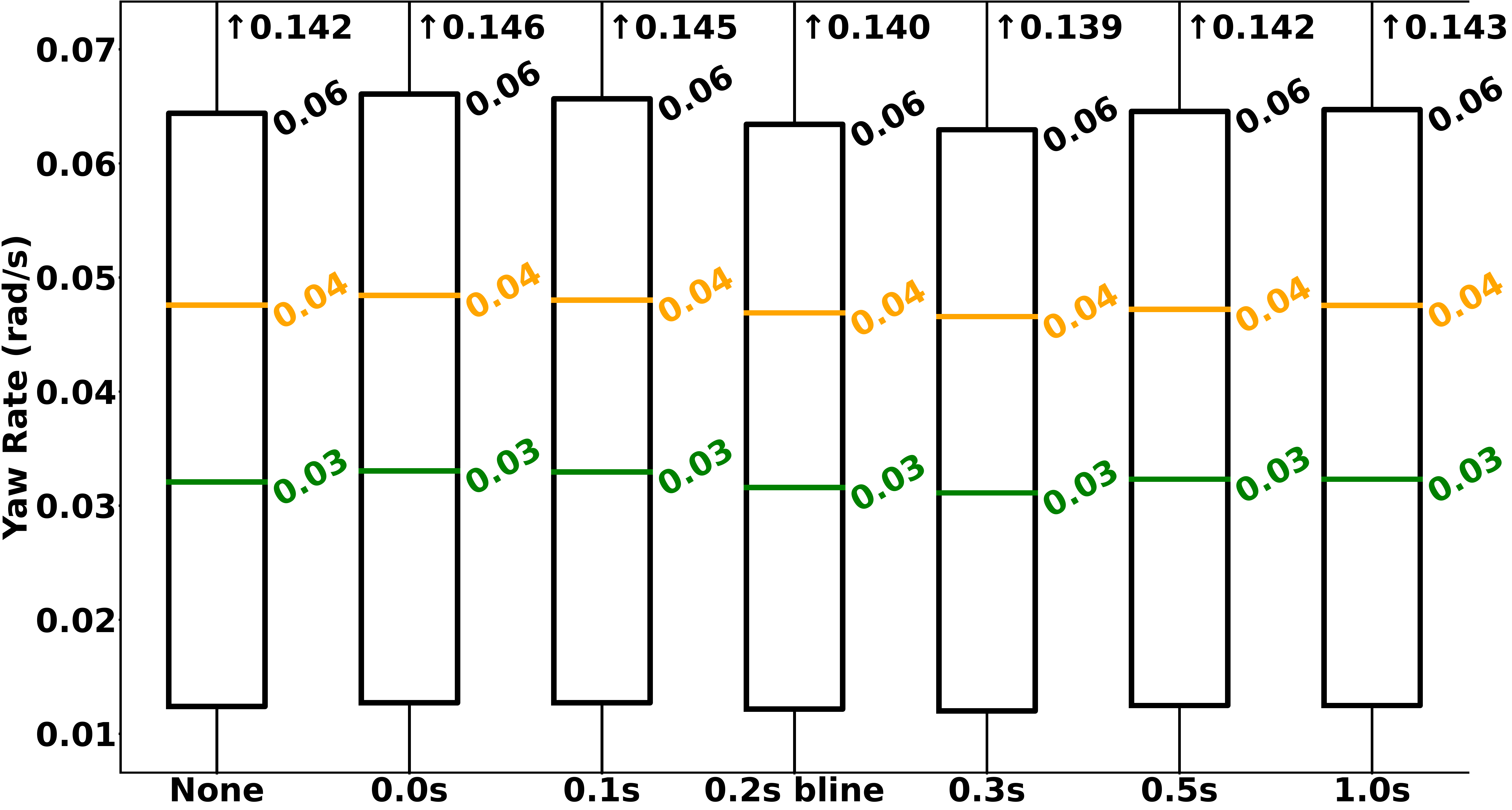} 
}
\end{subfloat}
\begin{subfloat}[Final Comparison \label{fig:final_comparison_yaw_rate}]{
\centering
\includegraphics[page=21, trim={0cm, 0cm, 0cm, 0cm}, clip, width=0.48\textwidth]{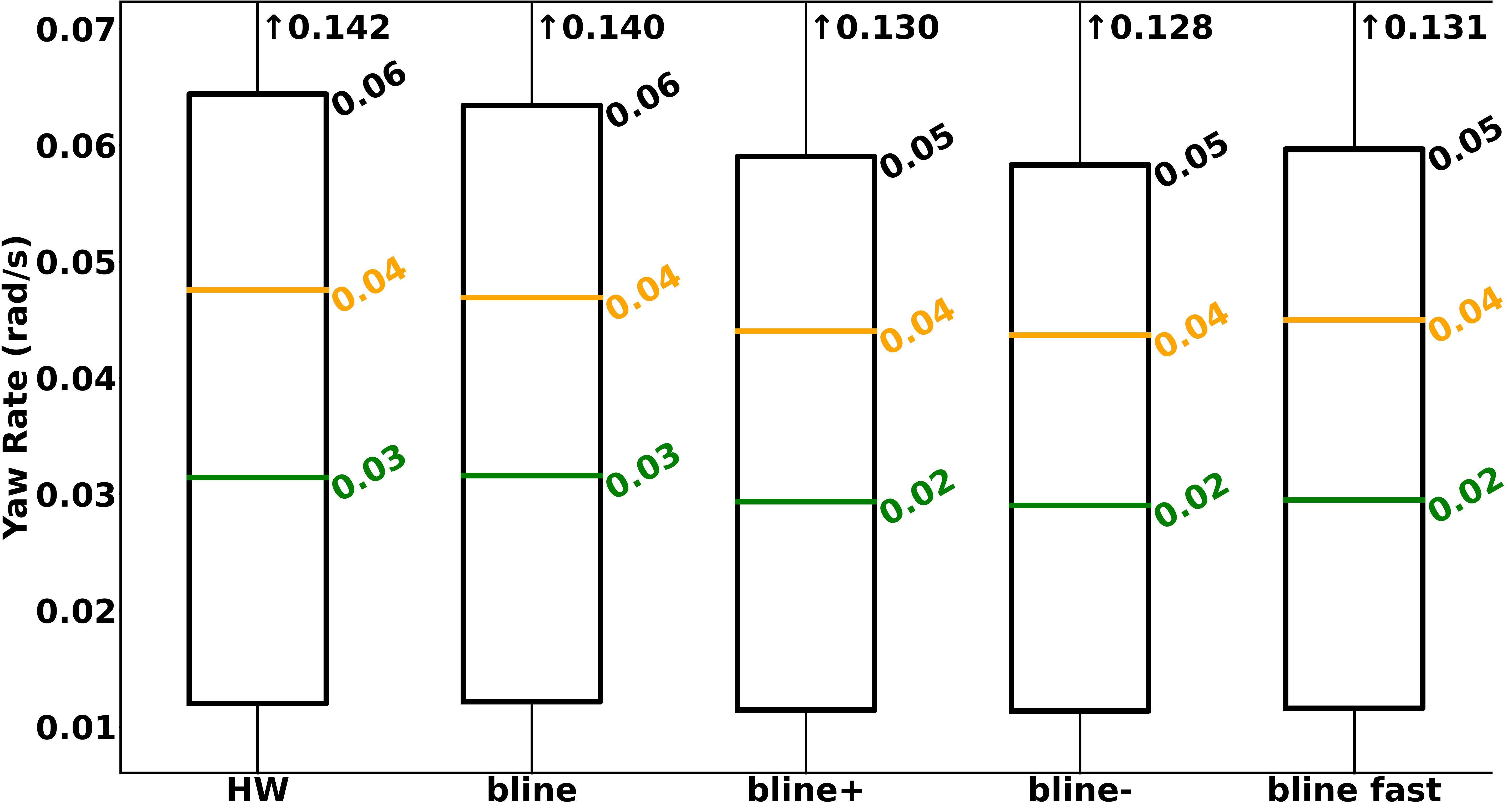}  
}
\end{subfloat}
\caption{
Box plots showing $\dot{\psi}$ error at the end of the $5s$ trajectory for various structural changes to the networks. 
Whiskers are defined as $\pm 1.5 IQR$ and is given with the arrows, the green line defines the median and the orange the mean value. 
The baseline model (bline) is kept consistent.
The explanations of the keys used in shown in the text \cref{subsubsec:structure_ablation}.
}
\label{fig:structure_ablation_results_yaw_rate}
\end{figure*}

\begin{figure*}[htbp]
\centering
\begin{subfloat}[Loss Structure \label{fig:structure_loss_shaft_angle}]{
\centering
\includegraphics[page=25, trim={0cm, 0cm, 0cm, 0cm}, clip, width=0.48\textwidth]{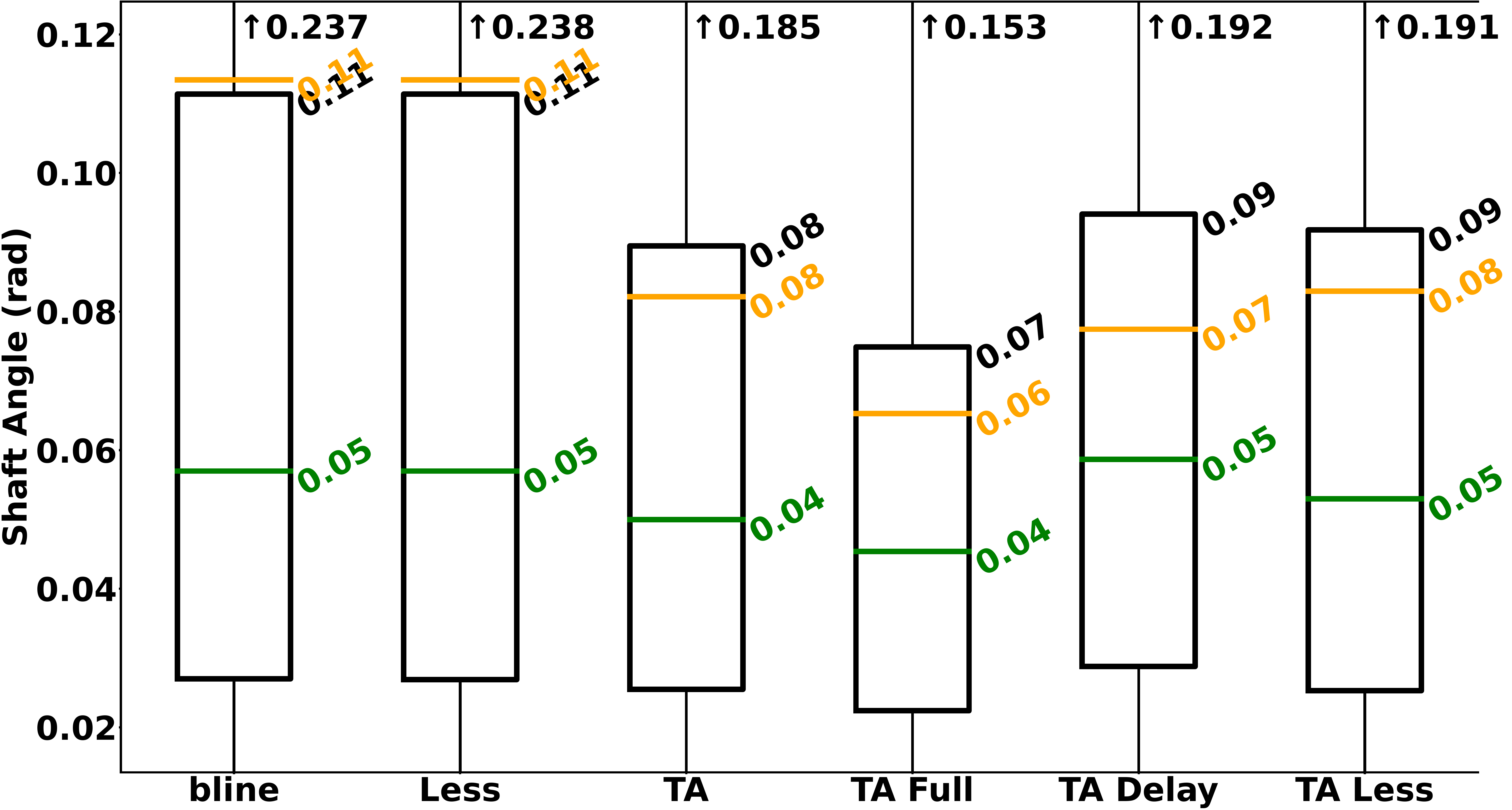}   
}
\end{subfloat}
\begin{subfloat}[Final Comparison \label{fig:final_comparison_shaft_angle}]{
\centering
\includegraphics[page=25, trim={0cm, 0cm, 0cm, 0cm}, clip, width=0.48\textwidth]{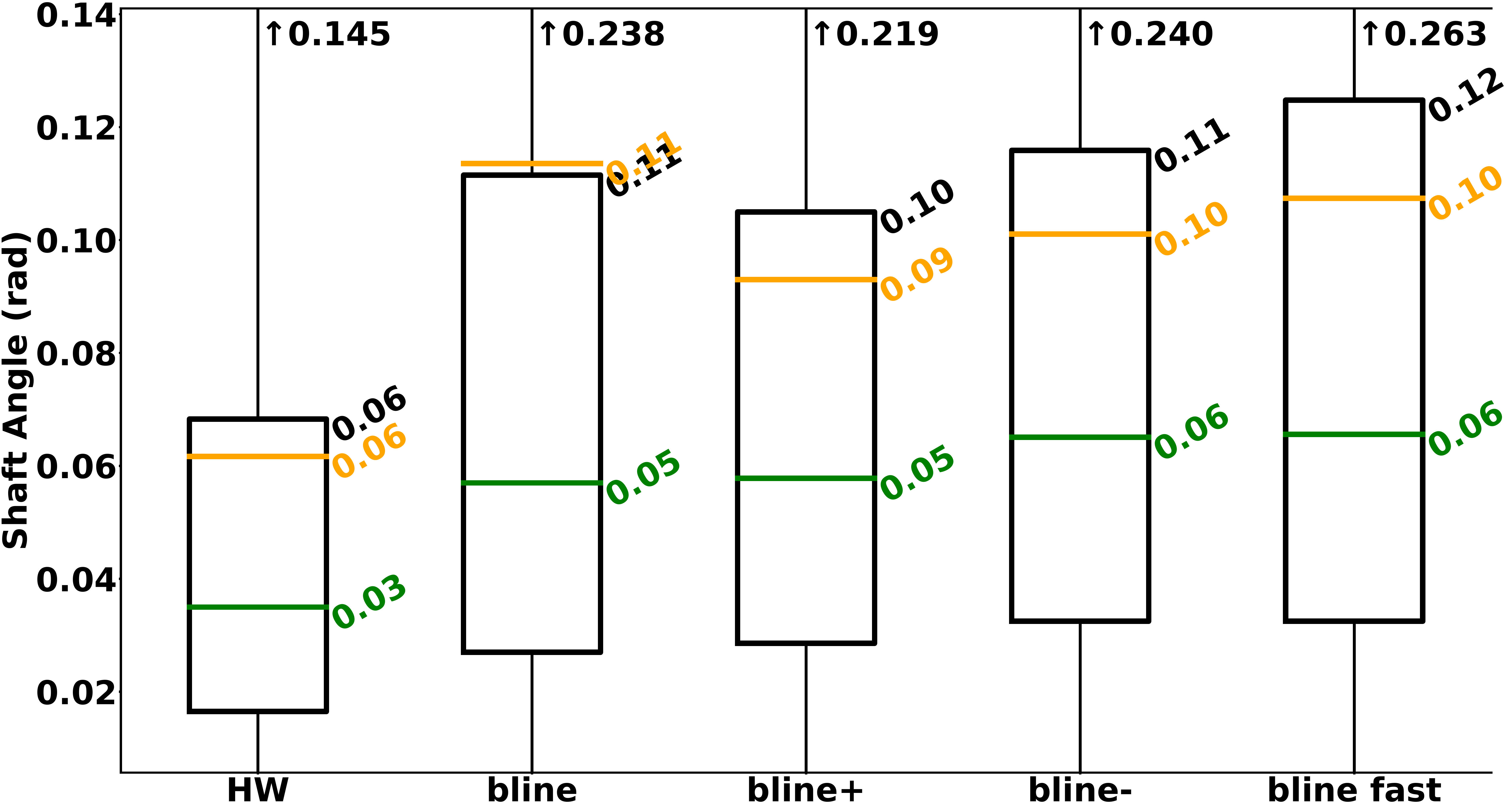}  
}
\end{subfloat}
\caption{
Box plots showing $\delta$ (steering angle, in range $-5, 5$) error at the end of the $5s$ trajectory for various structural changes to the networks. 
Whiskers are defined as $\pm 1.5 IQR$ and is given with the arrows, the green line defines the median and the orange the mean value. 
The baseline model (bline) is kept consistent.
The explanations of the keys used in shown in the text \cref{subsubsec:structure_ablation}.
}
\label{fig:structure_ablation_results_shaft_angle}
\end{figure*}

\begin{figure*}[htbp]
\centering
\begin{subfloat}[Structure Changes 1\label{fig:structure_bad_rpm}]{
\centering
\includegraphics[page=29, trim={0cm, 0cm, 0cm, 0cm}, clip, width=0.48\textwidth]{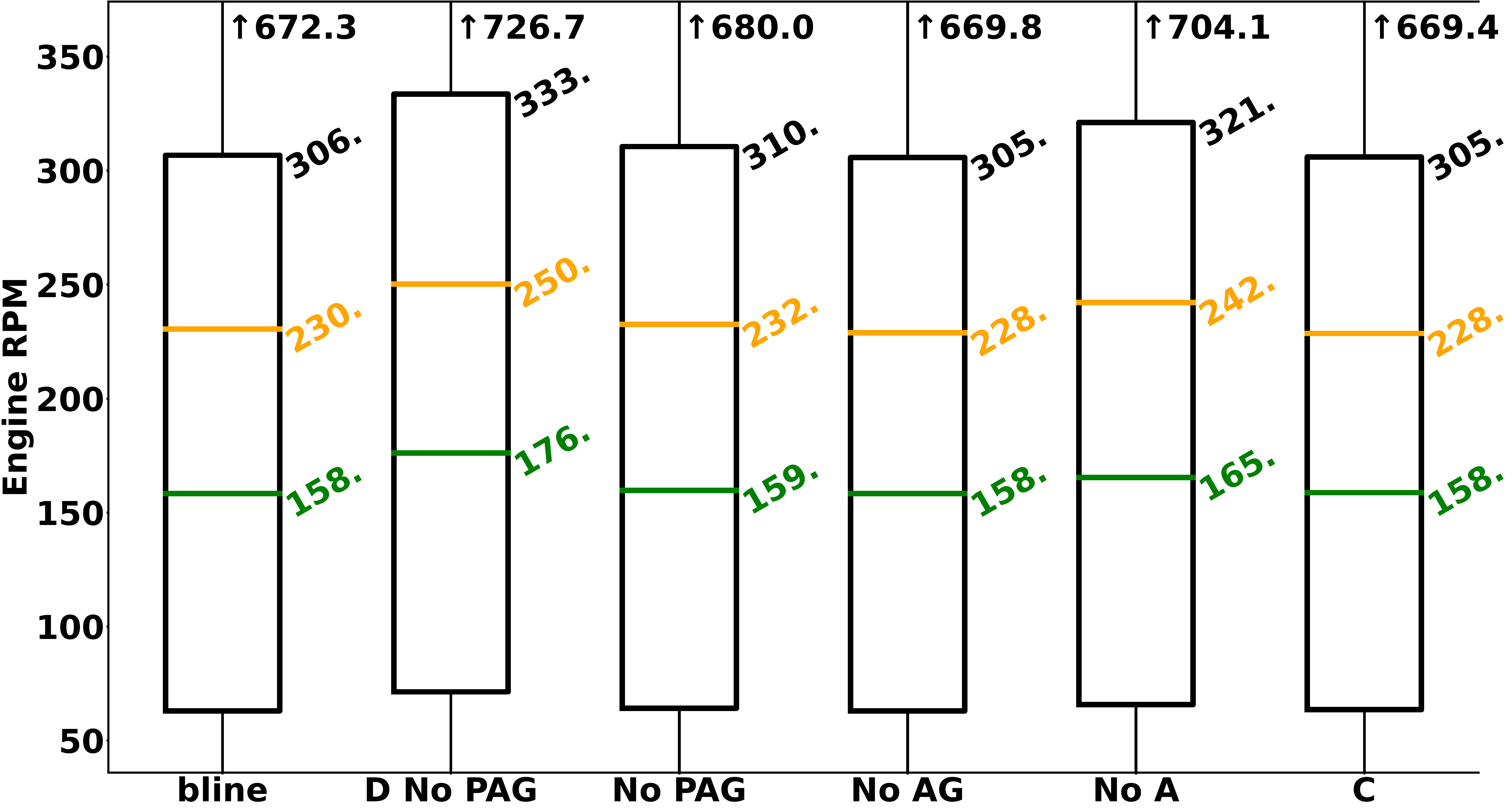}  
}
\end{subfloat}
\begin{subfloat}[Structure Changes 2\label{fig:structure_good_rpm}]{
\centering
\includegraphics[page=29, trim={0cm, 0cm, 0cm, 0cm}, clip, width=0.48\textwidth]{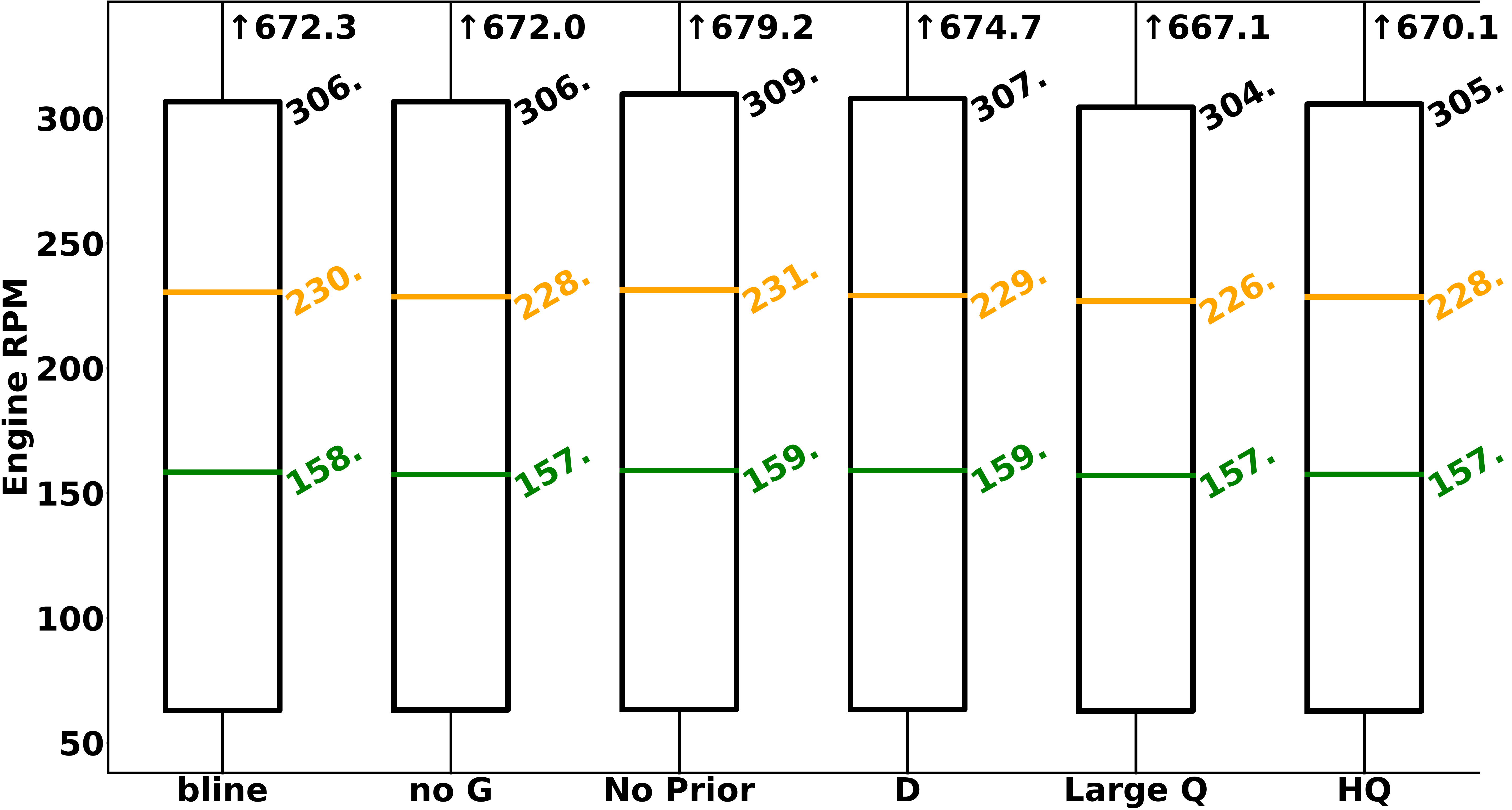}  
}
\end{subfloat}
\begin{subfloat}[Structure Changes 3 \label{fig:structure_weird_rpm}]{
\centering
\includegraphics[page=29, trim={0cm, 0cm, 0cm, 0cm}, clip, width=0.48\textwidth]{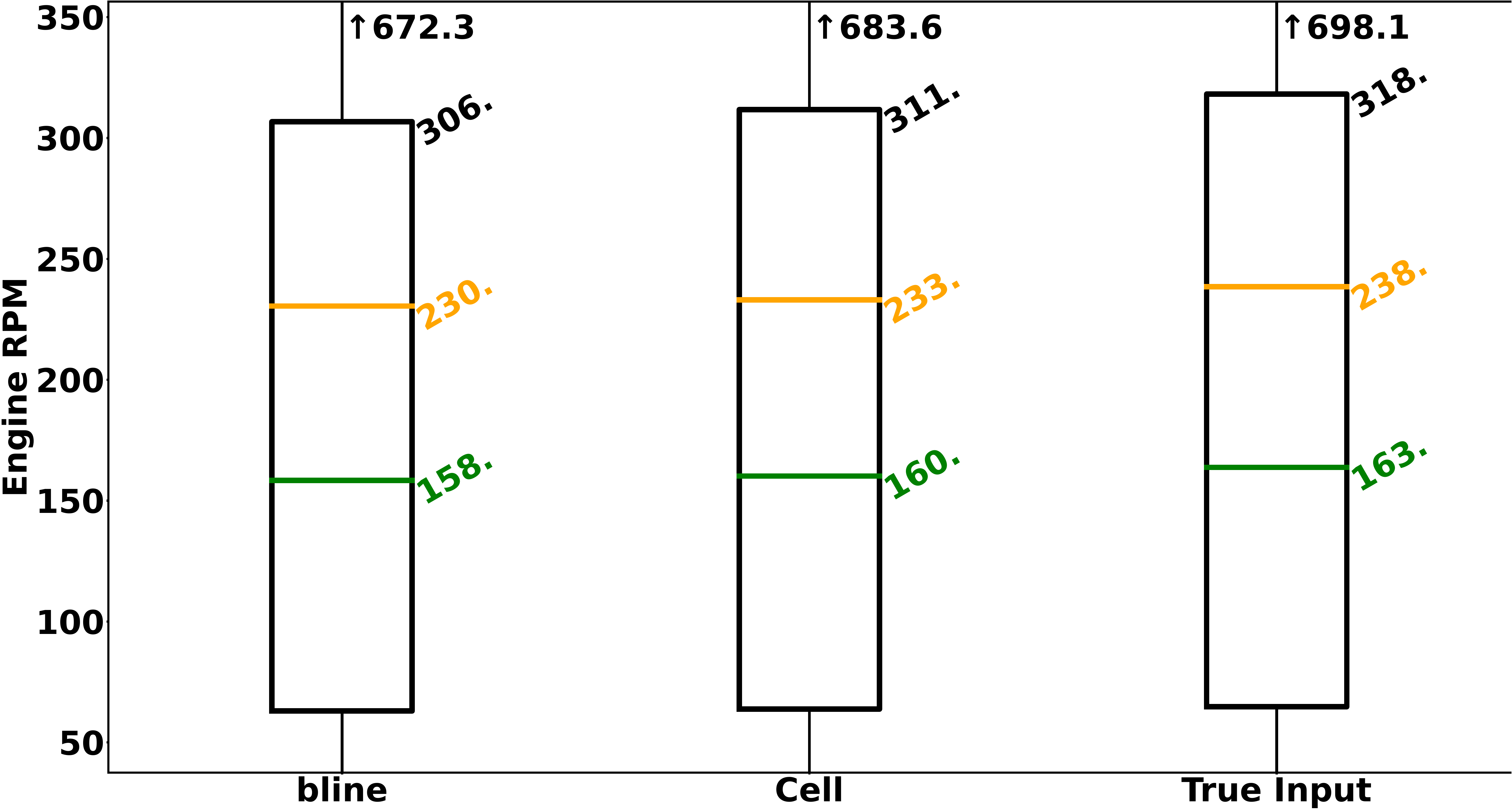}   
}
\end{subfloat}
\begin{subfloat}[Architecture \label{fig:architecture_rpm}]{
\centering
\includegraphics[page=29, trim={0cm, 0cm, 0cm, 0cm}, clip, width=0.48\textwidth]{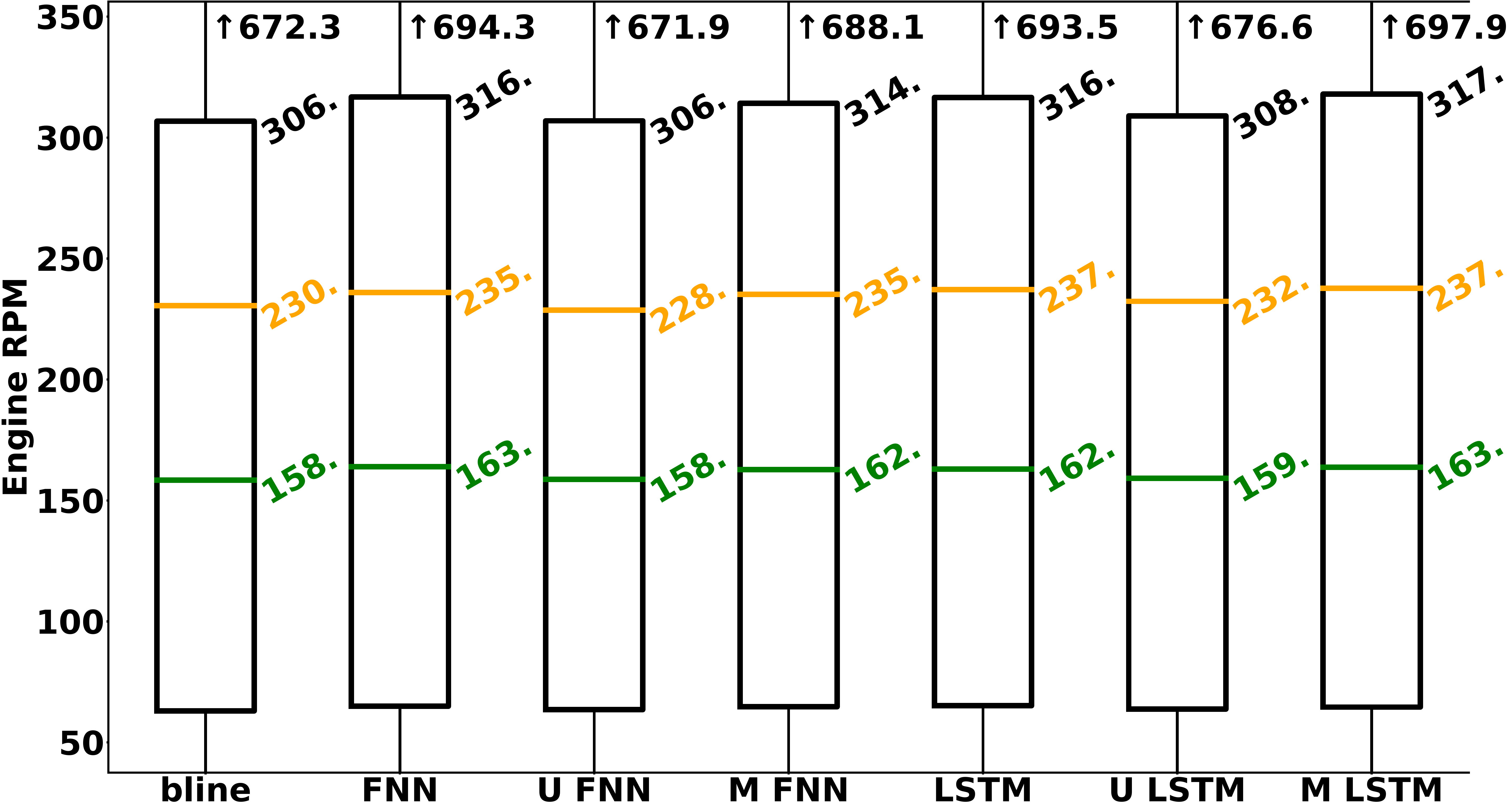}
}
\end{subfloat}
\begin{subfloat}[Initialization \label{fig:init_ablation_rpm}]{
\centering
\includegraphics[page=29, trim={0cm, 0cm, 0cm, 0cm}, clip, width=0.48\textwidth]{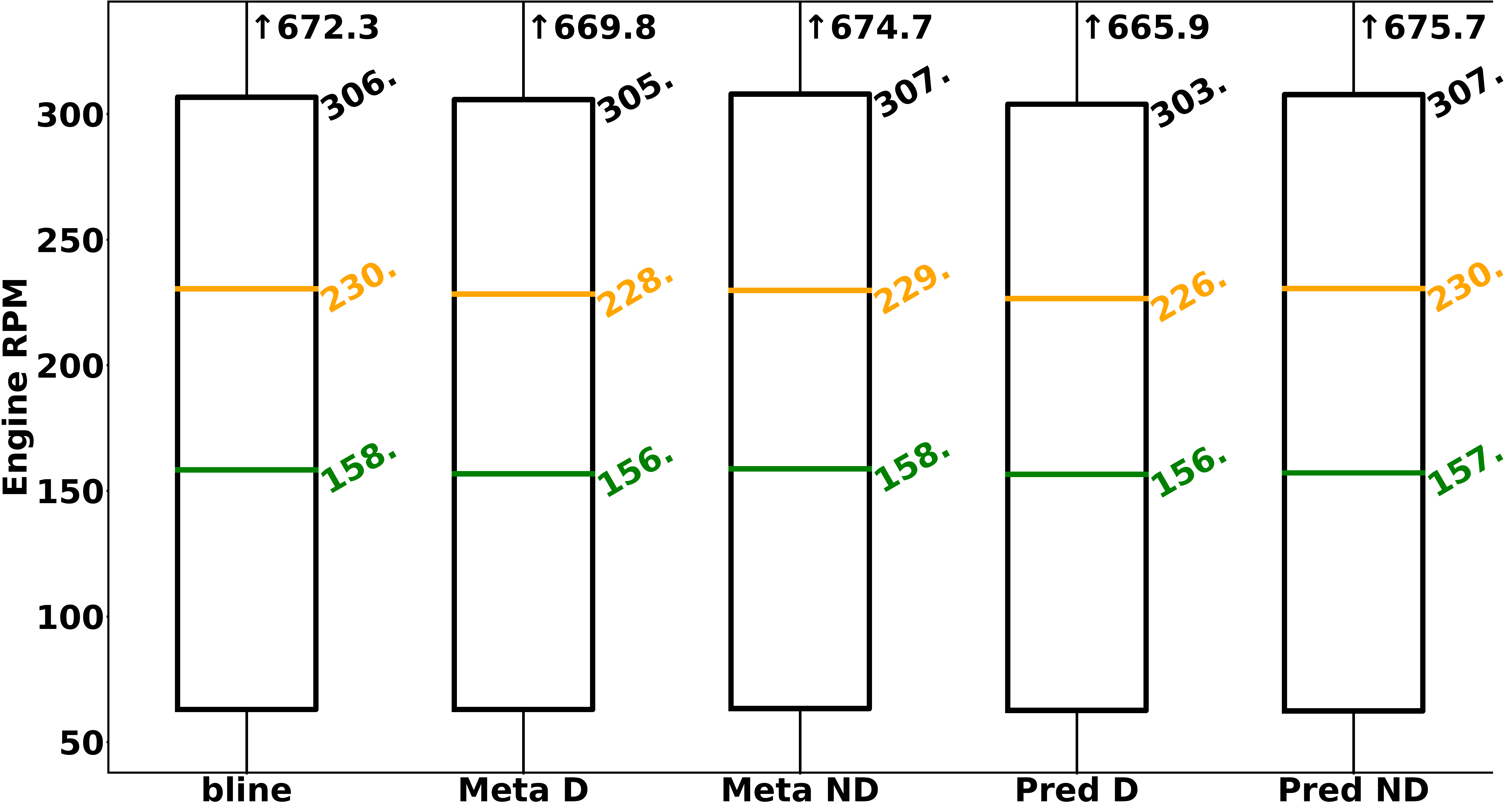}   
}
\end{subfloat}
\begin{subfloat}[Loss Structure \label{fig:structure_loss_rpm}]{
\centering
\includegraphics[page=29, trim={0cm, 0cm, 0cm, 0cm}, clip, width=0.48\textwidth]{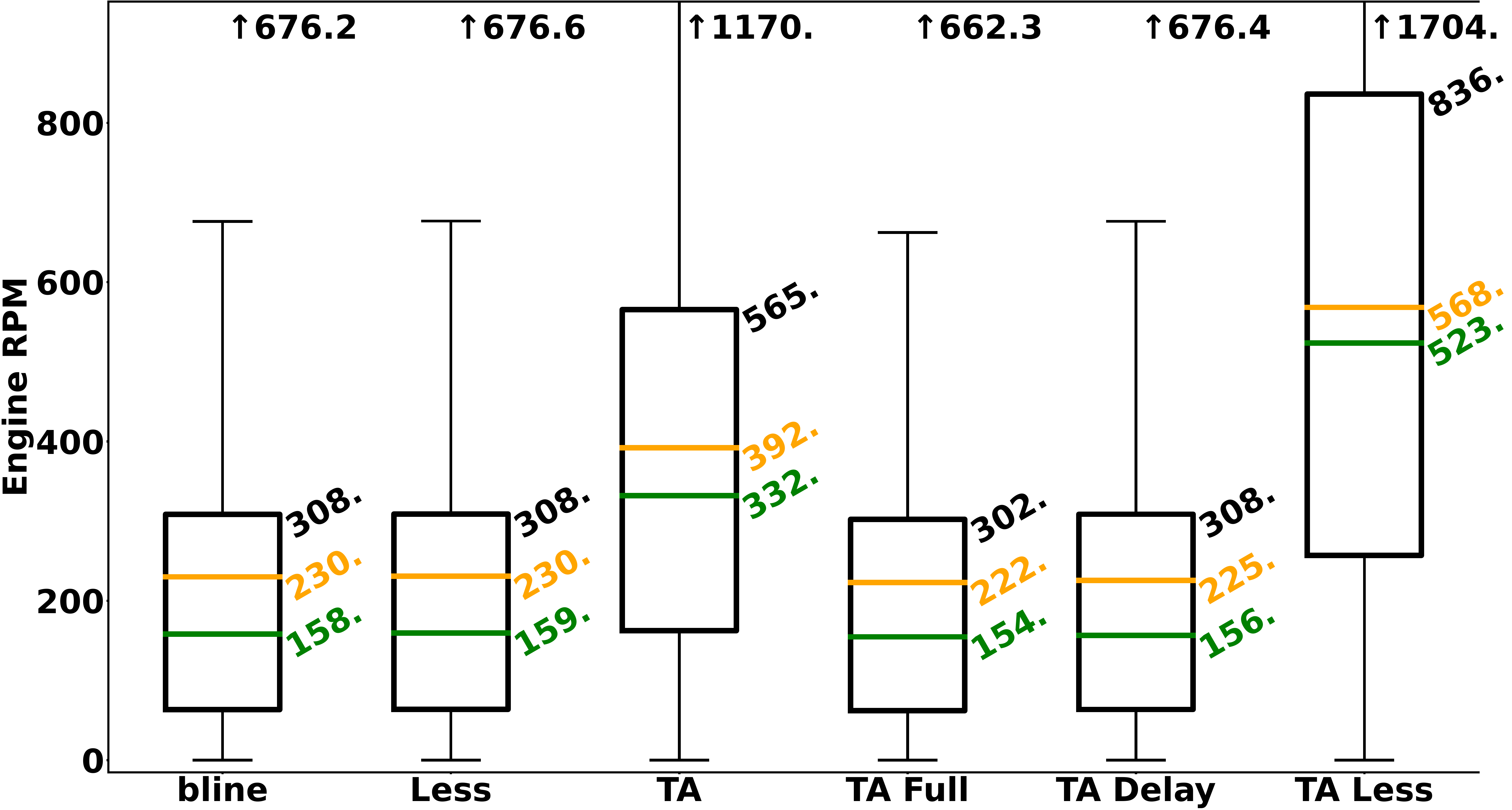}   
}
\end{subfloat}
\begin{subfloat}[Buffer History ($\tau$) \label{fig:tau_ablation_rpm}]{
\centering
\includegraphics[page=29, trim={0cm, 0cm, 0cm, 0cm}, clip, width=0.48\textwidth]{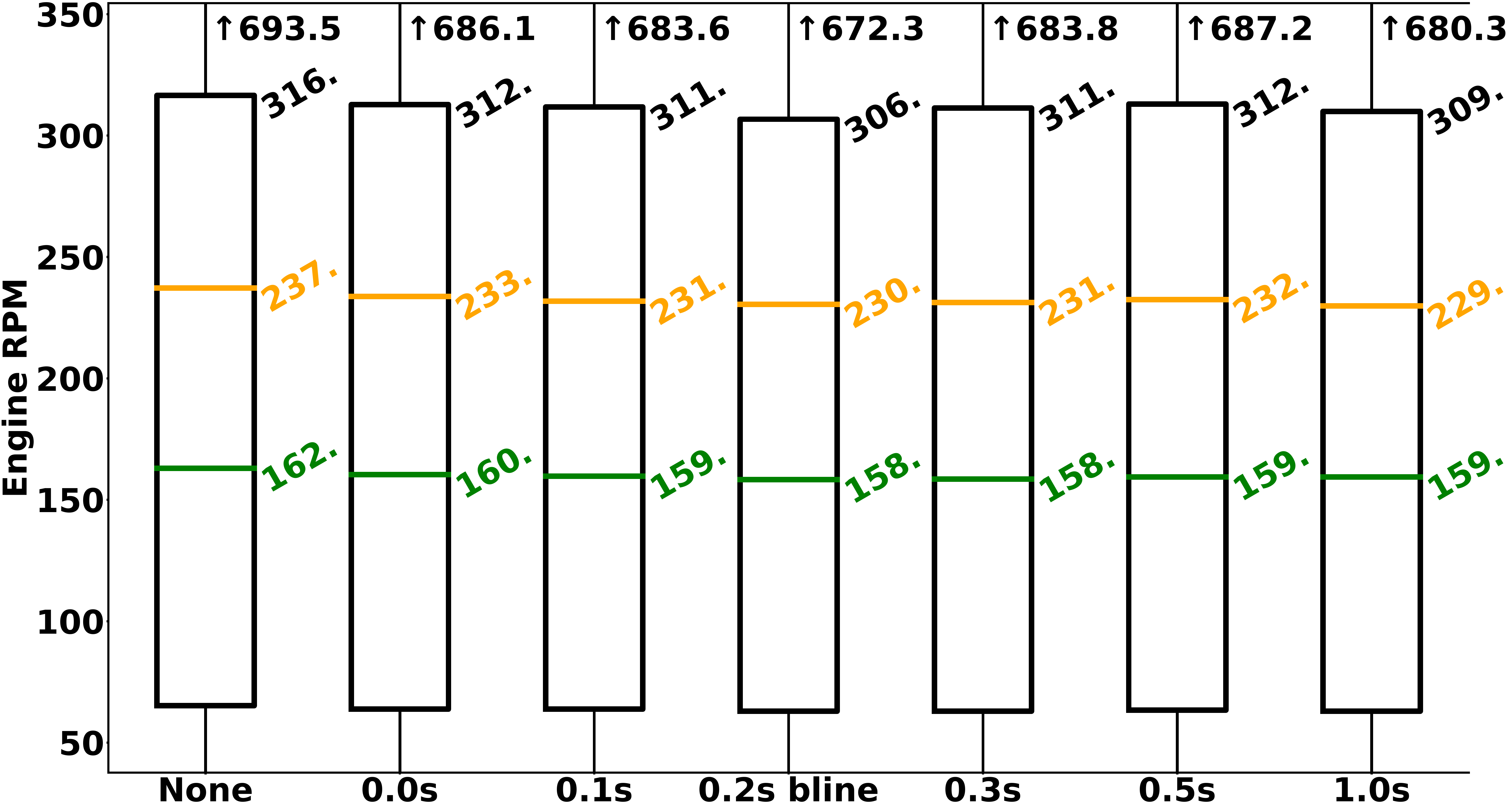} 
}
\end{subfloat}
\begin{subfloat}[Final Comparison \label{fig:final_comparison_rpm}]{
\centering
\includegraphics[page=29, trim={0cm, 0cm, 0cm, 0cm}, clip, width=0.48\textwidth]{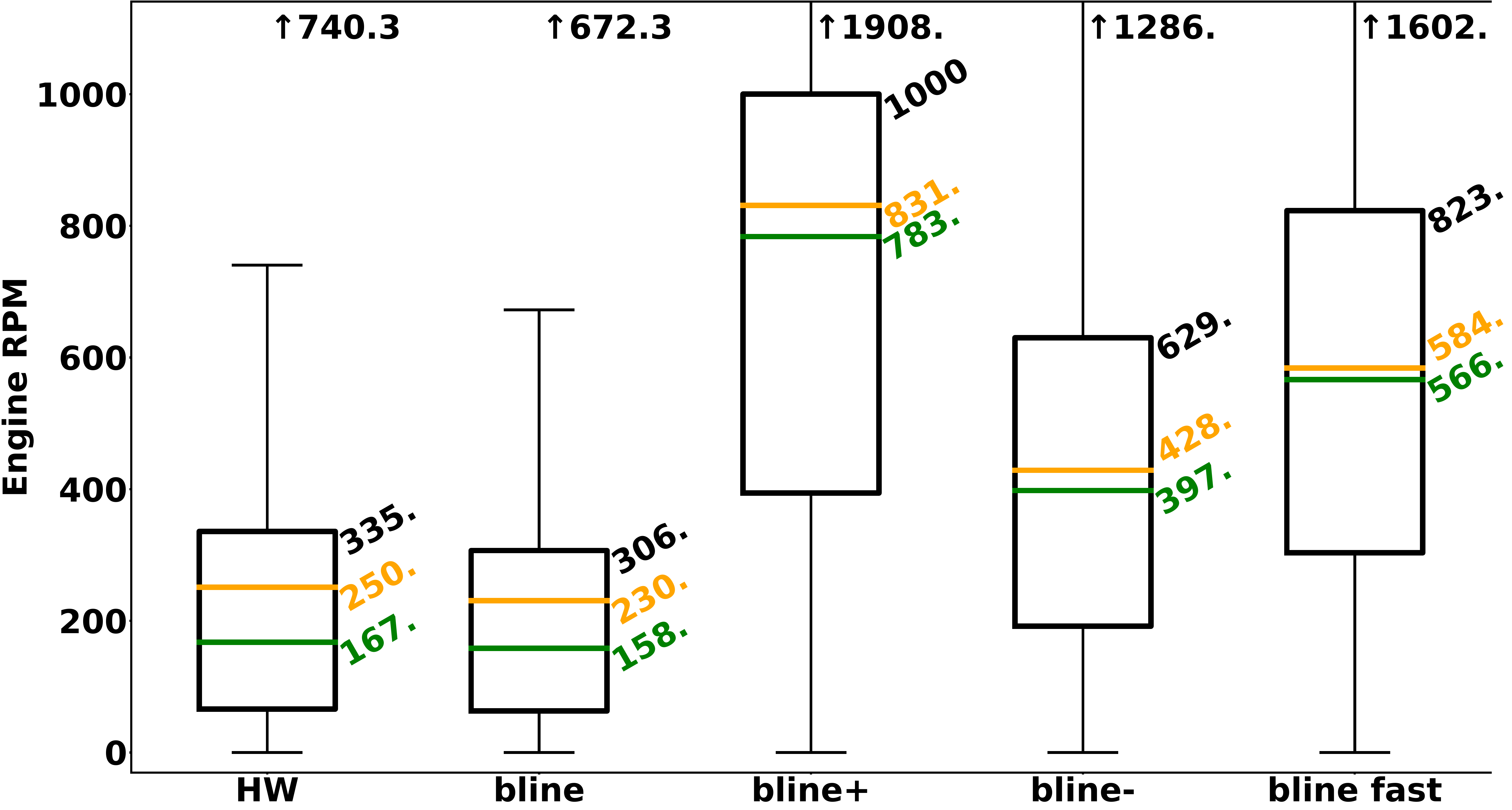}  
}
\end{subfloat}
\caption{
Box plots showing $e$ (engine RPM) error at the end of the $5s$ trajectory for various structural changes to the networks. 
Whiskers are defined as $\pm 1.5 IQR$ and is given with the arrows, the green line defines the median and the orange the mean value. 
The baseline model (bline) is kept consistent.
The explanations of the keys used in shown in the text \cref{subsubsec:structure_ablation}.
}
\label{fig:structure_ablation_results_rpm}
\end{figure*}

\clearpage

\subsection{Additional Hardware Result Images}

We also include still images from the hardware results in complete context to hopefully better explain the sequence of events.
The video is still recommended to get a better picture of how the system is working an evolving in real time.

\input{hardware_appendix}

\end{document}

%% file: packages.tex
\usepackage[ruled,vlined]{algorithm2e}
\usepackage{amsfonts}       
\usepackage{amssymb}
\usepackage{balance}
\usepackage{booktabs} 
\usepackage{graphicx}
\usepackage{float}
\usepackage{mathtools}
\usepackage{mathrsfs}
\usepackage{multirow}
\usepackage[caption=false, font=footnotesize]{subfig}
\usepackage{siunitx}
\usepackage[normalem]{ulem}
\usepackage{usebib}
\usepackage[dvipsnames]{xcolor}

\iffalse
\usepackage[backref=section]{hyperref}
\else
\usepackage[backref=page]{hyperref}
\fi

\hypersetup{
    colorlinks=true,
    linkcolor=OliveGreen,
    filecolor=magenta,      
    urlcolor=blue,
    citecolor=purple,
}
\usepackage[capitalise]{cleveref}

\usepackage[nolist]{acronym}

\DeclareGraphicsExtensions{.png,.pdf}
\makeatletter

%% file: commands.tex
\usepackage{amsfonts}       
\usepackage{amsmath}
\usepackage{amssymb}
\usepackage{amsthm}
\usepackage{balance}
\usepackage{booktabs} 

\usepackage{graphicx}
\usepackage{float}
\usepackage{mathtools}
\usepackage{mathrsfs}
\usepackage{multirow}
\usepackage{siunitx}
\usepackage[dvipsnames]{xcolor}
\usepackage[ruled,vlined]{algorithm2e}

\usepackage[capitalise]{cleveref}
\usepackage[caption=false, font=footnotesize]{subfig}

\usepackage[nolist]{acronym}
\usepackage{usebib}

\DeclareGraphicsExtensions{.png,.pdf}
\makeatletter 
\usepackage[normalem]{ulem}

\theoremstyle{definition}

\begin{acronym}
\acro{ADMM}{Alternating Direction Method of Multipliers}
\acro{AR}{AutoRally}
\acro{CAN}{Control Area Network}
\acro{CEM}{Cross-Entropy Method}
\acro{CNN}{Convolutional Neural Network}
\acro{ConvLSTM}{Convolutional Long Short Term Memory}
\acro{CS}{Covariance Steering}
\acro{CVaR}{Conditional Value-at-Risk}
\acro{DARPA}{Defense Advanced Research Projects Agency}
\acro{DDP}{Differential Dynamic Programming}
\acro{EKF}{Extended Kalman Filter}
\acro{UKF}{Unscented Kalman Filter}
\acro{EM}{Expectation Maximization}
\acro{ES}{Evolutionary Strategies}
\acro{FNN}{Feedforward Neural Network}
\acro{GMM}{Gaussian Mixture Model}
\acro{GP}{Gaussian Process}
\acroplural{GP}[GPs]{Gaussian Processes}
\acro{GRU}{Gated Recurrent Unit}
\acro{GPR}{Gaussian Process Regression}
\acro{FT}{Fourier Transform}
\acro{iCEM}{improved Cross-Entropy Method}
\acro{iFFT}{inverse Fast Fourier Transform}
\acro{I-LSTM}{Initialized LSTM}
\acro{IS}{Importance Sampler}
\acro{KL}{Kullback Leibler}
\acro{LSTM}{Long Short-term Memory}
\acro{MBRL}{Model-Based Reinforcement Learning}
\acro{MC}{Monte-Carlo}
\acro{MCS}{Monte-Carlo Samples}
\acro{MOO}{Multi-Objective Optimization}
\acro{MPC}{Model Predictive Control}
\acro{MPPI}{Model Predictive Path Integral}
\acro{MSE}{Mean Squared Error}
\acro{NLL}{Negative Log-Likelihood}
\acro{NLN}{Normal Log-Normal}
\acro{NN}{Neural Network}
\acro{PC}{Polynomial Chaos}
\acro{PCA}{Principle Component Analysis}
\acro{PDF}{Probability Density Function}
\acro{PSD}{Power Spectral Density}
\acro{RCNN}{Region-based Convolutional Neural Network}
\acro{RL}{Reinforcement Learning}
\acro{RNN}{Recurrent Neural Network}
\acro{RPN}{Region Proposal Network}
\acro{SQP}{Sequential Quadratic Programming}
\acro{VaR}{Value-at-Risk}
\acro{YOLO}{You Only Look Once}
\end{acronym}

\newcommand{\vc}{{\bf c}}
\newcommand{\vx}{{\bf x}}
\newcommand{\vy}{{\bf y}}

\newcommand{\vp}{{\bf p}}

\newcommand{\vq}{{\bf q}}

\newcommand{\vf}{{\bf f}}

\newcommand{\vu}{{\bf u}}
\newcommand{\vv}{{\bf  v}}

\newcommand{\vF}{{\bf F}}

\newcommand{\vpi}{{\mbox{\boldmath$\pi$}}}


%% file: hardware_results.tex
\begin{figure*}[ht]
  \centering
  \begin{tabular}{cc}
    \multirow{6}{*}[2.45cm]{
      \includegraphics[width=0.6\textwidth, height=0.9\textheight, keepaspectratio]{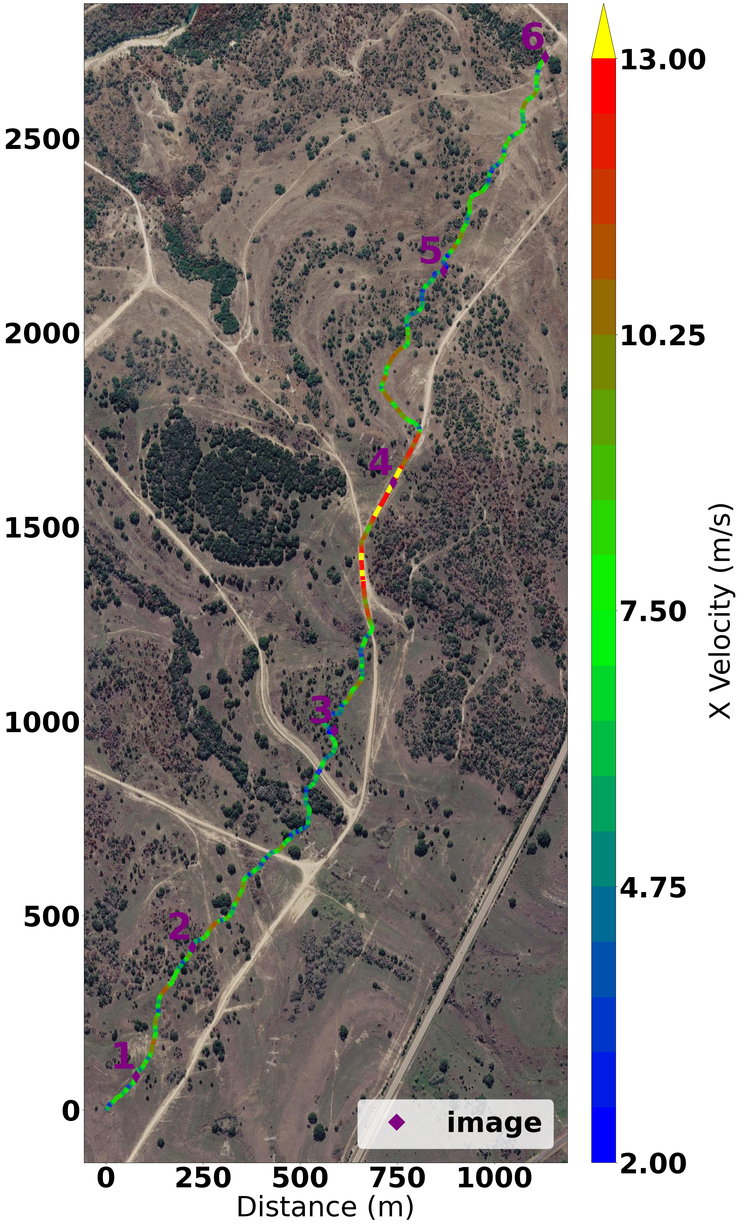}
    }
    &
    \includegraphics[trim={0cm, 0cm, 0cm, 0cm}, clip, height=0.12\textheight]{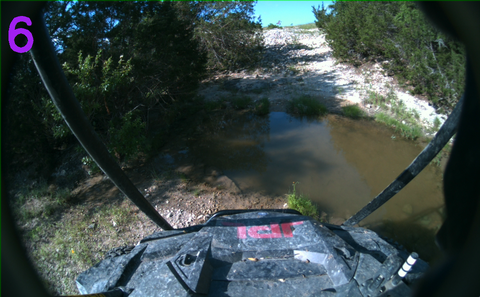} \\
    & \includegraphics[trim={0cm, 0cm, 0cm, 0cm}, clip, height=0.12\textheight]{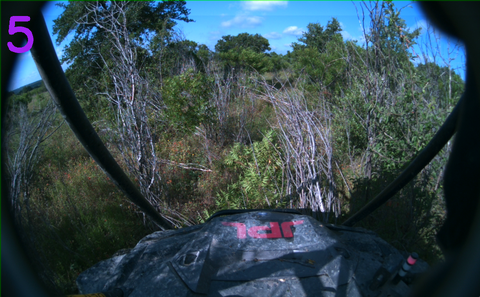} \\
    & \includegraphics[trim={0cm, 0cm, 0cm, 0cm}, clip, height=0.12\textheight]{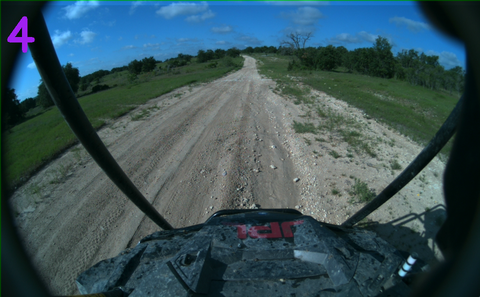} \\
    & \includegraphics[trim={0cm, 0cm, 0cm, 0cm}, clip, height=0.12\textheight]{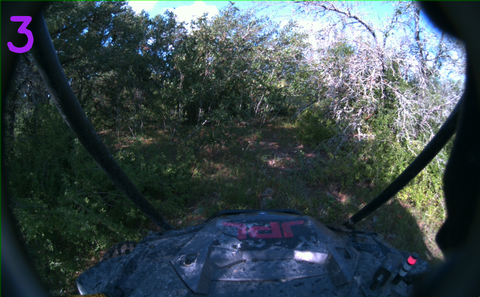} \\
    & \includegraphics[trim={0cm, 0cm, 0cm, 0cm}, clip, height=0.12\textheight]{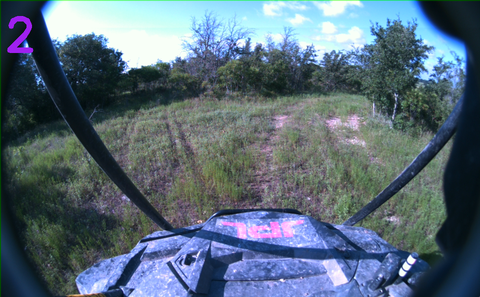} \\
    & \includegraphics[trim={0cm, 0cm, 0cm, 0cm}, clip, height=0.12\textheight]{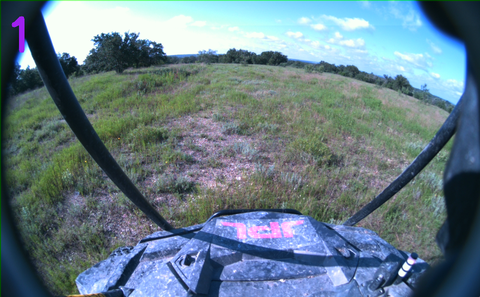} \\
  \end{tabular}
  \caption{
    A view into the long traverse discused in \cref{subsubsec:long_traverse}.
    The left shows the top down speed map of $v^x$ for the entire traverse while the left shows images pulled from the camera during the run.
    Images are marked in purple with the number that corresponds to the marker in the top down image where they were pulled from.
  }
  \label{fig:long_traverse}
\end{figure*}

\begin{figure*}[ht]
  \centering
  \begin{subfloat}[X Velocity Map\label{fig:Red_Slide_top_down}]{
    \centering
    \includegraphics[trim={0cm, 0cm, 0cm, 0cm}, clip, width=0.3\textwidth, height=0.4\textheight, keepaspectratio]{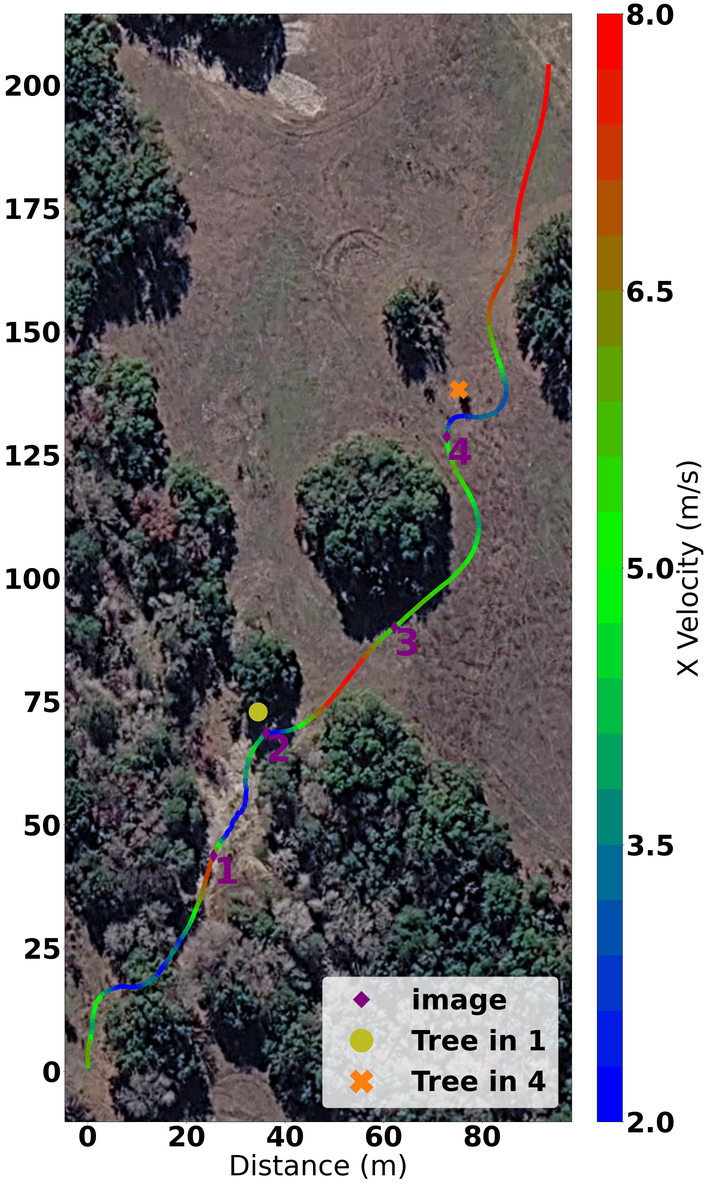}
  }
  \end{subfloat}
  \begin{subfloat}[Y Velocity Map\label{fig:Red_Slide_top_down2}]{
    \centering
    \includegraphics[trim={0cm, 0cm, 0cm, 0cm}, clip, width=0.3\textwidth, height=0.4\textheight, keepaspectratio]{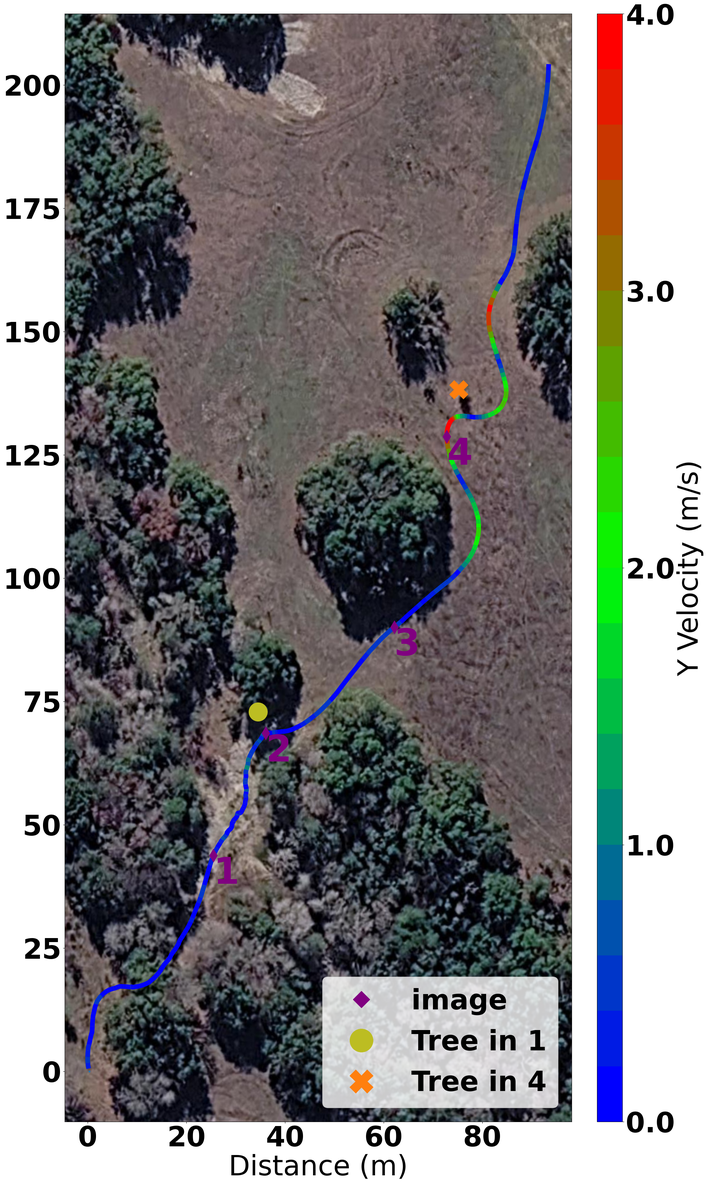}
  }
  \end{subfloat}
  \\
  \begin{subfloat}[Front Image 1\label{fig:Red_Slide_1}]{
    \centering
    \includegraphics[trim={0cm, 0cm, 0cm, 0cm}, clip, width=0.3\textwidth]{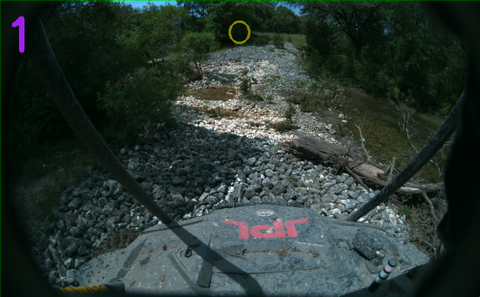}
  }
  \end{subfloat}
  \begin{subfloat}[Front Image 4\label{fig:Red_Slide_2}]{
    \centering
    \includegraphics[trim={0cm, 0cm, 0cm, 0cm}, clip, width=0.3\textwidth]{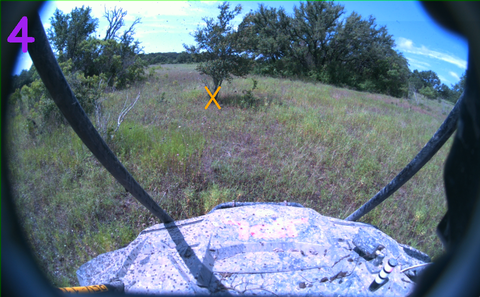}
  }
  \end{subfloat}
  \begin{subfloat}[Rear Image 4\label{fig:Red_Slide_2_rear}]{
    \centering
    \includegraphics[trim={0cm, 0cm, 0cm, 0cm}, clip, width=0.3\textwidth]{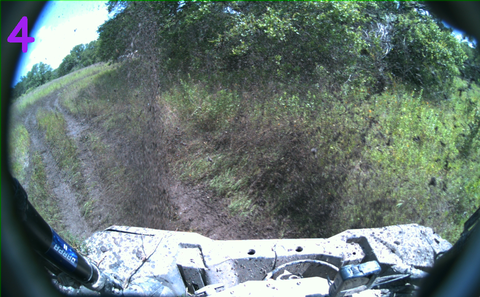}
  }
  \end{subfloat}
  \caption{
    Speed maps for the first case study discussed in \cref{subsubsec:case_studies}.
    \cref{fig:Red_Slide_top_down} shows the $v^x$ values while \cref{fig:Red_Slide_top_down2} shows the $v^y$ values.
    Relevant trees are marked in the images and in the top down map to indicate where events occur.
    Images are marked in purple with a number that corresponds to the top down image.
  }
  \label{fig:red_slide}
\end{figure*}

Our hardware results will be broken into three sections.
The first demonstrates the effectiveness of the entire architecture in a $3.5 km$ traverse over a variety of different terrains.
The second will explore specific case studies in more detail to show the efficacy of the approach in challenging situations.
Finally, the third will focus on the impact in aggressiveness of changing the size of the uncertainty ellipse by altering $c_\sigma$.

For all traverses the vehicle will be using the same perception stack and an identical planning stack unless otherwise noted.
For all runs the vehicle has no prior map of the area, just GPS waypoints and a goal acceptance distance of $1m$.
GPS accuracy is artificially degraded during the experiment so there will be some slowing down and swerving near waypoints that should be ignored.
The goal will be the fastest time to the waypoints, but for some maps we will not denote a waypoint since it is so far away that it does not impact the local behavior of the vehicle other than giving a direction.

Our results will be on the adaptability of the planning stack to diverse situations without explicit modes and not the inherent issues with the perception stack that influence planning.
Issues with perception are more clearly outlined in \cite{atha2024fewshotsemanticlearningrobust}.
The perception stack is sensitive to lighting conditions changes and can struggle with specific vegetation.
Throughout the testing of this vehicle we have noted that there is high variability in the path taken, unless it is following a trail.
This can come from many possible reasons, but the main two are perceptual differences and environmental change.
The vehicle changes the environment by driving through it.
These changes are more noticeable in vegetation since the bushes or grass are irreparably compressed by the wheels.
A light trail tends to form as the vehicle drives similar paths multiple times, so the speed on repeated courses without stack changes substantially.
Any prior mistake or success gives creates more open space that the vehicle will naturally follow to drive faster.
These issues together create a situation where repeated testing difficult to compare, hence our focus on long traverses and specific case studies of good behavior.
The examples provided in this paper are from areas where the perception stack is performing well to highlight the planning stack.

There are two concepts in the description of the wheel or body costs that are relevant for these results.
In general the perception system is trying to classify three types of traversable terrain or the space as unknown traversability.
The classes are, 1). obstacles that cannot be traversed at any speed like a tree 2). obstacles like bushes that can be traversed at low speed since perceptual uncertainty is high for what is undernearth or behind them, and finally 3). trails that might indicate a high likelihood of future traversability.
Trails can be seen in \cref{fig:trail_comparison} and take the form of previously driven tracks to fully paved roads \cite{atha2024fewshotsemanticlearningrobust}.
The vehicle can be incentivized to stay on the trail through the \textit{CostToGo} coming from a higher level planner and a small road centering cost applied in the cost map to encourage visibility of the trail in the future.
This feature is off unless otherwise noted.

Unknown space is defined as voxels where we have too few points to reliably estimate the ground plane or semantically classify the potential vegetation there.
A lack of points can be caused by occlusion or lack of point density at distance.
The most important type of obstacle unknown space is trying to classify are negative obstacles.
These tend to be things like smaller ditches or cliff edges that show up as only a lack of information.
Hitting ditches at too high of a speed was the main reason for substantial vehicle damage during the project.
Our system makes an assumption that if the ground can be semantically detected as trail, then we will not classify it as unknown space.
The stereo cameras have a lower latency and higher density than lidar, so this is what allows the vehicle to drive close to maximum speed on trail.
The depth from the cameras can cause issues with geometric obstacles so only the semantics of the points are used, stereo is not used to create the elevation map.
We further outline the importance and issues at high speeds with unknown space in \cref{subsubsec:trail_comparison}.

The vehicle is equipped with a speed limiter for safety that overrides the autonomous control.
It kicks in at $30 mph \approx 13.4 m/s$ which abruptly cuts the throttle based on wheel speed measurements of the rear wheels.
The throttle abruptly is cut until the speed of the vehicle drops below the threshold.
So while there may be places where the vehicle is able to drive faster than the limiter, there is a good chance the limiter is turning on and off in those moments.
This should not be viewed as a limitation of the algorithm but of the safety limits placed on the vehicle.
Typically, we will only encounter the limiter on open trails with the current perception limits on clearing of unknown space.
Speeds that occur near the limiter are marked in yellow in the speed maps.
%


\subsubsection{Long Traverse}
\label{subsubsec:long_traverse}

\begin{figure*}[ht]
  \centering
  \begin{tabular}{cc}
    \multirow{3}{*}[3.2cm]{
        \centering
        \includegraphics[width=0.5\textwidth, height=0.5\textheight, keepaspectratio]{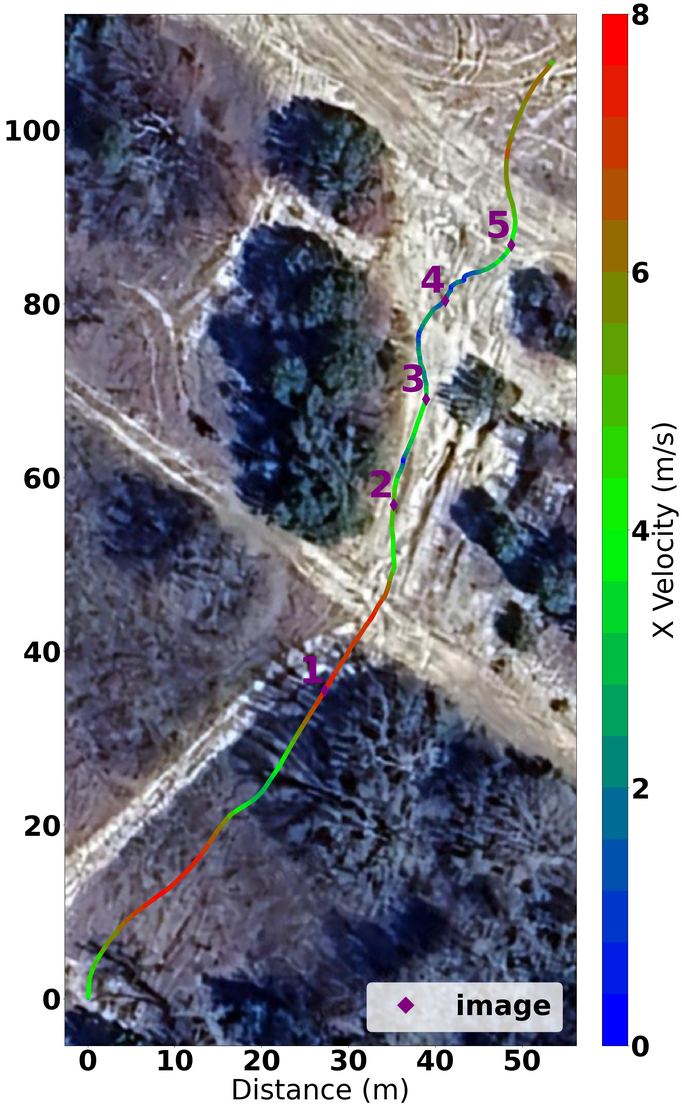}  
    } 
    &
    \begin{subfloat}[First Slope Approach Image\label{fig:Tight_Slope_Green_2}]{
      \centering
      \includegraphics[width=0.3\textwidth, height=0.25\textheight, keepaspectratio]{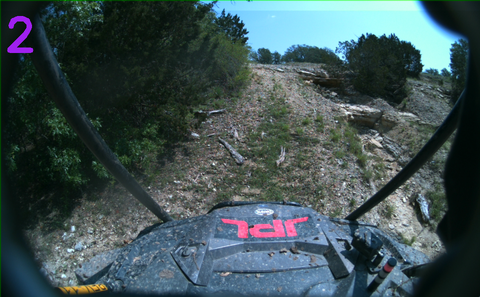}
    }
    \end{subfloat} \\
    &
    \begin{subfloat}[High Rollover Risk\label{fig:Tight_Slope_Green_3}]{
      \centering
      \includegraphics[width=0.3\textwidth, height=0.25\textheight, keepaspectratio]{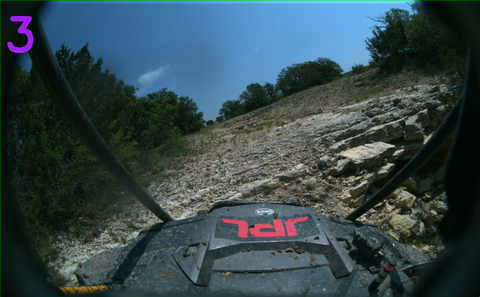}
    }
    \end{subfloat}\\
    &
    \begin{subfloat}[Second Slope\label{fig:Tight_Slope_Green_4}]{
      \centering
      \includegraphics[width=0.3\textwidth, height=0.25\textheight, keepaspectratio]{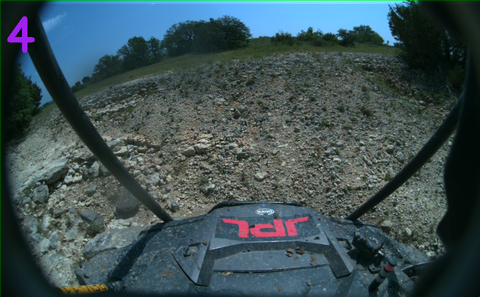}
    }
    \end{subfloat}\\
  \end{tabular}
  \caption{
    Shows top down view of $v^x$ and images for the second case study looked at in \cref{subsubsec:case_studies}.
    The top down map and images are marked by numbers in purple that correlate across the entire figure.
  }
  \label{fig:Tight_Slope_Green}
\end{figure*}

We can see a longer traverse in \cref{fig:long_traverse} with an average vehicle speed of $\approx 6.2 m/s$ and a total distance of $\approx 3.3km$ in $\approx 8.7$ minutes.
This run includes a variety of off-road terrain that the vehicle must navigate.
The subsection of the first three images goes from an open field (image 1) a sparse grove of trees (image 2) and finally a denser line of trees (image 3).
The slower sections in this area come from tall grasses not clearing out unknown space, negative obstacles, or navigating close to trees.
There are also sporadic ditches in the open fields between the trees the vehicle has to slow down for, the maximum roll angle it get to is $0.25 rad$ and the highest pitch is $0.25 rad$.
The section from image 4 shows the vehicle finding a trail and getting up to maximum speed (denoted by the yellow in figure).
There is a grove of trees and dense vegetation near image 5 and finally there is a creek near image 6.
The vehicle stops and reverses at the creek since the perception system created incorrect elevation that was hazardous due to lidar reflections on the water creating a deep ditch.
The run was stopped at this point due to an unrelated safety issue in the chase vehicle.

We emphasize the adaptation of the speed on these diverse environments that is entirely a function of the learned dynamical uncertainty.
The vehicle drives faster on open fields purely because it is open and drives slower in constrained areas because it is tight.
Overall the vehicle displays intelligent behavior in how it chose to navigate the terrain and at what speed on a large range traverse without environment specific tuning of the planning architecture or explicit modes.

\subsubsection{Interesting Case Studies}
\label{subsubsec:case_studies}

\begin{figure*}[ht]
  \centering
  \begin{subfloat}[T1\label{fig:YT_3-5_T1}]{
    \centering
    \includegraphics[trim={0cm, 0cm, 0cm, 0cm}, clip, height=0.2\textheight, width=0.32\textwidth, keepaspectratio]{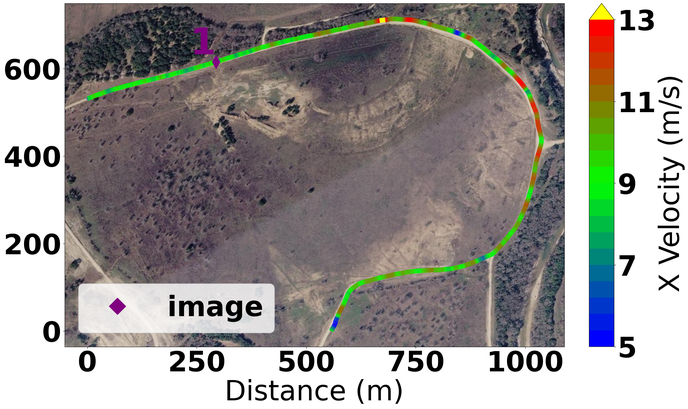}
  }
  \end{subfloat}
  \begin{subfloat}[T2\label{fig:YT_3-5_T2}]{
    \centering
    \includegraphics[trim={0cm, 0cm, 0cm, 0cm}, clip, height=0.2\textheight, width=0.32\textwidth, keepaspectratio]{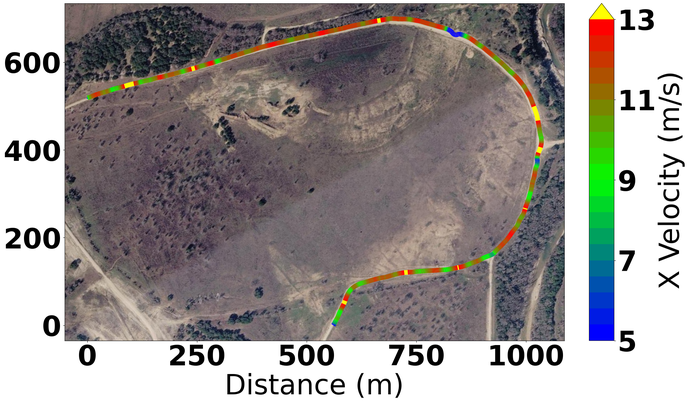}
  }
  \end{subfloat}
  \begin{subfloat}[Image 1\label{fig:YT_1}]{
    \centering
    \includegraphics[trim={0cm, 0cm, 0cm, 0cm}, clip, height=0.2\textheight, width=0.25\textwidth, keepaspectratio]{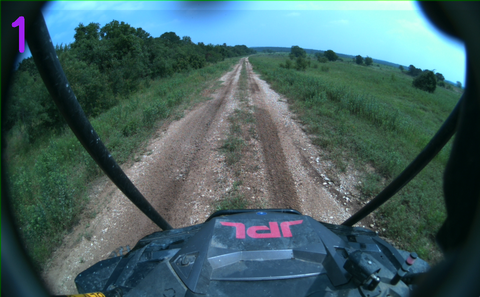}
  }
  \end{subfloat}
  \begin{subfloat}[T1\label{fig:YT_9-11_T1}]{
    \centering
    \includegraphics[trim={0cm, 0cm, 0cm, 0cm}, clip, height=0.2\textheight, width=0.32\textwidth, keepaspectratio]{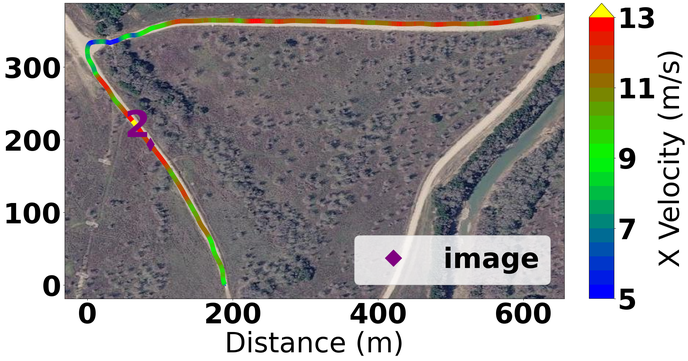}
  }
  \end{subfloat}
  \begin{subfloat}[T2\label{fig:YT_9_11_T2}]{
    \centering
    \includegraphics[trim={0cm, 0cm, 0cm, 0cm}, clip, height=0.2\textheight, width=0.32\textwidth, keepaspectratio]{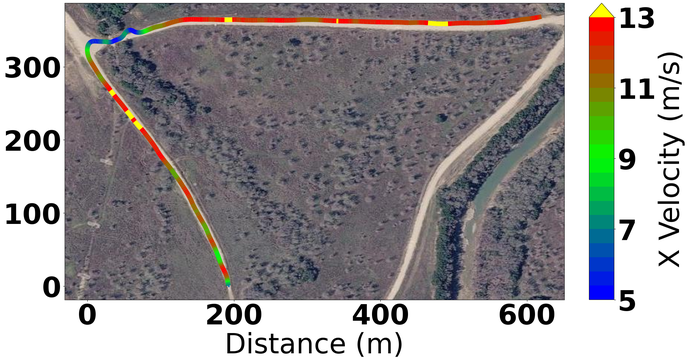}
  }
  \end{subfloat}
  \begin{subfloat}[Image 2\label{fig:YT_3}]{
    \centering
    \includegraphics[trim={0cm, 0cm, 0cm, 0cm}, clip, height=0.2\textheight, width=0.25\textwidth, keepaspectratio]{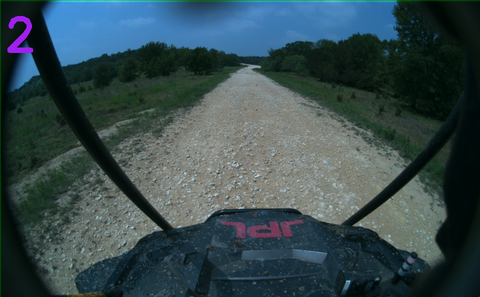}
  }
  \end{subfloat}
  \caption{
    Shows top down view of $v^x$ and images for the different runs of the trail course outlined in \cref{subsubsec:trail_comparison}.
    The top down map and images are marked by numbers in purple that correlate with the map shown to the left of it.
    \cref{fig:YT_3-5_T1} and \cref{fig:YT_3-5_T2} is the first and \cref{fig:YT_9-11_T1} and \cref{fig:YT_9_11_T2} is the second pair taken from the same location and can be compared between T1 and T2.
    T1 is using $c_\sigma = 2.0$ and T2 uses $c_\sigma = 1.0$ but otherwise they are identical stacks run around the same time of day.
  }
  \label{fig:trail_comparison}
\end{figure*}

Now we focus on smaller scale sections from different traverses that demonstrate the effectiveness of the dynamical uncertainty approach.
The focus here is to show the system makes intelligent decisions about speed and aggressiveness due to uncertainty.
We use the exact same planning stack for both case studies as we did for the long traverse in the previous section.

The first situation is a transition from slow driving in a grove of trees to an open muddy field \cref{fig:red_slide}.
\cref{fig:Red_Slide_1} shows the narrow opening between a downed tree and a tree in a creek bed the vehicle navigates through slowly.
The gap is very narrow so the vehicle slowly navigates with its right wheel over the downed tree.
The downed tree is marked as risky while the tree to the left is marked not traversable.
Right before marker 2 the vehicle has issues with traction on the rocks and ends up very close to the tree \cref{fig:Red_Slide_1} due to a hidden negative obstacle and applies the brake to avoid hitting that tree.
Finally in the open, the vehicle applies a throttle and begins to take a turn to the left around a group of trees seen in \cref{fig:Red_Slide_2_rear} and ends up in several aggressive slides in the mud \cref{fig:Red_Slide_top_down2}.
During the first slide the vehicle tries to turn to avoid the tree and bush shown in \cref{fig:Red_Slide_2} which creates the second slide.
The second slide creates the third slide when trying to return to the desired trajectory towards the waypoint.
The final peak of $v^y$ occurs after exiting the mud, so the vehicle is quickly able to regain control and speeds toward the waypoint.
At the peak of the second slide the vehicle is traveling at $\approx 5 m/s$ in the body y direction \cref{fig:Red_Slide_top_down2}.

The vehicle does not hit any obstacle during the slide and continues to apply a high throttle throughout.
While the behavior is not time optimal, it demonstrates the adaptability of the uncertainty based approach.
The system does not take visual features into account for the dynamics, so the slides are entirely unexpected.
This demonstrates that our approach can continue to drive aggressively even with the large uncertainty because the penalty comes from hitting obstacles, not having a large uncertainty.
Furthermore, the size and shape of the uncertainty captures the lateral slide direction and forces the trajectory to include a wide margin in the body y direction when sliding.






The second situation to explore is the vehicle slowing down to navigate a tight passage between a tree and a vehicle sized ditch \cref{fig:Tight_Slope_Green}.
First the vehicle starts driving slower before marker 1, since there is taller vegetation that creates unknown space.
Next it decides to go up the challenging path shown in \cref{fig:Tight_Slope_Green_2}.
At the narrow opening there is a small amount of clearance on either side of the vehicle with the left hitting a tree and the right falling into a large ditch that is $\approx 4ft$ deep.
Furthermore, at the narrowest part of the gap the vehicle goes through is at a roll angle of $0.35 rad$.
The vehicle has to orient the front wheels precisely to the left to prevent from tipping over with the ledge seen on the right in \cref{fig:Tight_Slope_Green_3}.
After navigating through the narrow gap, there is a second steep slope shown in \cref{fig:Tight_Slope_Green_4}.
The vehicle wants to take a hard left after getting up the slope to follow an erroneous trail detection to the left that clears almost immediately as it crests the top of the second slope.
Unfortunately, the wheels are turned to the left with a high throttle to get up the last part of the slope which causes a short slide to occur.
The slide is executed safely and only occurs since the top of the ridge is open and the vehicle is not close to rolling over.
This contrasts with the first slope where the throttle and steering applied is much more conservative.

Again the vehicle demonstrates intelligent speed control by slowing down to navigate the narrow opening.
The uncertainty ellipse can only be fit through the smaller gap at low speeds, so the vehicle slows down as a function of the uncertainty and environment.
Prior experiments without the uncertainty based planning were unable to precisely control speed and steering in the narrow gap and rolled over into the ditch.
The old system relied on heuristics for maximum speed that were not able to account for the diversity of terrain.
The uncertainty approach is more robust in overly constrained areas and also drives faster overall with fewer constraints.

\subsubsection{Trail Driving Comparison}
\label{subsubsec:trail_comparison}


In this section we show two different runs on the same set of waypoints shown in \cref{fig:trail_comparison}.
Repeatability issues are mitigated since the entire course is on clear trails.
The only change between the two attempts are altering $c_\sigma = 2.0$ in the T1 run and $c_\sigma = 1.0$ in the second run.
This demonstrates how tuning the risk profile of the system is possible and that a smaller covariance would drive faster, motivating the changes outlined in \cref{subsec:ablation_studies}.
We turn on the additional feature to encourage staying on trail when the trail appears to be moving in the direction of the waypoint.
The short range planner is making the tradeoff of when to exit the trail based on if it sees the current trail as generally making progress towards the waypoint.
This is still based entirely off of local information.
The determination on if we should follow the trail is based on the angle between the direction of the waypoint and the direction the trail looks like it will continue in.
We will follow a trail that is within $90 \deg$ of the waypoint direction as a heuristic.
Generally, this results in the vehicle staying on trail when it is moving towards the waypoint and taking a more direct route otherwise.

For the first section of trail in \cref{fig:YT_3-5_T1} takes $204s$ at an average speed of $9.5 m/s$ compared to \cref{fig:YT_3-5_T2} where it takes $179s$ at an average speed of $10.75 m/s$.
Both paths are about $1.92 km$ of total distance with T1 being $16m$ longer.
The median speed on T1 is $9.5 m/s$ while on T2 it is $11.2 m/s$ showing that not only is the average higher but the distribution is shifted as well.
The limiter is triggered for $0.4\%$ for T1 compared to $4.6 \%$ for T2.
The second section in \cref{fig:YT_9-11_T1} takes $99s$ at an average speed of $10.3 m/s$ compared to \cref{fig:YT_9_11_T2} that takes $93s$ at an average speed of $11 m/s$.
Path length is within $1m$ for both paths and is just over $1km$.
T2 struggles more with perceiving the water as traversable than in T1 at location 4, giving a similar time but with a higher speed otherwise.
This is better shown in the median speed which is $11 m/s$ for T1 and $11.8m/s$ for T2 and the time spent at maximum speed which is $1\%$ for T1 and $6.6\%$ for T2.
The maximum $v^y$ for any run is $2 m/s$, and the slip is negligible.

Speed on trail is mainly limited by unknown space in the forward direction and trail width in the lateral direction.
Trail width is the limiting factor in \cref{fig:YT_1} while forward is the limiter in \cref{fig:YT_3}.
The width of the trail limits speed when the edges of the trail remain unknown space but the uncertainty ellipse cannot fit through it.
Vegetation on the edges of the trail can create false ridges in the elevation map when we detect only the top of the vegetation and can prevent unknown space from being cleared.
For the forward direction the entire $5s$ \ac{MPPI} trajectory, and sigma points, must remain inside safe spaces on the map to ensure safety.
Unknown space is treated as a lethal obstacle since the perception system in unsure what is in that voxel.
We chose $5s$ horizon since this is beyond the stopping time of the vehicle at maximum speeds, so our planner must ensure the vehicle can come to a complete and safe stop without entering unknown space.
As speed increases, the $5s$ trajectory is looking farther and farther forward meaning a longer perception distance is required to maintain speed.

The issue with this approach is when moving at close to maximum speed the lidar density at the ends of the trajectory generated by \ac{MPPI} approaches unknown space.
The closed loop uncertainty aids in this by controlling the size of the uncertainty, especially for yaw which impacts wheel query locations.
This is the main limiter of speed on trail, but is a function of the perception stack more than the planning stack.
Increasing the density of the lidar or decreasing latency of the map would result in higher average speeds without changing the planning stack at all, since it would push unknown space farther away from the vehicle.
Alternatively, we could also use the changes outlined in \cref{subsec:ablation_studies} to improve prediction accuracy and better calibrate the uncertainty size and shape.
It should be noted that, while T1 is more conservative than T2, it can still drive aggressively and is not overly conservative.
T1 still reaches maximum speed, it just requires a wider trail and a clear view forward.

%% file: hardware_appendix.tex
\begin{figure*}
  \centering
  \begin{subfloat}[Long Traverse Image 1]{
    \centering
    \includegraphics[trim={0cm, 0cm, 0cm, 0cm}, clip, width=0.3\textwidth, height=0.4\textheight, keepaspectratio]{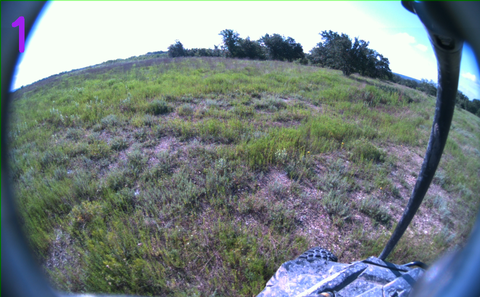}
    \includegraphics[trim={0cm, 0cm, 0cm, 0cm}, clip, width=0.3\textwidth, height=0.4\textheight, keepaspectratio]{Figures/hardware_results/Gabo_t6/image1}
    \includegraphics[width=0.3\textwidth, height=0.4\textheight, keepaspectratio]{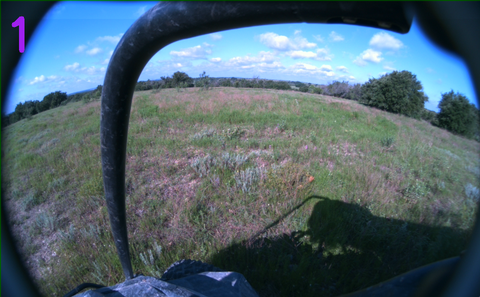}
  }
  \end{subfloat}
  \begin{subfloat}[Long Traverse Image 1 Rear]{
    \centering
    \includegraphics[width=0.3\textwidth, height=0.4\textheight, keepaspectratio]{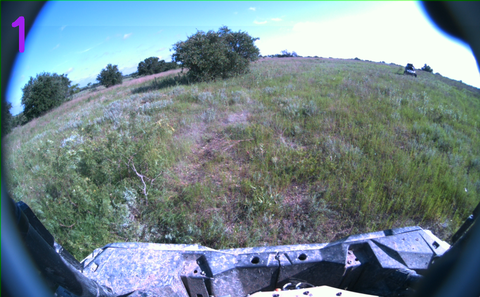}
  }
  \end{subfloat}
  \caption{
    Full set of images from marker 1 in \cref{fig:long_traverse}.
  }
\end{figure*}

\begin{figure*}
  \centering
  \begin{subfloat}[Long Traverse Image 2]{
    \centering
    \includegraphics[trim={0cm, 0cm, 0cm, 0cm}, clip, width=0.3\textwidth, height=0.4\textheight, keepaspectratio]{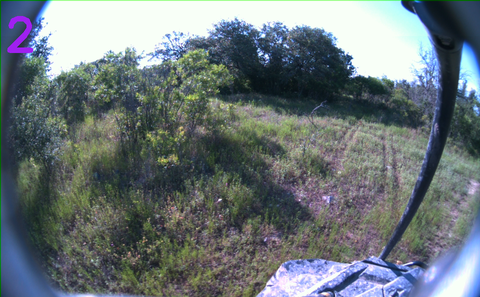}
    \includegraphics[trim={0cm, 0cm, 0cm, 0cm}, clip, width=0.3\textwidth, height=0.4\textheight, keepaspectratio]{Figures/hardware_results/Gabo_t6/image2}
    \includegraphics[width=0.3\textwidth, height=0.4\textheight, keepaspectratio]{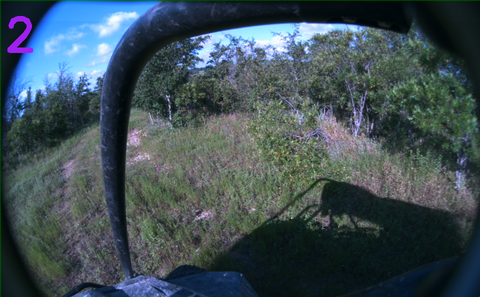}
  }
  \end{subfloat}
  \begin{subfloat}[Long Traverse Image 2 Rear]{
    \centering
    \includegraphics[width=0.3\textwidth, height=0.4\textheight, keepaspectratio]{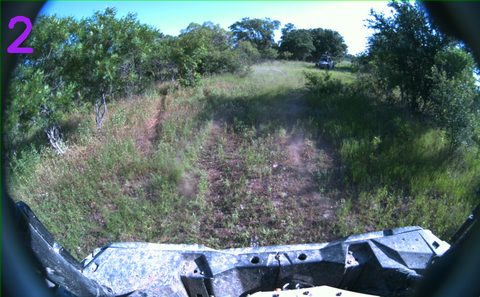}
  }
  \end{subfloat}
  \caption{
    Full set of images from marker 2 in \cref{fig:long_traverse}.
  }
\end{figure*}

\begin{figure*}
  \centering
  \begin{subfloat}[Long Traverse Image 3]{
    \centering
    \includegraphics[trim={0cm, 0cm, 0cm, 0cm}, clip, width=0.3\textwidth, height=0.4\textheight, keepaspectratio]{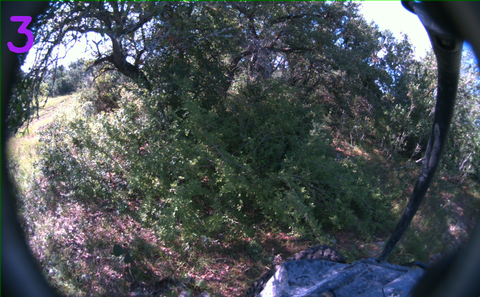}
    \includegraphics[trim={0cm, 0cm, 0cm, 0cm}, clip, width=0.3\textwidth, height=0.4\textheight, keepaspectratio]{Figures/hardware_results/Gabo_t6/image3}
    \includegraphics[width=0.3\textwidth, height=0.4\textheight, keepaspectratio]{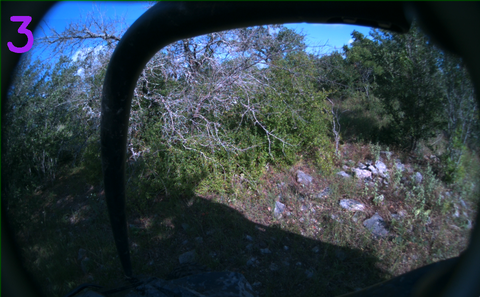}
  }
  \end{subfloat}
  \begin{subfloat}[Long Traverse Image 3 Rear]{
    \centering
    \includegraphics[width=0.3\textwidth, height=0.4\textheight, keepaspectratio]{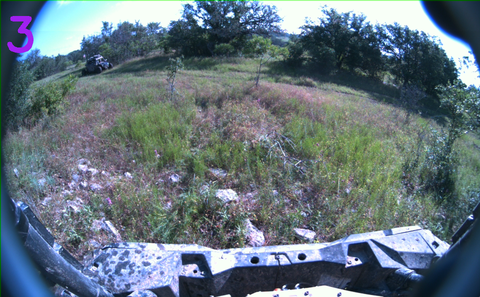}
  }
  \end{subfloat}
  \caption{
    Full set of images from marker 3 in \cref{fig:long_traverse}.
  }
\end{figure*}

\begin{figure*}
  \centering
  \begin{subfloat}[Long Traverse Image 4]{
    \centering
    \includegraphics[trim={0cm, 0cm, 0cm, 0cm}, clip, width=0.3\textwidth, height=0.4\textheight, keepaspectratio]{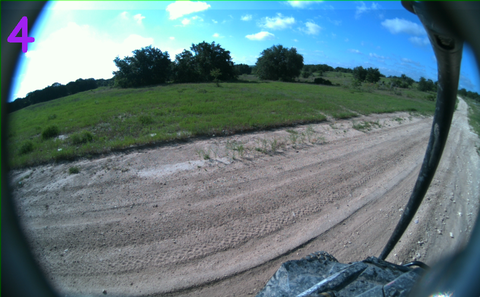}
    \includegraphics[trim={0cm, 0cm, 0cm, 0cm}, clip, width=0.3\textwidth, height=0.4\textheight, keepaspectratio]{Figures/hardware_results/Gabo_t6/image4}
    \includegraphics[width=0.3\textwidth, height=0.4\textheight, keepaspectratio]{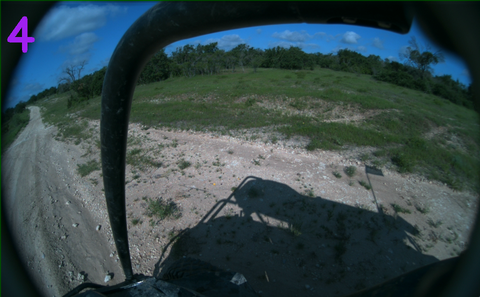}
  }
  \end{subfloat}
  \begin{subfloat}[Long Traverse Image 4 Rear]{
    \centering
    \includegraphics[width=0.3\textwidth, height=0.4\textheight, keepaspectratio]{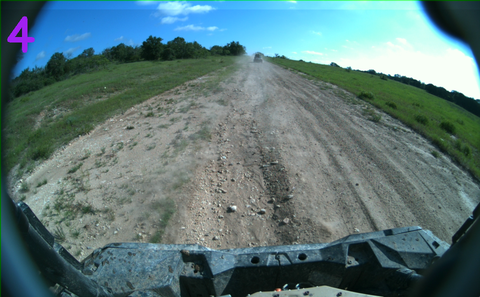}
  }
  \end{subfloat}
  \caption{
    Full set of images from marker 4 in \cref{fig:long_traverse}.
  }
\end{figure*}

\begin{figure*}
  \centering
  \begin{subfloat}[Long Traverse Image 5]{
    \centering
    \includegraphics[trim={0cm, 0cm, 0cm, 0cm}, clip, width=0.3\textwidth, height=0.4\textheight, keepaspectratio]{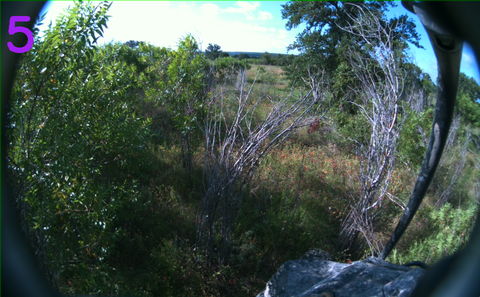}
    \includegraphics[trim={0cm, 0cm, 0cm, 0cm}, clip, width=0.3\textwidth, height=0.4\textheight, keepaspectratio]{Figures/hardware_results/Gabo_t6/image5}
    \includegraphics[width=0.3\textwidth, height=0.4\textheight, keepaspectratio]{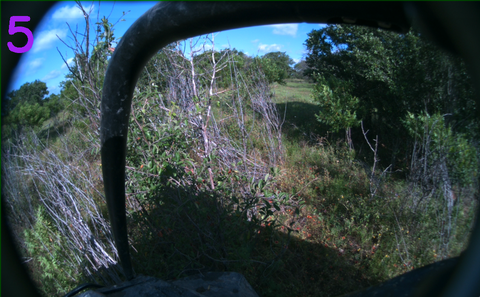}
  }
  \end{subfloat}
  \begin{subfloat}[Long Traverse Image 5 Rear]{
    \centering
    \includegraphics[width=0.3\textwidth, height=0.4\textheight, keepaspectratio]{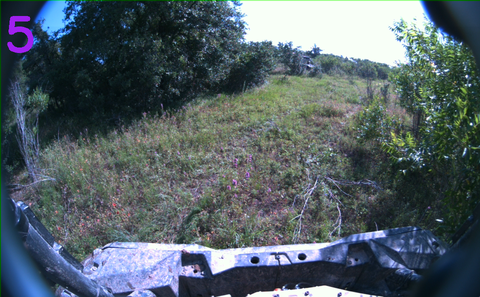}
  }
  \end{subfloat}
  \caption{
    Full set of images from marker 5 in \cref{fig:long_traverse}.
  }
\end{figure*}

\begin{figure*}
  \centering
  \begin{subfloat}[Long Traverse Image 6]{
    \centering
    \includegraphics[trim={0cm, 0cm, 0cm, 0cm}, clip, width=0.3\textwidth, height=0.4\textheight, keepaspectratio]{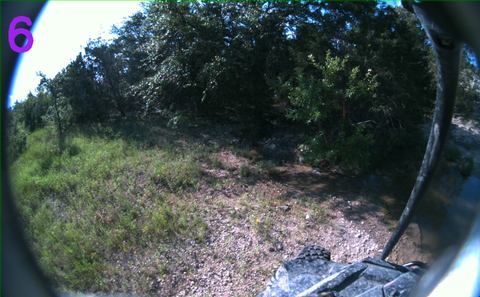}
    \includegraphics[trim={0cm, 0cm, 0cm, 0cm}, clip, width=0.3\textwidth, height=0.4\textheight, keepaspectratio]{Figures/hardware_results/Gabo_t6/image6}
    \includegraphics[width=0.3\textwidth, height=0.4\textheight, keepaspectratio]{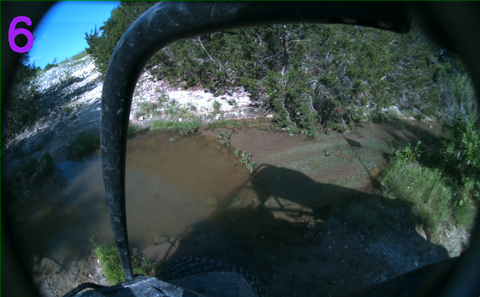}
  }
  \end{subfloat}
  \begin{subfloat}[Long Traverse Image 6 Rear]{
    \centering
    \includegraphics[width=0.3\textwidth, height=0.4\textheight, keepaspectratio]{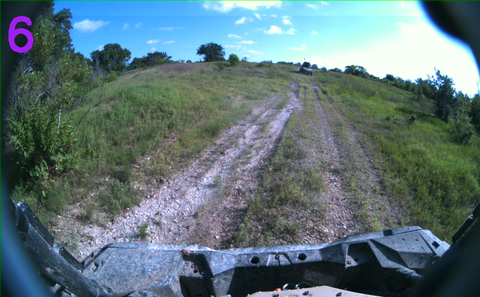}
  }
  \end{subfloat}
  \caption{
    Full set of images from marker 6 in \cref{fig:long_traverse}.
  }
\end{figure*}

\begin{figure*}
  \centering
  \begin{subfloat}[Slide Image 1]{
    \centering
    \includegraphics[trim={0cm, 0cm, 0cm, 0cm}, clip, width=0.3\textwidth, height=0.4\textheight, keepaspectratio]{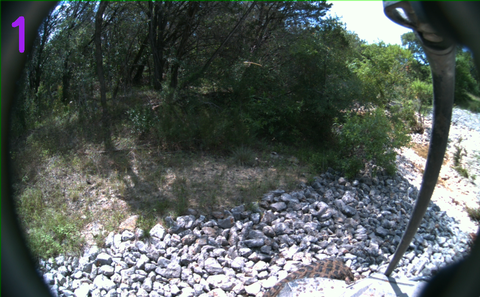}
    \includegraphics[trim={0cm, 0cm, 0cm, 0cm}, clip, width=0.3\textwidth, height=0.4\textheight, keepaspectratio]{Figures/hardware_results/Red_Slide_2/image1}
    \includegraphics[width=0.3\textwidth, height=0.4\textheight, keepaspectratio]{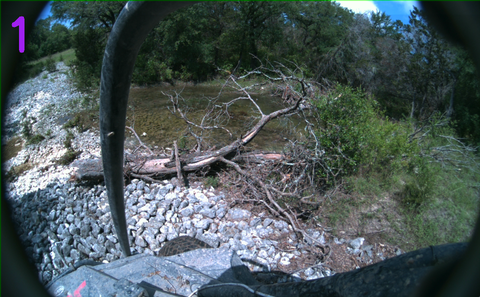}
  }
  \end{subfloat}
  \begin{subfloat}[Slide Image 1 Rear]{
    \centering
    \includegraphics[width=0.3\textwidth, height=0.4\textheight, keepaspectratio]{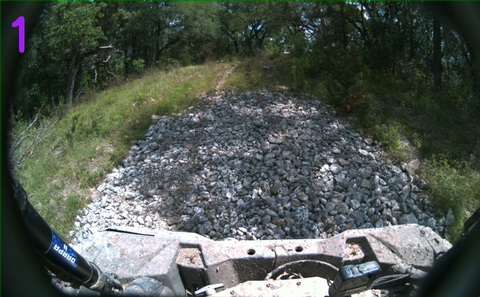}
  }
  \end{subfloat}
  \caption{
    Full set of images from marker 1 in \cref{fig:red_slide}.
  }
\end{figure*}

\begin{figure*}
  \centering
  \begin{subfloat}[Slide Image 2]{
    \centering
    \includegraphics[trim={0cm, 0cm, 0cm, 0cm}, clip, width=0.3\textwidth, height=0.4\textheight, keepaspectratio]{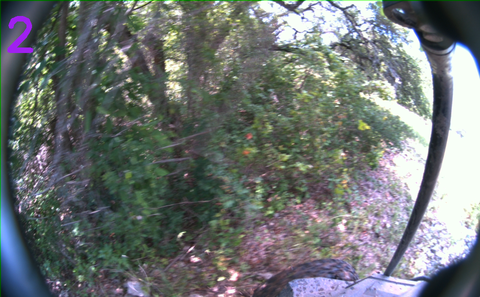}
    \includegraphics[trim={0cm, 0cm, 0cm, 0cm}, clip, width=0.3\textwidth, height=0.4\textheight, keepaspectratio]{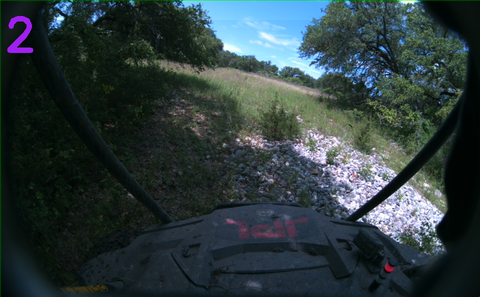}
    \includegraphics[width=0.3\textwidth, height=0.4\textheight, keepaspectratio]{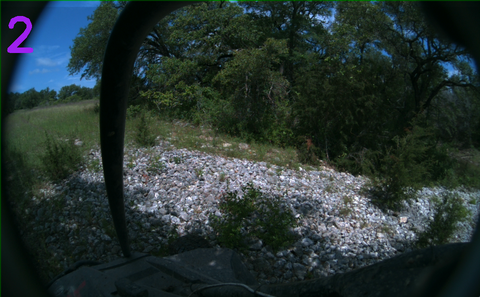}
  }
  \end{subfloat}
  \begin{subfloat}[Slide Image 2 Rear]{
    \centering
    \includegraphics[width=0.3\textwidth, height=0.4\textheight, keepaspectratio]{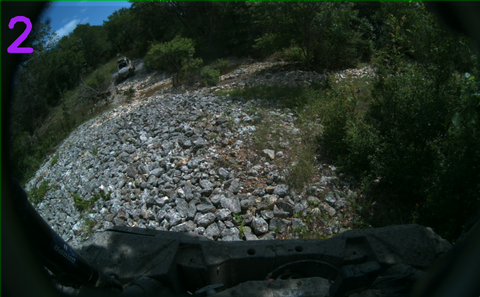}
  }
  \end{subfloat}
  \caption{
    Full set of images from marker 2 in \cref{fig:red_slide}.
  }
\end{figure*}

\begin{figure*}
  \centering
  \begin{subfloat}[Slide Image 3]{
    \centering
    \includegraphics[trim={0cm, 0cm, 0cm, 0cm}, clip, width=0.3\textwidth, height=0.4\textheight, keepaspectratio]{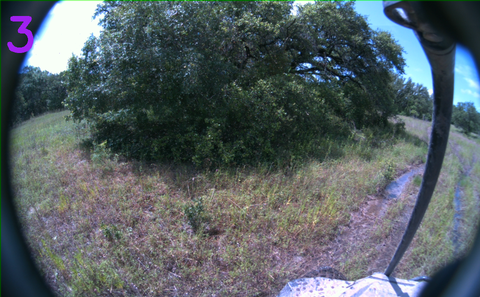}
    \includegraphics[trim={0cm, 0cm, 0cm, 0cm}, clip, width=0.3\textwidth, height=0.4\textheight, keepaspectratio]{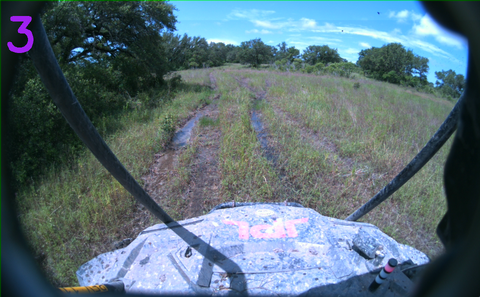}
    \includegraphics[width=0.3\textwidth, height=0.4\textheight, keepaspectratio]{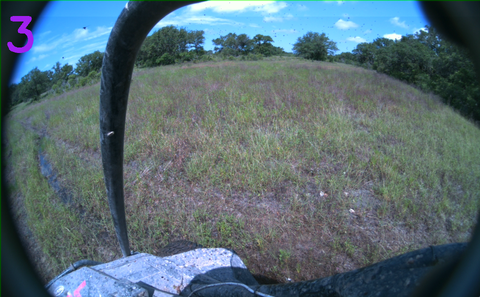}
  }
  \end{subfloat}
  \begin{subfloat}[Slide Image 3 Rear]{
    \centering
    \includegraphics[width=0.3\textwidth, height=0.4\textheight, keepaspectratio]{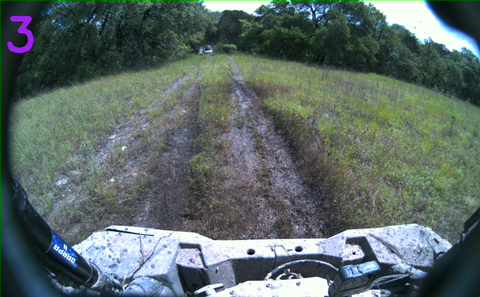}
  }
  \end{subfloat}
  \caption{
    Full set of images from marker 3 in \cref{fig:red_slide}.
  }
\end{figure*}

\begin{figure*}
  \centering
  \begin{subfloat}[Slide Image 4]{
    \centering
    \includegraphics[trim={0cm, 0cm, 0cm, 0cm}, clip, width=0.3\textwidth, height=0.4\textheight, keepaspectratio]{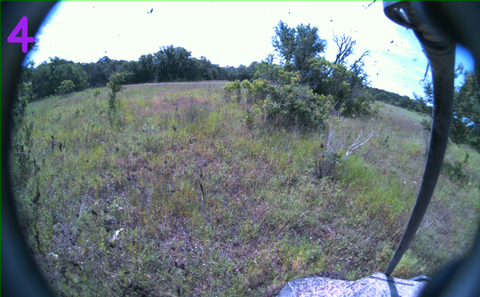}
    \includegraphics[trim={0cm, 0cm, 0cm, 0cm}, clip, width=0.3\textwidth, height=0.4\textheight, keepaspectratio]{Figures/hardware_results/Red_Slide_2/image4}
    \includegraphics[width=0.3\textwidth, height=0.4\textheight, keepaspectratio]{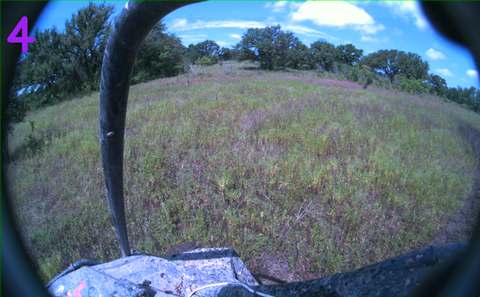}
  }
  \end{subfloat}
  \begin{subfloat}[Slide Image 4 Rear]{
    \centering
    \includegraphics[width=0.3\textwidth, height=0.4\textheight, keepaspectratio]{Figures/hardware_results/Red_Slide_2/rear_image4}
  }
  \end{subfloat}
  \caption{
    Full set of images from marker 4 in \cref{fig:red_slide}.
  }
\end{figure*}

\begin{figure*}
  \centering
  \begin{subfloat}[Tight Slope Image 1]{
    \centering
    \includegraphics[trim={0cm, 0cm, 0cm, 0cm}, clip, width=0.3\textwidth, height=0.4\textheight, keepaspectratio]{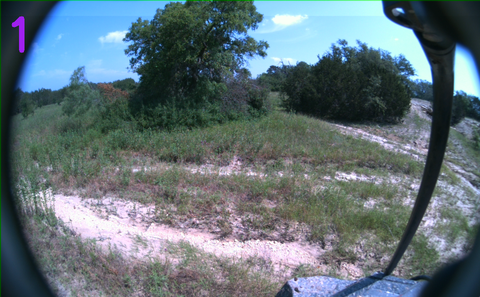}
    \includegraphics[trim={0cm, 0cm, 0cm, 0cm}, clip, width=0.3\textwidth, height=0.4\textheight, keepaspectratio]{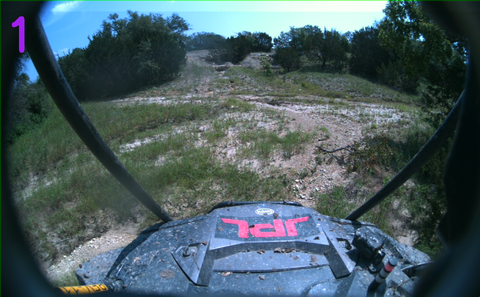}
    \includegraphics[width=0.3\textwidth, height=0.4\textheight, keepaspectratio]{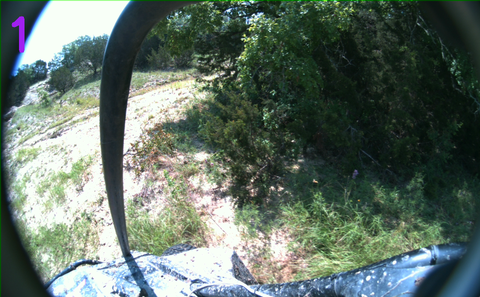}
  }
  \end{subfloat}
  \begin{subfloat}[Tight Slope Image 1 Rear]{
    \centering
    \includegraphics[width=0.3\textwidth, height=0.4\textheight, keepaspectratio]{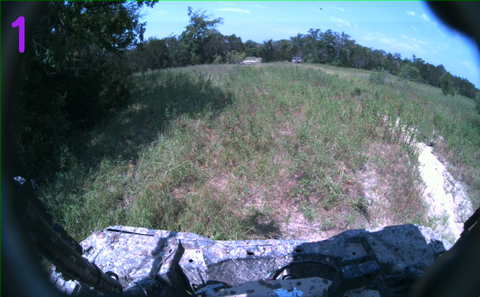}
  }
  \end{subfloat}
  \caption{
    Full set of images from marker 1 in \cref{fig:Tight_Slope_Green}.
  }
\end{figure*}

\begin{figure*}
  \centering
  \begin{subfloat}[Tight Slope Image 2]{
    \centering
    \includegraphics[trim={0cm, 0cm, 0cm, 0cm}, clip, width=0.3\textwidth, height=0.4\textheight, keepaspectratio]{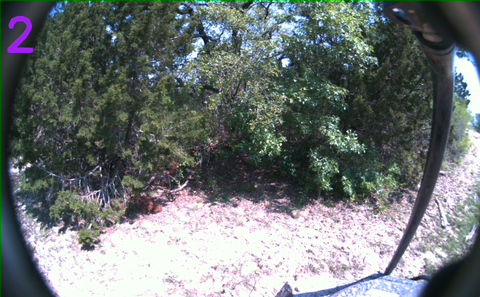}
    \includegraphics[trim={0cm, 0cm, 0cm, 0cm}, clip, width=0.3\textwidth, height=0.4\textheight, keepaspectratio]{Figures/hardware_results/Tight_Slope_Green/image2}
    \includegraphics[width=0.3\textwidth, height=0.4\textheight, keepaspectratio]{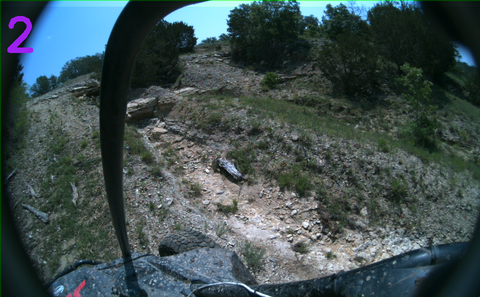}
  }
  \end{subfloat}
  \begin{subfloat}[Tight Slope Image 2 Rear]{
    \centering
    \includegraphics[width=0.3\textwidth, height=0.4\textheight, keepaspectratio]{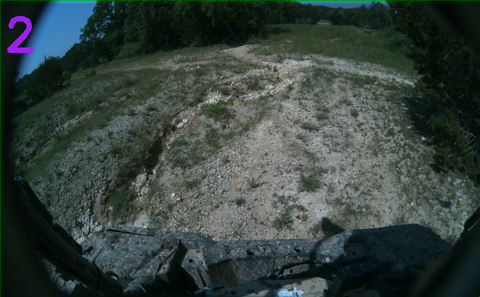}
  }
  \end{subfloat}
  \caption{
    Full set of images from marker 2 in \cref{fig:Tight_Slope_Green}.
  }
\end{figure*}

\begin{figure*}
  \centering
  \begin{subfloat}[Tight Slope Image 3]{
    \centering
    \includegraphics[trim={0cm, 0cm, 0cm, 0cm}, clip, width=0.3\textwidth, height=0.4\textheight, keepaspectratio]{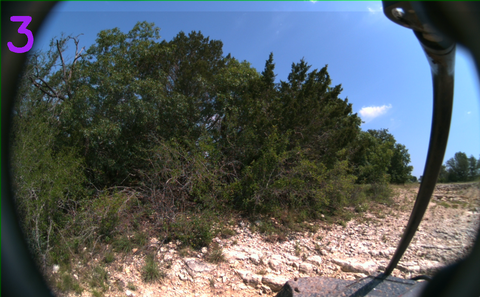}
    \includegraphics[trim={0cm, 0cm, 0cm, 0cm}, clip, width=0.3\textwidth, height=0.4\textheight, keepaspectratio]{Figures/hardware_results/Tight_Slope_Green/image3}
    \includegraphics[width=0.3\textwidth, height=0.4\textheight, keepaspectratio]{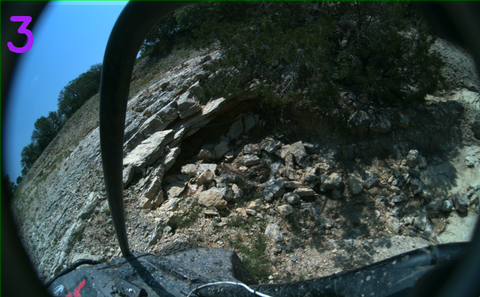}
  }
  \end{subfloat}
  \begin{subfloat}[Tight Slope Image 3 Rear]{
    \centering
    \includegraphics[width=0.3\textwidth, height=0.4\textheight, keepaspectratio]{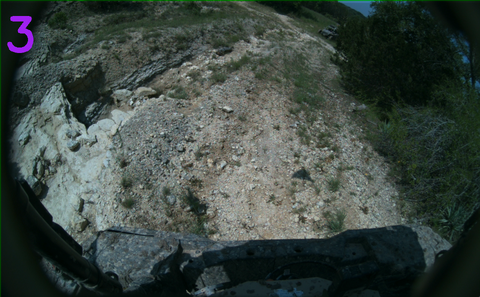}
  }
  \end{subfloat}
  \caption{
    Full set of images from marker 3 in \cref{fig:Tight_Slope_Green}.
  }
\end{figure*}

\begin{figure*}
  \centering
  \begin{subfloat}[Tight Slope Image 4]{
    \centering
    \includegraphics[trim={0cm, 0cm, 0cm, 0cm}, clip, width=0.3\textwidth, height=0.4\textheight, keepaspectratio]{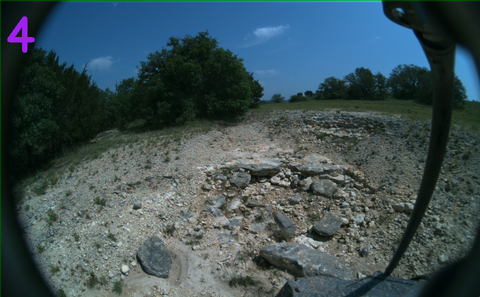}
    \includegraphics[trim={0cm, 0cm, 0cm, 0cm}, clip, width=0.3\textwidth, height=0.4\textheight, keepaspectratio]{Figures/hardware_results/Tight_Slope_Green/image4}
    \includegraphics[width=0.3\textwidth, height=0.4\textheight, keepaspectratio]{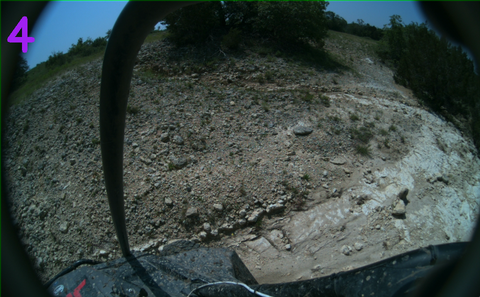}
  }
  \end{subfloat}
  \begin{subfloat}[Tight Slope Image 4 Rear]{
    \centering
    \includegraphics[width=0.3\textwidth, height=0.4\textheight, keepaspectratio]{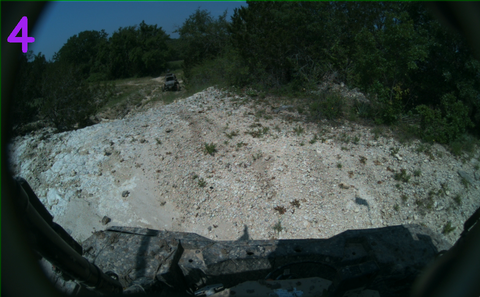}
  }
  \end{subfloat}
  \caption{
    Full set of images from marker 4 in \cref{fig:Tight_Slope_Green}.
  }
\end{figure*}

\begin{figure*}
  \centering
  \begin{subfloat}[Tight Slope Image 5]{
    \centering
    \includegraphics[trim={0cm, 0cm, 0cm, 0cm}, clip, width=0.3\textwidth, height=0.4\textheight, keepaspectratio]{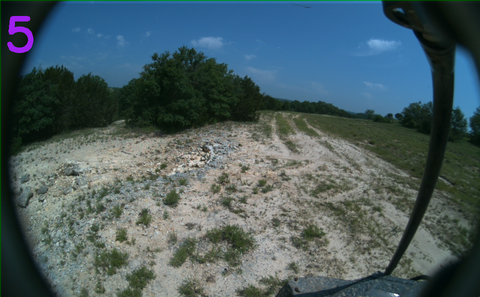}
    \includegraphics[trim={0cm, 0cm, 0cm, 0cm}, clip, width=0.3\textwidth, height=0.4\textheight, keepaspectratio]{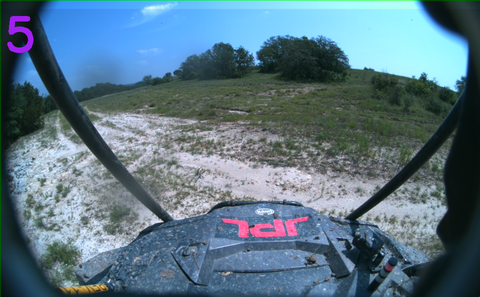}
    \includegraphics[width=0.3\textwidth, height=0.4\textheight, keepaspectratio]{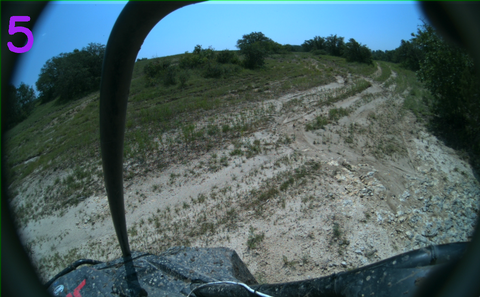}
  }
  \end{subfloat}
  \begin{subfloat}[Tight Slope Image 5 Rear]{
    \centering
    \includegraphics[width=0.3\textwidth, height=0.4\textheight, keepaspectratio]{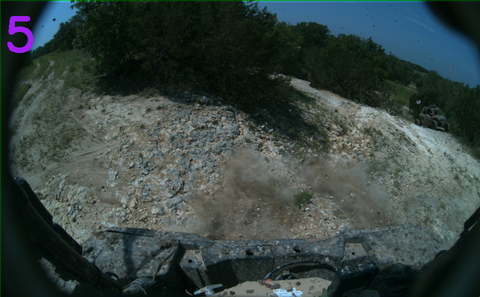}
  }
  \end{subfloat}
  \caption{
    Full set of images from marker 5 in \cref{fig:Tight_Slope_Green}.
  }
\end{figure*}

\begin{figure*}
  \centering
  \begin{subfloat}[Trail T1 Image 1]{
    \centering
    \includegraphics[trim={0cm, 0cm, 0cm, 0cm}, clip, width=0.3\textwidth, height=0.4\textheight, keepaspectratio]{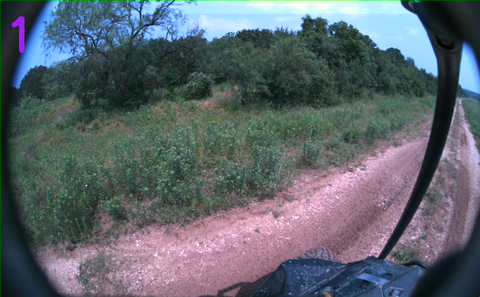}
    \includegraphics[trim={0cm, 0cm, 0cm, 0cm}, clip, width=0.3\textwidth, height=0.4\textheight, keepaspectratio]{Figures/hardware_results/YT_3-5t1/image1}
    \includegraphics[width=0.3\textwidth, height=0.4\textheight, keepaspectratio]{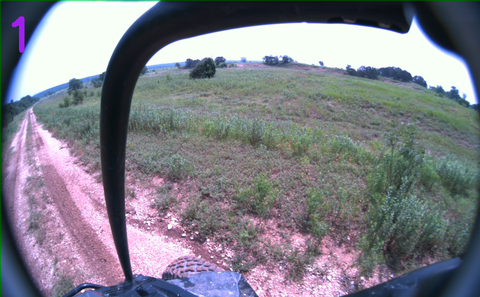}
  }
  \end{subfloat}
  \begin{subfloat}[Trail T1 Image 1 Rear]{
    \centering
    \includegraphics[width=0.3\textwidth, height=0.4\textheight, keepaspectratio]{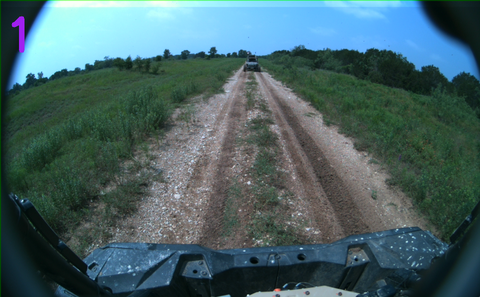}
  }
  \end{subfloat}
  \caption{
    Full set of images from marker 1 in \cref{fig:trail_comparison}.
  }
\end{figure*}

\begin{figure*}
  \centering
  \begin{subfloat}[Trail T2 Image 2]{
    \centering
    \includegraphics[trim={0cm, 0cm, 0cm, 0cm}, clip, width=0.3\textwidth, height=0.4\textheight, keepaspectratio]{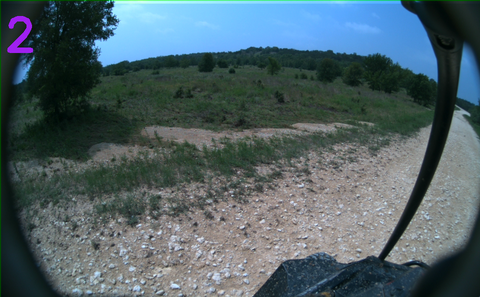}
    \includegraphics[trim={0cm, 0cm, 0cm, 0cm}, clip, width=0.3\textwidth, height=0.4\textheight, keepaspectratio]{Figures/hardware_results/YT_9-11t1/image2}
    \includegraphics[width=0.3\textwidth, height=0.4\textheight, keepaspectratio]{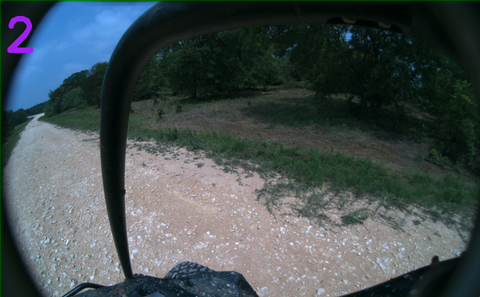}
  }
  \end{subfloat}
  \begin{subfloat}[Trail T2 Image 2 Rear]{
    \centering
    \includegraphics[width=0.3\textwidth, height=0.4\textheight, keepaspectratio]{Figures/hardware_results/YT_9-11t1/image2}
  }
  \end{subfloat}
  \caption{
    Full set of images from marker 2 in \cref{fig:trail_comparison}.
  }
\end{figure*}

%% file: references.bib
@INPROCEEDINGS{Gibson2022,
  author={Gibson, Jason and Vlahov, Bogdan and Fan, David and Spieler, Patrick and Pastor, Daniel and Agha-mohammadi, Ali-akbar and Theodorou, Evangelos A.},
  booktitle={2023 IEEE International Conference on Robotics and Automation (ICRA)}, 
  title={A Multi-step Dynamics Modeling Framework For Autonomous Driving In Multiple Environments}, 
  year={2023},
  volume={},
  number={},
  pages={7959-7965},
  doi={10.1109/ICRA48891.2023.10161330}}

@ARTICLE{Vlahov2023,
  author={Vlahov, Bogdan and Gibson, Jason and Fan, David D. and Spieler, Patrick and Agha-mohammadi, Ali-akbar and Theodorou, Evangelos A.},
  journal={IEEE Robotics and Automation Letters}, 
  title={Low Frequency Sampling in Model Predictive Path Integral Control}, 
  year={2024},
  volume={9},
  number={5},
  pages={4543-4550},
  doi={10.1109/LRA.2024.3382530}}

@article{Fakoorian2022,
author = {Fakoorian, Seyed and Otsu, Kyon and Khattak, Shehryar and Palieri, Matteo and A. A. Agha-mohammadi},
doi = {10.1162/089976605774320557},
issn = {08997667},
journal = {International Foundation of Robotics Research
(ISRR)},
title = {{ROSE: Robust State Estimation via Online Covaraince Adaptation}},
year = {2022}
}

@inproceedings{nubert2022graph,
  title={Graph-based Multi-sensor Fusion for Consistent Localization of Autonomous Construction Robots},
  author={Nubert, Julian and Khattak, Shehryar and Hutter, Marco},
  booktitle={IEEE International Conference on Robotics and Automation (ICRA)},
  year={2022},
  organization={IEEE}
}

@inproceedings{Fan2021,
  title={STEP: Stochastic Traversability Evaluation and Planning for Risk-Aware Off-road Navigation},
  author={Fan, David D and Otsu, Kyohei and Kubo, Yuki and Dixit, Anushri and Burdick, Joel and Agha-Mohammadi, Ali-Akbar},
  booktitle={Robotics: Science and Systems},
  pages={1--21},
  year={2021},
  organization={RSS Foundation}
}

@INPROCEEDINGS{GVOM,
  author={Overbye, Timothy and Saripalli, Srikanth},
  booktitle={2022 IEEE Intelligent Vehicles Symposium (IV)}, 
  title={G-VOM: A GPU Accelerated Voxel Off-Road Mapping System}, 
  year={2022},
  volume={},
  number={},
  pages={1480-1486},
  doi={10.1109/IV51971.2022.9827107}}

@incollection{PyTorch,
title = {PyTorch: An Imperative Style, High-Performance Deep Learning Library},
author = {Paszke, Adam and Gross, Sam and Massa, Francisco and Lerer, Adam and Bradbury, James and Chanan, Gregory and Killeen, Trevor and Lin, Zeming and Gimelshein, Natalia and Antiga, Luca and Desmaison, Alban and Kopf, Andreas and Yang, Edward and DeVito, Zachary and Raison, Martin and Tejani, Alykhan and Chilamkurthy, Sasank and Steiner, Benoit and Fang, Lu and Bai, Junjie and Chintala, Soumith},
booktitle = {Advances in Neural Information Processing Systems 32},
editor = {H. Wallach and H. Larochelle and A. Beygelzimer and F. d\textquotesingle Alch\'{e}-Buc and E. Fox and R. Garnett},
pages = {8024--8035},
year = {2019},
publisher = {Curran Associates, Inc.},
url = {http://papers.neurips.cc/paper/9015-pytorch-an-imperative-style-high-performance-deep-learning-library.pdf}
}

@article{Mohajerin2019, 
archivePrefix = {arXiv}, 
arxivId = {1806.00526}, 
author = {Mohajerin, Nima and Waslander, Steven L.}, 
doi = {10.1109/TNNLS.2019.2891257}, 
eprint = {1806.00526}, 
issn = {21622388}, 
journal = {IEEE Transactions on Neural Networks and Learning Systems}, 
number = {11}, 
pages = {3370--3383}, 
pmid = {30714932}, 
title = {{Multistep Prediction of Dynamic Systems with Recurrent Neural Networks}}, 
volume = {30}, 
year = {2019} 
}

@misc{mppi-generic,
      title={{MPPI-Generic:} A {CUDA} Library for Stochastic Optimization}, 
      author={Bogdan Vlahov and Jason Gibson and Manan Gandhi and Evangelos A. Theodorou},
      year={2024},
      eprint={2409.07563},
      archivePrefix={arXiv},
      primaryClass={cs.MS},
      url={https://arxiv.org/abs/2409.07563}, 
}

@misc{adam,
      title={Adam: A Method for Stochastic Optimization}, 
      author={Diederik P. Kingma and Jimmy Ba},
      year={2017},
      eprint={1412.6980},
      archivePrefix={arXiv},
      primaryClass={cs.LG},
      url={https://arxiv.org/abs/1412.6980}, 
}

@misc{gibson2024dynamicsmodelingusingvisual,
      title={Dynamics Modeling using Visual Terrain Features for High-Speed Autonomous Off-Road Driving}, 
      author={Jason Gibson and Anoushka Alavilli and Erica Tevere and Evangelos A. Theodorou and Patrick Spieler},
      year={2024},
      eprint={2412.00581},
      archivePrefix={arXiv},
      primaryClass={cs.RO},
      url={https://arxiv.org/abs/2412.00581}, 
}

@inproceedings{2010Shakouri,
author = {Shakouri, Payman and Ordys, A. and Askari, M. and Laila, Dina Shona},
year = {2010},
month = {01},
pages = {1-6},
title = {Longitudinal vehicle dynamics using Simulink/Matlab},
volume = {2010},
doi = {10.1049/ic.2010.0410}
}

@ARTICLE{2021Subosits,
  author={Subosits, John K. and Gerdes, J. Christian},
  journal={IEEE Transactions on Intelligent Vehicles}, 
  title={Impacts of Model Fidelity on Trajectory Optimization for Autonomous Vehicles in Extreme Maneuvers}, 
  year={2021},
  volume={6},
  number={3},
  pages={546-558},
  doi={10.1109/TIV.2021.3051325}}

@phdthesis{2020Metzler,
author = {Metzler, Mathias},
year = {2020},
month = {11},
pages = {},
title = {Automotive applications of explicit non-linear model predictive control},
doi = {10.15126/thesis.00853174}
}

@inproceedings{2000UKF, address={Lake Louise, Alta., Canada}, title={The unscented Kalman filter for nonlinear estimation}, ISBN={978-0-7803-5800-3}, url={http://ieeexplore.ieee.org/document/882463/}, DOI={10.1109/ASSPCC.2000.882463}, abstractNote={The Extended Kalman Filter (EKF) has become a standard technique used in a number of nonlinear estimation and machine learning applications. These include estimating the state of a nonlinear dynamic system, estimating parameters for nonlinear system identiﬁcation (e.g., learning the weights of a neural network), and dual estimation (e.g., the Expectation Maximization (EM) algorithm) where both states and parameters are estimated simultaneously. This paper points out the ﬂaws in using the EKF, and introduces an improvement, the Unscented Kalman Filter (UKF), proposed by Julier and Uhlman [5]. A central and vital operation performed in the Kalman Filter is the propagation of a Gaussian random variable (GRV) through the system dynamics. In the EKF, the state distribution is approximated by a GRV, which is then propagated analytically through the ﬁrst-order linearization of the nonlinear system. This can introduce large errors in the true posterior mean and covariance of the transformed GRV, which may lead to sub-optimal performance and sometimes divergence of the ﬁlter. The UKF addresses this problem by using a deterministic sampling approach. The state distribution is again approximated by a GRV, but is now represented using a minimal set of carefully chosen sample points. These sample points completely capture the true mean and covariance of the GRV, and when propagated through the true nonlinear system, captures the posterior mean and covariance accurately to the 3rd order (Taylor series expansion) for any nonlinearity. The EKF, in contrast, only achieves ﬁrst-order accuracy. Remarkably, the computational complexity of the UKF is the same order as that of the EKF.}, booktitle={Proceedings of the IEEE 2000 Adaptive Systems for Signal Processing, Communications, and Control Symposium (Cat. No.00EX373)}, publisher={IEEE}, author={Wan, E.A. and Van Der Merwe, R.}, year={2000}, pages={153–158}, language={en} }

@INPROCEEDINGS{williams2017,
  author={Williams, Grady and Wagener, Nolan and Goldfain, Brian and Drews, Paul and Rehg, James M. and Boots, Byron and Theodorou, Evangelos A.},
  booktitle={2017 IEEE International Conference on Robotics and Automation (ICRA)}, 
  title={Information theoretic MPC for model-based reinforcement learning}, 
  year={2017},
  volume={},
  number={},
  pages={1714-1721},
  doi={10.1109/ICRA.2017.7989202}}

@misc{atha2024fewshotsemanticlearningrobust,
      title={Few-shot Semantic Learning for Robust Multi-Biome 3D Semantic Mapping in Off-Road Environments}, 
      author={Deegan Atha and Xianmei Lei and Shehryar Khattak and Anna Sabel and Elle Miller and Aurelio Noca and Grace Lim and Jeffrey Edlund and Curtis Padgett and Patrick Spieler},
      year={2024},
      eprint={2411.06632},
      archivePrefix={arXiv},
      primaryClass={cs.CV},
      url={https://arxiv.org/abs/2411.06632}, 
}

@InProceedings{choleskyDecomp,
author="Walenty{\'{n}}ski, Ryszard A.",
editor="Bubak, Marian
and van Albada, Geert Dick
and Sloot, Peter M. A.
and Dongarra, Jack",
title="Choleski-Banachiewicz Approach to Systems with Non-positive Definite Matrices with Mathematica®",
booktitle="Computational Science - ICCS 2004",
year="2004",
publisher="Springer Berlin Heidelberg",
address="Berlin, Heidelberg",
pages="311--318",
isbn="978-3-540-25944-2"
}

@misc{han2024dynamicsmodelsaggressiveoffroad,
      title={Dynamics Models in the Aggressive Off-Road Driving Regime}, 
      author={Tyler Han and Sidharth Talia and Rohan Panicker and Preet Shah and Neel Jawale and Byron Boots},
      year={2024},
      eprint={2405.16487},
      archivePrefix={arXiv},
      primaryClass={cs.RO},
      url={https://arxiv.org/abs/2405.16487}, 
}

@article{Goh2020TowardAutomatedVehicleControl,
    author = {Goh, Jonathan Y. and Goel, Tushar and Christian Gerdes, J.},
    title = {Toward Automated Vehicle Control Beyond the Stability Limits: Drifting Along a General Path},
    journal = {Journal of Dynamic Systems, Measurement, and Control},
    volume = {142},
    number = {2},
    pages = {021004},
    year = {2019},
    month = {11},
    issn = {0022-0434},
    doi = {10.1115/1.4045320},
    url = {https://doi.org/10.1115/1.4045320},
    eprint = {https://asmedigitalcollection.asme.org/dynamicsystems/article-pdf/142/2/021004/6472363/ds\_142\_02\_021004.pdf},
}

@ARTICLE{Lee2023TerrainAwareKinedynamic,
  author={Lee, Hojin and Kim, Taekyung and Mun, Jungwi and Lee, Wonsuk},
  journal={IEEE Robotics and Automation Letters}, 
  title={Learning Terrain-Aware Kinodynamic Model for Autonomous Off-Road Rally Driving With Model Predictive Path Integral Control}, 
  year={2023},
  volume={8},
  number={11},
  pages={7663-7670},
  doi={10.1109/LRA.2023.3318190}}

@ARTICLE{Russell2021MultivariateUncertainty,
  author={Russell, Rebecca L. and Reale, Christopher},
  journal={IEEE Transactions on Neural Networks and Learning Systems}, 
  title={Multivariate Uncertainty in Deep Learning}, 
  year={2022},
  volume={33},
  number={12},
  pages={7937-7943},
  doi={10.1109/TNNLS.2021.3086757}}

@inproceedings{Chua2018DeepReinforcementLearning, title={Deep Reinforcement Learning in a Handful of Trials using Probabilistic Dynamics Models}, volume={31}, url={https://proceedings.neurips.cc/paper_files/paper/2018/hash/3de568f8597b94bda53149c7d7f5958c-Abstract.html}, abstractNote={Model-based reinforcement learning (RL) algorithms can attain excellent sample efficiency, but often lag behind the best model-free algorithms in terms of asymptotic performance. This is especially true with high-capacity parametric function approximators, such as deep networks. In this paper, we study how to bridge this gap, by employing uncertainty-aware dynamics models. We propose a new algorithm called probabilistic ensembles with trajectory sampling (PETS) that combines uncertainty-aware deep network dynamics models with sampling-based uncertainty propagation. Our comparison to state-of-the-art model-based and model-free deep RL algorithms shows that our approach matches the asymptotic performance of model-free algorithms on several challenging benchmark tasks, while requiring significantly fewer samples (e.g. 8 and 125 times fewer samples than Soft Actor Critic and Proximal Policy Optimization respectively on the half-cheetah task).}, booktitle={Advances in Neural Information Processing Systems}, publisher={Curran Associates, Inc.}, author={Chua, Kurtland and Calandra, Roberto and McAllister, Rowan and Levine, Sergey}, year={2018} }

@article{Kim2023BridgingActiveExploration, title={Bridging Active Exploration and Uncertainty-Aware Deployment Using Probabilistic Ensemble Neural Network Dynamics}, url={http://arxiv.org/abs/2305.12240}, abstractNote={In recent years, learning-based control in robotics has gained significant attention due to its capability to address complex tasks in real-world environments. With the advances in machine learning algorithms and computational capabilities, this approach is becoming increasingly important for solving challenging control problems in robotics by learning unknown or partially known robot dynamics. Active exploration, in which a robot directs itself to states that yield the highest information gain, is essential for efficient data collection and minimizing human supervision. Similarly, uncertainty-aware deployment has been a growing concern in robotic control, as uncertain actions informed by the learned model can lead to unstable motions or failure. However, active exploration and uncertainty-aware deployment have been studied independently, and there is limited literature that seamlessly integrates them. This paper presents a unified model-based reinforcement learning framework that bridges these two tasks in the robotics control domain. Our framework uses a probabilistic ensemble neural network for dynamics learning, allowing the quantification of epistemic uncertainty via Jensen-Renyi Divergence. The two opposing tasks of exploration and deployment are optimized through state-of-the-art sampling-based MPC, resulting in efficient collection of training data and successful avoidance of uncertain state-action spaces. We conduct experiments on both autonomous vehicles and wheeled robots, showing promising results for both exploration and deployment.}, note={arXiv:2305.12240}, number={arXiv:2305.12240}, publisher={arXiv}, author={Kim, Taekyung and Mun, Jungwi and Seo, Junwon and Kim, Beomsu and Hong, Seongil}, year={2023}, month=may }

@article{pacejkaModel,
 ISSN = {0096736X},
 URL = {http://www.jstor.org/stable/44470677},
 author = {Egbert Bakker and Lars Nyborg and Hans B. Pacejka},
 journal = {SAE Transactions},
 pages = {190--204},
 publisher = {SAE International},
 title = {Tyre Modelling for Use in Vehicle Dynamics Studies},
 urldate = {2025-01-10},
 volume = {96},
 year = {1987}
}

@BOOK{Pacejka2005-ed,
  title     = "Tyre and vehicle dynamics",
  author    = "Pacejka, H B and Pacejka, Hans",
  publisher = "Butterworth-Heinemann",
  edition   =  2,
  month     =  dec,
  year      =  2005,
  address   = "Woburn, MA"
}

@article{Chrosniak2024DeepDynamics, title={Deep Dynamics: Vehicle Dynamics Modeling With a Physics-Constrained Neural Network for Autonomous Racing}, volume={9}, ISSN={2377-3766}, DOI={10.1109/LRA.2024.3388847}, abstractNote={Autonomous racing is a critical research area for autonomous driving, presenting significant challenges in vehicle dynamics modeling, such as balancing model precision and computational efficiency at high speeds (>280 km/h), where minor errors in modeling have severe consequences. Existing physics-based models for vehicle dynamics require elaborate testing setups and tuning, which are hard to implement, time-intensive, and cost-prohibitive. Conversely, purely data-driven approaches do not generalize well and cannot adequately ensure physical constraints on predictions. This letter introduces Deep Dynamics, a physics-constrained neural network (PCNN) for autonomous racecar vehicle dynamics modeling. It merges physics coefficient estimation and dynamical equations to accurately predict vehicle states at high speeds. A unique Physics Guard layer ensures internal coefficient estimates remain within their nominal physical ranges. Open-loop and closed-loop performance assessments, using a physics-based simulator and full-scale autonomous Indy racecar data, highlight Deep Dynamics as a promising approach for modeling racecar vehicle dynamics.}, number={6}, journal={IEEE Robotics and Automation Letters}, author={Chrosniak, John and Ning, Jingyun and Behl, Madhur}, year={2024}, month=jun, pages={5292–5297} }

@article{Djeumou2023AutonomousDrifting, title={Autonomous Drifting with 3 Minutes of Data via Learned Tire Models}, url={http://arxiv.org/abs/2306.06330}, abstractNote={Near the limits of adhesion, the forces generated by a tire are nonlinear and intricately coupled. Efficient and accurate modelling in this region could improve safety, especially in emergency situations where high forces are required. To this end, we propose a novel family of tire force models based on neural ordinary differential equations and a neural-ExpTanh parameterization. These models are designed to satisfy physically insightful assumptions while also having sufficient fidelity to capture higher-order effects directly from vehicle state measurements. They are used as drop-in replacements for an analytical brush tire model in an existing nonlinear model predictive control framework. Experiments with a customized Toyota Supra show that scarce amounts of driving data -- less than three minutes -- is sufficient to achieve high-performance autonomous drifting on various trajectories with speeds up to 45mph. Comparisons with the benchmark model show a $4 times$ improvement in tracking performance, smoother control inputs, and faster and more consistent computation time.}, note={arXiv:2306.06330 [cs, eess]}, number={arXiv:2306.06330}, publisher={arXiv}, author={Djeumou, Franck and Goh, Jonathan Y. M. and Topcu, Ufuk and Balachandran, Avinash}, year={2023}, month=jun }

@inproceedings{Trivedi2023ProbabilisticDynamicModeling, title={Probabilistic Dynamic Modeling and Control for Skid-Steered Mobile Robots in Off-Road Environments}, DOI={10.1109/ICAA58325.2023.00016}, abstractNote={Skid-Steered Mobile Robots (SSMRs) are commonly deployed for autonomous navigation across challenging off-road terrains due to their high maneuverability. However, modeling the tire-terrain interactions for these robots when operating at their dynamic limits is challenging, since slipping and skidding govern their movement. During nominal operation, the data collected from the deviation of the robot’s measured states from their commanded values can be informative of these hard-to-model dynamics. In this work-in-progress paper, we propose a probabilistic motion model for SSMRs by leveraging least squares and Sparse Gaussian Process Regression (SGPR) algorithms. This model allows for a nonlinear stochastic Model Predictive Control (MPC) formulation that can be solved in real-time. Initial results on the application of GPR to account for unmodeled dynamics of a physics-simulated quadrotor are shown, suggesting that it can similarly be put to good use for off-road autonomy applications. We explain how these results reinforce the promising application of an SGPR model to risk-averse motion planning for SSMRs.}, booktitle={2023 IEEE International Conference on Assured Autonomy (ICAA)}, author={Trivedi, Ananya and Bazzi, Salah and Zolotas, Mark and Padır, Taşkın}, year={2023}, month=jun, pages={57–60} }

@article{Revach2022KalmanNet, title={KalmanNet: Neural Network Aided Kalman Filtering for Partially Known Dynamics}, volume={70}, ISSN={1941-0476}, DOI={10.1109/TSP.2022.3158588}, abstractNote={State estimation of dynamical systems in real-time is a fundamental task in signal processing. For systems that are well-represented by a fully known linear Gaussian state space (SS) model, the celebrated Kalman filter (KF) is a low complexity optimal solution. However, both linearity of the underlying SS model and accurate knowledge of it are often not encountered in practice. Here, we present KalmanNet, a real-time state estimator that learns from data to carry out Kalman filtering under non-linear dynamics with partial information. By incorporating the structural SS model with a dedicated recurrent neural network module in the flow of the KF, we retain data efficiency and interpretability of the classic algorithm while implicitly learning complex dynamics from data. We demonstrate numerically that KalmanNet overcomes non-linearities and model mismatch, outperforming classic filtering methods operating with both mismatched and accurate domain knowledge.}, journal={IEEE Transactions on Signal Processing}, author={Revach, Guy and Shlezinger, Nir and Ni, Xiaoyong and Escoriza, Adrià López and van Sloun, Ruud J. G. and Eldar, Yonina C.}, year={2022}, pages={1532–1547} }

@inproceedings{Ko2007GP-UKF, title={GP-UKF: Unscented kalman filters with Gaussian process prediction and observation models}, ISSN={2153-0866}, url={https://ieeexplore.ieee.org/document/4399284/?arnumber=4399284}, DOI={10.1109/IROS.2007.4399284}, abstractNote={This paper considers the use of non-parametric system models for sequential state estimation. In particular, motion and observation models are learned from training examples using Gaussian process (GP) regression. The state estimator is an unscented Kalman filter (UKF). The resulting GP-UKF algorithm has a number of advantages over standard (parametric) UKFs. These include the ability to estimate the state of arbitrary nonlinear systems, improved tracking quality compared to a parametric UKF, and graceful degradation with increased model uncertainty. These advantages stem from the fact that GPs consider both the noise in the system and the uncertainty in the model. If an approximate parametric model is available, it can be incorporated into the GP; resulting in further performance improvements. In experiments, we show how the GP-UKF algorithm can be applied to the problem of tracking an autonomous micro-blimp.}, booktitle={2007 IEEE/RSJ International Conference on Intelligent Robots and Systems}, author={Ko, Jonathan and Klein, Daniel J. and Fox, Dieter and Haehnel, Dirk}, year={2007}, month=oct, pages={1901–1907} }

@article{Trivedi2024DataDrivenSampling, title={Data-Driven Sampling Based Stochastic MPC for Skid-Steer Mobile Robot Navigation}, url={http://arxiv.org/abs/2411.03289}, DOI={10.48550/arXiv.2411.03289}, abstractNote={Traditional approaches to motion modeling for skid-steer robots struggle with capturing nonlinear tire-terrain dynamics, especially during high-speed maneuvers. In this paper, we tackle such nonlinearities by enhancing a dynamic unicycle model with Gaussian Process (GP) regression outputs. This enables us to develop an adaptive, uncertainty-informed navigation formulation. We solve the resultant stochastic optimal control problem using a chance-constrained Model Predictive Path Integral (MPPI) control method. This approach formulates both obstacle avoidance and path-following as chance constraints, accounting for residual uncertainties from the GP to ensure safety and reliability in control. Leveraging GPU acceleration, we efficiently manage the non-convex nature of the problem, ensuring real-time performance. Our approach unifies path-following and obstacle avoidance across different terrains, unlike prior works which typically focus on one or the other. We compare our GP-MPPI method against unicycle and data-driven kinematic models within the MPPI framework. In simulations, our approach shows superior tracking accuracy and obstacle avoidance. We further validate our approach through hardware experiments on a skid-steer robot platform, demonstrating its effectiveness in high-speed navigation. The GPU implementation of the proposed method and supplementary video footage are available at https: //stochasticmppi.github.io.}, note={arXiv:2411.03289 [cs]}, number={arXiv:2411.03289}, publisher={arXiv}, author={Trivedi, Ananya and Prajapati, Sarvesh and Shirgaonkar, Anway and Zolotas, Mark and Padir, Taskin}, year={2024}, month=nov }

@inproceedings{Luo2022LearningBasedNoiseTracking, title={A Learning-based Noise Tracking Method of Adaptive Kalman Filter for UAV Positioning}, url={https://ieeexplore.ieee.org/document/9922222/?arnumber=9922222}, DOI={10.1109/ITSC55140.2022.9922222}, abstractNote={The Unmanned Aerial Vehicle (UAV) is widely used in transportation tasks, disaster rescue, reconnaissance and so on, all of which require high-accuracy positioning. The Global Positioning System (GPS) and the Inertial Navigation System (INS) are two commonly used positioning systems, but both have their shortcomings. When using GPS alone, it will cause large errors due to environmental interference. While when using INS alone, there will be cumulative errors increase with time. Kalman Filter (KF) can overcome the shortcomings of a single GPS or INS system by fusing them together. We propose a learning-based noise tracking method for KF (which is simply called RLKF). The RLKF learns the rule of noise change based on reinforcement learning. The output action of RLKF adjusts noise covariance matrix to the most suitable value, thereby strengthening the adaptive ability of KF. We make strategies output from continuous space by constructing the reinforcement learning model by Deep Determined Policy Gradient (DDPG). By comparing with the adaptive extended KF, experimental results show that the RLKF has better performance, and navigation accuracy is improved.}, booktitle={2022 IEEE 25th International Conference on Intelligent Transportation Systems (ITSC)}, author={Luo, Haohang and Luo, Ying and Han, Bin and Zeng, Min}, year={2022}, month=oct, pages={440–445} }

@article{OHagan2013PC, title={Polynomial Chaos: A Tutorial and Critique from a Statistician’s Perspective}, abstractNote={This article is written in the spirit of helping recent e¤orts to build bridges between the community of researchers in …elds such as applied mathematics and engineering, where the term UQ began, and the community of statisticians who work on problems of uncertainty in the predictions of mechanistic models. It is addressed to researchers and practitioners in both communities.}, author={O’Hagan, Anthony}, language={en}, year={2013} }

@misc{zhang2020quantilepropagation,
      title={Quantile Propagation for Wasserstein-Approximate Gaussian Processes}, 
      author={Rui Zhang and Christian J. Walder and Edwin V. Bonilla and Marian-Andrei Rizoiu and Lexing Xie},
      year={2020},
      eprint={1912.10200},
      archivePrefix={arXiv},
      primaryClass={cs.LG},
      url={https://arxiv.org/abs/1912.10200}, 
}

@inproceedings{Yin2023RiskAwareMPPICVAR, title={Risk-Aware Model Predictive Path Integral Control Using Conditional Value-at-Risk}, url={https://ieeexplore.ieee.org/document/10161100/?arnumber=10161100}, DOI={10.1109/ICRA48891.2023.10161100}, abstractNote={In this paper, we present a novel Model Predictive Control method for autonomous robot planning and control subject to arbitrary forms of uncertainty. The proposed Risk-Aware Model Predictive Path Integral (RA-MPPI) control utilizes the Conditional Value-at-Risk (CVaR) measure to generate optimal control actions for safety-critical robotic applications. Different from most existing Stochastic MPCs and CVaR optimization methods that linearize the original dynamics and formulate control tasks as convex programs, the proposed method directly uses the original dynamics without restricting the form of the cost functions or the noise. We apply the novel RA-MPPI controller to an autonomous vehicle to perform aggressive driving maneuvers in cluttered environments. Our simulations and experiments show that the proposed RA-MPPI controller can achieve similar lap times with the baseline MPPI controller while encountering significantly fewer collisions. The proposed controller performs online computation at an update frequency of up to 80 Hz, utilizing modern Graphics Processing Units (GPUs) to multi-thread the generation of trajectories as well as the CVaR values.}, booktitle={2023 IEEE International Conference on Robotics and Automation (ICRA)}, author={Yin, Ji and Zhang, Zhiyuan and Tsiotras, Panagiotis}, year={2023}, month=may, pages={7937–7943} }

@article{Parwana2024RiskAwareMPPIHybrid, title={Risk-aware MPPI for Stochastic Hybrid Systems}, url={http://arxiv.org/abs/2411.09198}, abstractNote={Path Planning for stochastic hybrid systems presents a unique challenge of predicting distributions of future states subject to a state-dependent dynamics switching function. In this work, we propose a variant of Model Predictive Path Integral Control (MPPI) to plan kinodynamic paths for such systems. Monte Carlo may be inaccurate when few samples are chosen to predict future states under state-dependent disturbances. We employ recently proposed Unscented Transform-based methods to capture stochasticity in the states as well as the state-dependent switching surfaces. This is in contrast to previous works that perform switching based only on the mean of predicted states. We focus our motion planning application on the navigation of a mobile robot in the presence of dynamically moving agents whose responses are based on sensor-constrained attention zones. We evaluate our framework on a simulated mobile robot and show faster convergence to a goal without collisions when the robot exploits the hybrid human dynamics versus when it does not.}, note={arXiv:2411.09198}, number={arXiv:2411.09198}, publisher={arXiv}, author={Parwana, Hardik and Black, Mitchell and Hoxha, Bardh and Okamoto, Hideki and Fainekos, Georgios and Prokhorov, Danil and Panagou, Dimitra}, year={2024}, month=nov }

@inproceedings{Hakobyan2023DistributionallyRobust, title={Distributionally Robust Optimization with Unscented Transform for Learning-Based Motion Control in Dynamic Environments}, DOI={10.1109/ICRA48891.2023.10161246}, abstractNote={Safety is one of the main challenges when applying learning-based motion controllers to practical robotic systems, especially when the dynamics of the robots and their surrounding dynamic environments are unknown. This issue is further exacerbated when the learned information is unreliable and inaccurate. In this paper, we aim to enhance the safety of learning-enabled mobile robots in dynamic environments from the perspective of distributionally robust optimization (DRO) and the unscented transform (UT). Our method infers the unknown dynamics of both the robot and the environment by adopting Gaussian process regression with an uncertainty propagation scheme based on UT to improve prediction accuracy. This leads to a novel learning-based model predictive control (MPC) method in which state information about both the robot and the environment is propagated via UT. The proposed method uses DRO to proactively limit the risk of collisions or other unsafe events in the presence of learning errors. However, the distributionally robust risk constraint is intractable because it involves a separate infinite-dimensional optimization problem. To overcome this challenge, we exploit UT with modern DRO techniques to replace the risk constraint with its simple upper bound. The performance and the utility of our method are demonstrated through simulations in autonomous driving scenarios, showing its capability to enhance safety and computational efficiency.}, booktitle={2023 IEEE International Conference on Robotics and Automation (ICRA)}, author={Hakobyan, Astghik and Yang, Insoon}, year={2023}, month=may, pages={3225–3232} }

@article{Wang2021AdaptiveRiskSensitive, title={Adaptive Risk Sensitive Model Predictive Control with Stochastic Search}, abstractNote={We present a general framework for optimizing the Conditional Value-at-Risk for dynamical systems using stochastic search. The framework is capable of handling the uncertainty from the initial condition, stochastic dynamics, and uncertain parameters in the model. The algorithm is compared against a risk-sensitive distributional reinforcement learning framework and demonstrates improved performance on a simulated pendulum and cartpole with stochastic dynamics. We also showcase the applicability of the framework to robotics as an adaptive risk-sensitive controller by optimizing with respect to the fully nonlinear belief provided by a particle ﬁlter on a pendulum, cartpole, and quadcopter in simulation.}, author={Wang, Ziyi and So, Oswin and Lee, Keuntaek and Theodorou, Evangelos A}, pages={13}, language={en} }

@article{Mohamed2023TowardsEfficientMPPI, title={Towards Efficient MPPI Trajectory Generation with Unscented Guidance: U-MPPI Control Strategy}, url={http://arxiv.org/abs/2306.12369}, abstractNote={The classical Model Predictive Path Integral (MPPI) control framework lacks reliable safety guarantees since it relies on a risk-neutral trajectory evaluation technique, which can present challenges for safety-critical applications such as autonomous driving. Additionally, if the majority of MPPI sampled trajectories concentrate in high-cost regions, it may generate an infeasible control sequence. To address this challenge, we propose the U-MPPI control strategy, a novel methodology that can effectively manage system uncertainties while integrating a more efficient trajectory sampling strategy. The core concept is to leverage the Unscented Transform (UT) to propagate not only the mean but also the covariance of the system dynamics, going beyond the traditional MPPI method. As a result, it introduces a novel and more efficient trajectory sampling strategy, significantly enhancing state-space exploration and ultimately reducing the risk of being trapped in local minima. Furthermore, by leveraging the uncertainty information provided by UT, we incorporate a risk-sensitive cost function that explicitly accounts for risk or uncertainty throughout the trajectory evaluation process, resulting in a more resilient control system capable of handling uncertain conditions. By conducting extensive simulations of 2D aggressive autonomous navigation in both known and unknown cluttered environments, we verify the efficiency and robustness of our proposed U-MPPI control strategy compared to the baseline MPPI. We further validate the practicality of U-MPPI through real-world demonstrations in unknown cluttered environments, showcasing its superior ability to incorporate both the UT and local costmap into the optimization problem without introducing additional complexity.}, note={arXiv:2306.12369 [cs, eess]}, number={arXiv:2306.12369}, publisher={arXiv}, author={Mohamed, Ihab S. and Xu, Junhong and Sukhatme, Gaurav S. and Liu, Lantao}, year={2023}, month=oct }

@inproceedings{Yin2022TrajectoryDistributionControl, title={Trajectory Distribution Control for Model Predictive Path Integral Control using Covariance Steering}, url={https://ieeexplore.ieee.org/document/9811615/?arnumber=9811615}, DOI={10.1109/ICRA46639.2022.9811615}, abstractNote={This paper presents a novel control approach for autonomous systems operating under uncertainty. We combine Model Predictive Path Integral (MPPI) control with Covariance Steering (CS) theory to obtain a robust controller for general nonlinear systems. The proposed Covariance-Controlled Model Predictive Path Integral (CC-MPPI) controller addresses the performance degradation observed in some MPPI implementations owing to unexpected disturbances and uncertainties. Namely, in cases where the environment changes too fast or the simulated dynamics during the MPPI rollouts do not capture the noise and uncertainty in the actual dynamics, the baseline MPPI implementation may lead to divergence. The proposed CC-MPPI controller avoids divergence by controlling the dispersion of the rollout trajectories at the end of the prediction horizon. Furthermore, the CC-MPPI has adjustable trajectory sampling distributions that can be changed according to the environment to achieve efficient sampling. Numerical examples using a ground vehicle navigating in challenging environments demonstrate the proposed approach.}, booktitle={2022 International Conference on Robotics and Automation (ICRA)}, author={Yin, Ji and Zhang, Zhiyuan and Theodorou, Evangelos and Tsiotras, Panagiotis}, year={2022}, month=may, pages={1478–1484} }

@article{Chen2020QuantileSurvey, title={A Survey of Approximate Quantile Computation on Large-scale Data (Technical Report)}, volume={8}, ISSN={2169-3536}, DOI={10.1109/ACCESS.2020.2974919}, abstractNote={As data volume grows extensively, data proﬁling helps to extract metadata of large-scale data. However, one kind of metadata, order statistics, is difﬁcult to be computed because they are not mergeable or incremental. Thus, the limitation of time and memory space does not support their computation on large-scale data. In this paper, we focus on an order statistic, quantiles, and present a comprehensive analysis of studies on approximate quantile computation. Both deterministic algorithms and randomized algorithms that compute approximate quantiles over streaming models or distributed models are covered. Then, multiple techniques for improving the efﬁciency and performance of approximate quantile algorithms in various scenarios, such as skewed data and high-speed data streams, are presented. Finally, we conclude with coverage of existing packages in different languages and with a brief discussion of the future direction in this area.}, note={arXiv:2004.08255 [cs]}, journal={IEEE Access}, author={Chen, Zhiwei and Zhang, Aoqian}, year={2020}, pages={34585–34597}, language={en} }

@article{Han2023ModelPredictiveControl, title={Model Predictive Control for Aggressive Driving Over Uneven Terrain}, url={http://arxiv.org/abs/2311.12284}, abstractNote={Terrain traversability in off-road autonomy has traditionally relied on semantic classification or resource-intensive dynamics models to capture vehicle-terrain interactions. However, our experiences in the development of a high-speed off-road platform have revealed several critical challenges that are not adequately addressed by current methods at our operating speeds of 7--10 m/s. This study focuses particularly on uneven terrain geometries such as hills, banks, and ditches. These common high-risk geometries are capable of disabling the vehicle and causing severe passenger injuries if poorly traversed. We introduce a physics-based framework for identifying traversability constraints on terrain dynamics. Using this framework, we then derive two fundamental constraints, with a primary focus on mitigating rollover and ditch-crossing failures. In addition, we present the design of our planning and control system, which uses Model Predictive Control (MPC) and a low-level controller to enable the fast and efficient computation of these constraints to meet the demands of our aggressive driving. Through real-world experimentation and traversal of hills and ditches, our approach is tested and benchmarked against a human expert. These results demonstrate that our approach captures fundamental elements of safe and aggressive control on these terrain features.}, note={arXiv:2311.12284 [cs]}, number={arXiv:2311.12284}, publisher={arXiv}, author={Han, Tyler and Liu, Alex and Li, Anqi and Spitzer, Alex and Shi, Guanya and Boots, Byron}, year={2023}, month=nov }

@article{Dauner2023PartingWithMisconceptions, title={Parting with Misconceptions about Learning-based Vehicle Motion Planning}, url={http://arxiv.org/abs/2306.07962}, DOI={10.48550/arXiv.2306.07962}, abstractNote={The release of nuPlan marks a new era in vehicle motion planning research, offering the first large-scale real-world dataset and evaluation schemes requiring both precise short-term planning and long-horizon ego-forecasting. Existing systems struggle to simultaneously meet both requirements. Indeed, we find that these tasks are fundamentally misaligned and should be addressed independently. We further assess the current state of closed-loop planning in the field, revealing the limitations of learning-based methods in complex real-world scenarios and the value of simple rule-based priors such as centerline selection through lane graph search algorithms. More surprisingly, for the open-loop sub-task, we observe that the best results are achieved when using only this centerline as scene context (i.e., ignoring all information regarding the map and other agents). Combining these insights, we propose an extremely simple and efficient planner which outperforms an extensive set of competitors, winning the nuPlan planning challenge 2023.}, note={arXiv:2306.07962 [cs]}, number={arXiv:2306.07962}, publisher={arXiv}, author={Dauner, Daniel and Hallgarten, Marcel and Geiger, Andreas and Chitta, Kashyap}, year={2023}, month=nov }

@inproceedings{Cobbe2019QuantifyingGeneralization, title={Quantifying Generalization in Reinforcement Learning}, ISSN={2640-3498}, url={https://proceedings.mlr.press/v97/cobbe19a.html}, abstractNote={In this paper, we investigate the problem of overfitting in deep reinforcement learning. Among the most common benchmarks in RL, it is customary to use the same environments for both training and testing. This practice offers relatively little insight into an agent’s ability to generalize. We address this issue by using procedurally generated environments to construct distinct training and test sets. Most notably, we introduce a new environment called CoinRun, designed as a benchmark for generalization in RL. Using CoinRun, we find that agents overfit to surprisingly large training sets. We then show that deeper convolutional architectures improve generalization, as do methods traditionally found in supervised learning, including L2 regularization, dropout, data augmentation and batch normalization.}, booktitle={Proceedings of the 36th International Conference on Machine Learning}, publisher={PMLR}, author={Cobbe, Karl and Klimov, Oleg and Hesse, Chris and Kim, Taehoon and Schulman, John}, year={2019}, month=may, pages={1282–1289}, language={en} }

@ARTICLE{AutoRally,
  author={Goldfain, Brian and Drews, Paul and You, Changxi and Barulic, Matthew and Velev, Orlin and Tsiotras, Panagiotis and Rehg, James M.},
  journal={IEEE Control Systems Magazine}, 
  title={AutoRally: An Open Platform for Aggressive Autonomous Driving}, 
  year={2019},
  volume={39},
  number={1},
  pages={26-55},
  doi={10.1109/MCS.2018.2876958}}

@article{Nahavandi2025,
author = {Nahavandi, Saeid and Alizadehsani, Roohallah and Nahavandi, Darius and Mohamed, Shady and Mohajer, Navid and Rokonuzzaman, Mohammad and Hossain, Ibrahim},
title = {A Comprehensive Review on Autonomous Navigation},
year = {2025},
issue_date = {September 2025},
publisher = {Association for Computing Machinery},
address = {New York, NY, USA},
volume = {57},
number = {9},
issn = {0360-0300},
url = {https://doi.org/10.1145/3727642},
doi = {10.1145/3727642},
journal = {ACM Comput. Surv.},
month = may,
articleno = {234},
numpages = {67}
}

@Article{Jorge2019,
AUTHOR = {Jorge, Vitor A. M. and Granada, Roger and Maidana, Renan G. and Jurak, Darlan A. and Heck, Guilherme and Negreiros, Alvaro P. F. and dos Santos, Davi H. and Gonçalves, Luiz M. G. and Amory, Alexandre M.},
TITLE = {A Survey on Unmanned Surface Vehicles for Disaster Robotics: Main Challenges and Directions},
JOURNAL = {Sensors},
VOLUME = {19},
YEAR = {2019},
NUMBER = {3},
ARTICLE-NUMBER = {702},
URL = {https://www.mdpi.com/1424-8220/19/3/702},
PubMedID = {30744069},
ISSN = {1424-8220},
DOI = {10.3390/s19030702}
}

@article{Osman2010, address={US}, title={Controlling uncertainty: A review of human behavior in complex dynamic environments}, volume={136}, ISSN={1939-1455}, DOI={10.1037/a0017815}, abstractNote={Complex dynamic control (CDC) tasks are a type of problem-solving environment used for examining many cognitive activities (e.g., attention, control, decision making, hypothesis testing, implicit learning, memory, monitoring, planning, and problem solving). Because of their popularity, there have been many findings from diverse domains of research (economics, engineering, ergonomics, human–computer interaction, management, psychology), but they remain largely disconnected from each other. The objective of this article is to review theoretical developments and empirical work on CDC tasks, and to introduce a novel framework (monitoring and control framework) as a tool for integrating theory and findings. The main thesis of the monitoring and control framework is that CDC tasks are characteristically uncertain environments, and subjective judgments of uncertainty guide the way in which monitoring and control behaviors attempt to reduce it. The article concludes by discussing new insights into continuing debates and future directions for research on CDC tasks. (PsycInfo Database Record (c) 2025 APA, all rights reserved)}, number={1}, journal={Psychological Bulletin}, publisher={American Psychological Association}, author={Osman, Magda}, year={2010}, pages={65–86} }
